\documentclass{article}

\usepackage[preprint]{neurips_2026}

\usepackage[utf8]{inputenc} 
\usepackage[T1]{fontenc}    
\usepackage[colorlinks,citecolor={blue!50!black},linkcolor={red!50!black}]{hyperref}
\usepackage{url}            
\usepackage{booktabs}       
\usepackage{amsfonts}       
\usepackage{nicefrac}       
\usepackage{microtype}      
\usepackage{xcolor}         
\usepackage[table]{xcolor}
\usepackage{amsmath, amssymb}
\usepackage{graphicx}
\usepackage{tcolorbox}
\usepackage[framemethod=tikz]{mdframed}
\usepackage{algorithm,algpseudocode}
\usepackage{bm}
\usepackage{multirow}
\usepackage{subcaption}
\usepackage{tikz}
\usepackage{amsthm}
\usepackage{subfiles}
\usepackage[nolist]{acronym}

\newcommand{\tball}[1]{\textcolor{#1}{\Large$\bullet$}}
\newtheorem{proposition}{Proposition}
\newcommand{\tilS}{\tilde{S}^t(x)}
\newcommand{\tilr}{\tilde{\rho}^t(x)}
\newcommand{\tilz}{\tilde{z}^t(x)}

\definecolor{darkgraybg}{gray}{0.9}
\definecolor{dodgerblue}{rgb}{0.12, 0.365, 1.0}
\definecolor{darkgreen}{rgb}{0.0, 0.39, 0.0}
\definecolor{darkyellow}{rgb}{.78, 0.43, 0.1}
\definecolor{lightgraybg}{gray}{0.92}
\definecolor{v3_1}{RGB}{68,1,84}
\definecolor{v3_2}{RGB}{33,145,140}
\definecolor{v3_3}{RGB}{253,231,37}
\definecolor{v4_1}{RGB}{68,1,84}
\definecolor{v4_2}{RGB}{59,82,139}
\definecolor{v4_3}{RGB}{33,145,140}
\definecolor{v4_4}{RGB}{253,231,37}
\definecolor{v5_1}{RGB}{68,1,84}
\definecolor{v5_2}{RGB}{59,82,139}
\definecolor{v5_3}{RGB}{33,145,140}
\definecolor{v5_4}{RGB}{94,201,97}
\definecolor{v5_5}{RGB}{253,231,37}
\definecolor{v7_1}{RGB}{68,1,84}
\definecolor{v7_2}{RGB}{72,40,120}
\definecolor{v7_3}{RGB}{62,73,137}
\definecolor{v7_4}{RGB}{49,104,142}
\definecolor{v7_5}{RGB}{38,130,142}
\definecolor{v7_6}{RGB}{94,201,97}
\definecolor{v7_7}{RGB}{253,231,37}
\definecolor{v8_1}{RGB}{68,1,84}
\definecolor{v8_2}{RGB}{71,24,106}
\definecolor{v8_3}{RGB}{59,50,133}
\definecolor{v8_4}{RGB}{44,84,142}
\definecolor{v8_5}{RGB}{33,113,141}
\definecolor{v8_6}{RGB}{39,138,129}
\definecolor{v8_7}{RGB}{92,200,99}
\definecolor{v8_8}{RGB}{253,231,37}
\definecolor{lightred}{rgb}{0.8, 0.22, 0.29}
\definecolor{middlegreen}{rgb}{0.0, 0.5, 0.0}
\definecolor{darkpurple}{rgb}{0.42, 0.2, 0.63}

\DeclareRobustCommand{\blackline}{\raisebox{2pt}{\tikz{\draw[black,solid,line width = 1.1pt](0,0) -- (4mm,0);}}}
\DeclareRobustCommand{\purpleline}{\raisebox{2pt}{\tikz{\draw[darkpurple,solid,line width = 1.1pt](0,0) -- (4mm,0);}}}

\usepackage{pifont}

\begin{acronym}[paper]
    \acro{UOT}[uOT]{unbalanced Optimal Transport}
    \acro{CUOT}[CuOT]{Contact unbalanced Optimal Transport}
    \acro{WFR}[WFR]{Wasserstein--Fisher--Rao}
    \acro{OT}[OT]{Optimal Transport}
\end{acronym}

\title{Contact Geometry for Generative Models: \\\ An Unbalanced Optimal Transport Formulation}

\author{%
  Andrea Testa$^{1}$ \ S\o ren Hauberg$^{2}$ \ Andras Gabor Kupcsik$^{1}$ \ Tamim Asfour$^{3}$ \ Leonel Rozo$^{4}$ \\[0.4em]
  {\small $^{1}$Bosch Center for Artificial Intelligence (BCAI), } \
  {\small $^{2}$Technical University of Denmark (DTU), } \\
  {\small $^{3}$Karlsruhe Institute of Technology (KIT), } \
  {\small $^{4}$The Italian Institute of Artificial Intelligence (AI4I)} \\[0.3em]
  {\footnotesize \texttt{\{Andrea.Testa,AndrasGabor.Kupcsik\}@de.bosch.com} \
  \texttt{sohau@dtu.dk}} \\
  {\footnotesize
  \texttt{asfour@kit.edu} \
  \texttt{leonel.rozo@ai4i.it}}
}

\begin{document}

\maketitle
\vspace{-5mm}

\begin{abstract}
  Real-world generative problems in biology, medical imaging, or robotics, rarely come with perfectly paired source and target distributions.
  While \ac{UOT} provides a robust framework to bridge unpaired datasets via minimum-length paths in probability space, existing formulations based on the \ac{WFR} geometry introduce a structural bias toward high density regions, often triggering mode collapse. 
  We introduce \emph{\ac{CUOT}}, a novel \ac{UOT} formulation grounded in \emph{contact geometry} that decouples density transport direction from mass growth to eliminate structural bias.
  This allows \ac{CUOT} to adapt sampling steps to the local data structure, naturally reducing step sizes in high-density regions to prevent overshooting and ensuring stable convergence. 
  The result is a stable, scalable path-length minimization solver that consistently outperforms \ac{WFR}-based methods on biological dynamics reconstruction, unpaired image-to-image translation, and video generation. \href{https://sites.google.com/view/contactunbalancedot/home}{Project website}.\looseness-1
\end{abstract}

\vspace{-5mm}
\section{Introduction}\label{sec:intro}
\begin{figure}[h]
\vspace{-2mm}
\centering
\begin{subfigure}{0.365\textwidth}
    \centering
    \includegraphics[width=\textwidth]{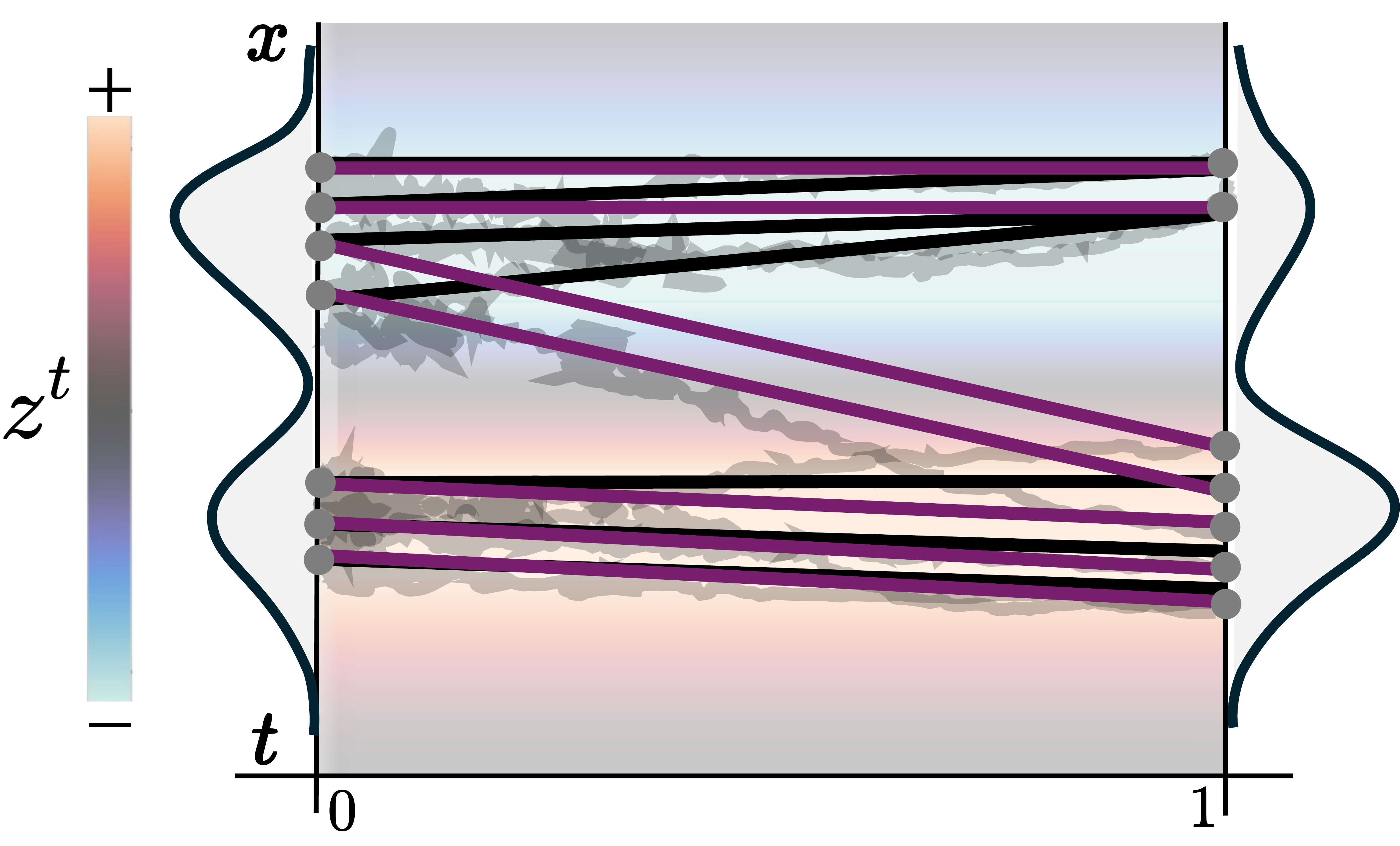}
\end{subfigure}%
\hspace{0pt}%
\begin{subfigure}{0.635\textwidth}
    \centering
    \includegraphics[width=\textwidth]{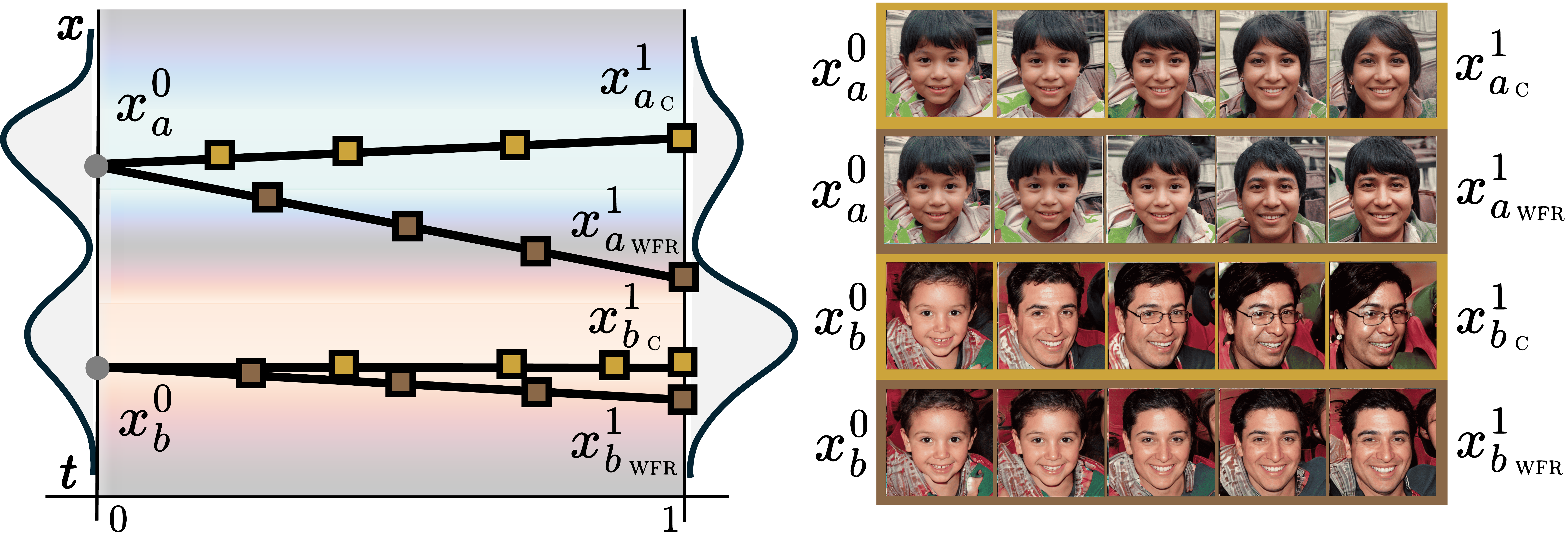}
\end{subfigure}
\vspace{-6mm}
\caption{{\textbf{Left}: Two marginals connected by a diffusion process (noisy gray path), an OT map (\purpleline), and a \ac{UOT} map (\blackline), the latter preventing mode leakage by adapting sample weights according to the growth rate $z^t$. \textbf{Middle-Right}: Comparison between \ac{UOT} methods, namely \ac{WFR} (brown squares) and \ac{CUOT} (gold squares), in a dataset pairing experiment (App.~\ref{app:dataset-pairing}). Our \ac{CUOT} method avoids bias toward high-growth regions and adapts the step size to the data structure.\looseness-1}}
\label{fig:teaser}
\vspace{-6mm}
\end{figure}

\vspace{-2mm}
Generative modeling synthesizes new data by transforming probability distributions, either from noise or more generally between unpaired datasets, resulting in models that capture the temporal evolution of their underlying densities~\citep{yang2022generative,tong2024simulation, neklyudov2023action, ding2022temporal}. 
\emph{The unpaired distributions problem} naturally arises in modeling real-world stochastic dynamics from sparse measurements, like in biology~\citep{yeo2021generative, moon2019visualizing, zhang2024scnode}, climate science~\citep{rasp2021data, verma2024climode}, and economics~\citep{kazakevivcius2021understanding, alouadi2025robust}.
It also plays a central role in domain adaptation, tackling distribution shifts in applications such as synthetic-to-real image translation~\citep{peng2017visda}, medical imaging~\citep{wang2024structure}, and systems control~\citep{li2026recogdrive, li2025towards, xu2025deal}.
Figure~\ref{fig:teaser}-\emph{left} contrasts some leading approaches for connecting probability distributions.
Diffusion processes~\citep{ho2020denoising, karczewski2026the} require an intermediate Gaussian noise
state~\citep{mokrov2021largescale}, whereas \ac{OT} instead enforce direct, minimum-transport connections, yielding minimum-length paths under the Wasserstein metric~\citep{vargas2021solving}.
While often easier to learn and integrate, \ac{OT} approaches are brittle under class imbalance or outliers~\citep{chizat2018, gazdieva2024light}.
Unbalanced OT (\ac{UOT}) addresses this by allowing mass variations during transport, so that semantically related but differently sized modes can be matched without cross-mode leakage~\citep{fatras2021unbalanced}.
Yet, current \ac{UOT} solutions (Sec.~\ref{sec:related_works}) based on the Wasserstein--Fisher--Rao (WFR) distance~\citep{liero2016optimal} introduce a new pathology: they bias the transport dynamics toward regions of maximal mass growth, resulting in soft mode collapse~\citep{chapel2021unbalanced, zhang2025learning}.\looseness-1

\textbf{This paper} overcomes the foregoing limitations by introducing a novel contact geometric formulation of \ac{UOT} (Sec.~\ref{sec:methodology}), which decouples the direction of sample transport dynamics from its weight growth. 
The contact geometry (Sec.~\ref{sec:background}) naturally induces a bias whereby sampling step sizes decrease in regions of increasing weight (i.e., higher data density). 
Since these regions exhibit richer variability~\citep{kumar2026flow}, requiring finer resolution to avoid overshooting~\citep{kharitenko2026landing, li2026kinetic, weed2019sharp}, the \ac{WFR} geometry falls short. 
In contrast, our \emph{Contact Unbalanced Optimal Transport} method successfully adapts generative dynamics by producing smaller steps in high-density regions, as shown in Fig.~\ref{fig:teaser}-\emph{middle-right}.
Furthermore, our contact geometric interpretation of the \ac{UOT} problem enables a stable and efficient solver via path-length minimization (Sec.~\ref{sec:implementation}).
We validate our method on reconstructing stochastic dynamics first through an illustrative ablation study and then on biological datasets, unpaired image-to-image translation, and ensemble regression for video generation, consistently outperforming both OT- and \ac{WFR}-based \ac{UOT} approaches (Sec.~\ref{sec:results}). \looseness-1

\begin{mdframed}[hidealllines=true,backgroundcolor=blue!7,innerleftmargin=.2cm,
  innerrightmargin=.2cm, roundcorner=5pt]
In summary, \textbf{our contributions} are: (1) A contact-geometric formulation of unbalanced optimal transport, where we recast \ac{UOT} through the lens of the geometry of non-conservative dynamics, resulting in a data-informed regularization that adapts sampling step sizes to the local data density and decouples it from mass growth;
(2) A geometry-aware efficient solver, the Contact Unbalanced Optimal Transport (CuOT), which treats generative dynamics as minimum-length curves in probability space. \ac{CUOT} delivers provably stable computation while maintaining an efficient profile: scaling linearly with dimensionality and achieving near-linear complexity in sample size.
\looseness-1
\end{mdframed}

\vspace{-4mm}
\section{Related Works}\label{sec:related_works}
\vspace{-2mm}
\begin{table}[b!]
\vspace{-5mm}
\centering
\setlength{\tabcolsep}{2pt}
\renewcommand{\arraystretch}{1.0}
\begin{tabular}{p{5.5cm}cccc}
\toprule
\textbf{Method} 
    & \textbf{Unbalanced} 
    & \textbf{Geometry} 
    & \textbf{Networks} 
    & \textbf{Complexity} \\
\midrule
\rowcolor{lightgraybg}
\small\citet{schiebinger2019optimal, ventre2023trajectory, lavenant2024toward}
    & \color{lightred}\ding{55}  
    & Wasserstein    
    & $\varphi_\theta^t, \phi_\theta^t$
    & $ \mathcal{O}(I_\text{ref} M D N^2)$ \\
    \rowcolor{lightgraybg}
\small\citet{chizat2022trajectory}         
    & \color{lightred}\ding{55}   
    & Wasserstein   
    & $\varphi_\theta^t, \phi_\theta^t$
    & $ \mathcal{O}(K M D N^2)$ \\
    \rowcolor{lightgraybg}
\small\citet{pariset2023unbalanced}      
    & \color{lightred}\ding{55}   
    & Wasserstein   
    & $\varphi_\theta^t, \phi_\theta^t$
    & $ \mathcal{O}(I_\text{IPF} KDN)$ \\
    \rowcolor{lightgraybg}
\textbf{\small\citet{tong2023improving, tong2024simulation}}
    & \color{lightred}\ding{55}   
    & Wasserstein                          
    & $v^t_\theta$     
    & $ \mathcal{O}(I_\text{ref} M D N)$ \\
    \small \textbf{\small\citet{sha2024reconstructing}};
    \citet{neklyudov2023action, neklyudov2023computational}
    & \color{middlegreen}\ding{51} 
    & WFR                          
    & $v^t_\theta, z^t_\theta$, $\rho^t_\theta$                  
    & $\mathcal{O}(K D N)$ \\
\small\citet{peng2024stvcr}              
    & \color{middlegreen}\ding{51} 
    & WFR                          
    & $v^t_\theta, z^t_\theta$     
    & $\mathcal{O}(M D N^2)$ \\
\small\citet{wang2025joint, wang2026wfr}
    & \color{middlegreen}\ding{51} 
    & WFR                          
    & $v^t_\theta, z^t_\theta$     
    & $ \mathcal{O}(I_\text{ref} M D N^2)$ \\
    \rowcolor{lightgraybg}
\textbf{\small\citet{zhang2025learning}}
    & \color{middlegreen}\ding{51} 
    & WFR*                      
    & $v^t_\theta, z^t_\theta$, $s^t_\theta$
    & $\mathcal{O}(M D N^2)$ \\
    \rowcolor{lightgraybg}
\textbf{\small\citet{sun2025variational}} 
    & \color{middlegreen}\ding{51} 
    & WFR*                      
    & $z^t_\theta$                 
    & $\mathcal{O}(M D N^2)$ \\
\small\citet{testa2026contact}
    & \color{lightred}\ding{55} 
    & Contact                      
    & $v^t_\theta$     
    & $\mathcal{O}(K D N\log{N})$ \\
\textbf{{\small{\ac{CUOT} (ours)}}}
    & \color{middlegreen}\ding{51} 
    & Contact                      
    & $v^t_\theta, z^t_\theta$     
    & $\mathcal{O}(M D N \log{N})$ \\
\bottomrule
\end{tabular}
\vspace{2mm}
\small\caption{
Comparison of OT methods. \textbf{Bold} entries are evaluated in Sec.~\ref{sec:results}.
\textbf{Unbalanced} indicates whether mass variation is explicitly modeled and learned. 
\textbf{Geometry} denotes the structure of the transport problem. 
WFR* indicates that the formulation recovers WFR geometry only for a specific choice of the penalty $\psi(z_\theta^t)$, and otherwise lacks a clear geometric interpretation.
\textbf{Networks} lists the parameterized quantities: $(\varphi_\theta^t, \phi_\theta^t)$ denote the Schrödinger potentials, $v_\theta^t$ the drift, $z_\theta^t$ the growth rate, $\rho_\theta^t$ the density, and $s_\theta^t$ the score function.
\textbf{Complexity} reports the training computational cost: $N$ is the number of samples, $D$ the data dimensionality, $K$ the number of time steps, $M$ the number of intermediate marginals, $I_\text{IPF}$ the number of IPF iterations, and $I_\text{ref}$ the number of reference refinement iterations.\looseness-1
}
\vspace{-5mm}
\label{tab:comparison}
\end{table}
Table~\ref{tab:comparison} surveys recent approaches for learning OT maps between stochastic distributions. 
Using OT solvers during training to measure the discrepancy between predicted and reference marginals has become standard in the field. 
Despite their super-linear complexity in the number of samples $N$ (ranging from  $\mathcal{O}(N\log N)$ to $\mathcal{O}(N^2)$), OT solvers achieve high reconstruction accuracy via explicit sample matching, compared to density-estimation methods based on direct parameterization or normalizing flows~\citep{sha2024reconstructing, tong2023improving, neklyudov2023computational}.
A key feature of the OT's geometric structure is the coupling between the transport drift $v^t$ and the local growth rate $z^t$. 
Balanced OT approaches~\citep{tong2023improving}  assume zero mass change, or hardcode the growth rate using task-specific knowledge~\citep{chizat2022trajectory, lavenant2024toward, pariset2023unbalanced, ventre2023trajectory, schiebinger2019optimal}, which fails to extend to broader cases.

More generally, in unbalanced settings, mass variation is controlled via a penalty $\psi(z^t)$, inducing the constraint $v^t = \nabla \partial_z \psi(z^t)$~\citep{zhang2025learning, sun2025variational, wang2025joint}. 
This constraint is enforced either by parameterizing the growth rate alone, which can yield noisy gradient estimates, or through a regularization term in the loss.
The quadratic penalty $\psi(z^t)=\frac{1}{2}\|z^t\|^2$~\citep{sha2024reconstructing, peng2024stvcr, wang2026wfr, neklyudov2023action, neklyudov2023computational} yields a tighter coupling $v^t = \nabla z^t$, which corresponds to a WFR geometry that admits a distance-based regularization that is more stable than Hamilton--Jacobi--Bellman-based (HJB) formulations~\citep{buzun2026hota}.
Instead, contact geometry relaxes the constraint between the direction of $v^t$ and $z^t$. 
Building on~\citet{testa2026contact} (App. \ref{app:non-conservative_OT}), who used a fixed time schedule for $z^t$ to modulate $\Vert v^t \Vert$ in balanced OT, we learn $z^t$ as a spatially-varying growth rate field. 
This enables general \ac{UOT} while adapting $\Vert{}v^t\Vert{}$ directly to the data.
\looseness-1
\vspace{-3mm}
\section{Geometric Background}\label{sec:background}
\vspace{-1mm}
We introduce the necessary notation throughout the following sections, and summarize it in App. \ref{app:notation}.

\textbf{Dynamics on the probability manifold.~~}
Let $\mathcal{P}^+(\mathcal{M})$ be the space of smooth, positive densities $\rho(x) : \mathcal{M} \to \mathbb{R}^+$ with total mass $\int_{\mathcal{M}} \rho(x)\,dx$, where $x \in \mathcal{M}$ denotes the sample space. 
A time-evolving density is described by a trajectory $\{\rho^t\}_{t \geq 0} \subset \mathcal{P}^+(\mathcal{M})$, whose infinitesimal variation is given by $\partial_t \rho^t(x) \in \mathcal{T}_{\rho^t}\mathcal{P}^+(\mathcal{M})$, with $\mathcal{T}_{\rho^t}\mathcal{P}^+(\mathcal{M})$ being the tangent space.
The pair $(\rho^t, \partial_t \rho^t)$ thus defines a trajectory on the tangent bundle $\mathcal{T}\mathcal{P}^+(\mathcal{M})$, the collection of all tangent spaces.
The evolution of $\rho^t$ is typically modeled by the Fokker--Planck (FP) equation, combining deterministic density transport with its associated velocity field $v^t$ and diffusion of strength $\varepsilon$,\looseness-1
\begin{equation}
\label{eq:fokker-planck}
    \partial_t \rho^t(x) = -\nabla_x \! \cdot \! \big(\rho^t(x)\, v^t(x)\big) + \varepsilon \Delta_x \rho^t(x).
\end{equation}
The classic OT seeks the trajectory $\rho^t \subset \mathcal{P}^+(\mathcal{M})$ connecting two densities $\rho_a, \rho_b \in \mathcal{P}^+(\mathcal{M})$ that minimizes the kinetic energy $\frac12\Vert v^t(x)\Vert^2 \rho^t(x)$~\citep{benamou2000computational}. 
The diffusion term introduces an entropy regularization in the transport map, interpretable as a solution of the Schrödinger Bridge (SB) problem \citep{schrodinger1931umkehrung, schrodinger1932theorie}, or as a \emph{geodesic} in $\mathcal{P}^+(\mathcal{M})$ under the Wasserstein metric $g^{\mathcal{W}_{2}}$~\citep{ambrosio2005gradient}, which induces the corresponding distance $d^{\mathcal{W}_2}$. In particular,
\looseness-1
\begin{equation}
\label{eq:wasserstein_metric}
g^{\mathcal{W}_2}(\partial_t\rho^t, \partial_t\rho^t)
= \int_{\mathcal{M}} \sqrt{\partial_t\rho^t(x)\, (-\Delta_{\rho})^\dagger \partial_t\rho^t(x)} \, dx,
\end{equation}
where $(-\Delta_{\rho})^\dagger$ is the inverse of the weighted Laplacian operator $\Delta_{\rho} = -\nabla_x \! \cdot \! (\rho^t \nabla_x)$~\citep{chow2020wasserstein}
(see App. \ref{app:wass_equivalence} for details).
The density dynamics $(\rho^t, \partial_t \rho^t) \subset \mathcal{T}\mathcal{P}^+(\mathcal{M})$ can be equivalently expressed as $(\rho^t, S^t)$ on the cotangent bundle $\mathcal{T}^*\mathcal{P}^+(\mathcal{M})$, the dual of $\mathcal{T}\mathcal{P}^+(\mathcal{M})$. Here, the covector $S^t(x) := (-\Delta_{\rho})^\dagger \partial_t \rho^t(x)$ is defined to transform covariantly with $\partial_t \rho^t$, ensuring invariance of the Wasserstein metric \eqref{eq:wasserstein_metric}. In the OT setting, $S^t$ is the potential generating the flow via $v^t = \nabla_x S^t(x)$.
\looseness-1

\textbf{Symplectic and contact geometries.~~}On the cotangent bundle $\mathcal{T}^*\mathcal{P}^+(\mathcal{M})$, the vector field $(\partial_t \rho^t, \partial_t S^t)$, which is the solution of the OT problem, respects the \emph{symplectic} constraint $\omega\big((\partial_t \rho^t, \partial_t S^t),\cdot\big) = d \mathcal{H}(\partial_t \rho^t, \partial_t S^t)$.
That is, the density dynamics evolve along level sets of the \emph{Hamiltonian} function $\mathcal{H}$, corresponding to constant volume sections of $\mathcal{T}^*\mathcal{P}^+(\mathcal{M})$~\citep{chow2020wasserstein}. 
This \emph{symplectic form} $\omega$, a geometric structure measuring a generalized notion of area, is preserved along the dynamics in traditional OT settings.
\ac{UOT} relaxes this constraint, which can be geometrically achieved by augmenting the cotangent bundle with an additional scalar variable $z^t$, governing deviations from Hamiltonian level sets. 
The resulting dynamics evolve on the extended space $\mathcal{T}^*\mathcal{P}^+(\mathcal{M}) \times \mathbb{R}$ and are described by a \emph{contact structure} satisfying $d\eta\big((\partial_t \rho^t, \partial_t S^t, \partial_t z^t),\cdot\big) - \mathcal{L}_{(\partial_t \rho^t, \partial_t S^t, \partial_t z^t)}\eta(\cdot)= d \mathcal{H}(\partial_t \rho^t, \partial_t S^t, \partial_t z^t)$. 
Here, $\eta$ denotes the \emph{contact form}, a generalized signed length quantifying the component of the vector field transversing Hamiltonian flow lines.
$\mathcal{L}_{(\partial_t \rho^t, \partial_t S^t,\partial_t z^t)}\eta(\cdot)$ denotes the Lie derivative of the contact form along the vector field. In general, this term is nonzero, implying that the induced dynamics do not preserve the volume element, but instead evolve across sections of $\mathcal{T}^*\mathcal{P}^+(\mathcal{M})$ with varying volume (see App. \ref{app:sympletcic_contact}). 
While the symplectic dynamics correspond to a Wasserstein geodesic, the contact dynamics project to geodesics w.r.t. a dynamically varying metric (see App. \ref{app:riemannian}).
As shown in Sec.~\ref{sec:methodology}, this change in the local volume form across the contact dynamics induces density changes that correspond to mass variations on $\mathcal{P}^+(\mathcal{M})$, thereby moving from symplectic OT to the more general \ac{UOT} setting.\looseness-1
\vspace{-5mm}
\section{Contact Unbalanced Optimal Transport}\label{sec:methodology}
\vspace{-2mm}
\textbf{Problem definition.~~} We propose an entropy-regularized uOT rooted in contact geometry. By treating the growth rate $z^t$ as an independent dynamical variable, we decouple it from the spatial transport velocity $v^t$. In contrast, WFR formulations build on symplectic variables $(\rho^t, S^t)$, constraining the growth rate to the action potential via $z^t := S^t$, and thus enforcing coupled dynamics where $v^t = \nabla_x z^t$.
Let $\mathcal{M}$ be a domain and let $\{\rho^m\}_{m=0}^M$ be a collection of densities observed at times $\{t_m\}_{m=0}^M$. We seek a continuous trajectory $\{\rho^t\}_{t\in[t_0,t_M]}$, interpolating these marginals, together with a velocity field $v^t$ and a scalar field $z^t$ governing local mass variation. The problem is defined as,\looseness-1
\begin{subequations}
\label{eq:contact_schrodinger_bridge}
\vspace{-1mm}
\begin{align}
    \min_{v^t, z^{t_0}} \quad 
    & J = \int_{t_0}^{t_M} \!\! \int_\mathcal{M} \partial_t z^t(x)\, dx \, dt, 
    \label{eq:CSB:J} \\
    \text{s.t.} \quad 
    & \partial_t z^t(x) 
    = \tfrac{1}{2} \|v^t(x)\|^2 \rho^t(x) 
    - \tfrac{1}{2} \|v^{t_0}(x)\|^2 \rho^{t_0}(x)
    +  \tfrac12 w_z\, z^t(x)^2, 
    \label{eq:CSB:z} \\
    & \partial_t \rho^t(x) 
    + \nabla_x \cdot \big(\rho^t(x)\, v^t(x)\big) 
    = \varepsilon \Delta_x \rho^t(x) 
    + w_z\, z^t(x)\, \rho^t(x), 
    \label{eq:CSB:FP} \\
    & \rho^{t_m} = \rho^m, 
    \quad \forall m \in \{0, \dots, M\}.
    \label{eq:CSB:M}
\end{align}
\end{subequations}
The objective functional~\eqref{eq:CSB:J} minimizes the terminal value of the action $z^t$, whose evolution is governed by the dynamics~\eqref{eq:CSB:z}.
Equation~\eqref{eq:CSB:FP} is a generalized FP equation of the state $\rho^t$, extending the balanced FP~\eqref{eq:fokker-planck} with the term $w_z z^t(x)\rho^t(x)$ to account for mass modulation.
In the limit $w_z \to 0$, the problem reduces to minimizing the kinetic energy of the transport map, recovering the balanced OT formulation (details in Sec. \ref{sec:comparison}). 
However, for $w_z\! \neq\! 0$, the dynamics introduce a nonlinear recursive dependence on $z^t$, such that it is neither a static action nor reducible to a standard state variable.
In particular, the system cannot be cast into a symplectic form $\big((\rho^t, z^t), (S^t, \lambda^t)\big)$ with a conjugate variable $\lambda^t$ acting as dual potential of $z^t$, and thus, falls outside the scope of classical Hamiltonian formulations on even-dimensional phase spaces. 
Instead, the system~\eqref{eq:contact_schrodinger_bridge} is naturally described within the Herglotz variational framework~\citep{herglotz1930}, leading to a contact Hamiltonian system on an odd-dimensional contact manifold.

\textbf{Contact Hamiltonian solution.~~}As demonstrated in App.~\ref{app:Contact-sb_demons}, the optimal solution of problem~\eqref{eq:contact_schrodinger_bridge} follows a contact Hamiltonian dynamics with Hamiltonian $\mathcal{H}$ defined as,\looseness-1
\begin{equation}
\label{eq:contact_hamiltonian_red}
\vspace{-1mm}
\mathcal{H}(\rho^t, S^t, z^t) 
=  \mathcal{K}(\rho^t, S^t) + \varepsilon \mathcal{F}(\rho^t, S^t)
- w_z \int_\mathcal{M} \tfrac{1}{2} z^t(x)^2 dx + \mathcal{H}_0.
\end{equation}
At optimality, $v^t = \nabla_x S^t$, so the term $\mathcal{K}=\int_\mathcal{M} \tfrac{1}{2} \|\nabla_x S^t(x)\|^2 \, \rho^t(x)\, dx$ represents the kinetic energy. 
The term $\mathcal{F}={\int_\mathcal{M} S^t(x)\,\Delta_x \rho^t(x)\,dx}$ introduces an entropic coupling, scaled by the diffusion coefficient $\varepsilon$, which adjusts the transport map to preserve the entropy level imposed by the FP constraint~\eqref{eq:CSB:FP}.
Finally, the term $\tfrac{1}{2} z^t(x)^2$ governs energy variation, while the constant $\mathcal{H}_0= \int_\mathcal{M} \tfrac12 \Vert v^{t_0}(x)\Vert^2 \rho^{t_0}(x) dx$ fixes the reference energy level. 
The contact optimality conditions are,\looseness-1
\begin{subequations}
\label{eq:contact_hamiltonian_dynamics}
\begin{align}
\partial_t{\rho}^t(x) &= \partial_S \mathcal{H}(\cdot) - \tfrac{1}{2} \rho^t(x) \partial_z \mathcal{H}(\cdot) \notag \\ &= -\nabla_x \cdot \left(\rho^t(x)\nabla_x S^t(x)\right) + \varepsilon \Delta_x \rho^t(x) +w_z \, z^t(x)\rho^t(x), \quad \color{darkgray} \text{(via the FP eq. \eqref{eq:CSB:FP})}
\label{eq:contact_rho}
\\
\partial_t{S}^t(x) &= -\partial_\rho \mathcal{H}(\cdot) - \tfrac{1}{2} S^t(x) \partial_z \mathcal{H}(\cdot) \notag \\ &= -\tfrac{1}{2}\|\nabla_x S^t(x)\|^2 
    - \varepsilon \Delta_x S^t(x) +w_z \, z^t(x)S^t(x), \label{eq:contact_S}
\\
\partial_t{z}^t(x) &= \tfrac{1}{2} \left( S^t(x) \partial_S \mathcal{H}(\cdot) + \rho^t(x) \partial_\rho \mathcal{H}(\cdot) \right) - \mathcal{H}(\cdot) \notag \\ &= \tfrac12 \Vert \nabla_x S^t(x) \Vert^2 \rho^t(x) + \tfrac12 w_z \, z^t(x)^2 - \tfrac12\Vert v^{t_0}\Vert^2 \rho^t(x),
\hspace{0.2cm} \color{darkgray} \text{(via the $z^t$ dynamics \eqref{eq:CSB:z})}
\label{eq:contact_z}
\end{align}
\end{subequations}
where the partial derivatives of the contact Hamiltonian $\mathcal{H}$ provide the local coordinate form of the contact geometric constraints (see Sec.~\ref{sec:background}).
The explicit dependence on $z^t$ encodes the trajectory traversed so far, distinguishing the contact Hamiltonian framework from the classical Hamiltonian setting. 
This enables modeling non-conservative, path-dependent forces, allowing the system’s energy to evolve over time~\citep{testa2026contact}.
In particular, this system induces a source or sink term in the generalized FP equation~\eqref{eq:contact_rho}, along with a corresponding amplification or damping term in the HJB equation~\eqref{eq:contact_S}. 
More generally, the non-conservative dynamics implies that the volume of regions in $\mathcal{T}^*\mathcal{P}^+(\mathcal{M})$ explored by $(\rho^t, S^t, z^t)$ is not preserved. 
As a result, both the transport velocity and the density are affected: the former adapts to traverse varying-volume regions, while the latter varies according to the local expansion or contraction of the support. 
This geometric structure is fundamental, as it couples mass variation and transport while keeping them as distinct, independently evolving quantities. 
\looseness-1

\textbf{Geodesic interpretation.~~}
The optimization~\eqref{eq:contact_schrodinger_bridge} can be reformulated on the probability space $\mathcal{P}^+(\mathcal{M})$ by reparametrizing time $t$ with an arc-length variable $s$, such that $ds = \sqrt{\mathcal{K}}\, dt$. 
Consequently, the problem reduces to finding a distance-minimizing curve $\rho^s \in \mathcal{P}^+(\mathcal{M})$ w.r.t.\@ a suitable metric~\citep{udricste2000geometric, abraham2008foundations,testa2025geometric}. 
In the limit $w_z, \varepsilon \to 0$, the cost functional \eqref{eq:CSB:J} is equivalent to the Wasserstein distance (see App.~\ref{app:wass_equivalence}), and the problem reduces to a standard Wasserstein geodesic computation.
In our formulation, however, two aspects reshape this geometric picture. 
First, $\mathcal{F}$ introduces an entropy regularization component, so that the resulting metric increases for distributions $\rho^s$ deviating from the reference entropy measure. 
Sinkhorn-based methods~\citep{feydy2019interpolating} can efficiently incorporate this effect when approximating the distributional distance.
Second, the explicit dependence on $z^t$ drives energy variations along the trajectories in $\mathcal{T}^*\mathcal{P}^+(\mathcal{M})$, reflecting density volume changes in the underlying space. 
To account for this, we adjust the Wasserstein metric with a conformal scaling factor $\sqrt{\mathcal{K}}$, yielding the \emph{contact Wasserstein metric} $g^{\mathcal{C}} = \sqrt{\mathcal{K}}\, g^{\mathcal{W}_2}$. 
Thus, the solution $\rho^s$ is a piecewise entropy-regularized geodesic w.r.t. the dynamically varying metric $g^{\mathcal{C}}$.
Computing the solution as a minimum-length path is advantageous, as it promotes optimization stability (see App.~\ref{app:stability}), avoiding the need to enforce the HJB equation~\eqref{eq:contact_S} in the loss, whose direct minimization is prone to instabilities~\citep{buzun2026hota}.
\looseness-1

\vspace{-2mm}
\section{CuOT: Solving the Contact uOT}\label{sec:implementation}
\vspace{-2mm}
\textbf{Geodesic discretization.~~} Our objective is to compute the geodesic induced by problem~\eqref{eq:contact_schrodinger_bridge}, i.e., a path $\rho^s \subset \mathcal{P}^+(\mathcal{M})$ consistent with the contact Hamiltonian dynamics and interpolating the observed marginals $\{\rho^m\}_{m=0}^M$.
These constraints naturally lead to a discretization of $\rho^s$ as a sequence of nodes $\{\rho^{s_k}\}_{k=0}^K \subset \mathcal{P}^+(\mathcal{M})$, each represented by a weighted particle ensemble $\rho^{s_k}=\{(x_n^{s_k}, w_n^{s_k})\}_{n=0}^N$ approximating the corresponding density.
Therefore, each discrete marginal is 
obtained by integrating: \emph{(1)} a parameterized \color{darkgreen}{ transport network $v_\theta(x^{s_k}, s_k)$}\color{black}, representing the velocity field; and \emph{(2)} a \color{dodgerblue} growth network $z_\theta(x^{s_k}, s_k)$\color{black}, representing the local mass growth rate.
Both networks take as input a batch of particles $\{x_n^{s_k}\}_{n=0}^N \sim \rho^{s_k}$ and the current arc-length parameter $s_k$. 
For a given step size $\tau$, this generative process, representing the discrete unbalanced FP equation~\eqref{eq:CSB:FP}, is defined as follows,
\begin{subequations}
\label{eq:discrete_geodesic}
\vspace{-1mm}
\begin{align}
    x_n^{s_{k+1}} &= x_n^{s_{k}} + \tau \, \color{darkgreen} v_\theta(x_n^{s_{k}}, s_{k}), \\
    w_n^{s_{k+1}} &= w_n^{s_{k}} + \tau \, w_z \color{dodgerblue} z_\theta(x_n^{s_{k}}, s_{k})\color{black}\, w_n^{s_{k}}.
\end{align}
\end{subequations}
\textbf{Training procedure.~~} Training these networks requires discretizing the rest of the contact uOT~\eqref{eq:CSB:J}- \eqref{eq:CSB:z}-\eqref{eq:CSB:M}. The objective minimization~\eqref{eq:CSB:J}-\eqref{eq:CSB:z} is addressed by the \emph{optimality} loss term $\ell_\text{opt}$, which 
minimizes the total geodesic arc-length and penalizes deviations of the simulated trajectory~\eqref{eq:discrete_geodesic} from the optimality conditions~\eqref{eq:contact_hamiltonian_dynamics}. Adherence to the marginal constraints~\eqref{eq:CSB:M}, in terms of total mass and reconstruction accuracy, is then measured by the \emph{feasibility} loss $\ell_\text{fea}$.
Thus, our training objective is,\looseness-1
\begin{subequations}
\vspace{-1mm}
\begin{align}
\ell &= w_{\text{fea}} \, \ell_\text{fea} + w_{\text{opt}} \, \ell_\text{opt} ,  \quad \text{where}, \label{eq:loss}\\
\ell_\text{fea} &=\sum_{m=1}^{M} 
{
\left(d^{\mathcal{W}_2}(w^{s_m},x^{s_m},w^m,x^m)
+  w_{\text{mass}} \Vert W^{s_m}-W^m\Vert^2 \right)
}, \label{eq:loss-fea} \\
\ell_\text{opt} &= \sum_{k=0}^{K-1} \sum_{n=0}^{N-1} \bigg({
({\nu_n^{s_k}})^2w_n^{s_k}\,
\Vert x_n^{s_{k+1}}-x_n^{s_{k}}\Vert^2 
}+ {\Vert z_n^{s_{k+1}}- z_\theta (x^{s_k}_n, s_{k+1})\Vert^2}\bigg) . \label{eq:loss-opt}
\end{align}
\end{subequations}
Minimizing the geodesic arc-length requires computing the contact Wasserstein metric $g^{\mathcal{C}} = \sqrt{\mathcal{K}}\, g^{\mathcal{W}_2}$, obtained by scaling the Wasserstein metric by the kinetic energy (Sec. \ref{sec:methodology}). 
To this end, we integrate the optimality conditions~\eqref{eq:contact_hamiltonian_dynamics} forward, discretized along the estimated path (see App.~\ref{app:implementation} for details), \looseness-1
\begin{subequations}
\label{eq:discrete_optimality}
\vspace{-1mm}
\begin{align}
\nu_n^{s_{k+1}} &= \nu^{s_k}_n +\frac{1}{1 + \nu^{s_k}_n} \left( \frac\tau2 w_z z_\theta(x^{s_k}_n, s_k) \,\nu_n^{s_k}\,
+ w_z \left((z_\theta(x^{s_{k+1}}_n, s_k)-z_\theta(x^{s_{k}}_n, s_k) \right) S_n^{s_k} \right),
\label{eq:discrete_nu}\\
S_n^{s_{k+1}} &= S_n^{s_k}
- \frac{\tau}{2}(\nu_n^{s_k})^2
+ \frac\tau2 w_z \, z_\theta(x_n^{s_k}, s_k) \, S_n^{s_k},
\label{eq:discrete_S}\\
z_n^{s_{k+1}} &= z_n^{s_k}
+ \frac{\tau}{2} {(\nu_n^{s_k})}^2w^{s_k}_n
+ \frac\tau2 w_z \, {z_\theta(x_n^{s_k}, s_k)}^2 - \frac{\tau}{2} {(\nu_n^{s_0})}^2 w^{s_0}_n.
\label{eq:discrete_z}
\end{align}
\end{subequations}
Here, $\nu_n^{s_k}$ denotes the scalar speed of the particle $x_n^{s_k}$, evolving jointly with the dual potential $S_n^{s_k}$ under the learned growth rate $z_\theta$.
The per-sample kinetic energy $(\nu_n^{s_k})^2 w_n^{s_k}$ acts as a conformal scaling factor for particle displacements, weighting the distance $\|x_n^{s_{k+1}} - x_n^{s_k}\|^2$ in the optimality loss~\eqref{eq:loss-opt}, to provide a discrete realization of the contact metric $g^\mathcal{C}$. 
High metric values, occurring in regions of positive growth $z_\theta$, penalize large displacements, forcing $v_\theta$ to predict smaller steps.
As empirically shown in Sec. \ref{sec:results}, this mechanism effectively refines the sampling resolution in high-density regions (see App.~\ref{app:geodesic-parameterization}).
In the optimality loss~\eqref{eq:loss-opt}, we also regularize the inter-marginal evolution of the learned growth rate $z_\theta$ via the optimal action dynamics~\eqref{eq:discrete_z}.
Reconstruction accuracy is enforced by requiring that the learned weighted ensemble $ \rho^{s_k} = \{(x_n^{s_k}, w_n^{s_k})\}_{n=1}^{N}$ remains close to the reference marginal $\rho^m=\{(x_n^{m}, w_n^{m})\}_{n=1}^{N_m}$ at each observation step $s_k = m$. 
We achieve this by defining the feasibility loss~\eqref{eq:loss-fea} as a function of the total mass $W^{s_k} = \int_\mathcal{M} \rho^{s_k}(x) dx \approx  \sum_{n=0}^{N} w_n^{s_k}$, and the distributional accuracy quantified by the weighted Wasserstein distance $d^{\mathcal{W}_2}$. 
The weighted distance is necessary as the number $N_m$ of available samples from the marginal $\rho^m$ may differ from the batch size $N$ used for the sample discretization of the geodesic.
The training phase is summarized in Alg.~\ref{alg:training}, while implementation details and a computational complexity analysis are given in App.~\ref{app:implementation}.\looseness-1
\vspace{-2mm}
\section{Contact Geometry Generalizes Optimal Transport Formulations}\label{sec:comparison}
\vspace{-2mm}
\textbf{Symplectic geometry of OT.~~}
Since symplectic dynamics is a special case of contact dynamics in which the action $z^t$ does not influence the evolution of $(\rho^t, S^t)$, the entropy-regularized OT problem is recovered as a limiting case of the contact formulation~\eqref{eq:contact_schrodinger_bridge} when $w_z \to 0$, as, \looseness-1
\begin{subequations}
\label{eq:schrodinger_bridge}
\vspace{-1mm}
\begin{align}
    \min_{v^t} \quad & J(v^t, \rho^t)
    = \int_{t_0}^{t_M} \!\! \int_{\mathcal{M}} \frac{1}{2} \|v^t(x)\|^2 \, \rho^t(x) \, dx \, dt, \label{eq:SB:J} \\
    \text{s.t.} \quad 
    & \partial_t \rho^t(x) = - \nabla_x \cdot \big(\rho^t(x)\, v^t(x)\big) + \varepsilon \Delta_x \rho^t(x), \label{eq:SB:FP} \\
    & \rho^{t_m} = \rho^m, \quad \forall m \in \{0, \dots, M\}. \label{eq:SB:M}
\end{align}
\end{subequations}
The Hamiltonian of the optimization problem~\eqref{eq:schrodinger_bridge} is $\mathcal{H}(\rho^t, S^t) = \mathcal{K}(\rho^t, S^t) + \varepsilon \mathcal{F}(S^t, \rho^t)$, and consists solely of the kinetic energy and an entropy regularization term~\citep{chen2014relationoptimaltransportschrodinger} (see App.~\ref{app:sb_demons}). 
The optimality conditions define a Hamiltonian system evolving on the symplectic manifold $(\mathcal{T}^*\mathcal{P}^+(\mathcal{M}), \omega)$ \citep{cui2024stochastic, chow2020wasserstein, neklyudov2023computational}. 
As a consequence, the flow preserves volume within $\mathcal{T}^*\mathcal{P}^+(\mathcal{M})$, and the density $\rho^t \in \mathcal{P}^+(\mathcal{M})$ conserves its total mass $\int_{\mathcal{M}} \rho^t(x)\, dx$ throughout the evolution, reflecting the balanced nature of classical OT.\looseness-1

\textbf{Wasserstein-Fisher-Rao geometry of uOT.~~}
To connect distributions with different mass, several related works (Sec.~\ref{sec:related_works}) propose WFR-based extensions of the OT problem~\eqref{eq:schrodinger_bridge}, such as,\looseness-1
\begin{subequations}
\label{eq:fisher-rao-schrodinger_bridge}
\vspace{-1mm}
\begin{align}
    \min_{v^t, z^t} \quad & J(v^t, z^t, \rho^t)
    = \int_{t_0}^{t_M} \!\! \int_{\mathcal{M}} \frac{1}{2} \big( \|v^t(x)\|^2 + w_z \color{darkyellow}\psi({z^t(x)}\color{black}) \big) \, \rho^t(x) \, dx \, dt, \label{eq:FRSB:J} \\
    \text{s.t.} \quad 
    & \partial_t \rho^t(x) = - \nabla_x \cdot \big(\rho^t(x)\, v^t(x)\big) + \varepsilon \Delta_x \rho^t(x) + \color{darkyellow}z^t(x)\rho^t(x), \label{eq:FRSB:FP} \\
    & \rho^{t_m} = \rho^m, \quad \forall m \in \{0, \dots, M\}.
\end{align}
\end{subequations}
The dynamics constraint~\eqref{eq:FRSB:FP} extends the FP equation~\eqref{eq:SB:FP} by introducing a \color{darkyellow}local mass growth rate $z^t$\color{black}~\citep{lu2019accelerating}, yielding a formulation equivalent to~\eqref{eq:CSB:FP}.
The growth rate \color{darkyellow}$z^t$ \color{black} enters the cost functional~\eqref{eq:FRSB:J} through a penalty term $\color{darkyellow}\psi(z^t)$, but unlike the contact Hamiltonian case, it does not correspond to the minimized action. 
The optimization problem~\eqref{eq:fisher-rao-schrodinger_bridge} therefore retains the standard variational form and still yields a classical Hamiltonian structure.\looseness-1

When setting $\color{darkyellow}\psi(z^t) = z^t(x)^2$, the Hamiltonian takes form $\mathcal{H}({\rho}^t,{S}^t)\!=\!\mathcal{K}(\rho^t, S^t)\!+\!\varepsilon \mathcal{F}(S^t, \rho^t)$, where the kinetic energy is given by $\mathcal{K}(\rho^t, S^t)\!=\!\tfrac{1}{2}\!\int_\mathcal{M}\!\left( \|\nabla_x{S}^t(x)\|^2 \!+ \!\tfrac{1}{w_z}{S}^t(x) ^2\right)\!{\rho}^t(x)dx$, and defines the WFR metric, which reinterprets the symplectic dynamics on $\mathcal{T}^*\mathcal{P}^+(\mathcal{M})$ as a tractable geodesic $\rho^s \subset \mathcal{P}^+(\mathcal{M})$~\citep{kondratyev2016new, neklyudov2023computational}. This yields a conic space that converges to the Wasserstein manifold as $w_z \! \to\! \infty$, where mass variation is suppressed and the WFR-based uOT reduces to the classic OT~\eqref{eq:schrodinger_bridge}.
The optimality condition $v^t(x)=\nabla_x z^t(x)$ (see App.~\ref{app:FR-sb_demons} for details) biases the transport dynamics toward regions of maximal mass growth. This induces a structural bias that can lead to mode collapse by discouraging the reconstruction of low-mass target modes~\citep{chapel2021unbalanced, zhang2025learning}. 
The same mechanism also introduces instability, as the transport velocity becomes sensitive to gradients of the growth field, which are often noisy in real datasets. This limitation is inherent to the volume-preserving symplectic dynamics that can only accommodate mass variation only through a rigid coupling between $z^t$ and $v^t$. Notably, \ac{CUOT} \eqref{eq:contact_schrodinger_bridge} overcomes these structural constraints by leveraging contact geometry to decouple transport from mass growth, providing a principled framework for more accurate and controlled adaptation to mass variations.
\looseness-1

\vspace{-3mm}
\section{Results}\label{sec:results}
\vspace{-3mm}
We benchmark the proposed CuOT framework against established OT and uOT baselines summarized in Table \ref{tab:comparison}. Additionally, to isolate the individual sources of performance gains, we conduct an ablation study comparing our full model against variants that systematically omit key components. \looseness-1

\begin{figure}[H]
\centering
\begin{subfigure}{0.24\textwidth}
    \centering
    \includegraphics[width=\linewidth]{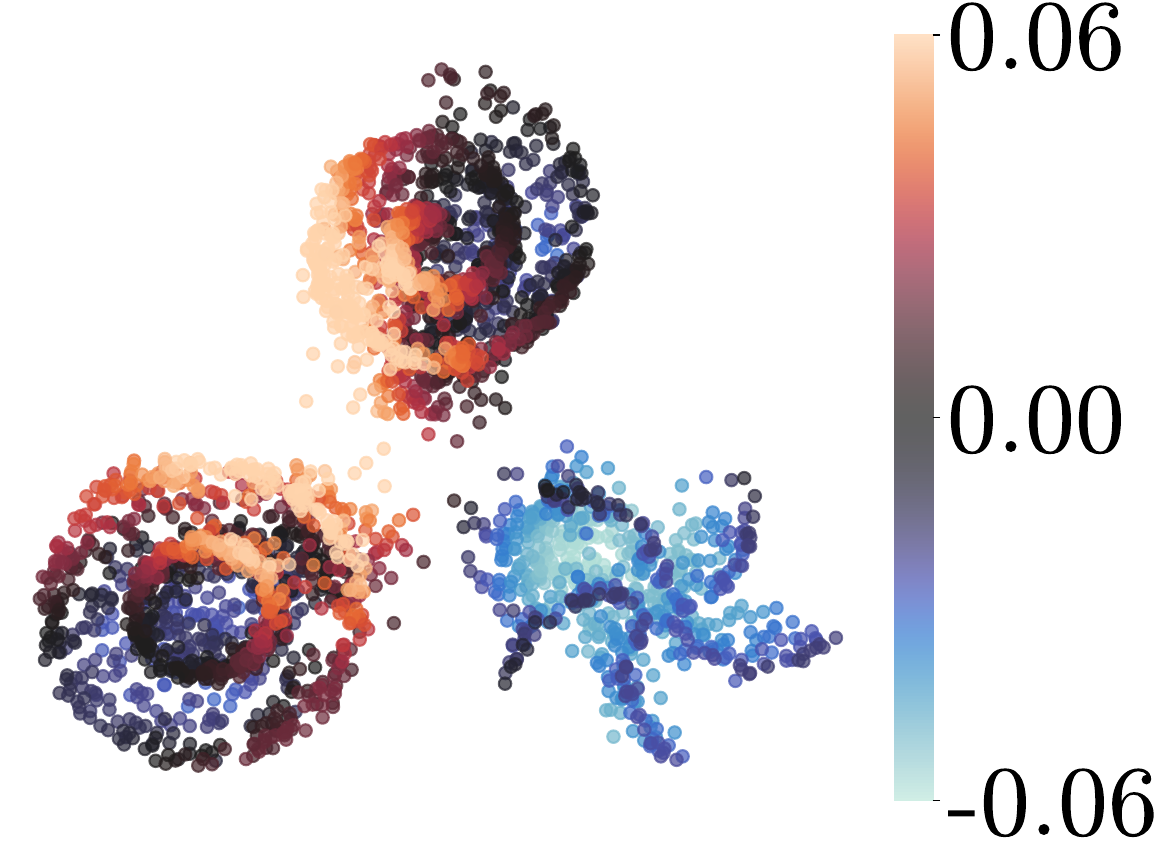}
    \caption{} 
    \label{fig:ablation_c}
\end{subfigure}
\hfill
\begin{subfigure}{0.24\textwidth}
    \centering
    \includegraphics[width=\linewidth]{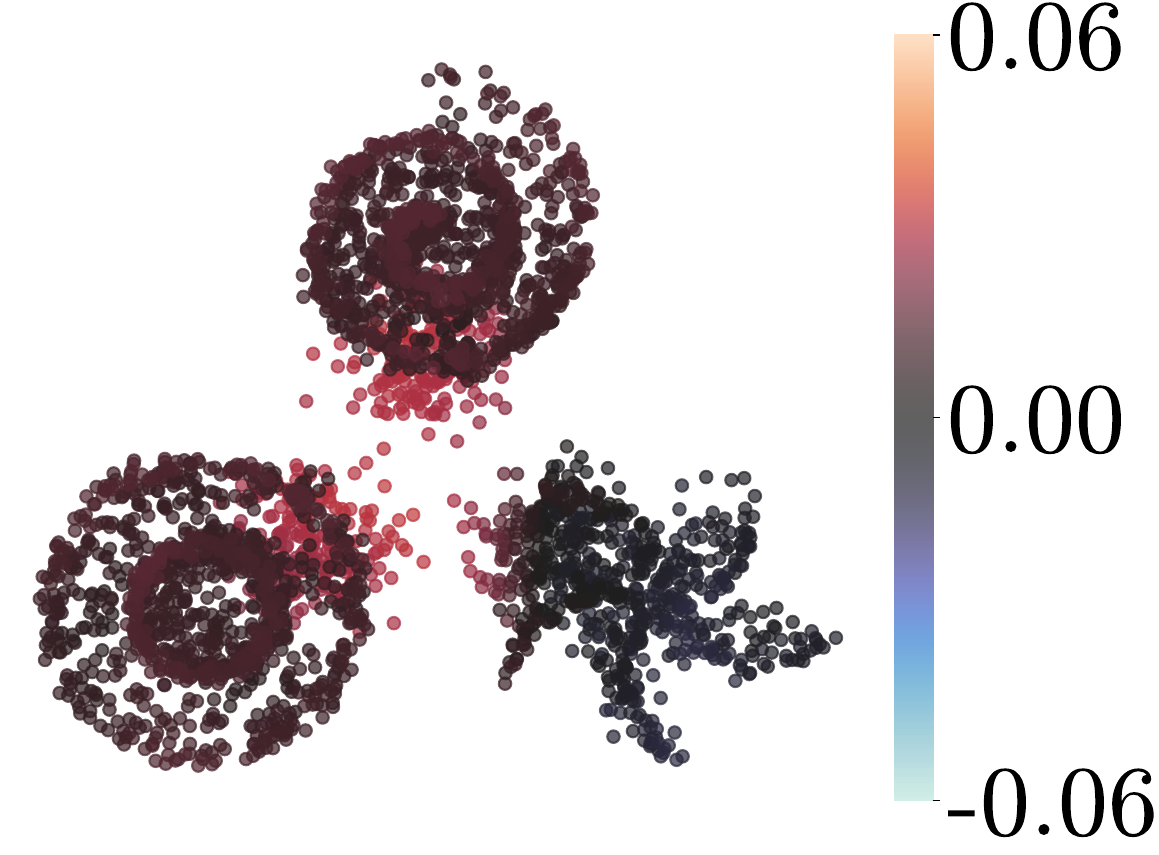}
    \caption{}
    \label{fig:ablation_fr}
\end{subfigure}
\hfill
\begin{subfigure}{0.24\textwidth}
    \centering
    \includegraphics[width=\linewidth]{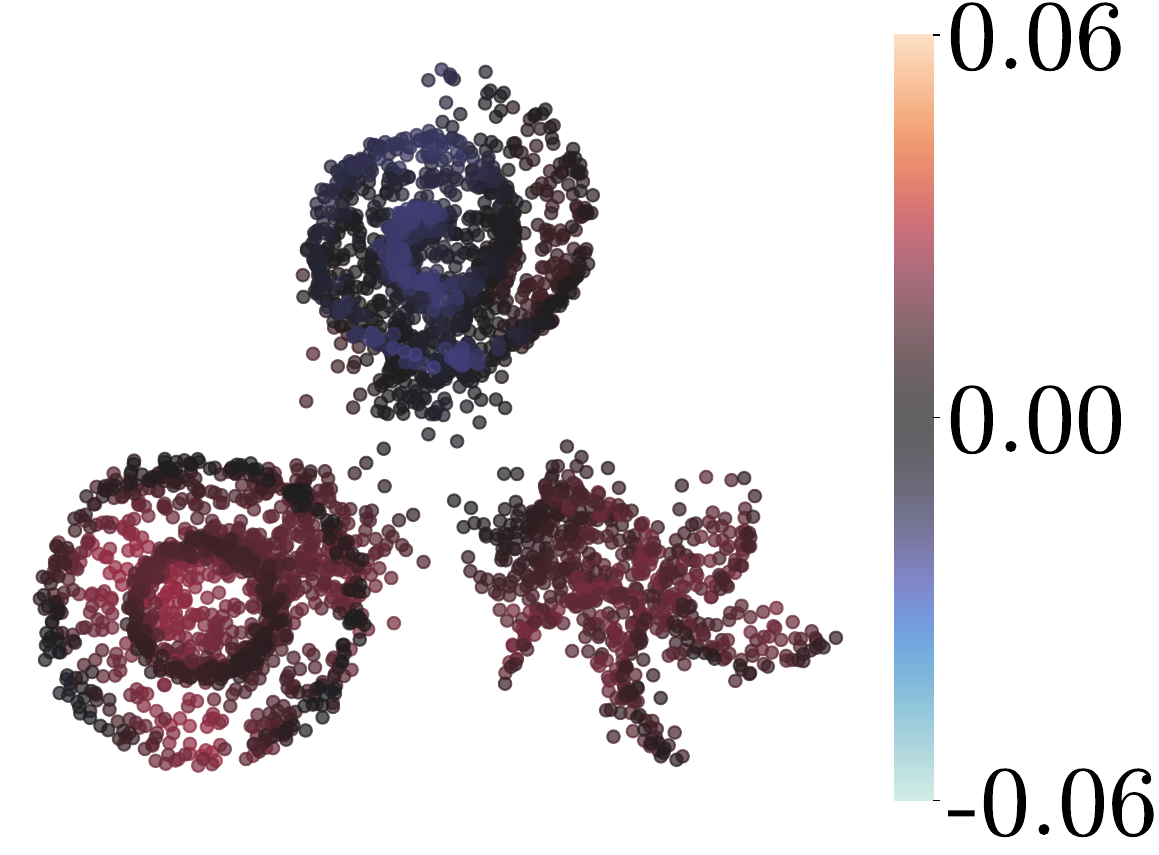}
    \caption{}
    \label{fig:ablation_ot}
\end{subfigure}
\hfill
\begin{subfigure}{0.24\textwidth}
    \centering
    \includegraphics[width=\linewidth]{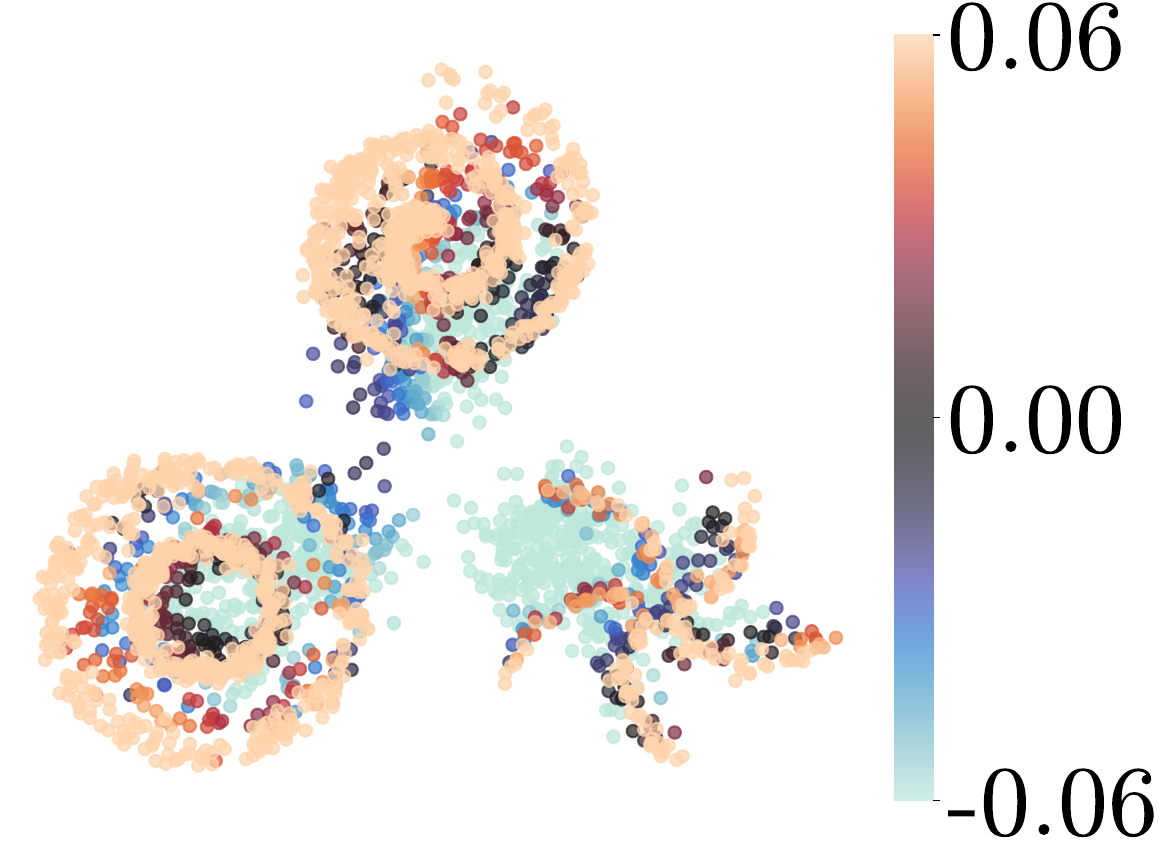}
    \caption{}
    \label{fig:ablation_w}
\end{subfigure}
\caption{Learned growth field $z_\theta(x^{s_k}, s_k)$ across variants in the ablation study.
\subref{fig:ablation_c}~{CuOT} (Complete), 
\subref{fig:ablation_fr}~WFR Geometry, 
\subref{fig:ablation_ot}~Standard OT, and 
\subref{fig:ablation_w}~w/ Time-Only Growth.}
\label{fig:ablation}
\end{figure}

\begin{figure}[t!]
\vspace{-3mm}
\centering
\begin{subfigure}{0.24\textwidth}
    \centering
    \caption{\textbf{CuOT (ours)}}
    \includegraphics[width=\linewidth]{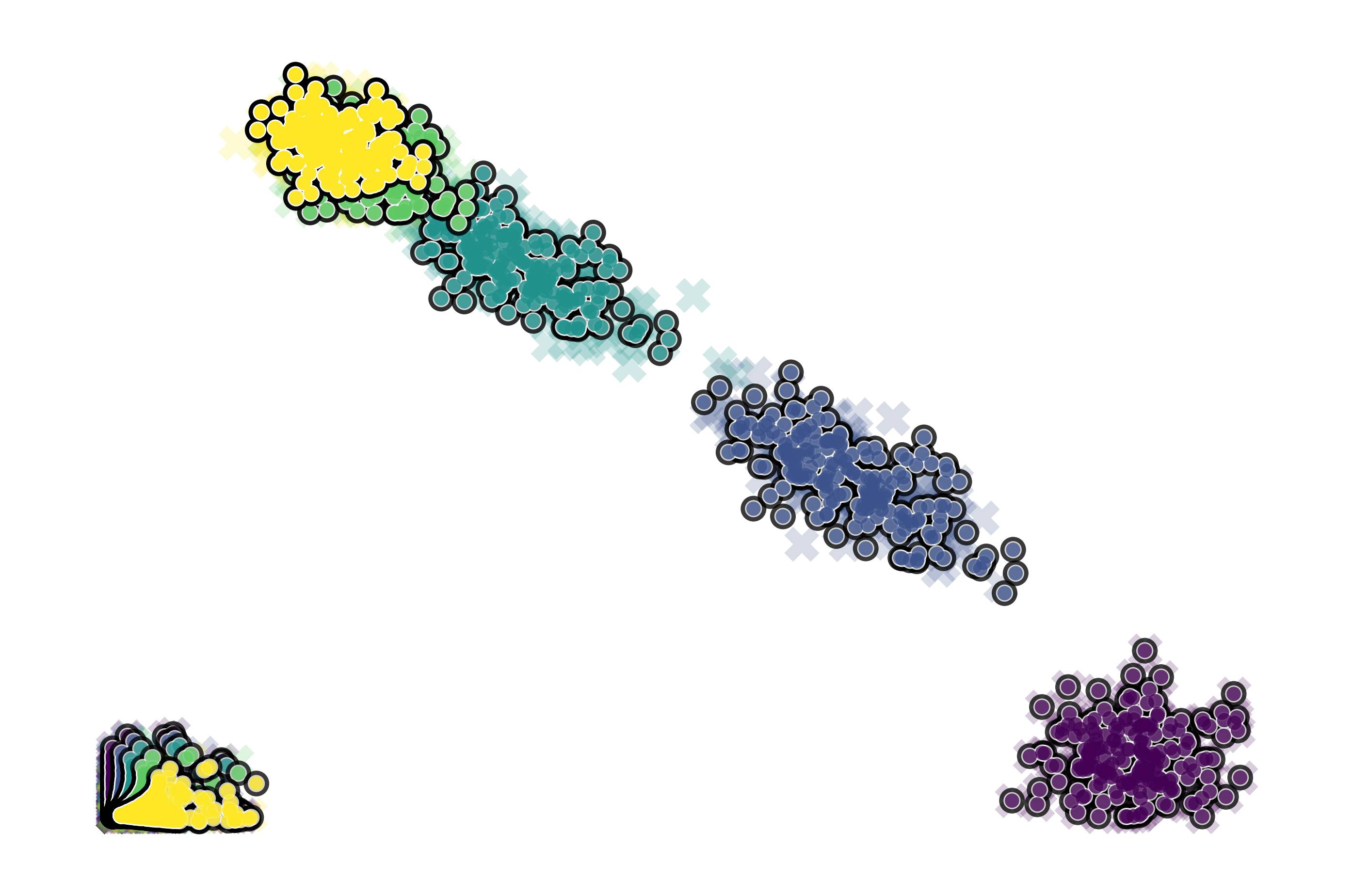}
\end{subfigure}
\hfill
\begin{subfigure}{0.24\textwidth}
    \centering
    \caption{VarRUOT}
    \includegraphics[width=\linewidth]{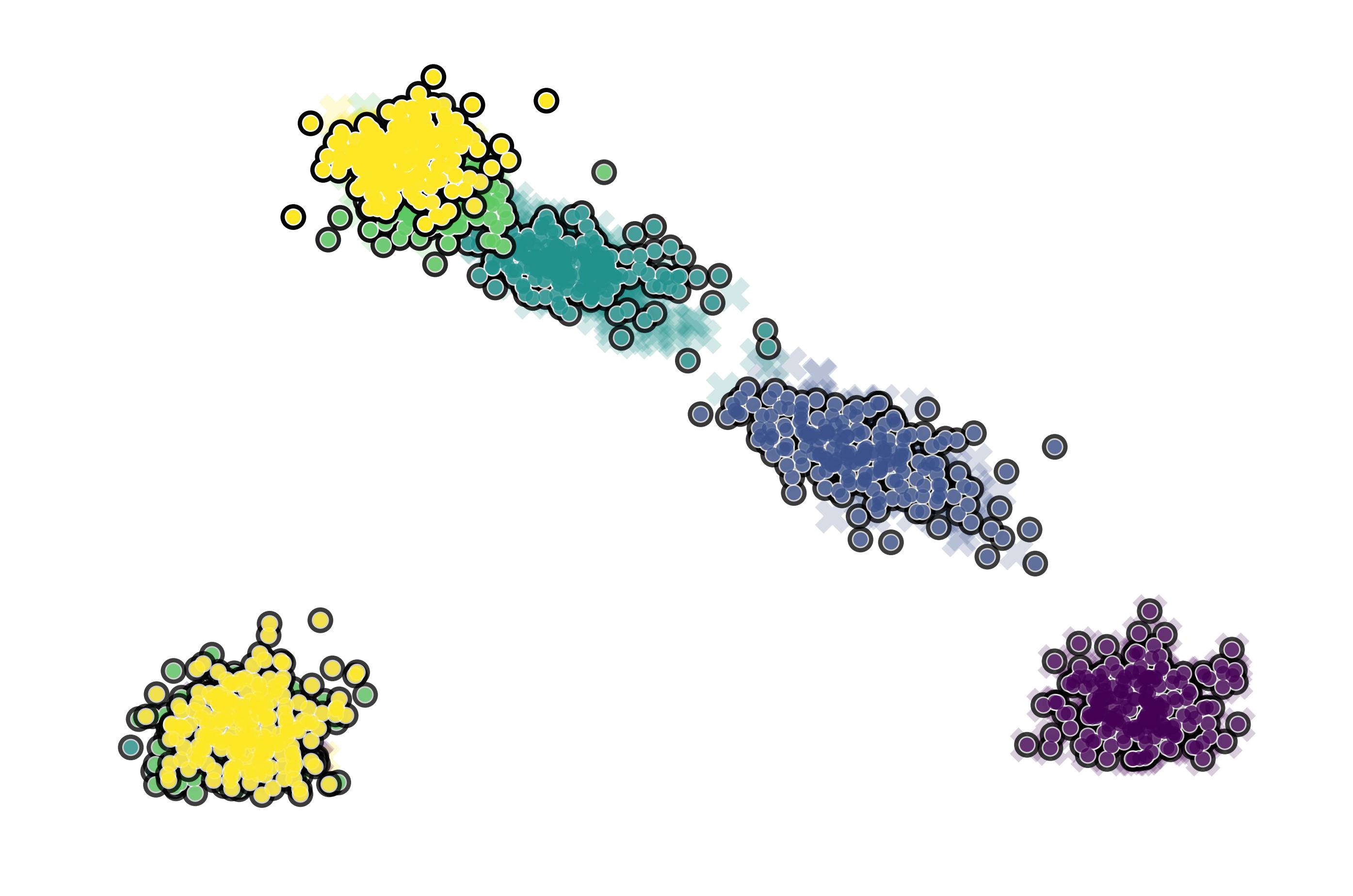}
\end{subfigure}
\hfill
\begin{subfigure}{0.24\textwidth}
    \centering
    \caption{DeepRUOT}
    \includegraphics[width=\linewidth]{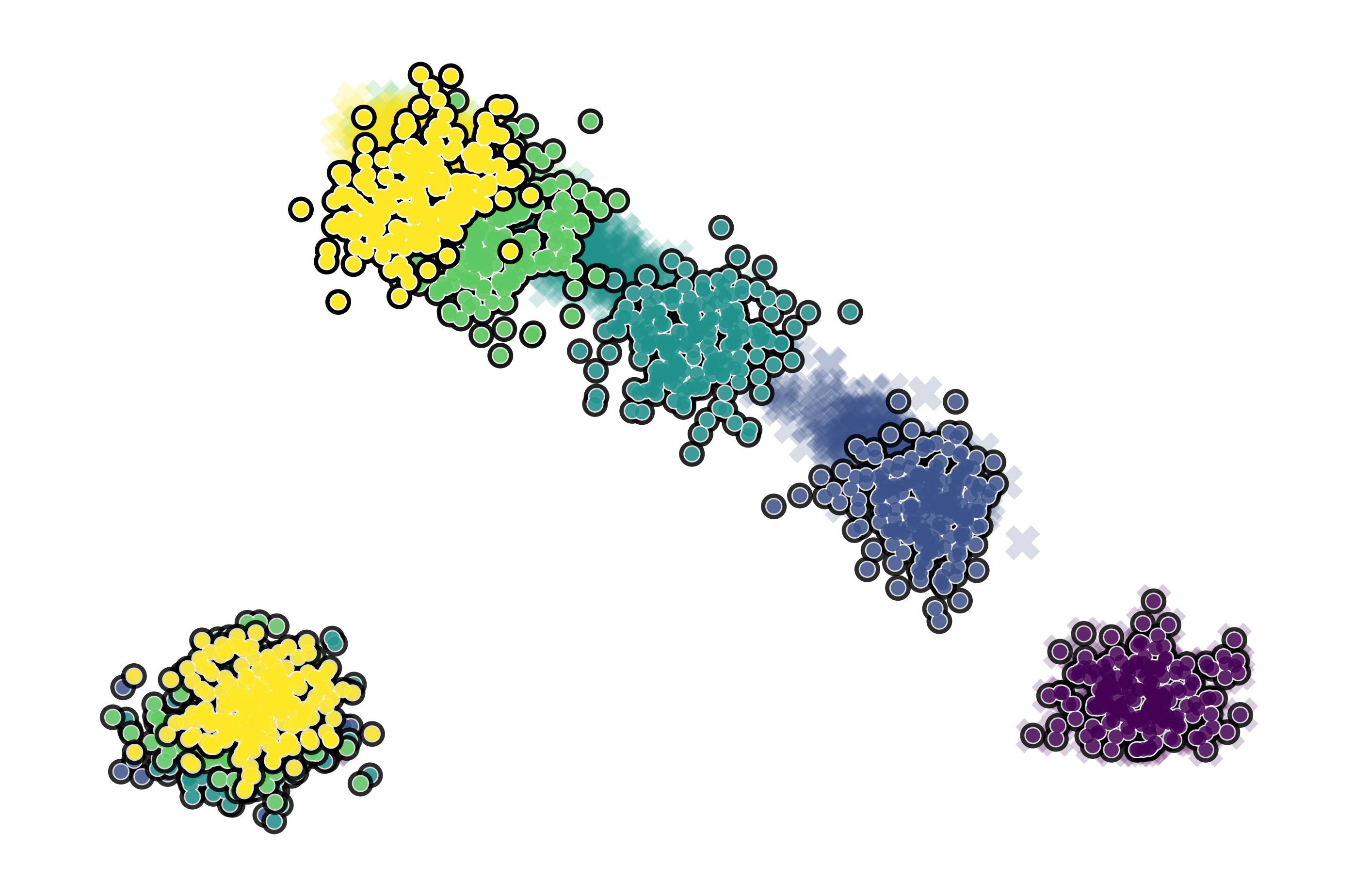}
\end{subfigure}
\hfill
\begin{subfigure}{0.24\textwidth}
    \centering
    \caption{OTCFM}
    \includegraphics[width=\linewidth, height=2.25cm]{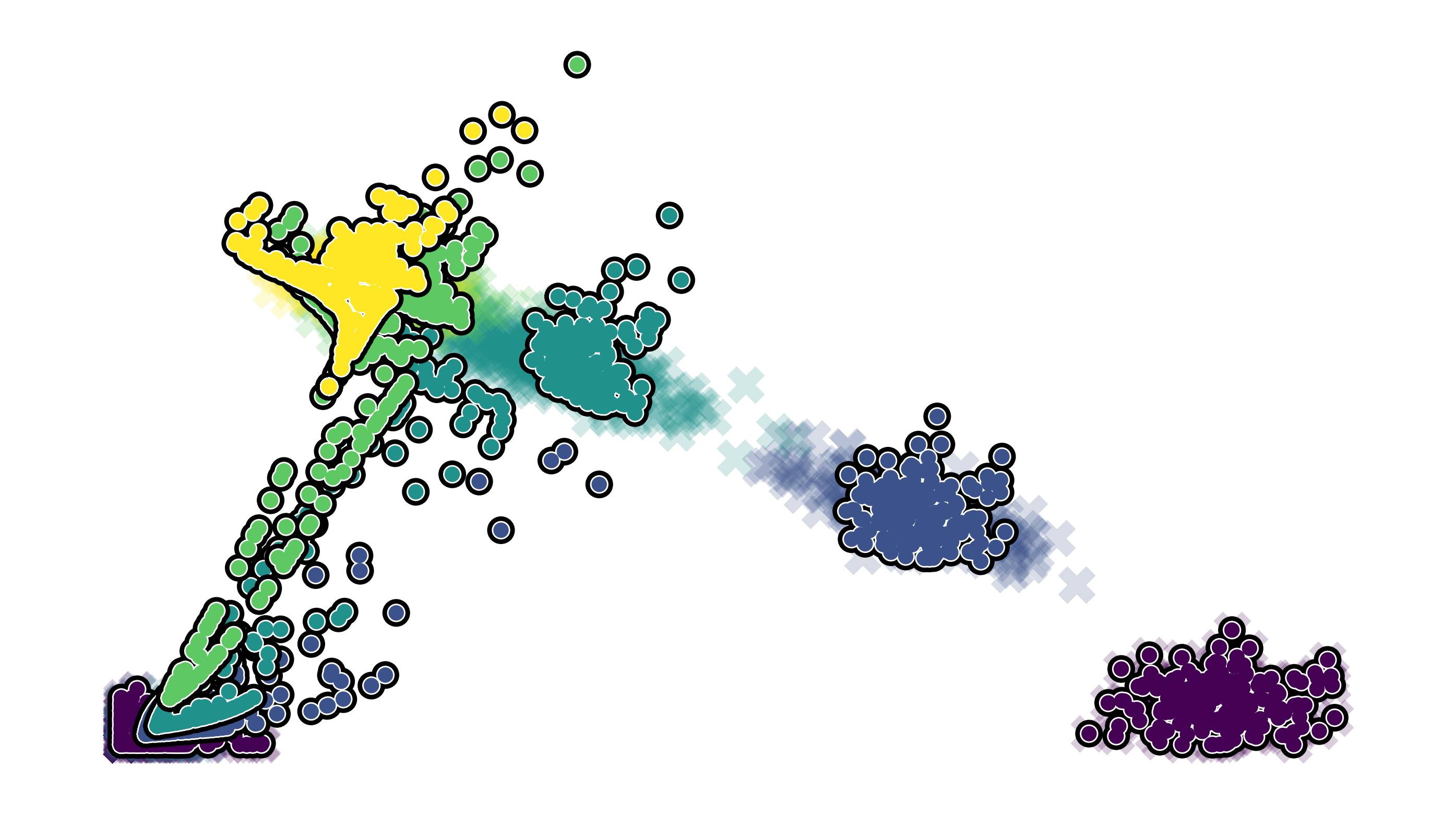}
\end{subfigure}

\vspace{0.0cm}

\begin{subfigure}{0.24\textwidth}
    \centering
    \includegraphics[width=\linewidth]{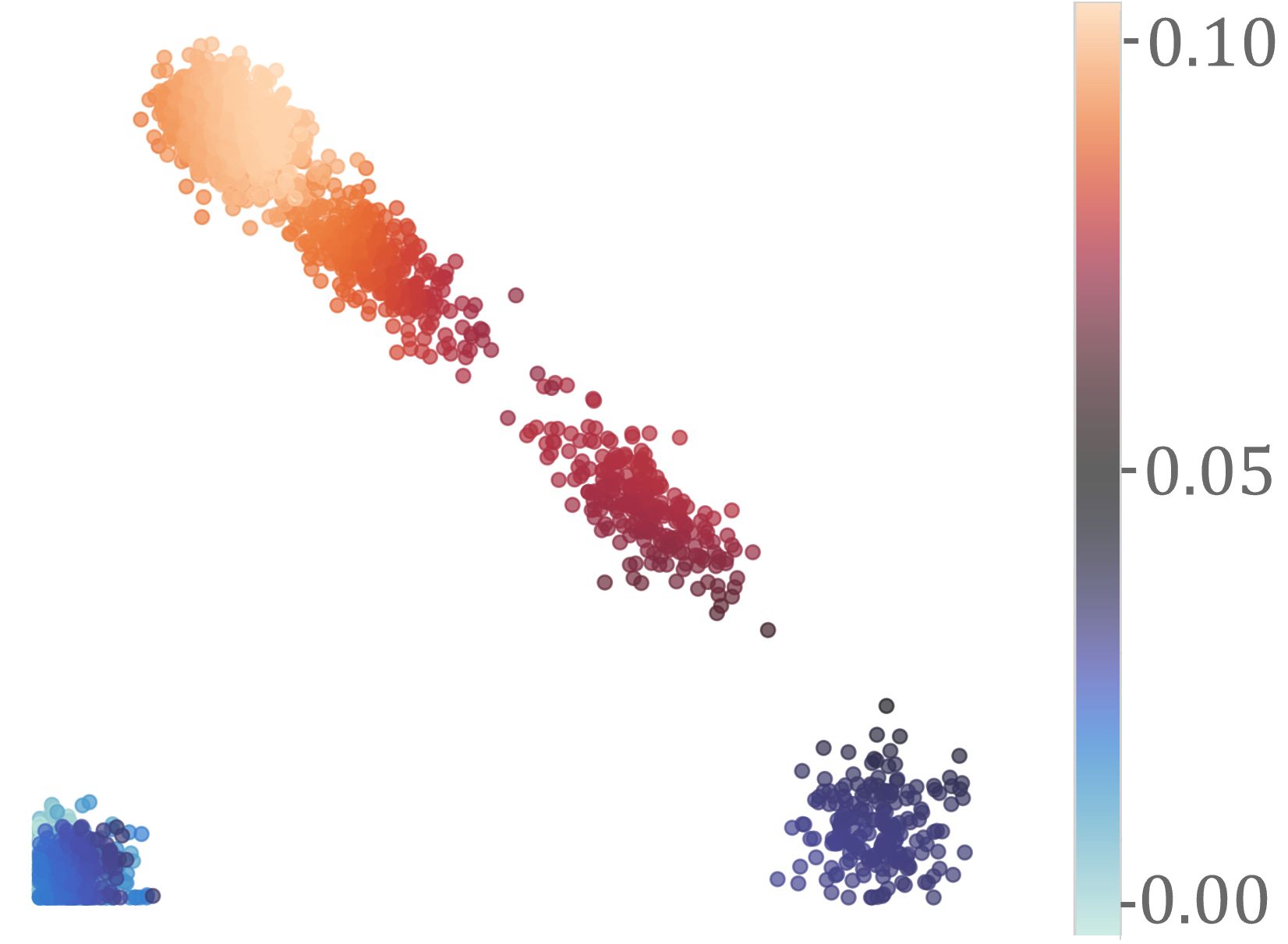}
\end{subfigure}
\hfill
\begin{subfigure}{0.24\textwidth}
    \centering
    \includegraphics[width=\linewidth]{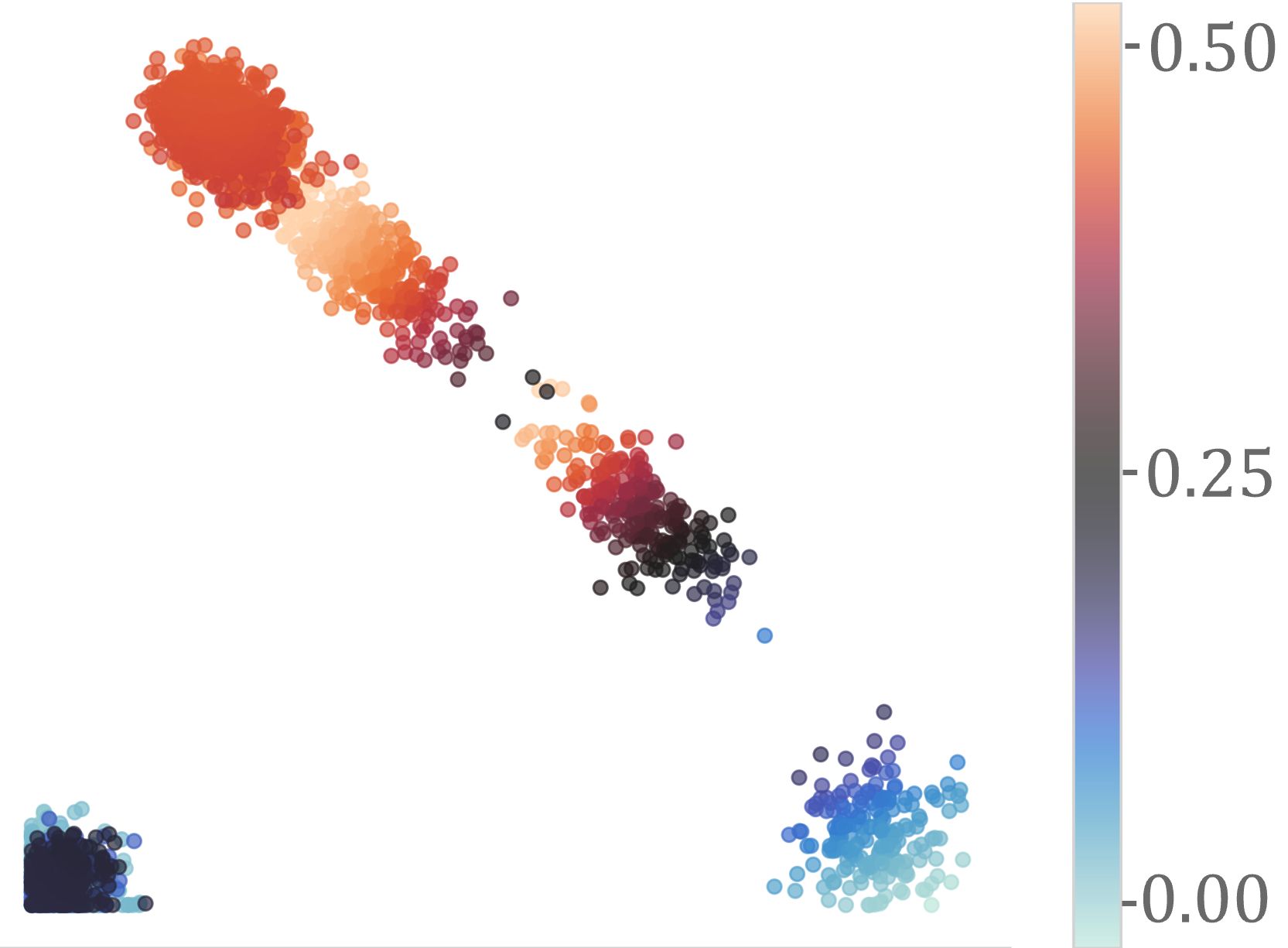}
\end{subfigure}
\hfill
\begin{subfigure}{0.24\textwidth}
    \centering
    \includegraphics[width=\linewidth]{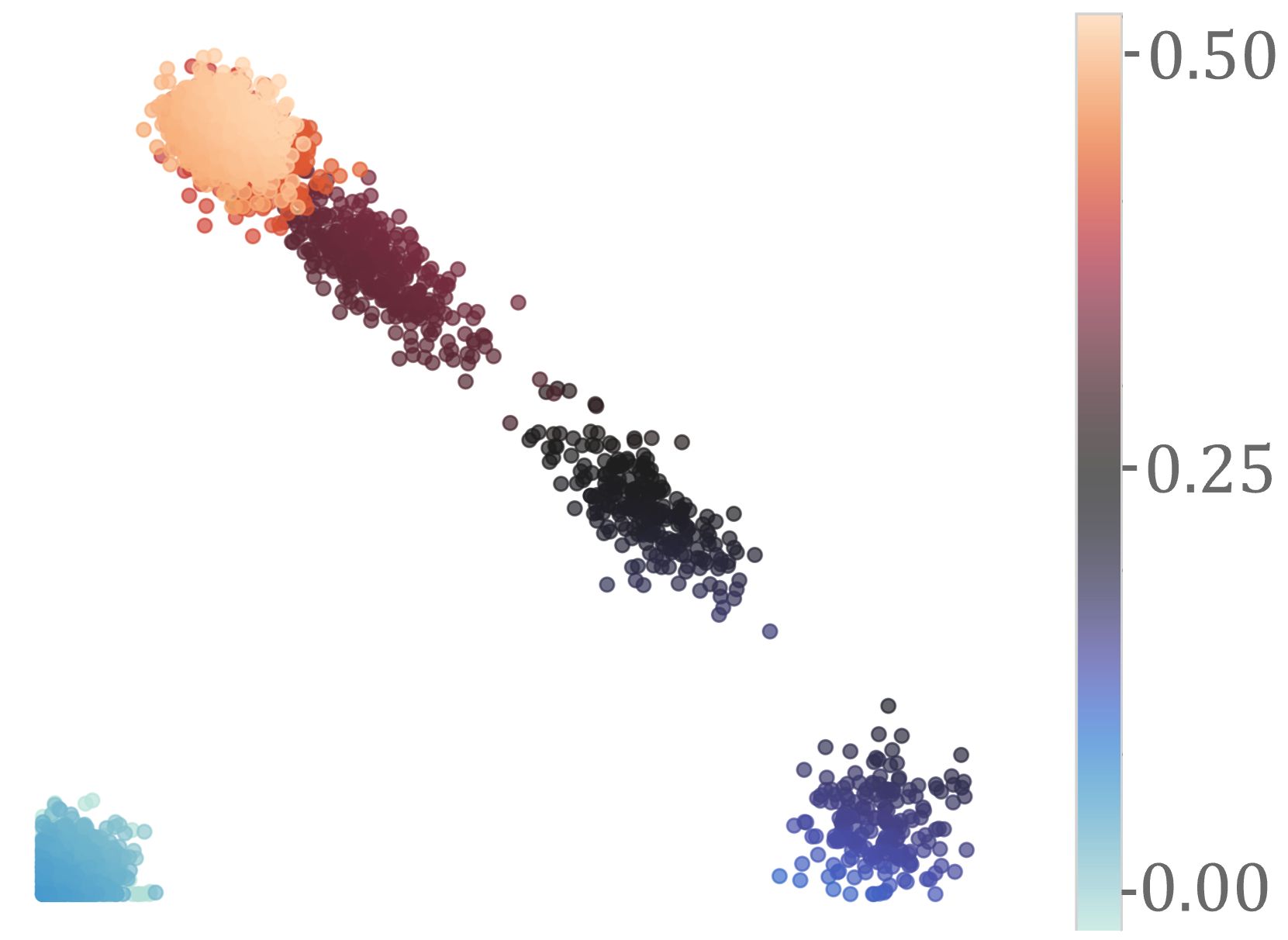}
\end{subfigure}
\hfill
\begin{subfigure}{0.24\textwidth}
    \centering
    \includegraphics[width=\linewidth]{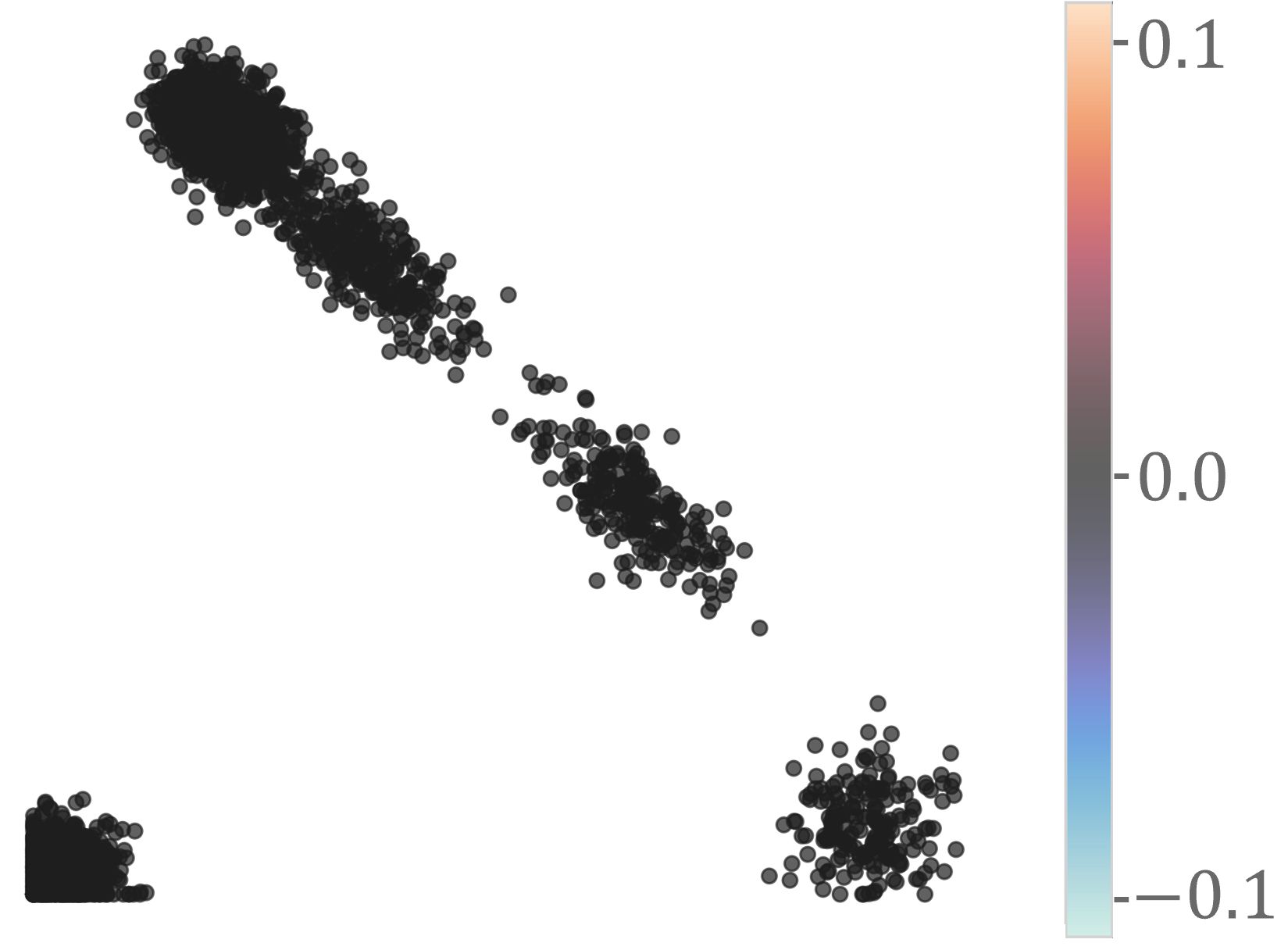}
\end{subfigure}
\caption{\small{\textbf{Top:} Dynamics 
 $\big(t_0$\tball{v5_1},\, $t_1$\tball{v5_2},\, $t_2$\tball{v5_3},\, $t_3$\tball{v5_4},\, $t_4$\tball{v5_5}$\big)$. \textbf{Bottom:} Growth rate (bottom) for \citet{zhang2025learning}.}\looseness-1}
\vspace{-3mm}
\label{fig:simulation}
\end{figure}

\begin{figure}[t!]
\centering

\begin{subfigure}{0.24\textwidth}
    \centering
    \caption{\textbf{CuOT (ours)}}
    \includegraphics[width=\linewidth]{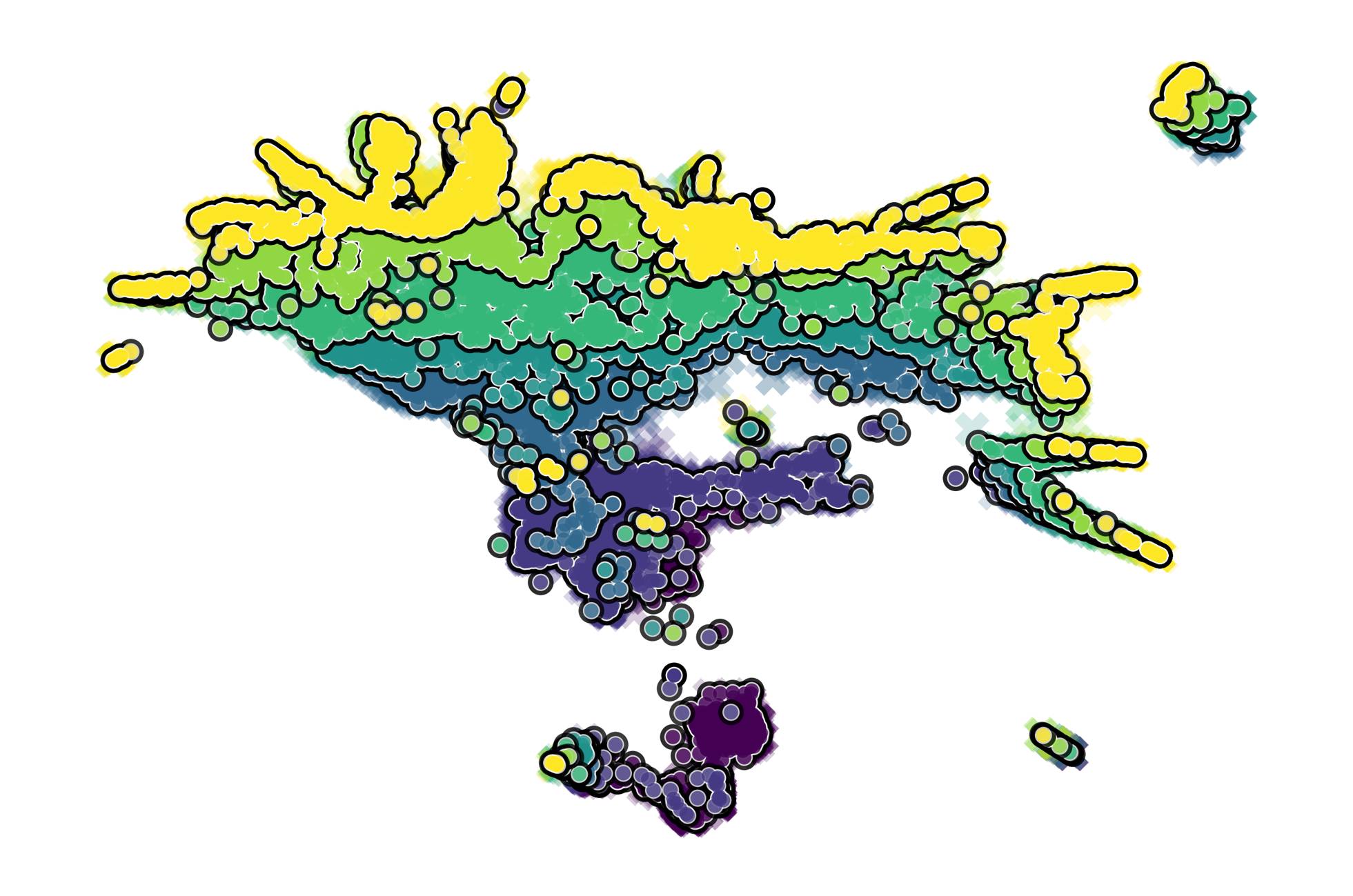}
\end{subfigure}
\hfill
\begin{subfigure}{0.24\textwidth}
    \centering
    \caption{VarRUOT}
    \includegraphics[width=\linewidth]{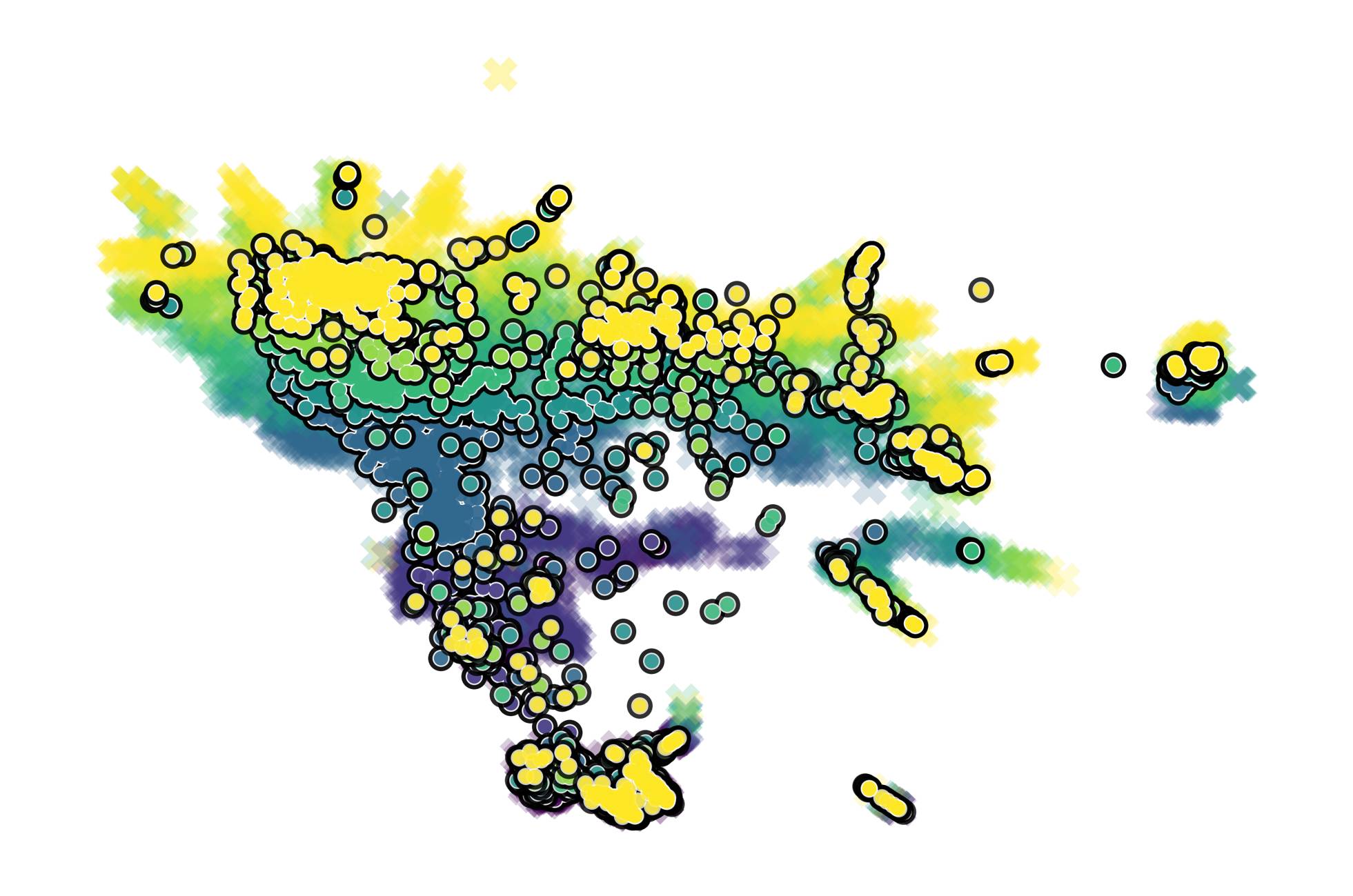}
\end{subfigure}
\hfill
\begin{subfigure}{0.24\textwidth}
    \centering
    \caption{DeepRUOT}
    \includegraphics[width=\linewidth]{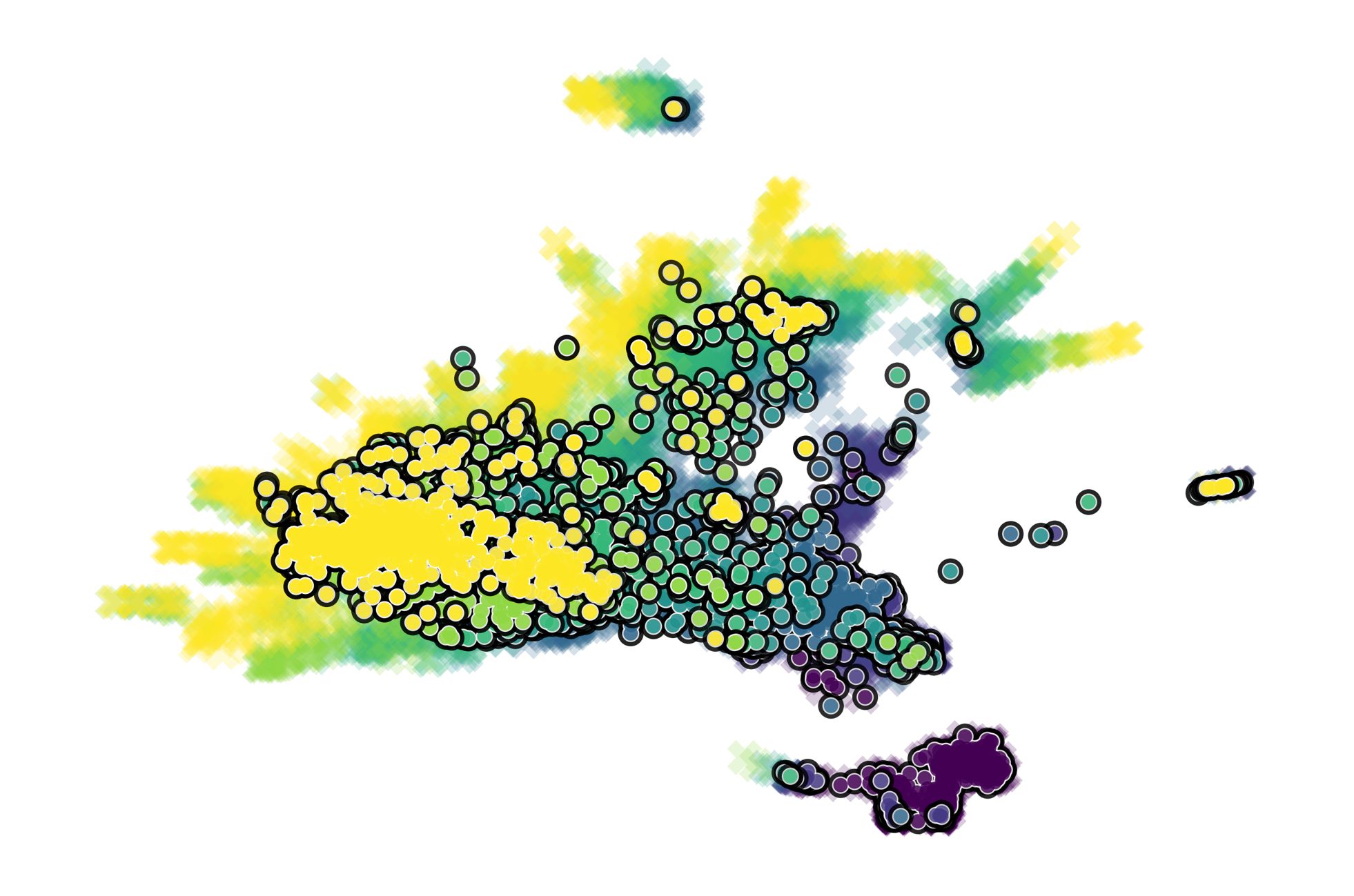}
\end{subfigure}
\hfill
\begin{subfigure}{0.24\textwidth}
    \centering
    \caption{OTCFM}
    \includegraphics[width=\linewidth, height=2.25cm]{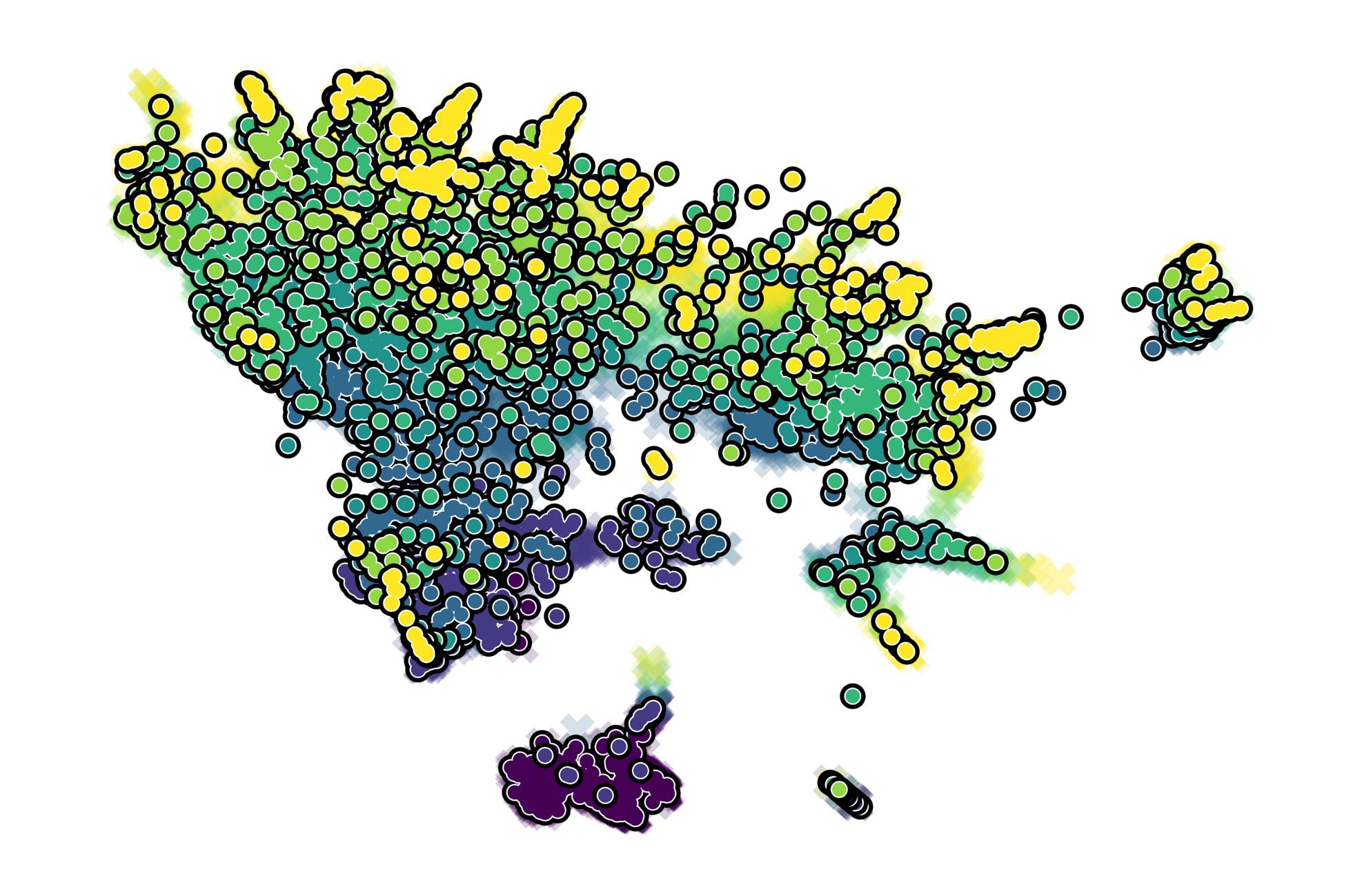}
\end{subfigure}

\vspace{0.0cm}

\begin{subfigure}{0.24\textwidth}
    \centering
    \includegraphics[width=\linewidth]{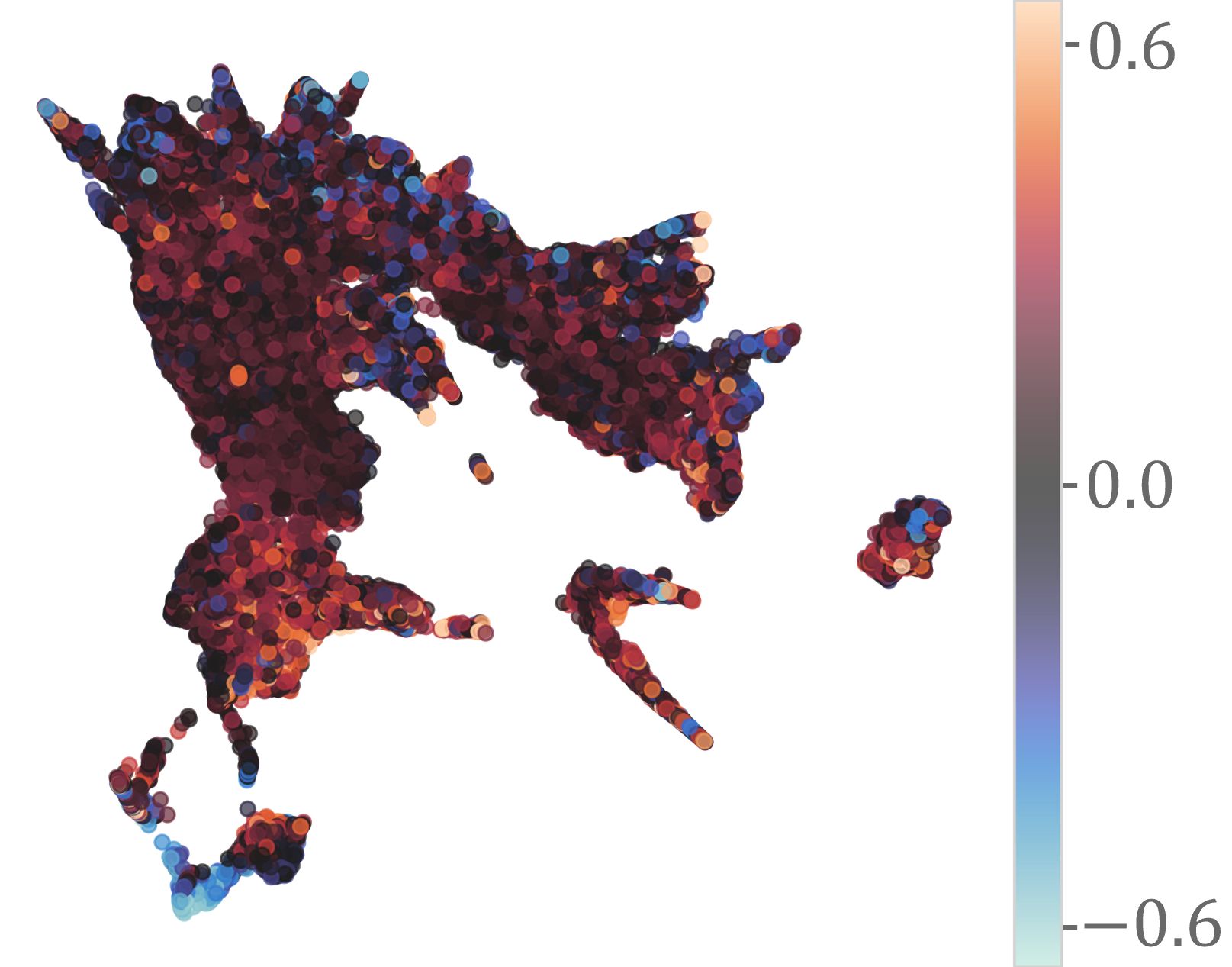}
\end{subfigure}
\hfill
\begin{subfigure}{0.24\textwidth}
    \centering
    \includegraphics[width=\linewidth]{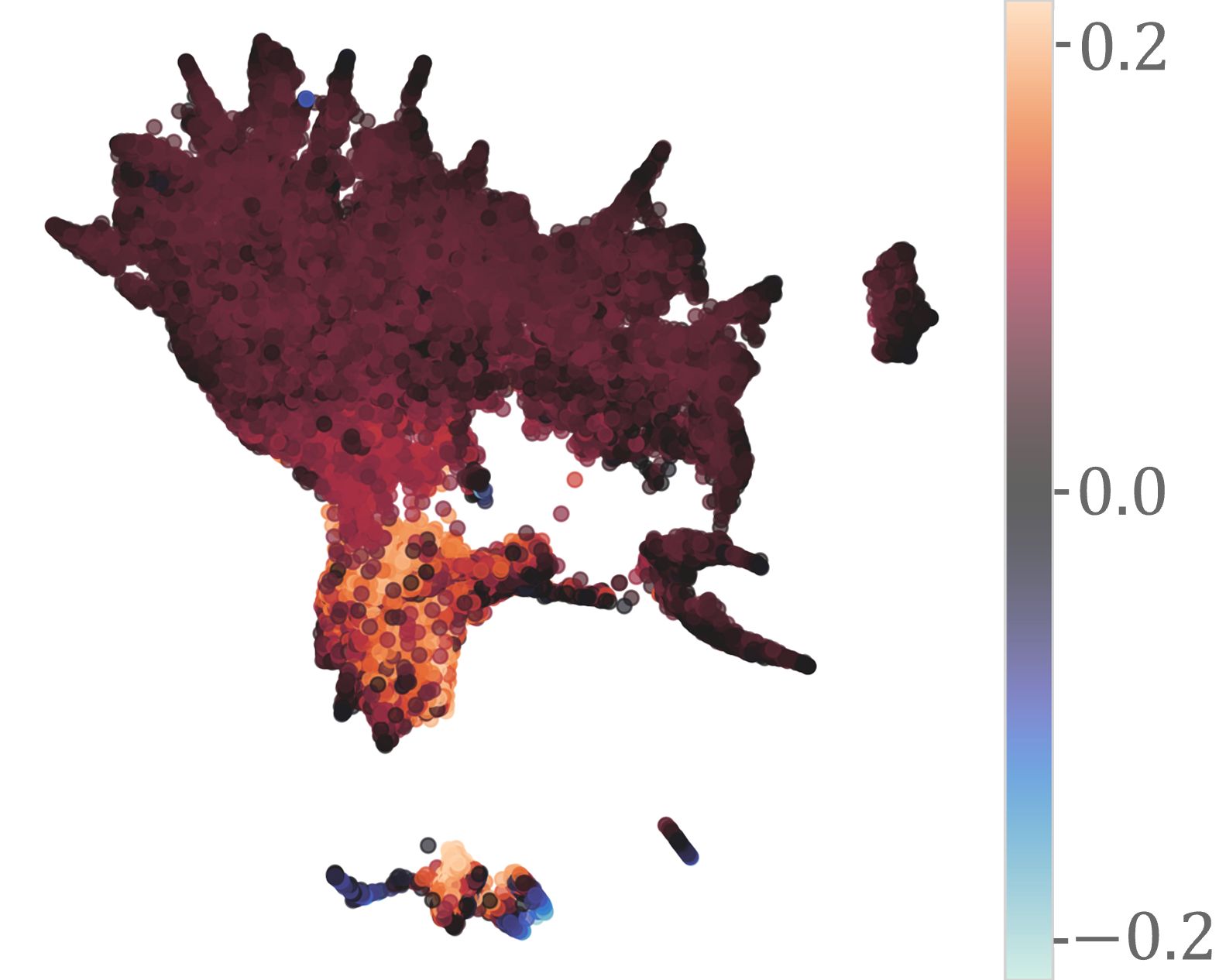}
\end{subfigure}
\hfill
\begin{subfigure}{0.24\textwidth}
    \centering
    \includegraphics[width=\linewidth]{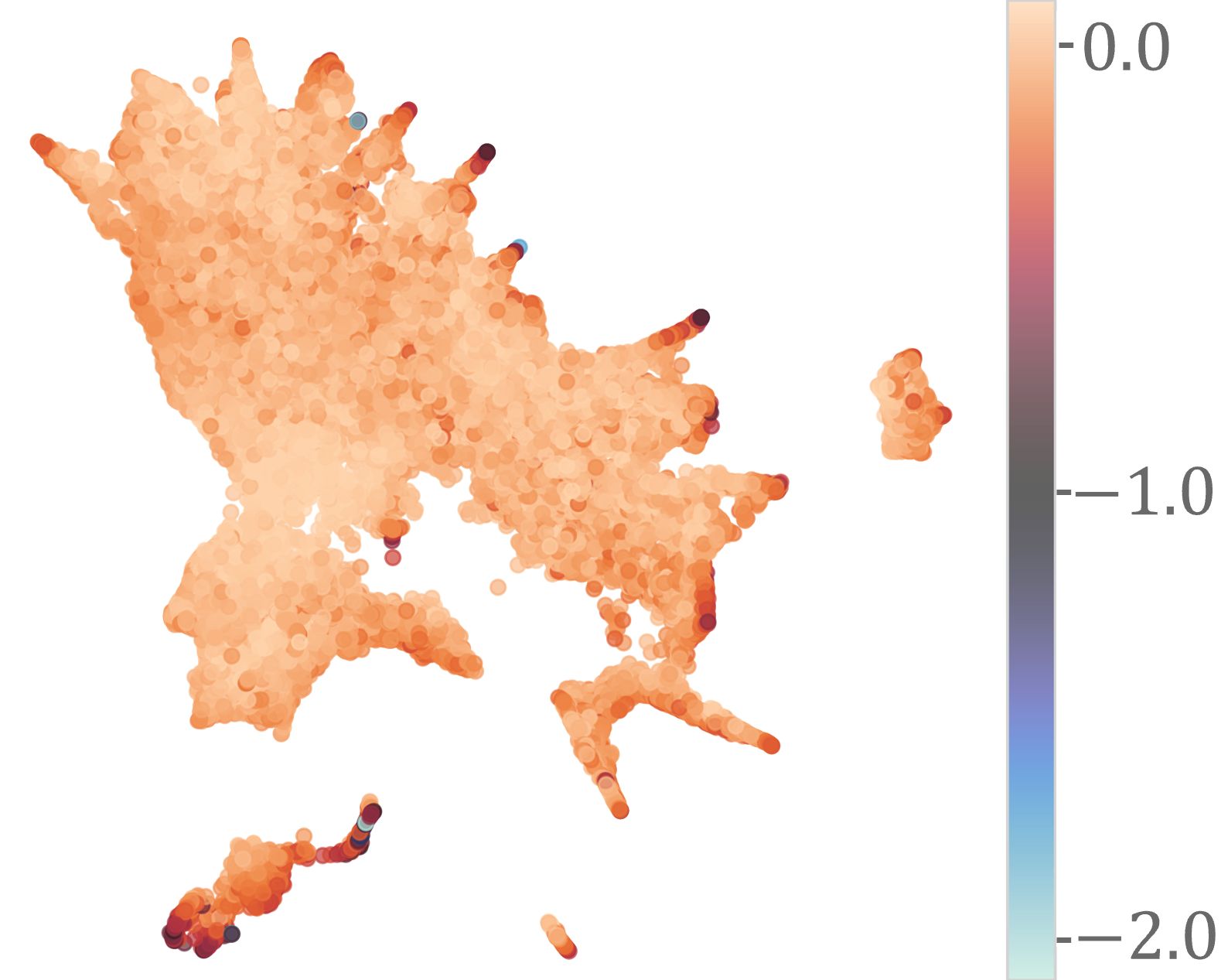}
\end{subfigure}
\hfill
\begin{subfigure}{0.24\textwidth}
    \centering
    \includegraphics[width=\linewidth]{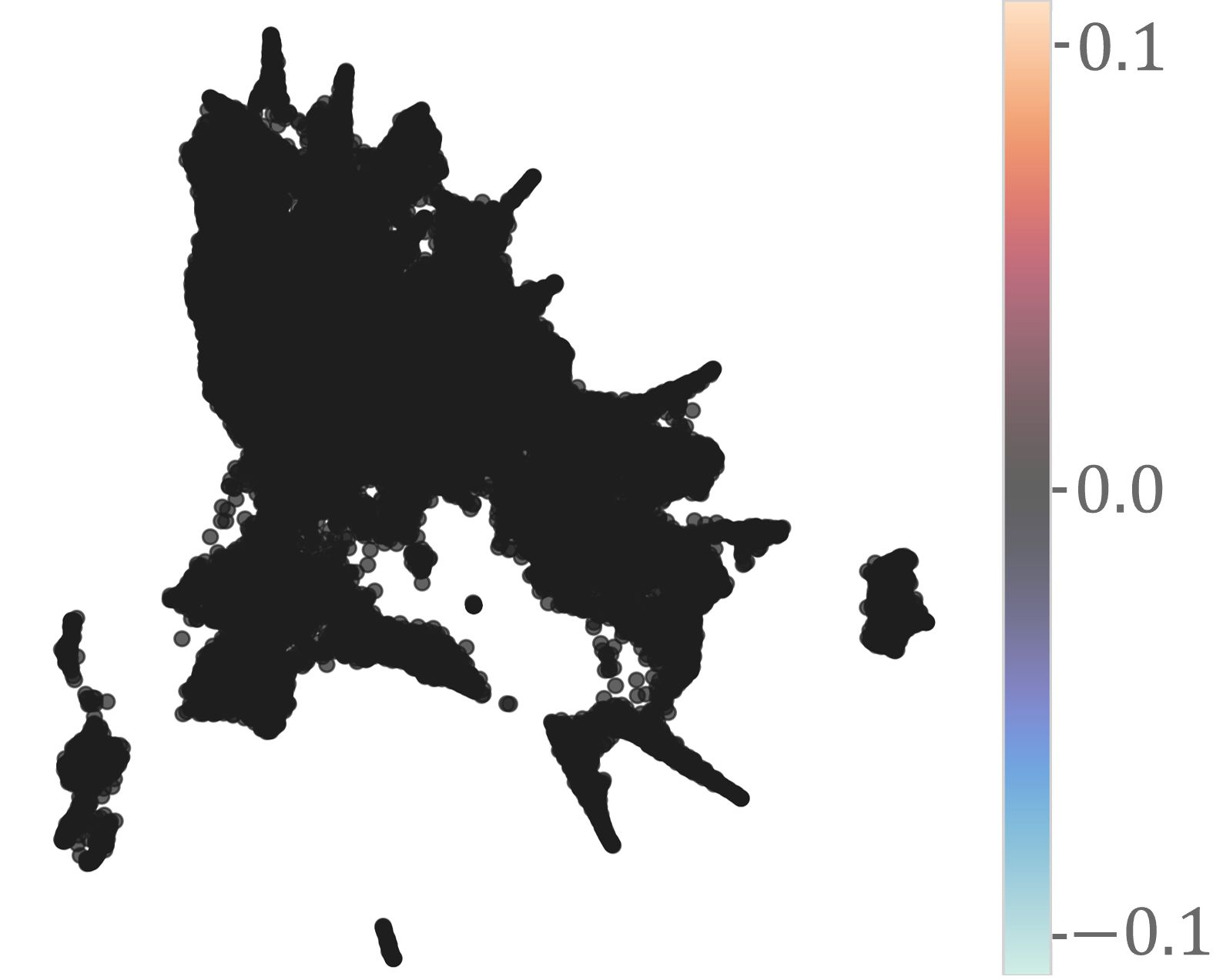}
\end{subfigure}
\caption{Qualitative analysis on the zebrafish embryogenesis dataset~\citep{farrell2018single}. \textbf{Top:} Predicted dynamics $\big(t_0$\tball{v7_1},\, $t_1$\tball{v7_2},\, $t_2$\tball{v7_3},\, $t_3$\tball{v7_4},\, $t_4$\tball{v7_5},\, $t_5$\tball{v7_6},\, $t_6$\tball{v7_7}$\big)$. \textbf{Bottom:} Growth rate.}
\label{fig:zebra}
\vspace{-5mm}
\end{figure}

\paragraph{Ablation study}
We first evaluate CuOT on a synthetic 2D distribution-bridging task with known mode correspondences and mass dynamics. The source distribution comprises three Gaussian modes ($400$ samples each), and the target distribution features three distinct shapes: a Swiss roll, concentric rings, and a pinwheel ($600$, $600$, and $200$ samples, respectively). Matching the ground truth requires mass expansion ($z_\theta > 0$) on the first two modes and decay on the third, posing a  challenge for OT frameworks. Furthermore, the deliberate misalignment between the transport direction and the growth gradient on the third mode highlights key failure modes in WFR. Table~\ref{tab:synthetic_ablation} summarizes ablation results and App.~\ref{app:ablations} details all variants.
The CuOT variant w/o Adaptive Step differs from the full model only in Equation~\eqref{eq:loss-opt}, where the contact metric $g^\mathcal{C}$ is replaced by the standard Wasserstein metric $g^{\mathcal{W}_2}$. Although this produces a virtually identical growth field (Figure~\ref{fig:ablation_c}), removing the adaptive step reduces overall transport accuracy ($\mathcal{W}_2$: $0.10 \to 0.13$; $\mathcal{W}_1$: $0.08 \to 0.10$). The variant with growth only dependent on time $z_\theta(s_k)$ performs the worst because it cannot capture spatially varying dynamics (Figure~\ref{fig:ablation_w}). WFR-based variants suffer from mode leakage as forward transport directions are incompatible with negative growth rates (Figure~\ref{fig:ablation_fr}). Vanishing growth regularization ($w_z \to 0$) eliminates leakage but results in a high Chamfer distance ($>0.3$) and mode collapse, as the WFR geometry enforces positive growth across all modes.
Finally, OT-based frameworks with constant sample weights cannot prevent misrouting, even when learning to scale global mass. This yields the highest mode leakage ($17.7\%$) by assigning positive growth to samples in the low-mass cluster (Figure~\ref{fig:ablation_ot}).
\looseness-1
\vspace{-3mm}
\begin{table}[H]
\centering
\caption{Quantitative ablation study on the 2D multi-mode synthetic transport problem. Mode coverage is evaluated using the Chamfer distance (CD) between the target and generated distributions.}
\label{tab:synthetic_ablation}
\small
\setlength{\tabcolsep}{2pt}
\resizebox{\textwidth}{!}{
\begin{tabular}{lccccc}
\toprule
{Method} & $\mathcal{W}_2$ & $\mathcal{W}_1 $ & {Mass} & {Leakage (\%)} & {CD} \\
\midrule
CuOT (Complete) & $\bm{0.10_{\pm 0.01}}$ & $\bm{0.08_{\pm 0.01}}$ & \underline{$0.0006_{\pm 0.0001}$} & $\bm{0.2\%}$ & \underline{$0.09_{\pm 0.02}$} \\
CuOT w/ Time-Only Growth & $0.27_{\pm 0.02}$ & $0.19_{\pm 0.02}$ & $0.1484_{\pm 0.0104}$ & $14.1\%$ & $0.31_{\pm 0.03}$ \\
CuOT w/ WFR Geometry & $0.16_{\pm 0.02}$ & $0.13_{\pm 0.01}$ & $0.0009_{\pm 0.0001}$ & $10.4\%$ & $0.11_{\pm 0.02}$ \\
CuOT w/o Adaptive Step & \underline{$0.13_\pm{0.01}$} & \underline{$0.10_{\pm 0.01}$} & $0.0009_{\pm 0.0001}$ & $\bm{0.2\%}$ & $0.10_{\pm 0.01}$ \\
{VarRUOT} & $0.16_{\pm 0.01}$ & $0.13_{\pm 0.01}$ & \underline{$0.0006_{\pm 0.0001}$} & $6.5\%$ & $0.13_{\pm 0.02}$ \\
{OTCFM} & $0.20_{\pm 0.01}$ & $0.13_{\pm 0.01}$ & $\bm{0.0001_{\pm 0.0001}}$ & $17.7\%$ & $\bm{0.07_{\pm 0.01}}$ \\
\bottomrule
\end{tabular}
}
\end{table}
\vspace{-4mm}
\vspace{-4mm}
\textbf{Single-cell reconstruction.~~}
We next benchmark the proposed \ac{CUOT} framework on the trajectory inference from temporal snapshots of single-cell data, considering $5$ well-established benchmarks (see Table~\ref{tab:experimental_overview}). 
Analysis on these datasets reveals a fundamental mismatch: the Wasserstein distance $d^{\mathcal{W}_2}$ between successive marginals $\{\rho^m\}_{m=0}^M$ is not proportional to the growth rate gradient $\nabla_x z^t(x)$.
This means that real-world biological processes often violate the assumption underlying WFR-based methods.
Even in the simulated three-gene dataset~\citep{zhang2025learning}, the distance between successive distributions decreases monotonically, while $\nabla_x z^t(x)$ remains nearly constant (see Table~\ref{tab:3gene_dataset} in App.~\ref{app:single-cell}), exposing a structural decoupling that WFR cannot accommodate.
Figure~\ref{fig:simulation} shows the 2D projection of the gene dynamics, alongside the predicted growth rate. 
The ground-truth dynamics consist of two distinct modes: a stationary primary mode and a dynamically evolving secondary mode.
Balanced OT methods fail to model this dynamics, suffering from spurious cross-mode leakage.
While uOT approaches successfully isolate the evolving mode,  
WFR-based solutions are hindered by the structural biases that degrade their reconstruction accuracy (see Table~\ref{tab:simulation}). 
In contrast, our CuOT approach achieves superior performance across all metrics.
On real-world datasets, the gradient $\nabla_x z^t$ is highly variable and noisy, rendering the WFR-based estimation of the transport term $v^t$ unstable. 
This is evident in the zebrafish embryogenesis dynamics~\citep{farrell2018single}, where a single progenitor population branches into multiple lineage modes. 
As shown in Fig.~\ref{fig:zebra}, balanced OT methods achieve broad modality coverage, but fail to adapt sample weights, hindering precise matching. 
Conversely, although most uOT methods achieve higher scores in Table~\ref{tab:zebrafish} through weight adjustments, their coverage of the different modalities remains unsatisfactory. 
By avoiding these structural biases, our CuOT approach consistently outperforms the state-of-the-art baselines, successfully capturing both the diversity and the precision of the biological dynamics.
\looseness-1
\begin{table}[t!]
\centering
\vspace{-5mm}
\footnotesize{
\caption{Summary of single-cell reconstruction experiments (details in App.~\ref{app:single-cell}).}
\resizebox{\textwidth}{!}{
\begin{tabular}{lccll}
\toprule
Dataset & Dim. & Biological Process & Details & Results \\
\midrule
Three-gene simulation \citep{zhang2025learning} & 2D & Simulated gene expression dynamics & Table~\ref{tab:3gene_dataset} & Fig.~\ref{fig:simulation}, Table~\ref{tab:simulation} \\
Mouse Hematopoiesis \citep{weinreb2020lineage} & 2D & Blood stem cell specialization & Table~\ref{tab:mouse_dataset} & Fig.~\ref{fig:mouse}, Table~\ref{tab:mouse} \\
EMT \citep{cook2020context} & 10D & Cancer cell transition & Table~\ref{tab:emt_dataset} & Fig.~\ref{fig:emt}, Table~\ref{tab:emt} \\
Pancreatic $\beta$-cell \citep{veres2019charting} & 30D & Stem cell maturation & Table~\ref{tab:veres_dataset} & Fig.~\ref{fig:veres}, Tables~\ref{tab:veres},~\ref{tab:extended_veres} \\
Zebrafish Embryogenesis \citep{farrell2018single} & 50D & Cell specialization dynamics & Table~\ref{tab:zebra_dataset} & Fig.~\ref{fig:zebra}, Tables~\ref{tab:zebrafish},~\ref{tab:extended_zebrafish} \\
\bottomrule
\end{tabular}
\label{tab:experimental_overview}
}}
\end{table}
\begin{table}[t!]
\centering
\caption{Quantitative analysis on the three-gene simulated dataset \citep{zhang2025learning}.}
\small
\setlength{\tabcolsep}{4pt}
\resizebox{\textwidth}{!}{
\begin{tabular}{llcccc}
\toprule
Model & Metric & $t=1$ & $t=2$ & $t=3$ & $t=4$ \\
\midrule

\multirow{3}{*}{OTCFM \citep{tong2023improving}}
& $\mathcal{W}_1$ & 0.0889$_{\pm 0.0019}$ & 0.1149$_{\pm 0.0018}$ & 0.2086$_{\pm 0.0019}$ & 0.4054$_{\pm 0.0019}$ \\
& $\mathcal{W}_2$ & 0.1430$_{\pm 0.0028}$ & 0.1257$_{\pm 0.0021}$ & 0.2793$_{\pm 0.0027}$ & 0.5505$_{\pm 0.0022}$ \\
& Mass & 0.0110$_{\pm 0.0000}$ & 0.1056$_{\pm 0.0000}$ & 0.5256$_{\pm 0.0000}$ & 2.0235$_{\pm 0.0000}$ \\

\midrule

\multirow{3}{*}{TIGON \citep{sha2024reconstructing}}
& $\mathcal{W}_1$ & 0.0636$_{\pm 0.0012}$ & 0.0998$_{\pm 0.0011}$ & 0.1601$_{\pm 0.0022}$ & 0.1885$_{\pm 0.0023}$ \\
& $\mathcal{W}_2$ & ${0.1040_{\pm 0.0008}}$ & 0.1925$_{\pm 0.0025}$ & 0.2534$_{\pm 0.0030}$ & 0.2871$_{\pm 0.0027}$ \\
& Mass & 0.0263$_{\pm 0.0009}$ & 0.0490$_{\pm 0.0009}$ & 0.0527$_{\pm 0.0009}$ & 0.0531$_{\pm 0.0010}$ \\

\midrule

\multirow{3}{*}{DeepRUOT \citep{zhang2025learning}}
& $\mathcal{W}_1$ & $\underline{0.0448_{\pm 0.0063}}$ & 0.0832$_{\pm 0.0074}$ & 0.1229$_{\pm 0.0074}$ & 0.1450$_{\pm 0.0074}$ \\
& $\mathcal{W}_2$ & 0.0525$_{\pm 0.0061}$ & 0.1391$_{\pm 0.0096}$ & 0.2325$_{\pm 0.0103}$ & 0.2429$_{\pm 0.0105}$ \\
& Mass & 0.0005$_{\pm 0.0001}$ & 0.0012$_{\pm 0.0001}$ & $\underline{0.0004_{\pm 0.0001}}$ & $\underline{0.0002_{\pm 0.0001}}$ \\

\midrule

\multirow{3}{*}{VarRUOT \citep{sun2025variational}}
& $\mathcal{W}_1$ & $\underline{0.0476_{\pm 0.0039}}$ & $\underline{0.0518_{\pm 0.0049}}$ & $\underline{0.0524_{\pm 0.0055}}$ & $\underline{0.0569_{\pm 0.0053}}$ \\
& $\mathcal{W}_2$ & $\underline{0.0431_{\pm 0.0101}}$ & $\underline{0.0865_{\pm 0.0117}}$ & $\underline{0.0884_{\pm 0.0129}}$ & $\underline{0.1009_{\pm 0.0104}}$ \\
& Mass & $\underline{0.0002_{\pm 0.0001}}$ & $\underline{0.0005_{\pm 0.0001}}$ & 0.0065$_{\pm 0.0006}$ & 0.0032$_{\pm 0.0008}$ \\

\midrule

\multirow{3}{*}{\textbf{CuOT (Ours)}}
& $\mathcal{W}_1$ & $\bm{0.0337_{\pm 0.0018}}$ & $\bm{0.0466_{\pm 0.0064}}$ & $\bm{0.0381_{\pm 0.0044}}$ & $\bm{0.0225_{\pm 0.0008}}$ \\
& $\mathcal{W}_2$ & $\bm{{0.0296_{\pm 0.0020}}}$ & $\bm{0.0589_{\pm 0.0115}}$ & $\bm{0.0624_{\pm 0.0092}}$ & $\bm{0.0355_{\pm 0.0054}}$ \\
& Mass & $\bm{9\times 10^{-8}_{\pm 6\times 10^{-8}}}$ & $\bm{2\times 10^{-6}_{\pm 2\times 10^{-7}}}$ & $\bm{1\times 10^{-5}_{\pm 3\times 10^{-6}}}$ & $\bm{5\times 10^{-5}_{\pm 1\times 10^{-5}}}$ \\

\bottomrule
\end{tabular}
}
\label{tab:simulation}
\vspace{-4mm}
\end{table}
\begin{table}[t!]
\centering
\footnotesize{
\caption{Quantitative analysis on the dataset \citet{farrell2018single}. Results averaged over $t=1,\dots,6$.\looseness-1}
{
\begin{tabular}{lccc}
\toprule
Method & $\mathcal{W}_1$ & $\mathcal{W}_2$ & Mass \\
\midrule
OTCFM \citep{tong2023improving} & $6.8356_{\pm 2.2546}$ & $5.0909_{\pm 1.2883}$ & $0.8359_{\pm 0.4004}$ \\
TIGON \citep{sha2024reconstructing} & $8.3305_{\pm 1.5746}$ & $7.2321_{\pm 2.0083}$ & $0.5274_{\pm 0.1145}$ \\
DeepRUOT \citep{zhang2025learning} & $\underline{{4.2386}_{\pm 1.2133}}$ & $\underline{{4.9448}_{\pm 1.0268}}$ & $\underline{{0.1823}_{\pm 0.1875}}$ \\
VarRUOT \citep{sun2025variational} & $5.2476_{\pm 0.9197}$ & $\underline{4.8008_{\pm 0.7942}}$ & $0.4192_{\pm 0.2929}$ \\
\textbf{CuOT (Ours)} & $\bm{3.2689_{\pm 0.4946}}$ & $\bm{3.0357_{\pm 0.4140}}$ & $\bm{0.0243_{\pm 0.0205}}$ \\
\bottomrule
\end{tabular}
\label{tab:zebrafish}
}}
\vspace{-3mm}
\end{table}

\begin{figure}[t!]
\centering
\begin{tikzpicture}
    \node[anchor=south west, inner sep=0] (grid) at (0,0)
    {\includegraphics[width=0.7\linewidth]{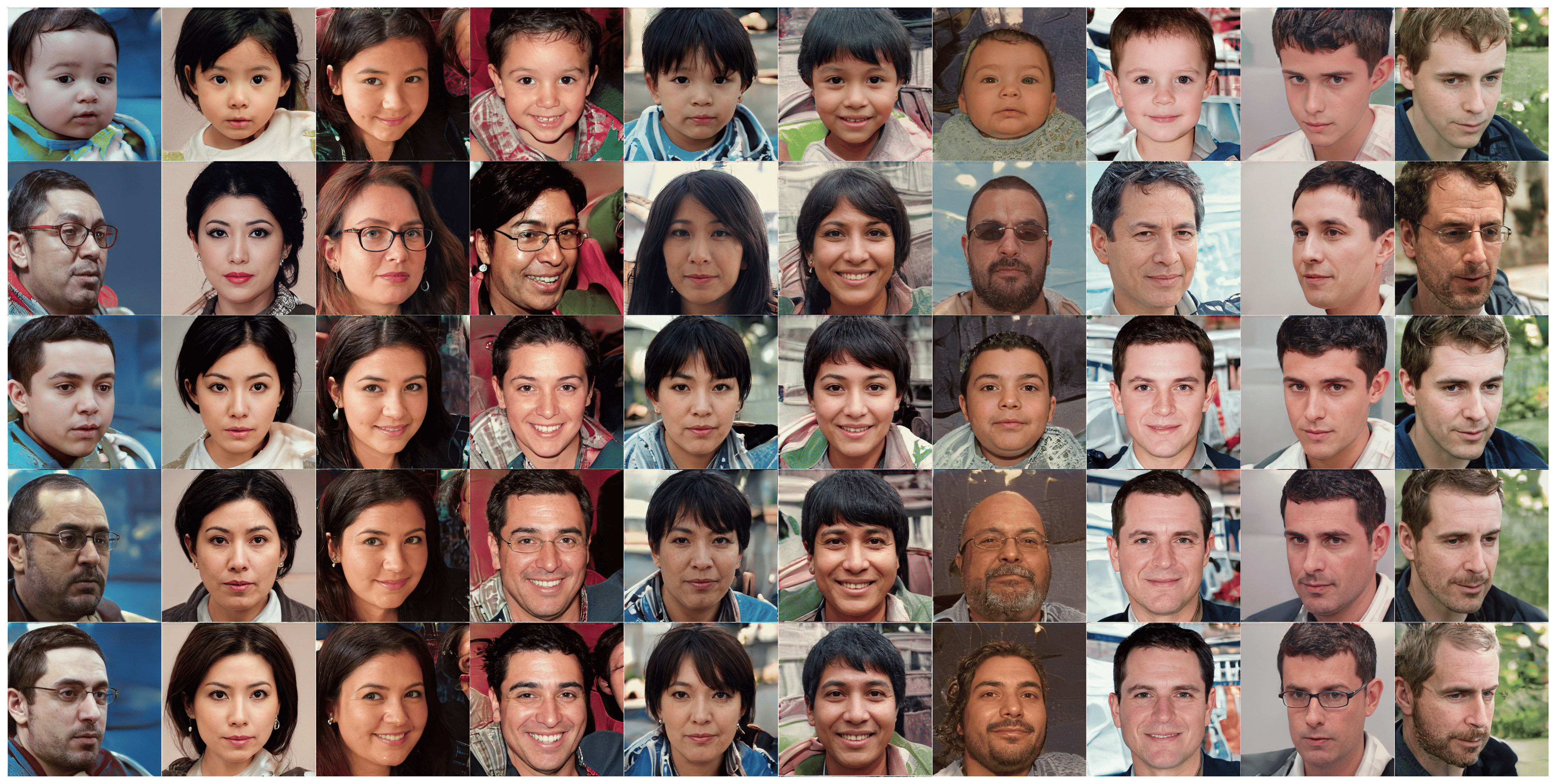}};
    \node[left] at (-0.0,4.5) { \footnotesize Input};
    \node[left] at (-0.0,3.4) {{\footnotesize \textbf{CuOT { (Ours)}}}};
    \node[left] at (-0.0,2.5) {\footnotesize VarRUOT};
    \node[left] at (-0.0,1.5) {\footnotesize DeepRUOT};
    \node[left] at (-0.0,0.6) {\footnotesize OTCFM};
\end{tikzpicture}
\vspace{-2mm}
\caption{Child $\to$ Adult generation on the FFHQ dataset \citep{karras2019style} with ALAE encoder.}
\label{fig:ffhq_c2a}
\vspace{-4mm}
\end{figure}

\textbf{Unpaired dataset matching.~~}
We then test CuOT on unpaired image-matching tasks using the \href{https://github.com/NVlabs/ffhq-dataset}{Flickr-Faces-HQ (FFHQ)} human face dataset \citep{karras2019style}. We employ pre-trained encoders to compute the stochastic dynamics in two distinct latent representations: a $512$-dimensional space provided by ALAE \citep{pidhorskyi2020adversarial} and a $9216$-dimensional space provided by e4e \citep{richardson2021encoding}.
We partition the dataset into two age-based groups: \textit{adults} ($45K$ samples) and \textit{children} ($15K$ samples). 
The Adult $\to$ Child task requires accurate early-stage dynamics in a highly-dense source to ensure correct transport to the target mode. 
Conversely, the Child $\to$ Adult task demands high-fidelity reconstructions and a correct coverage of the target distribution support.
Within the ALAE latent space (Table~\ref{tab:ffhq_avg}), balanced OT approaches yield larger distances from the target distribution than WFR-based uOT methods; however, they achieve the second-best FID scores on decoded images, which better measures mode coverage in the reconstruction. As illustrated in Figure~\ref{fig:ffhq_c2a}, although higher latent distances correlate with lower visual fidelity, balanced OT methods provide broader distribution coverage, producing images that faithfully reflect the target age group. In contrast, WFR-based uOT baselines frequently fail to complete the transport trajectory effectively and suffer from severe scalability bottlenecks in the higher-dimensional e4e latent space (Table~\ref{tab:ffhq_avg_e2e}). By leveraging contact geometry over the WFR structure, CuOT avoids mode collapse, achieving state-of-the-art performance across all metrics and latent representations with superior sample quality and target coverage.
Additional details are provided in App.~\ref{app:dataset-pairing}, where Figs.~\ref{fig:ffhq_contactubs}--\ref{fig:ffhq_otcfm} show how CuOT allocates more sampling steps in high-density regions to improve accuracy.\looseness-1
\vspace{-3mm}
\begin{table}[H]
\footnotesize{
\centering
\caption{Adult$\leftrightarrow$Child image generation on the FFHQ dataset \citep{karras2019style}, ALAE encoder.}
\resizebox{\textwidth}{!}{
\begin{tabular}{lcccc}
\toprule
Method & FID & $\mathcal{W}_1$ & $\mathcal{W}_2$ & Mass \\
\midrule
OTCFM \citep{tong2023improving} 
& $\underline{5.5307_{\pm 0.0816}}$ 
& $14.4850_{\pm 0.1087}$ 
& $10.3316_{\pm 0.1097}$ 
& $6.5052_{\pm 0.0000}$ \\

TIGON \citep{sha2024reconstructing} 
& $21.8858_{\pm 0.4533}$ 
& $16.4599_{\pm 0.1795}$ 
& $11.5193_{\pm 0.1489}$ 
& $0.7945_{\pm 0.0696}$ \\

DeepRUOT \citep{zhang2025learning} 
& $6.5654_{\pm 0.1801}$ 
& $\underline{13.7665_{\pm 0.1090}}$ 
& $\underline{9.8020_{\pm 0.1058}}$ 
& $\underline{0.0102_{\pm 0.0082}}$ \\

VarRUOT \citep{sun2025variational} 
& $27.2228_{\pm 0.2637}$ 
& $14.7688_{\pm 0.1265}$ 
& $10.5592_{\pm 0.1171}$ 
& $0.8849_{\pm 0.3058}$ \\

\textbf{CuOT (Ours)} 
& $\bm{5.1764_{\pm 0.0761}}$ 
& $\bm{12.4042_{\pm 0.1080}}$ 
& $\bm{8.8574_{\pm 0.1005}}$ 
& $\bm{0.0043_{\pm 0.0018}}$ \\

\bottomrule
\end{tabular}
}
\label{tab:ffhq_avg}
}
\end{table}
\vspace{-8mm}
\begin{table}[H]
\footnotesize{
\centering
\caption{Adult$\leftrightarrow$Child image generation on the FFHQ dataset \citep{karras2019style}, e2e encoder.}
\resizebox{\textwidth}{!}{
\begin{tabular}{lcccc}
\toprule
Method & FID & $\mathcal{W}_1$ & $\mathcal{W}_2$ & Mass \\
\midrule
OTCFM \citep{tong2023improving} 
& \underline{$9.3859_{\pm 0.1646}$} 
& \underline{$19.3396_{\pm 0.3013}$} 
& \underline{$14.8317_{\pm 0.1881}$} 
& $6.5049_{\pm 0.0000}$ \\

TIGON \citep{sha2024reconstructing} 
& $45.2745_{\pm 0.3889}$ 
& $28.8340_{\pm 0.3516}$ 
& $20.5483_{\pm 0.3577}$ 
& $1.5906_{\pm 0.1553}$ \\

DeepRUOT \citep{zhang2025learning} 
& $39.7057_{\pm 0.1952}$ 
& $27.7649_{\pm 0.2185}$ 
& $20.7265_{\pm 0.2127}$ 
& \underline{$0.0735_{\pm 0.0103}$} \\

VarRUOT \citep{sun2025variational} 
& $54.6319_{\pm 0.0800}$ 
& $31.3152_{\pm 0.2591}$ 
& $24.5797_{\pm 0.3155}$ 
& $1.1373_{\pm 0.4025}$ \\

\textbf{CuOT (Ours)} 
& $\bm{8.3972_{\pm 0.0972}}$ 
& $\bm{16.0868_{\pm 0.1131}}$ 
& $\bm{12.1086_{\pm 0.1408}}$ 
& $\bm{0.0153_{\pm 0.0144}}$ \\

\bottomrule
\end{tabular}
}
\label{tab:ffhq_avg_e2e}
}
\end{table}
\vspace{-3mm}

\begin{figure}[H]
\centering
\begin{tikzpicture}
    \node[anchor=south west, inner sep=0] (grid) at (0,0)
    {\includegraphics[width=0.7\linewidth]{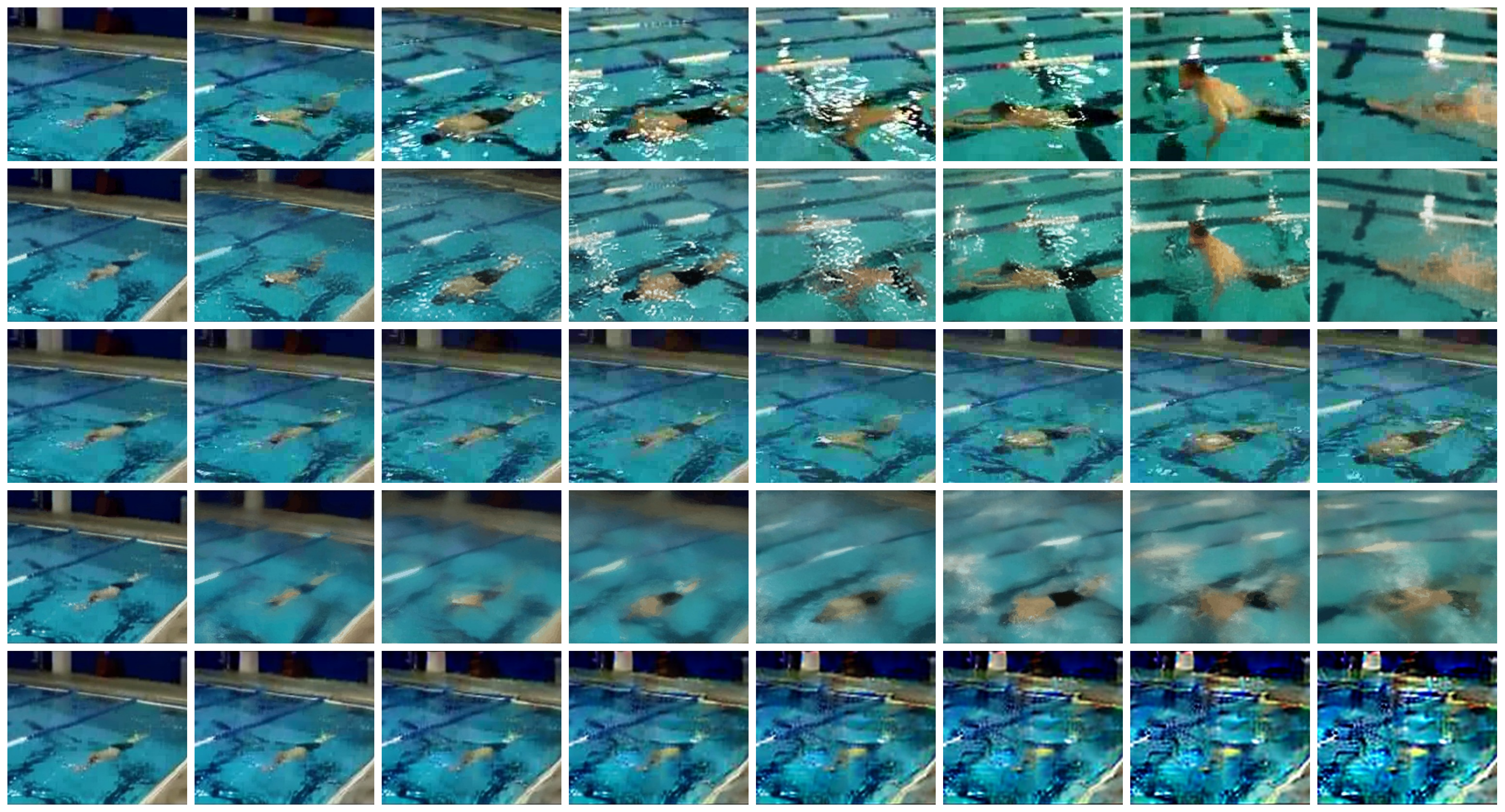}};
    \node[left] at (-0.0,4.8) { \footnotesize Reference};
    \node[left] at (-0.0,3.7) {{\footnotesize \textbf{CuOT (Ours)}}};
    \node[left] at (-0.0,2.6) {\footnotesize VarRUOT};
    \node[left] at (-0.0,1.5) {\footnotesize DeepRUOT};
    \node[left] at (-0.0,0.5) {\footnotesize OTCFM};
\end{tikzpicture}
\vspace{-1mm}
\caption{Image-to-Video generation on the UCF101 dataset \citep{soomro2012ucf101}, for DC encoder. \looseness-1}
\label{fig:ucf101}
\vspace{-4mm}
\end{figure}

\textbf{Image-to-Video generation.~~}
We further evaluate our method by performing video trajectory reconstruction on the \href{https://www.crcv.ucf.edu/data/UCF101.php}{UCF101} dataset \citep{soomro2012ucf101}. Here, the learned transport map reconstructs temporal frame sequences across $M=60$ intermediate marginal distributions, with each marginal corresponding to a discrete time step along the video dynamics. 
Because the frame count per video remains constant, mass variation is negligible. Therefore, this benchmark primarily evaluates scalability and stability over high-dimensional, long-transport horizons.
We operate in pre-trained latent spaces to focus on semantic video dynamics over pixel redundancies. Specifically, we ablate performance across two  representations: (i) the Deep Compression Autoencoder \citep{chen2025deep}, which maps video frames into a $32 \times 8 \times 8 = 2048$-dimensional space, and (ii) the TiTok encoder \citep{yu2024an}, which compresses frames into a $128 \times 32$ sequence representation.
We consider three action classes with approximately $100$ videos each: \textit{BreastStroke}, \textit{BlowingCandles}, and \textit{CuttingInKitchen}.
As summarized in Table~\ref{tab:ucf101}, WFR-based uOT methods fail to scale to this regime, often producing collapsed trajectories that do not evolve beyond the initial frame, or exhibiting poor distributional alignment that leads to high noise and distortion. In contrast, CuOT successfully recovers the underlying video dynamics. Fig.~\ref{fig:ucf101} shows a qualitative example for the \textit{BreastStroke} class. In both latent representations, CuOT maintains high-fidelity reconstructions across the full temporal horizon. Detailed breakdowns across individual classes and latent representations are provided in App.~\ref{app:image-to-video}.\looseness-1
\vspace{-4mm}
\begin{table}[H]
\centering
\footnotesize{
\caption{Image-to-Video generation on the UCF101 dataset \citep{soomro2012ucf101}.}
\begin{tabular}{lccc}
\toprule
Method & FID & $\mathcal{W}_1$ & $\mathcal{W}_2$ \\
\midrule
OTCFM    & $208.5315_{\pm 19.7064}$ & $25.6506_{\pm 3.5847}$ & $17.3783_{\pm 2.3124}$ \\
TIGON    & $336.5115_{\pm 17.0147}$ & $31.0243_{\pm 4.9427}$ & $20.1477_{\pm 2.8925}$ \\
DeepRUOT & $\underline{138.2666_{\pm 4.4798}}$ & $\underline{20.6668_{\pm 2.8854}}$ & $\underline{14.3858_{\pm 2.1248}}$ \\
VarRUOT  & $325.7000_{\pm 13.2486}$ & $28.1000_{\pm 4.7682}$ & $18.3665_{\pm 3.3568}$ \\
\textbf{CuOT (Ours)} & $\bm{18.3783_{\pm 1.7926}}$ & $\bm{11.2006_{\pm 1.6584}}$ & $\bm{8.1669_{\pm 1.0897}}$ \\
\bottomrule
\end{tabular}
\label{tab:ucf101}
}
\end{table}
\vspace{-5mm}
\vspace{-3mm}
\section{Conclusions and Limitations}\label{sec:conclusions}
\vspace{-3mm}
We introduced Contact unbalanced Optimal Transport (CuOT), a stable and scalable generative model for unpaired dataset matching grounded in contact geometry. 
CuOT outperforms traditional uOT approaches by eliminating structural biases toward high-density regions and adapting the sampling step size to the data geometry. 
As detailed in App. \ref{app:limitations}, while CuOT is architecture-agnostic, validation on structured data with spatial or sequential relationships, where Transformers are the established backbone, remains open. Furthermore, establishing formal connections with diffusion methods that match data via Fisher--Rao geometry in noise space remains an open direction for future work.\looseness-1

\newpage
\bibliographystyle{apalike}
\bibliography{neurips_2026}

\section*{NeurIPS Paper Checklist}

\begin{enumerate}

\item {\bf Claims}
    \item[] Question: Do the main claims made in the abstract and introduction accurately reflect the paper's contributions and scope?
    \item[] Answer: \answerYes{} 
    \item[] Justification: Abstract and Section \ref{sec:intro}.

\item {\bf Limitations}
    \item[] Question: Does the paper discuss the limitations of the work performed by the authors?
    \item[] Answer: \answerYes{} 
    \item[] Justification: Section \ref{sec:results}, Appendix \ref{app:limitations}.

\item {\bf Theory assumptions and proofs}
    \item[] Question: For each theoretical result, does the paper provide the full set of assumptions and a complete (and correct) proof?
    \item[] Answer: \answerYes{} 
    \item[] Justification: Section \ref{sec:methodology}, \ref{sec:comparison}, Appendix \ref{app:wass_equivalence}, \ref{app:Contact-sb_demons}, \ref{app:sb_demons}, \ref{app:FR-sb_demons}.

    \item {\bf Experimental result reproducibility}
    \item[] Question: Does the paper fully disclose all the information needed to reproduce the main experimental results of the paper to the extent that it affects the main claims and/or conclusions of the paper (regardless of whether the code and data are provided or not)?
    \item[] Answer: \answerYes{} 
    \item[] Justification: Section \ref{sec:implementation}, Appendix \ref{app:implementation}, \ref{app:results}.

\item {\bf Open access to data and code}
    \item[] Question: Does the paper provide open access to the data and code, with sufficient instructions to faithfully reproduce the main experimental results, as described in supplemental material?
    \item[] Answer: \answerNo{} 
    \item[] Justification: We will release code for full reproducibility upon publication of the paper.

\item {\bf Experimental setting/details}
    \item[] Question: Does the paper specify all the training and test details (e.g., data splits, hyperparameters, how they were chosen, type of optimizer) necessary to understand the results?
    \item[] Answer: \answerYes{} 
    \item[] Justification: Appendix \ref{app:implementation}, \ref{app:results}.

\item {\bf Experiment statistical significance}
    \item[] Question: Does the paper report error bars suitably and correctly defined or other appropriate information about the statistical significance of the experiments?
    \item[] Answer: \answerYes{} 
    \item[] Justification: Section \ref{app:results}, Appendix \ref{app:results}.

\item {\bf Experiments compute resources}
    \item[] Question: For each experiment, does the paper provide sufficient information on the computer resources (type of compute workers, memory, time of execution) needed to reproduce the experiments?
    \item[] Answer: \answerYes{} 
    \item[] Justification: Appendix \ref{app:results}.
    
\item {\bf Code of ethics}
    \item[] Question: Does the research conducted in the paper conform, in every respect, with the NeurIPS Code of Ethics \url{https://neurips.cc/public/EthicsGuidelines}?
    \item[] Answer: \answerYes{} 
    \item[] Justification: N/A

\item {\bf Broader impacts}
    \item[] Question: Does the paper discuss both potential positive societal impacts and negative societal impacts of the work performed?
    \item[] Answer: \answerYes{} 
    \item[] Justification: Appendix \ref{app:impacts}.
    
\item {\bf Safeguards}
    \item[] Question: Does the paper describe safeguards that have been put in place for responsible release of data or models that have a high risk for misuse (e.g., pre-trained language models, image generators, or scraped datasets)?
    \item[] Answer: \answerNA{} 
    \item[] Justification:  N/A

\item {\bf Licenses for existing assets}
    \item[] Question: Are the creators or original owners of assets (e.g., code, data, models), used in the paper, properly credited and are the license and terms of use explicitly mentioned and properly respected?
    \item[] Answer: \answerYes{} 
    \item[] Justification: Appendix \ref{app:results}.

\item {\bf New assets}
    \item[] Question: Are new assets introduced in the paper well documented and is the documentation provided alongside the assets?
    \item[] Answer: \answerNo{} 
    \item[] Justification: We will release code along with examples and documentation for full reproducibility upon publication of the paper.

\item {\bf Crowdsourcing and research with human subjects}
    \item[] Question: For crowdsourcing experiments and research with human subjects, does the paper include the full text of instructions given to participants and screenshots, if applicable, as well as details about compensation (if any)? 
    \item[] Answer: \answerNA{} 
    \item[] Justification: N/A

\item {\bf Institutional review board (IRB) approvals or equivalent for research with human subjects}
    \item[] Question: Does the paper describe potential risks incurred by study participants, whether such risks were disclosed to the subjects, and whether Institutional Review Board (IRB) approvals (or an equivalent approval/review based on the requirements of your country or institution) were obtained?
    \item[] Answer: \answerNA{} 
    \item[] Justification: N/A

\item {\bf Declaration of LLM usage}
    \item[] Question: Does the paper describe the usage of LLMs if it is an important, original, or non-standard component of the core methods in this research? Note that if the LLM is used only for writing, editing, or formatting purposes and does \emph{not} impact the core methodology, scientific rigor, or originality of the research, declaration is not required.
    \item[] Answer: \answerNA{} 
    \item[] Justification: N/A

\end{enumerate}
\newpage
\appendix
\section{Notation}\label{app:notation}
\begin{table}[H]
\centering
\caption{Summary of mathematical notation.}
\label{tab:notation}
\renewcommand{\arraystretch}{1.4}
\begin{tabular}{|l|p{9cm}|l|}
\hline
\textbf{Symbol} & \textbf{Description} & \textbf{Reference} \\
\hline
\multicolumn{3}{|l|}{\textbf{Probability Space}} \\
\hline
$\mathcal{M}$ & Data domain (sample space). & Sec.~\ref{sec:background} \\
$\mathcal{P}^{+}(\mathcal{M})$ & Space of smooth, positive probability densities supported on $\mathcal{M}$ (Wasserstein manifold). The associated tangent and cotangent bundles are $\mathcal{T}\mathcal{P}^{+}(\mathcal{M})$ and $\mathcal{T}^*\mathcal{P}^{+}(\mathcal{M})$, respectively. & Sec.~\ref{sec:background} \\
$\rho^t(x)$ & Time-dependent probability density; $\rho^t \in \mathcal{P}^+(\mathcal{M})$. & Sec.~\ref{sec:background} \\
$\{\rho^m\}_{m=0}^{M}$ & Observed marginal densities at times $\{t_m\}_{m=0}^{M}$. & Eq.~\eqref{eq:contact_schrodinger_bridge} \\
\hline
\multicolumn{3}{|l|}{\textbf{Transport Dynamics}} \\
\hline
$v^t(x)$ & Velocity field governing sample transport; $v^t = \nabla_x S^t$. & Eq.~\eqref{eq:fokker-planck} \\
$z^t(x)$ & Local mass growth rate field. & Eq.~\eqref{eq:contact_schrodinger_bridge} \\
$S^t(x)$ & Dual potential (canonical momentum); generates the velocity via $\nabla_x S^t(x) = v^t(x)$. & Sec.~\ref{sec:background} \\
$\varepsilon$ & Diffusion coefficient for entropy regularization. & Eq.~\eqref{eq:fokker-planck} \\
$w_z$ & Weighting factor for the influence of mass variation. & Eq.~\eqref{eq:contact_schrodinger_bridge} \\
$J$ & Objective functional minimized. & Eq.~\eqref{eq:contact_schrodinger_bridge} \\
\hline
\multicolumn{3}{|l|}{\textbf{Hamiltonian Functions}} \\
\hline
$\mathcal{H}(\rho^t, S^t)$ & Symplectic Hamiltonian; $\mathcal{H} = \mathcal{K} + \varepsilon\mathcal{F}$. & Sec.~\ref{sec:comparison} \\
$\mathcal{H}(\rho^t, S^t, z^t)$ & Contact Hamiltonian; $\mathcal{H} = \mathcal{K} + \varepsilon\mathcal{F} + \frac{w_z}{2}\int (z^t)^2 dx + \mathcal{H}_0$. & Eq.~\eqref{eq:contact_hamiltonian_red} \\
$\mathcal{K}(\rho^t, S^t)$ & Kinetic energy; $\mathcal{K} = \int_\mathcal{M} \frac{1}{2}\|\nabla_x S^t\|^2 \rho^t\, dx$. & Sec.~\ref{sec:methodology} \\
$\mathcal{F}(\rho^t, S^t)$ & Entropic coupling term (scaled by $\varepsilon$). & Sec.~\ref{sec:methodology} \\
$\mathcal{H}_0$ & Reference energy constant; $\mathcal{H}_0 = \int_\mathcal{M} \frac{1}{2}\|v^{t_0}\|^2 \rho^{t_0}\, dx$. & Eq.~\eqref{eq:contact_hamiltonian_red} \\
\hline
\multicolumn{3}{|l|}{\textbf{Geometric Structure}} \\
\hline
$g^{\mathcal{W}_2}$ & Wasserstein metric tensor on $\mathcal{P}^+(\mathcal{M})$, defining distance $d^{\mathcal{W}_2}$. & Eq.~\eqref{eq:wasserstein_metric} \\
$g^C$ & Contact Wasserstein metric, $g^C = \sqrt{\mathcal{K}}\, g^{\mathcal{W}_2}$, defining distance $d^{\mathcal{C}}$.. & Sec.~\ref{sec:methodology} \\
$\Delta_{\rho^t}$ & Weighted Laplacian, $\Delta_{\rho^t} = -\nabla_x \cdot (\rho^t \nabla_x)$. & Eq.~\eqref{eq:wasserstein_metric} \\
$\omega$ & Symplectic form on $\mathcal{T}^*\mathcal{P}^+(\mathcal{M})$; preserved under OT. & Sec.~\ref{sec:background} \\
$\eta$ & Contact form on $\mathcal{T}^*\mathcal{P}^+(\mathcal{M}) \times \mathbb{R}$. & Sec.~\ref{sec:background} \\
\hline
\multicolumn{3}{|l|}{\textbf{Discretization and Networks}} \\
\hline
$\{x^{s_k}_n, w^{s_k}_n\}_{n=1}^N$ & Weighted particle ensemble approximating $\rho^{s_k}$. & Eq.~\eqref{eq:discrete_geodesic} \\
$v_\theta(x^{s_k}, s_k)$ & Neural network parameterizing the velocity field. & Sec.~\ref{sec:implementation} \\
$z_\theta(x^{s_k}, s_k)$ & Neural network parameterizing the  mass growth rate. & Sec.~\ref{sec:implementation} \\
$\nu^{s_k}_n$ & Scalar speed of particle $x^{s_k}_n$; conformal scaling factor for displacement. & Eq.~\eqref{eq:discrete_nu} \\
$S^{s_k}_n$ & Per-particle dual potential at arc-length step $s_k$. & Eq.~\eqref{eq:discrete_S} \\
$\tau$ & Geodesic discretization step size. & Eq.~\eqref{eq:discrete_optimality} \\
$K,\, M,\, N,\, D$ & Number of geodesic steps, intermediate marginals, samples, and data dimensions. & Sec.~\ref{sec:implementation} \\
\hline
\end{tabular}
\end{table}
\newpage
\section{Geometric details}\label{app:geometry}
\vspace{-2mm}
\subsection{Symplectic and Contact Geometries}\label{app:sympletcic_contact}
\vspace{-2mm}
Hamiltonian and contact Hamiltonian dynamics are governed by specific energy constraints that can be analyzed via differential geometry as flows on specialized manifolds. Hamiltonian dynamics is energy-conserving and evolves on a \emph{symplectic manifold}. Contact Hamiltonian dynamics is more general, allowing for variable energy levels, and takes place on a \emph{contact manifold}. Their formal definitions and key differences are discussed next.\looseness-1

\begin{figure}[ht]
\vspace{-4mm}
  \centering
\includegraphics[width=0.45\textwidth]{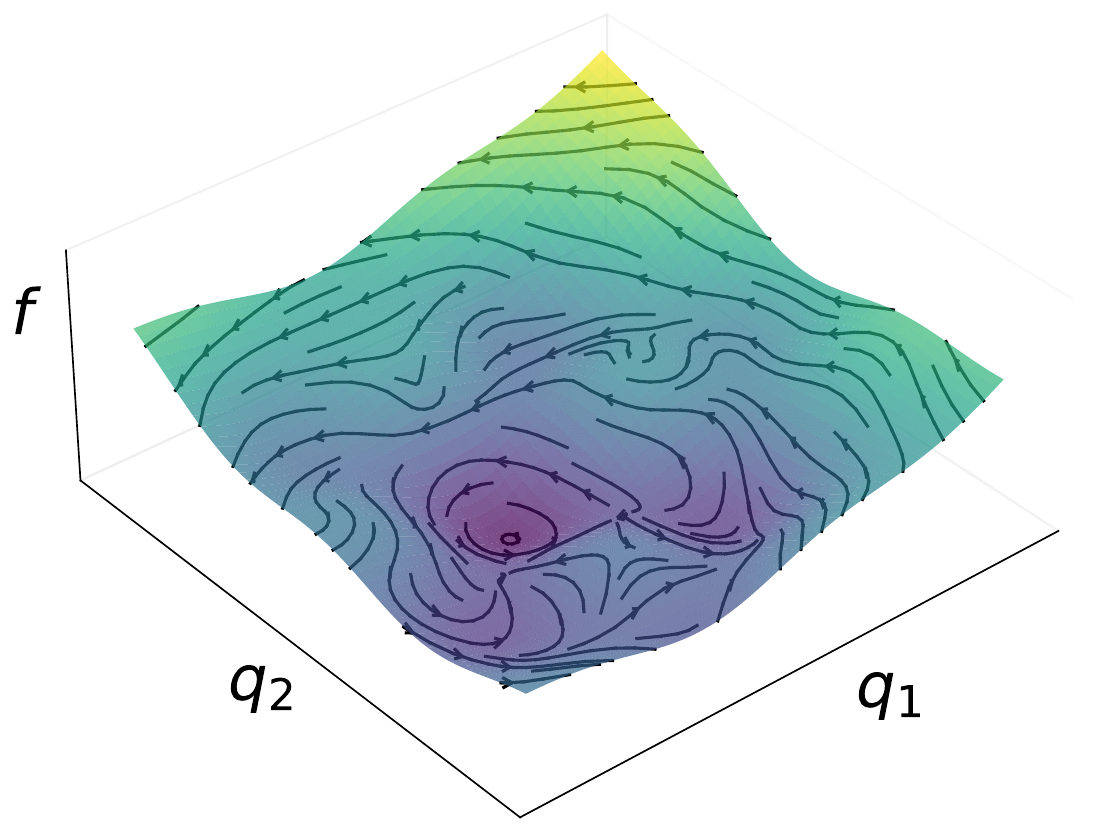}
\includegraphics[width=0.45\textwidth]{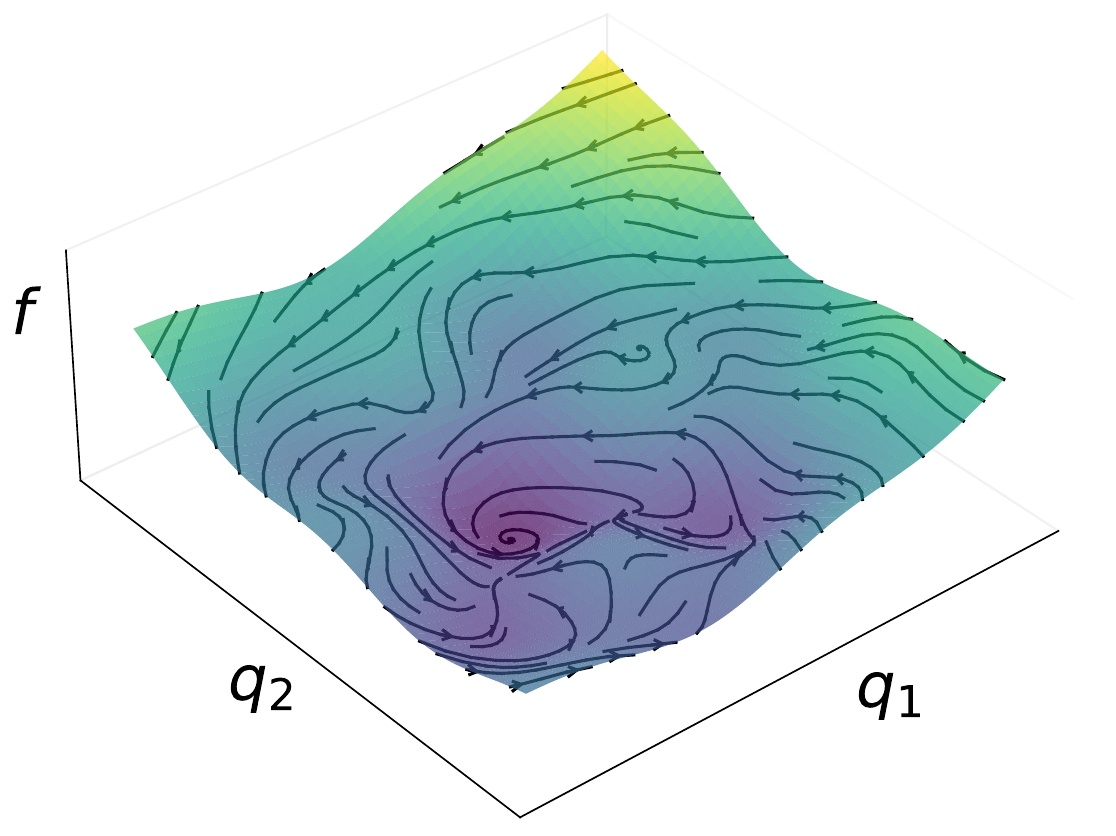}
  \caption{The same scalar function $f$, associated with the 1-form $\alpha=df$, gives rise to two distinct vector fields under the symplectic (left) and contact (right) geometric structures. The streamlines of these vector fields are illustrated on a representation of the state manifold. In symplectic geometry, the streamlines are tangent to the level curves of $f$, representing isoenergetic trajectories where $f$ remains constant,  thus describing the dynamics of conservative systems. In contrast, in contact geometry, a single flow line can traverse different energy levels.}
  \label{fig:geometric_connections}
\vspace{-2mm}
\end{figure}

\textbf{Symplectic and contact structures.~~}\label{app:riemann_pre}
Let $\mathcal{N}$ be a smooth compact manifold, and let $\mathcal{T}_x\mathcal{N}$ denote the tangent space at $x \in \mathcal{N}$. The collection of all the tangent spaces identifies the tangent bundle $\mathcal{T}\mathcal{N} = \cup_{x \in \mathcal{N}} \mathcal{T}_x\mathcal{N}$. A vector field $X:\mathcal{N} \rightarrow \mathcal{T}\mathcal{N}$ assigns a tangent vector $v$ to each point $x \in \mathcal{N}$. The set of all the vector fields over $\mathcal{T}\mathcal{N}$ is denoted as $\Gamma(\mathcal{T}\mathcal{N})$. A differential 1-form $\alpha : \mathcal{T}\mathcal{N} \rightarrow \mathbb{R}$ is a smooth map field acting on vectors of the tangent bundle. For a smooth function $f:\mathcal{N}\rightarrow\mathbb{R}$, the 1-form $\alpha = df$ generalizes the gradient from Euclidean spaces.
Specifically, $df$ measures the variation of $f$ under an infinitesimal displacement on $\mathcal{N}$.
This displacement is locally described by a starting point $x$ and a direction $v$, such that $(x, v) \in \mathcal{T}\mathcal{N}$. Alternatively, it can be globally expressed by a vector field $X$. 
The variation of $f$ along the vector field $X$ is given by $df(X)$. This variation is independent of the choice of reference frame. To preserve this invariance, $df$ must transform covariantly with $X$. Consequently, the 1-form $\alpha=df$ resides in the cotangent bundle $\mathcal{T}^*\mathcal{N}$, the dual space to $\mathcal{T}\mathcal{N}$.
The symplectic and contact structures provide two distinct mechanisms for associating a 1-form to a vector field, thereby establishing connections between the tangent and cotangent bundles. By considering the dynamics governed by the vector field and the scalar function defining the 1-form, a relationship between these elements emerges, as illustrated in Fig. \ref{fig:geometric_connections}.\looseness=-1

\textbf{Symplectic geometry.~~}
A differential 2-form $\omega : \mathcal{T}\mathcal{N} \times \mathcal{T}\mathcal{N} \rightarrow \mathbb{R}$ is a skew-symmetric, bilinear, and smooth field of maps acting on pairs of tangent vectors. A 2-form is called \emph{symplectic} if it is both closed ($d\omega=0$) and non-degenerate.
The symplectic form lacks the properties required to define an inner product. However, it still establishes a fundamental relation between differential 1-forms and vector fields: Given a 1-form $df$, the symplectic form $\omega$ uniquely determines a vector field $X_f$ that is tangent to the level sets of $f$. This relation is defined by,
\begin{equation}
\label{eq:symplectic_ham}
    df(X)=\omega(X_f, X), \quad \forall X \in \Gamma(\mathcal{T}\mathcal{N}).
\end{equation}
By definition, $f$ remains constant along the flow of $X_f$, which in turn preserves the symplectic form $\omega$, i.e., $\mathcal{L}_{X_f} \omega=0$ where $\mathcal{L}_{X_f}$ denotes the Lie derivative~\citep{silva2001lectures}.
Since the symplectic form captures a generalized notion of signed area on pairs of tangent vectors, its invariance under the flow implies the preservation of volume.
In this framework, the function $f$ is interpreted as a conserved energy, or equivalently, as a Hamiltonian $\mathcal{H}$. The symplectic structure thereby endows $\mathcal{N}$ with a natural geometric framework for formulating Hamiltonian dynamics~\citep{tokasi2022symplectic}.
The pair $(\mathcal{N}, \omega)$ is referred to as a symplectic manifold. Notably, the non-degeneracy of $\omega$ implies that $\mathcal{N}$ must be even-dimensional.\looseness=-1

\textbf{Contact geometry.~~}
While symplectic manifolds provide a geometric framework for modeling the dynamics of volume-preserving, conservative systems in classical mechanics, a more general approach is required to describe non-conservative systems. This is addressed by contact manifolds, the odd-dimensional counterparts of symplectic manifolds~\citep{Geiges2001:ContactHistory,bravetti2017contact}.
A \emph{contact manifold} is defined as $(\mathcal{N}, \eta)$, where $\mathcal{N}$ is an odd-dimensional smooth manifold, and $\eta$ is a non-degenerate 1-form known as the contact form \citep{geiges2008introduction}. 
The contact form satisfies the \emph{maximal non-integrability} condition, meaning that the top-degree differential form $\eta \wedge (d\eta)^d\neq 0$ is nowhere vanishing on $\mathcal{N}$. This form is constructed by taking the exterior product of $\eta$ with the $d$-fold wedge product of its exterior derivative $d \eta$, i.e.,
\begin{equation}
    (d\eta)^d = \underbrace{d\eta \wedge \cdots \wedge d\eta}_{d \text{ times}}.
\end{equation}
The $(2d+1)$-form defines a volume form on $\mathcal{N}$, ensuring that the hyperplanes $\text{ker}(\eta) \subset \mathcal{TN}$, constraining the dynamics on the contact manifold, do not form a foliation, i.e., they do not partition the manifold into lower-dimensional submanifolds \citep{Geiges2001:ContactHistory, geiges2008introduction}.
Geometrically, this means that the contact distribution imposes \emph{non-holonomic constraints}: it restricts the admissible directions of motion at each point without confining the dynamics to a fixed submanifold or energy level. This property is crucial for modeling systems where energy can change over time, enabling constraints on energy behavior without enforcing conservation.
\looseness-1

Like symplectic geometry, contact geometry connects scalar functions to vector fields, enabling the description of dynamical systems~\citep{zadrathesis}.
Given an energy function $\mathcal{H} : \mathcal{N} \to \mathbb{R}$, the dynamics on a contact manifold are defined by a contact Hamiltonian vector field $X_\mathcal{H}$, as follows,
\begin{equation}
\label{eq:contact_ham}
    d\mathcal{H}(X) = d\eta (X_\mathcal{H}, X) \!-\! \mathcal{L}_{X_\mathcal{H}} \eta(X) , \,\, \forall X \in \Gamma(\mathcal{T}\mathcal{N}).
\end{equation}
Unlike symplectic geometry, where dynamics are confined to energy-preserving flows along the level sets of the Hamiltonian, contact geometry allows for an additional component of motion. Specifically, the dynamics on a contact manifold are not restricted to the term $d\eta (X_\mathcal{H}, X)$, which lies tangent to the level sets of $\mathcal{H}$, but also include a transverse component $\mathcal{L}_{X_\mathcal{H}} \eta(X)$, arising from the non-degeneracy of the contact form.
Consequently, while in symplectic geometry the symplectic form $\omega$ is strictly preserved, contact geometry allows the contact form $\eta$ to be preserved only up to a scaling factor $a \in \mathbb{R}$~\citep{bravetti2017contact}. Consequently, the volume form $\eta \wedge (d\eta)^d$ need not be conserved along the flow, permitting the dynamics to unfold on regions of the contact manifold $\mathcal{N}$ of varying measure.
\vspace{-2mm}
\subsection{Riemannian Geometry}\label{app:riemannian}
\vspace{-2mm}
\textbf{The Riemannian metric.~~}
Let $\mathcal{M}$ be a smooth compact manifold.
A Riemannian metric $g : \mathcal{T}\mathcal{M} \times \mathcal{T}\mathcal{M} \rightarrow \mathbb{R}$ is a smooth, symmetric, and positive-definite bilinear field of maps defined on pairs of vectors in the tangent bundle. This enables the introduction of an inner product on the tangent spaces of the manifold, allowing us to measure distances and curve lengths. For a smooth curve $x(t) : [t_0, t_1] \to \mathcal{M}$, the length $l$ w.r.t. the metric $g$ is $l=\int_{t_0}^{t_1} \sqrt{g(\dot{x}(t), \dot{x}(t))}dt$,
where $\dot{x}(t) \in \mathcal{T}_{x(t)}\mathcal{M}$ is the vector tangent to the curve at $x(t)$. The curve minimizing this length between two points $x(t_0)$ and $x(t_1)$ on $\mathcal{M}$ is called a \emph{geodesic}.
Geodesics generalize straight lines in Euclidean space to curved spaces, representing the shortest paths in the geometry induced by $g$.

\textbf{Hamiltonian geodesics.~~}
The geodesic flow $x(t)$ on a Riemannian manifold $(\mathcal{M}, g)$ lifts to the joint evolution of coordinates $\left(x(t),\alpha(x(t), \dot{x}(t)) \right)$ on the cotangent bundle $\mathcal{T}^*\mathcal{M}$~\citep{abraham2008foundations}. 
This extended dynamics is governed by an energy function $\mathcal{H}(x, \alpha):\mathcal{T}^*\mathcal{M} \rightarrow \mathbb{R} = g^{-1}(\alpha, \alpha)$, which remains constant along the flow. A reparameterization $ds = \sqrt{\mathcal{H}} dt$ links the trajectory of the integrated dynamical system at time $t$ on $\mathcal{T}^*\mathcal{M}$ with the length of the corresponding geodesic on $\mathcal{M}$. This framework reveals a fundamental connection between geodesic flows and Hamiltonian dynamics in the special case where the Hamiltonian consists solely of a kinetic energy term $\mathcal{H}=\mathcal{K}$. The cotangent bundle $\mathcal{T}^*\mathcal{M}$ is naturally equipped with a symplectic structure, making it a symplectic manifold $(\mathcal{T}^*\mathcal{R},\omega)$.
\looseness-1

\textbf{Contact Hamiltonian geodesics.~~}
This formulation admits a further generalization within the framework of contact Hamiltonian systems. In this setting, the contact Hamiltonian $\mathcal{H}(x, \alpha, z)$ depends explicitly on the action-like variable $z \in \mathbb{R}$, which evolves along the flow on the augmented space $(\mathcal{T}^*\mathcal{R} \times \mathbb{R}, \eta)$. Since the reparameterization $ds = \sqrt{\mathcal{K}} dt$ is no longer constant, the Hamiltonian does not define a unique metric, rather inducing a family of Riemannian metrics $\{ \sqrt{\mathcal{K}} g \}$ scaled by a $z$-dependent conformal factor. Consequently, the resulting dynamics can be interpreted as a piecewise geodesic flow on a manifold where the metric evolves according to the value of $z$.
\vspace{-2mm}
\subsection{Geometry of the Probability Space}
\label{app:geometry_probability_space}
\vspace{-2mm}
\textbf{Symplectic and Contact Structures on the Wasserstein Manifold.~~}
Denoting the Wasserstein manifold by $\mathcal{P}^+(\mathcal{M})$, its cotangent bundle
$\mathcal{T}^*\mathcal{P}^+(\mathcal{M})$ is an even-dimensional space whose state is described
by the local coordinate chart $(\rho^t, S^t)$. The balanced OT problem
induces a volume-preserving Hamiltonian dynamics on $(\mathcal{T}^*\mathcal{P}^+(\mathcal{M}),
\omega)$, with symplectic form given by
\begin{equation}
    \omega\big((\partial_t \rho_1^t, \partial_t S_1^t),\, (\partial_t \rho_2^{t}, \partial_t S_2^{t})\big)
    := \int_{\mathcal{M}} \partial_t S_1^{t}(x)\, \partial_t \rho_2^{t}(x)\, \mathrm{d}x
    - \int_{\mathcal{M}} \partial_t S_2^{t}(x)\, \partial_t \rho_1^{t}(x)\, \mathrm{d}x,
\end{equation}
where $(\partial_t \rho_1^t, \partial_t S_1^t),\, (\partial_t \rho_2^t, \partial_t S_2^t)
\in \mathcal{T}_{(\rho^t, S^t)}\!\big(\mathcal{T}^*\mathcal{P}^+(\mathcal{M})\big)$
are admissible tangent vectors. The conservation properties of this dynamics are
evident from the associated evolution equations (Appendix~\ref{app:sb_demons}), which
preserve both the kinetic energy, as the velocity field remains constant along
trajectories, and the total probability mass $\int_{\mathcal{M}} \rho^t(x)\, dx$.
These conservation properties can be relaxed by augmenting the state space with an
additional scalar variable $z^t \in \mathbb{R}$, yielding a \emph{contact dynamical
system} with extended state $(\rho^t, S^t, z^t) \in \mathcal{T}^*\mathcal{P}^+(\mathcal{M})
\times \mathbb{R}$.
The additional degree of freedom $z^t$ tracks the accumulated action of the dynamics and defines a direction transverse to $\mathcal{T}^*\mathcal{P}^+(\mathcal{M})$, foliating the contact manifold into a family of symplectic hyperplanes parametrized by the value of $z^t$.
The contact manifold $(\mathcal{T}^*\mathcal{P}^+(\mathcal{M}) \times \mathbb{R}, \eta)$ is endowed with a form $\eta$, representing a signed length, specifically the projection of the velocity $(\partial_t\rho^t, \partial_t S^t, \partial_t z^t) \in \mathcal{T}_{(\rho^t, S^t, z^t)}\big(\mathcal{T}^*\mathcal{P}^+(\mathcal{M}) \times \mathbb{R}\big)$ onto the transverse direction, i.e., the part of the dynamics that escapes the symplectic hyperplane.
The expression of $\eta$ in local coordinates depends on the choice of coordinate system.
The two most common options are the canonical coordinates, leading to $\eta_{\mathrm{std}_1}$,
and the Legendre-variant coordinates, leading to $\eta_{\mathrm{std}_2}$
\citep{bravetti2023bregman},
\begin{subequations}
\begin{align}
    \eta_{\mathrm{std}_1} &= \partial_t \tilde{z}^t
        - \int_{\mathcal{M}} \tilde{S}^t(x)\, \partial_t \tilde{\rho}^t(x)\, \mathrm{d}x, \\
    \eta_{\mathrm{std}_2} &= \partial_t z^t
        - \frac{1}{2} \int_{\mathcal{M}} S^t(x)\, \partial_t \rho^t(x)\, \mathrm{d}x
        + \frac{1}{2} \int_{\mathcal{M}} \rho^t(x)\, \partial_t S^t(x)\, \mathrm{d}x.
\end{align}
\end{subequations}
As the contact form governs the relationship between the scalar contact Hamiltonian
$\mathcal{H}$ and the associated vector field, each of the two forms gives rise to a
distinct set of evolution equations,
\begin{subequations}
\begin{align}
    \partial_t \tilde{\rho}^t &= \partial_{\tilde{S}}\, \mathcal{H}(\cdot),
    \notag \\
    \partial_t \tilde{S}^t &= -\partial_{\tilde{\rho}}\, \mathcal{H}(\cdot)
        - \tilde{S}^t\, \partial_{\tilde{z}}\, \mathcal{H}(\cdot),
    \label{eq:contact_std1} \\
    \partial_t \tilde{z}^t &= \tilde{S}^t\, \partial_{\tilde{S}}\, \mathcal{H}(\cdot)
        - \mathcal{H}(\cdot),
    \notag \\[5pt]
    \partial_t \rho^t &= \partial_{S}\, \mathcal{H}(\cdot)
        - \tfrac{1}{2}\,\rho^t\, \partial_{z}\, \mathcal{H}(\cdot),
    \notag \\
    \partial_t S^t &= -\partial_{\rho}\, \mathcal{H}(\cdot)
        - \tfrac{1}{2}\,S^t\, \partial_{z}\, \mathcal{H}(\cdot),
    \label{eq:contact_std2}\\
    \partial_t z^t &= \tfrac{1}{2}\!\left(S^t + \rho^t\,\partial_{\rho}\,\mathcal{H}(\cdot)\right)
        - \mathcal{H}(\cdot).
    \notag
\end{align}
\end{subequations}
A contact dynamics is characterized by $\eta$ being rescaled by a constant factor along its flow, a condition strictly weaker than that of a symplectic dynamics: the symplectic form $\omega$ on each hyperplane need not be preserved, allowing the contact flow to traverse hyperplanes of varying volume. As shown in Appendix~\ref{app:Contact-sb_demons}, this volume variation on the augmented cotangent bundle induces mass variation on the probability space, providing a geometrically principled framework for the uOT without complicating the metric structure of the problem.

\textbf{Geodesics on the Wasserstein Manifold.~~}
Analogously to the framework detailed in Appendix \ref{app:riemannian}, the symplectic dynamics $(\rho^t, S^t) \subset (\mathcal{T}^*\mathcal{P}^+(\mathcal{M}), \omega)$ project to a geodesic on the Wasserstein manifold $\mathcal{P}^+(\mathcal{M})$. Since the Hamiltonian function $\mathcal{H} = \int_\mathcal{M} \frac{1}{2} \Vert \nabla_x S^t(x)\Vert^2 \rho^t(x) dx$ represents the minimal constant kinetic energy $\mathcal{K}$ associated with the stochastic dynamics, the projected trajectory minimizes the Wasserstein distance, equivalent to the action functional (see Appendix \ref{app:wass_equivalence}), thereby constituting a geodesic.
In the contact Hamiltonian setting, the dynamics $(\rho^t, S^t, z^t) \subset (\mathcal{T}^*\mathcal{P}^+(\mathcal{M}) \times \mathbb{R}, \eta)$ are characterized by a contact Hamiltonian $\mathcal{H}(\rho^t, S^t, z^t)$ whose value evolves over time. This implies that the Wasserstein metric is no longer scaled uniformly. Instead, the geometry is governed by a dynamic scaling factor, resulting in a conformal transformation of the metric expressed as $\sqrt{\mathcal{K}} g^{\mathcal{W}_2}$.

\subsection{Equivalence of Wasserstein Metric Formulations}\label{app:wass_equivalence}
\begin{proposition}
Consider the dynamic formulation of OT~\citep{benamou2000computational},
\begin{equation}
\label{eq:kinetic}
J = \int_{t_0}^{t_M} \int_{\mathcal{M}} \|v^t(x)\|^2 \rho^t(x) \, dx \, dt.
\end{equation}
At optimality, this functional coincides with the Wasserstein distance between the endpoint densities.
\end{proposition}

\begin{proof}
\noindent
To interpret this functional as a distance, consider a curve $\rho^t$ connecting two probability densities. At optimality, the density evolution satisfies the continuity equation $\partial_t \rho^t + \nabla_x \cdot (\rho^t \nabla_x S^t(x))$ for some potential $S^t$, where $\nabla_x S^t(x)=v^t(x)$ represents a velocity field. Substituting this representation into~\eqref{eq:kinetic} yields a kinetic energy of the form,
\begin{equation}
\int_{t_0}^{t_M}\int_{\mathcal{M}} \|\nabla_x S^t(x)\|^2 \rho^t(x) \, dx \, dt.
\end{equation}
By focusing on the temporal integrand and leveraging the scalar nature of $\rho^t$, we can write,
\begin{equation}
\int_{\mathcal{M}} \|\nabla_x S^t(x)\|^2 \rho^t(x) \, dx
= \int_{\mathcal{M}} \nabla_x S^t(x) \cdot \big(\rho^t(x) \nabla_x S^t(x)\big) \, dx.
\end{equation}
Applying integration by parts (assuming suitable boundary conditions),
\begin{equation}
\label{eq:equivalence_passage}
\int_{\mathcal{M}} S^t(x) \left( -\nabla_x \cdot (\rho^t(x) \nabla_x S^t(x)) \right) dx
= \int_{\mathcal{M}} S^t(x) \, \Delta_{\rho} \,  S^t(x) \, dx.
\end{equation}
The continuity equation connects the density variation $\partial_t \rho^t$ and the potential $S^t$ by the weighted Laplacian operator, 
$\Delta_{\rho} = -\nabla_x \cdot (\rho^t \nabla_x)$ \citep{chow2020wasserstein}, as 
\begin{equation}
\label{eq:continuity_op}
\partial_t \rho^t(x) = \Delta_{\rho} S^t(x), 
\quad \text{equivalently} \quad 
S^t(x) = \Delta_{\rho}^\dagger \partial_t \rho^t(x).
\end{equation}
Using \eqref{eq:continuity_op} in , this becomes
\begin{equation}
g^{\mathcal{W}_2}
= \int_{\mathcal{M}} S^t(x)\, \partial_t \rho^t(x) \, dx.
\end{equation}
Substituting $S^t(x) = \Delta_{\rho}^\dagger \partial_t \rho^t(x)$ in \eqref{eq:equivalence_passage}, we obtain
\begin{equation}
g^{\mathcal{W}_2^2}(\partial_t\rho^t, \partial_t\rho^t)
= \int_{\mathcal{M}} \partial_t\rho^t(x)\, (-\Delta_{\rho})^\dagger \partial_t\rho^t(x) \, dx,
\end{equation}
which matches the standard formulation of the Wasserstein metric \eqref{eq:wasserstein_metric}. Therefore, the kinetic energy functional~\eqref{eq:kinetic} can be equivalently expressed as
\begin{equation}
J = \int_{t_0}^{t_M} g^{\mathcal{W}^2_2}(\partial_t\rho^t, \partial_t\rho^t) \, dt,
\end{equation}
which corresponds to the energy of a curve in $\mathcal{P}^+(\mathcal{M})$ endowed with the Wasserstein metric. Minimizing $J$ thus amounts to computing a geodesic.
This interpretation can be made explicit by introducing an arc-length parametrization. Defining $ds = \sqrt{\Vert v^t(x) \Vert^2\rho^t(x)} dt=\sqrt{g^{\mathcal{W}_2}(\partial_t \rho^t, \partial_t \rho^t)}\, dt$, the functional reduces to $J = \int_{s_0}^{s_M} ds$, showing that the minimization of kinetic energy is equivalent to minimizing the length of the curve.
\end{proof}
\section{Contact Unbalanced Optimal Transport}

\subsection{Contact Hamiltonian Structure of Unbalanced Optimal Transport}\label{app:Contact-sb_demons}
This part presents the derivation of the Hamiltonian structure of the Contact uOT problem~(\ref{eq:contact_schrodinger_bridge}) (Section~\ref{sec:methodology}), obtained via the method of Lagrange multipliers.
To transform the constrained optimization problem~\eqref{eq:contact_schrodinger_bridge} 
into an unconstrained one, we incorporate the generalized Fokker--Planck equation, 
scaled by the Lagrange multiplier $S^t$, into the original running cost, yielding the 
augmented Lagrangian,
\begin{align}
J(v^t, \rho^t, S^t,z^t) &= \int_{t_0}^{t_M} \int_\mathcal{M} \mathcal{L}(v^t, \rho^t, S^t,z^t) \, dx \, dt; \label{eq:contact-aug_cost_fun}\\
\mathcal{L}(v^t, \rho^t, S^t, z^t)  &=  \tfrac{1}{2} 
    \|v^t(x)\|^2 + \tfrac{1}{2}w_z {z^t(x)}^2 +  S^t(x) \underbrace{\Big( 
    \partial_t \rho^t(x) 
    + \nabla_x \cdot \big(\rho^t(x)\, v^t(x)\big)  
\Big)}_{\text{Fokker--Planck equation}},\label{eq:contact-general_field_EL}
\end{align}
dependent on the optimization variable $v^t$ and the contact state $(\rho^t, S^t, z^t)$. The problem \eqref{eq:contact_schrodinger_bridge} falls outside the scope of classical variational calculus due to the explicit dependence of the cost functional $J$ on the evolving action $z^t$. Instead, it is naturally framed within non-conservative variational principles, where $J$ evolves dynamically with the system. This setting is naturally addressed by the Herglotz variational principle, which extends the Euler--Lagrange equations to systems with dissipative effects. Applying this framework to derive the optimality conditions leads to a contact Hamiltonian dynamical system expressed in canonical coordinates \citep{bravetti2023bregman}. For our purposes, however, it is particularly convenient to employ an equivalent non-canonical representation (Appendix \ref{app:sympletcic_contact}), as it simplifies the structure of the Hamiltonian.
To carry out the Hamiltonian derivation using Herglotz-type Euler–Lagrange equations, it is therefore necessary to transition from the non-canonical state $(\rho^t, S^t, z^t)$ to the canonical one $(\tilde{\rho}^t, \tilde{S}^t, \tilde{z}^t)$, via the substitution introduced in \citep{bravetti2023bregman},\looseness-1
\begin{subequations}
\label{eq:canonical_transformation}
\begin{align}
\rho^t(x) &= \tfrac12\tilr - \tilS, \\
S^t(x) &= \tilS + \tfrac12\tilr, \\
z^t(x) &= \tilz + \tfrac18\tilr^2 - \tfrac12\tilS^2.
\end{align}
\end{subequations}
Accordingly, the Lagrangian \eqref{eq:contact-general_field_EL} transforms as follow,
\begin{equation}
\label{eq:contact_canonical_lagrangian}
\begin{split}
\mathcal{L} = &\tfrac{1}{2} \|v^t\|^2 \left( \tfrac{1}{2}\tilr - \tilS \right) 
+ \tfrac12 w_z  \left( \tilz + \tfrac{1}{8}\tilr^2 - \tfrac{1}{2}\tilS^2 \right)^2 + \left( \tilS + \tfrac{1}{2}\tilr \right)\cdot 
\\
& \left( \tfrac{1}{2}\partial_t \tilr - \partial_t \tilS + \nabla_x \cdot \left(\left( \tfrac{1}{2}\tilr - \tilS \right) \cdot v^t \right) - \varepsilon \Delta_x  \left( \tfrac{1}{2}\tilr - \tilS \right)\right)
\end{split}
\end{equation}
The optimality conditions obtained from the variations of $\mathcal{L}$, namely, the Herglotz-type Euler--Lagrange equations, for a generic field argument $\psi^t(x)$ in this case take the form,
\begin{equation}
\label{eq:herglotz_EL}
    d_\psi \mathcal{L} =\partial_\psi \mathcal{L} - \partial_t( \partial_{\partial_t \psi} \mathcal{L}) - \nabla_x \cdot (\partial_{\nabla_x \psi} \mathcal{L}) + \Delta_x (\partial_{\Delta_x} \mathcal{L}) -\partial_z \mathcal{L} \ \partial_{\partial_t \psi} \mathcal{L}=0.
\end{equation}
Applying the Herglotz-type Euler–Lagrange equations~\eqref{eq:herglotz_EL} to the Lagrangian $\mathcal{L}$ defined in equation~\eqref{eq:contact_canonical_lagrangian}, with respect to the fields $v^t$, $\tilde{\rho}^t$, and $\tilde{S}^t$, yields the following optimality conditions,
\begin{subequations}
\begin{align}
    d_v \mathcal{L} &= v\left(\tfrac12\tilr- \tilS \right) - \nabla_x \left( \tilS + \tfrac12\tilr \right)\left(\tfrac12\tilr- \tilS \right)=0 \notag \\ &\Longrightarrow v^t= \nabla_x \left( \tilS + \tfrac12\tilr \right), \\[5pt]
    d_{\tilde{\rho}} \mathcal{L} &= \tfrac{1}{4} \|v\|^2 + w_z \left( \tilz + \tfrac{1}{8}\tilr^2 - \tfrac{1}{2}\tilS^2 \right) \tfrac14 \tilr \notag \\ &+ \tfrac12 \bigg( \tfrac{1}{2}\partial_t \tilr - \partial_t \tilS
    + \nabla_x \cdot \left(\left( \tfrac{1}{2}\tilr - \tilS \right) \cdot v^t\right) - \varepsilon \left( \tfrac12 \Delta_x \tilr - \Delta_x \tilS \right)\bigg) \notag
    \\ &
    -\tfrac12 \nabla_x (\tilS + \tfrac{1}{2}\tilr) v^t- \tfrac12\partial_t \tilS - \tfrac14 \partial_t \tilr
    - \tfrac12 \varepsilon \Delta_x \tilS - \tfrac14 \varepsilon \Delta_x \tilr \notag \\ &
    - \tfrac12 w_z (\tilS + \tfrac{1}{2}\tilr) \left(\tilz + \tfrac{1}{8}\tilr^2 - \tfrac{1}{2}\tilS^2\right)=0, \\[5pt]
    \partial_{\tilde{S}} \mathcal{L}&= -\tfrac12 \Vert v\Vert^2 - w_z\left( \tilz + \tfrac{1}{8}\tilr^2 - \tfrac{1}{2}\tilS^2 \right) \tilS \tilS \notag \\ & + \left( \tfrac{1}{2}\partial_t \tilr - \partial_t \tilS
    + \nabla_x \cdot \left(\left( \tfrac{1}{2}\tilr - \tilS \right) \cdot v^t\right) - \varepsilon \left( \tfrac12 \Delta_x \tilr - \Delta_x \tilS \right) \right) \notag
    \\
    &+ \nabla_x (\tilS +\tfrac12 \tilr)v^t+ \tfrac12 \partial_t \tilr + \partial_t \tilS + \varepsilon \Delta_x \tilS + \tfrac12 \varepsilon \Delta_x \tilr
    \notag \\ &
    + w_z (\tilS + \tfrac{1}{2}\tilr) \left(\tilz + \tfrac{1}{8}\tilr^2 - \tfrac{1}{2}\tilS^2\right)=0
\end{align}
\end{subequations}
Rearranging the expression and performing the necessary algebraic simplifications yields,
\begin{subequations}
\label{eq:sum_and_diff}
\begin{align}
\partial_t \tilS 
&= \tfrac{1}{2} \nabla_x \cdot \left( \left( \tfrac{1}{2} \tilr - \tilS \right) \nabla_x \left( \tilS + \tfrac{1}{2} \tilr \right) \right) 
- \tfrac{1}{4} \left\Vert \nabla_x \left( \tilS + \tfrac{1}{2} \tilr \right) \right\Vert^2 \notag \\
&\quad - \tfrac12 \varepsilon \Delta_x \tilr 
- \tfrac{1}{2} w_z\left( \tilz + \tfrac{1}{8} \tilr^2 - \tfrac{1}{2} \tilS^2\right)\tilS. \label{eq:canonical_S} \\[5pt]
\partial_t \tilr 
&= - \nabla_x \cdot \left( \left( \tfrac{1}{2} \tilr - \tilS \right) \nabla_x \left( \tilS + \tfrac{1}{2} \tilr \right) \right) 
- \tfrac{1}{2} \left\Vert \nabla_x \left( \tilS + \tfrac{1}{2} \tilr \right) \right\Vert^2 \notag \\
&\quad - 2 \varepsilon \Delta_x \tilS 
+ \tfrac{1}{2} w_z\left( \tilz + \tfrac{1}{8} \tilr^2 - \tfrac{1}{2} \tilS^2 \right) \tilr, \label{eq:canonical_rho}
\end{align}
\end{subequations}
Taking the sum and difference of equations \eqref{eq:canonical_rho} and \eqref{eq:canonical_S}, we obtain,
\begin{subequations}
\label{eq:canonical_optimality}
\begin{align}
\tfrac{1}{2} \partial_t \tilr - \partial_t \tilS
&= -\nabla_x \cdot \left( \left( \tfrac{1}{2} \tilr - \tilS \right) \nabla_x \left( \tilS + \tfrac{1}{2} \tilr \right) \right) \notag \\ &
+ \varepsilon \Delta_x \left( \tfrac{1}{2} \tilr - \tilS \right)
\notag \\ &+ \tfrac{1}{2} w_z\!\left( \tilz + \tfrac{1}{8} \tilr^2 - \tfrac{1}{2} \tilS^2  \right) \left( \tfrac{1}{2} \tilr - \tilS \right), \\
\partial_t \tilS + \tfrac{1}{2} \partial_t \tilr
&= -\tfrac{1}{2} \left\| \nabla_x \left( \tilS + \tfrac{1}{2} \tilr \right) \right\|^2
- \varepsilon \Delta_x \left( \tilS + \tfrac{1}{2} \tilr \right)
\notag \\ &+ \tfrac{1}{2} w_z\!\left( \tilz + \tfrac{1}{8} \tilr^2 - \tfrac{1}{2} \tilS^2 \right) \left( \tilS + \tfrac{1}{2} \tilr \right).
\end{align}
\end{subequations}
At this stage, we transform equations \eqref{eq:canonical_optimality} using the inverse of the canonical coordinate transformation \eqref{eq:canonical_transformation}, namely,
\begin{subequations}
\begin{align}
\tilr &= \rho^t(x) + S^t(x), \\
\tilS &= \tfrac12(S^t(x)-\rho^t(x)), \\
\tilz &= z^t(x) -\tfrac12 \rho^t(x)S^t(x),
\end{align}
\end{subequations}
and recover the dynamical equation for the action $z^t$ \eqref{eq:CSB:z}. This yields the optimality conditions in equation \eqref{eq:contact_hamiltonian_dynamics}, expressed in non-canonical coordinates,
\begin{subequations}
\begin{align*}
\partial_t{\rho}^t(x) &= \partial_S \mathcal{H}(\cdot) - \dfrac{1}{2} \rho^t(x) \partial_z \mathcal{H}(\cdot) \notag \\ &= -\nabla_x \cdot \left(\rho^t(x)\nabla_x S^t(x)\right) + \varepsilon \Delta_x \rho^t(x) +w_z \, z^t(x)\rho^t(x),
\\
\partial_t{S}^t(x) &= -\partial_\rho \mathcal{H}(\cdot) - \dfrac{1}{2} S^t(x) \partial_z \mathcal{H}(\cdot) \notag \\ &= -\tfrac{1}{2}\|\nabla_x S^t(x)\|^2 - \varepsilon \Delta_x S^t(x) +w_z \, z^t(x)S^t(x),
\\
\partial_t{z}^t(x) &= \dfrac{1}{2} \left( S^t(x) \partial_S \mathcal{H}(\cdot) + \rho^t(x) \partial_\rho \mathcal{H}(\cdot) \right) - \mathcal{H}(\cdot) \notag \\ &= \tfrac12 \Vert \nabla_x S^t(x) \Vert^2 \rho^t(x) + \tfrac{1}{2} w_z \, z^t(x)^2 - \tfrac12\Vert v_0\Vert^2 \rho^t(x),
\end{align*}
\end{subequations}
This defines a contact Hamiltonian system associated with the following contact Hamiltonian function,
\begin{equation}
\begin{split}
\mathcal{H}(\rho^t, S^t, z^t) = 
    &{\int_\mathcal{M} \tfrac{1}{2} \|\nabla_x S^t(x)\|^2  \rho^t(x)\,dx}
    + {\varepsilon \int_\mathcal{M} S^t(x)\,\Delta_x \rho^t(x)\,dx} \\ &- w_z \int_\mathcal{M} \tfrac12 z^t(x)^2 \,dx - {\int_\mathcal{M} \tfrac{1}{2} \|\nabla_x S^t(x)\|^0  \rho^0(x)\,dx}.
\end{split}
\label{eq:contact_hamiltonian}
\end{equation}
Here, the first term represents the kinetic energy of the flow, while the second term
introduces an entropic coupling scaled by the diffusion coefficient $\varepsilon$. The
latter modifies the drift of the transport map so as to maintain a level of entropy
consistent with the reference process in the Fokker--Planck equation \eqref{eq:CSB:FP}. The third term constitutes the damping factor, responsible for drawing or increasing energy. As direct consequences, this have the introduction of a damping or accelerating term in the Hamilton-Jacobi-Bellman equation \eqref{eq:contact_S}, and a source or sink term in the Fokker-Planck equation \eqref{eq:contact_rho}. The last term represents a constant fixing the reference energy level.
\section{Geometric Analysis of State-of-the-Art Methods}\label{app:geometry_state-of-the-art}

\subsection{Hamiltonian Structure of the  Optimal Transport}\label{app:sb_demons}
This part presents the derivation of the Hamiltonian structure of the entropy-regularized OT problem~(\ref{eq:schrodinger_bridge}) (Section~\ref{sec:background}), obtained via the method of Lagrange multipliers \citep{testa2026contact}.
To transform the constrained optimization problem into an unconstrained one, we incorporate the Fokker--Planck equation, scaled by the Lagrange multiplier $S^t$, into the original running cost $\mathcal{L}$ (i.e., the Lagrangian), as follows,
\begin{align}
    J(v^t, \rho^t, S^t) =& \int_{t_0}^{t_M} \mathcal{L}(v^t, \rho^t, S^t) \, dt; \label{eq:aug_cost_fun} \\
    \mathcal{L}(v^t, \rho^t, S^t) =& \int_\mathcal{M} \frac{1}{2} \Vert v^t (x)\Vert^2 \rho^t(x) \, dx \nonumber \\ 
    &+ \int_\mathcal{M} S^t(x)\underbrace{\left( \partial_t \rho^t(x) + \nabla_x \cdot (\rho^t(x) v^t(x)) - \varepsilon \Delta \rho^t(x) \right)}_{\text{Fokker-Planck equation}} \, dx. \label{eq:gen_lagrangian}
\end{align}
The optimality conditions resulting from the extremization of the cost functional $J$ in equation~(\ref{eq:aug_cost_fun}) follow from the Euler--Lagrange equations, generalized to the setting of classical field theory~\citep{blohmann2023lagrangian}. In this framework, the arguments of the Lagrangian $\mathcal{L}$, in equation~(\ref{eq:gen_lagrangian}), are viewed as smooth fields defined over space and time. By setting to zero the variations of $\mathcal{L}$ with respect to these fields, we obtain the stationarity conditions for $J$. For a generic field $\psi^t(x)$, the corresponding Euler--Lagrange equation takes the form, \looseness-1
\begin{equation}
\label{eq:general_field_EL}
    d_\psi \mathcal{L} =\partial_\psi \mathcal{L} - \partial_t( \partial_{\partial_t \psi} \mathcal{L}) - \nabla_x \cdot (\partial_{\nabla_x \psi} \mathcal{L}) + \Delta_x (\partial_{\Delta_x} \mathcal{L}) =0.
\end{equation}
Applying equation~(\ref{eq:general_field_EL}) to equation~(\ref{eq:gen_lagrangian}) for the fields $v^t$, $\rho^t$, and $S^t$, we obtain the following system of optimality conditions,
\begin{subequations}
\label{eq:gen_optimality_conditions}
\begin{align}
&d_v \mathcal{L} = v^t(x) \rho^t(x) - \rho^t(x) \nabla_x S^t(x) = 0 \ \  \Longrightarrow \ \ v^t(x) = \nabla_x S^t(x), \label{eq:demons01} \\
&d_\rho \mathcal{L} = \frac{1}{2} \Vert v^t(x)\Vert^2 - \partial_t S^t(x) - \nabla_x S^t(x) \cdot v^t(x) - \varepsilon \Delta_x S^t(x) = 0, \label{eq:demons02} \\
&d_S \mathcal{L} = \partial_t \rho^t(x) + \nabla_x \cdot (\rho^t(x) v(x)) - \varepsilon \Delta_x \rho^t(x) = 0. \label{eq:demons03}
\end{align}
\end{subequations}
Substituting the expression for the optimal velocity from equation~(\ref{eq:demons01}) into equations (\ref{eq:demons02}) and (\ref{eq:demons03}), we obtain the following Hamiltonian system,
\begin{subequations}
\label{eq:ge_nonlinear_hamiltonian_dynamics}
\begin{align}
&\partial_t \rho^t(x) = \partial_S \mathcal{H}(\cdot) = - \nabla_x \cdot (\rho^t(x) \nabla_x S^t(x)) + \varepsilon \Delta \rho^t(x), \\
&\partial_t S^t(x) = -\partial_\rho \mathcal{H}(\cdot) = -\frac{1}{2} \Vert \nabla_x S^t(x)\Vert^2 - \varepsilon \Delta_x S^t(x),
\end{align}
\end{subequations}
with the corresponding Hamiltonian function,
\begin{equation}
\label{eq:sb_hamiltonian}
\mathcal{H}(\rho^t, S^t) = \frac{1}{2} \int_\mathcal{M} \Vert \nabla_x S^t(x) \Vert^2 \rho^t(x) \, dx + \varepsilon \int_\mathcal{M} S^t(x) \Delta_x \rho^t(x) \, dx.
\end{equation}
Here, the first term represents the kinetic energy of the flow, while the second term
introduces an entropic coupling scaled by the diffusion coefficient $\varepsilon$. The
latter modifies the drift of the transport map so as to maintain a level of entropy
consistent with the Brownian motion used as the reference process. In the limiting case
$\varepsilon \to 0$, the entropic correction vanishes and one recovers a classical OT
plan, wherein the minimization of the Hamiltonian kinetic energy on
$\mathcal{T}^*\mathcal{P}^+(\mathcal{M})$ is equivalent to the computation of a geodesic
on the Wasserstein manifold $\mathcal{P}^+(\mathcal{M})$, with the kinetic energy
coinciding with the Wasserstein metric $g^{\mathcal{W}_2}$ defined in
equation~\eqref{eq:wasserstein_metric} (Appendix~\ref{app:wass_equivalence}). In the
general case $\varepsilon > 0$, the Hamiltonian governs the minimization of an
entropy-regularized transport cost, whose solution is equivalent to that obtained via
the Sinkhorn algorithm \citep{feydy2019interpolating}.

\subsection{Hamiltonian Structure of the Fisher-Rao Unbalanced Optimal Transport}\label{app:FR-sb_demons}
This part presents the derivation of the Hamiltonian structure of the Fisher--Rao uOT problem~(\ref{eq:fisher-rao-schrodinger_bridge}) (Section~\ref{sec:background}), obtained via the method of Lagrange multipliers.
To transform the constrained optimization problem~\eqref{eq:fisher-rao-schrodinger_bridge} 
into an unconstrained one, we incorporate the generalized Fokker--Planck equation, 
scaled by the Lagrange multiplier $S^t$, into the original running cost, yielding the 
augmented Lagrangian,
\begin{align}
J(v^t, z^t, \rho^t, S^t) &= \int_{t_0}^{t_M} \mathcal{L}(v^t, z^t, \rho^t, S^t) \, dt; \label{eq:FR-aug_cost_fun}\\
\mathcal{L}(v^t, z^t, \rho^t, S^t) &= \int_\mathcal{M} \frac{1}{2} 
    \Big( \|v^t(x)\|^2 + w_z \,{z^t(x)}^2 \Big) \rho^t(x) \, dx \nonumber \\
&\quad + \int_\mathcal{M} S^t(x) \underbrace{\Big( 
    \partial_t \rho^t(x) 
    + \nabla_x \cdot \big(\rho^t(x)\, v^t(x)\big) 
    - \varepsilon \Delta_x \rho^t(x) 
    - z^t(x)\rho^t(x) 
\Big)}_{\text{Generalized Fokker--Planck equation}} dx.\label{eq:FR-general_field_EL}
\end{align}

Applying the Euler--Lagrange equations~\eqref{eq:FR-general_field_EL} to 
$\mathcal{L}$ with respect to the fields $v^t$, $z^t$, $\rho^t$, and $S^t$, 
we obtain the following optimality conditions,
\begin{subequations}
\begin{align}
d_v \mathcal{L} &= v^t(x)\,\rho^t(x) - \rho^t(x)\,\nabla_x S^t(x) = 0 
    \quad \Longrightarrow \quad 
    v^t(x) = \nabla_x S^t(x), \label{eq:FR_demons01} \\
d_g \mathcal{L} &= w_z\, z^t(x)\,\rho^t(x) - S^t(x)\,\rho^t(x) = 0 
    \quad \Longrightarrow \quad 
    z^t(x) = \frac{S^t(x)}{w_z}, \label{eq:FR_demons02} \\
d_\rho \mathcal{L} &= \frac{1}{2}\|v^t(x)\|^2 + \frac{w_z}{2}{z^t(x)}^2 
    - \partial_t S^t(x) - \nabla_x S^t(x) \cdot v^t(x) 
    - \varepsilon \Delta_x S^t(x) - z^t(x) S^t(x) = 0, \label{eq:FR_demons03}\\
d_S \mathcal{L} &= \partial_t \rho^t(x) 
    + \nabla_x \cdot \big(\rho^t(x)\, v^t(x)\big) 
    - \varepsilon \Delta_x \rho^t(x) 
    - z^t(x)\rho^t(x) = 0. \label{eq:FR_demons04}
\end{align}
\end{subequations}
Substituting the optimal velocity~\eqref{eq:FR_demons01} and 
growth rate~\eqref{eq:FR_demons02} into equations~\eqref{eq:FR_demons03} 
and~\eqref{eq:FR_demons04}, we obtain the Hamiltonian system,
\begin{subequations}
\label{eq:FR_hamiltonian_dynamics}
\begin{align}
\partial_t \rho^t(x) &= \phantom{-}\partial_{S} \mathcal{H}(\cdot) 
    = -\nabla_x \cdot \big(\rho^t(x)\,\nabla_x S^t(x)\big) 
    + \varepsilon \Delta_x \rho^t(x) 
    + \frac{S^t(x)}{w_z}\rho^t(x), \label{eq:FR-hamiltonain_dyn-rho}\\
\partial_t S^t(x) &= -\partial_{\rho} \mathcal{H}(\cdot) 
    = -\frac{1}{2}\|\nabla_x S^t(x)\|^2 
    - \frac{1}{2w_z}{S^t(x)}^2 
    - \varepsilon \Delta_x S^t(x),
    \label{eq:FR-hamiltonain_dyn-S}
\end{align}
\end{subequations}
governed by the Hamiltonian function,
\begin{equation}
\label{eq:FR_hamiltonian}
\mathcal{H}(\rho^t, S^t) = 
    {\frac{1}{2} \int_\mathcal{M} \left( \|\nabla_x S^t(x)\|^2 + \frac{1}{w_z}{S^t(x)}^2\right) \rho^t(x)\,dx}
    + {\varepsilon \int_\mathcal{M} S^t(x)\,\Delta_x \rho^t(x)\,dx}.
\end{equation}
Compared to the OT Hamiltonian~\eqref{eq:sb_hamiltonian}, the Fisher--Rao unbalanced formulation introduces an explicit dependence on $S^t$ in the kinetic energy. This dependence arises from the deformation of the underlying conic space induced by the growth rate $S^t$.
Despite this structural complication, the quadratic Fisher--Rao penalty, $\psi(z^t(x)) = \bigl(z^t(x)\bigr)^2$, remains the most tractable choice within the class of $\psi$-divergences. Indeed, it is the only choice that preserves a polynomial structure of the Hamiltonian. By contrast, a generic convex $\psi$-divergence induces a transcendental dependence on $S^t$, thereby significantly increasing the analytical complexity of the Hamiltonian and hindering the explicit computation of geodesics~\citep{kondratyev2016new}.
In this setting, distance-based regularization cannot be recovered, requiring Hamilton-Jacobi-Bellman regularization~\citep{buzun2026hota}, whose PDE-based formulation lacks comparable optimization stability (see Appendix~\ref{app:stability}).\looseness-1

The dynamical equations~\eqref{eq:FR_hamiltonian_dynamics} describe a symplectic flow on $\mathcal{T}^*\mathcal{P}^+(\mathcal{M})$, corresponding to a geodesic with respect to the WFR metric. Indeed, as in problems~\eqref{eq:contact_schrodinger_bridge} and~\eqref{eq:schrodinger_bridge}, $\mathcal{K}$ defines a proper geometric metric that is linear in $\rho^t$, allowing the WFR geodesic length of $\rho^s \subset \mathcal{P}^+(\mathcal{M})$ to be computed as an expectation over samples ${x_n^{s_k}}_{n=0}^N$~\citep{neklyudov2023computational}.
In the limit $w_z \!\to\! \infty$, mass variation is suppressed and this formulation recovers the classical OT~\eqref{eq:schrodinger_bridge}, constituting a principled extension thereof.
However, the optimality conditions $v^t(x) = \nabla_x S^t(x)$ and $z^t(x) = S^t(x)$ drives the transport dynamics to regions of maximal mass growth. This structural bias leads to mode collapse by discouraging the reconstruction of small target modes~\citep{chapel2021unbalanced, zhang2025learning}, where sample weights must naturally decrease.
This bias is also unstable as it regulates the transport velocity w.r.t. the gradient of the weight growth, often noisy in real datasets.
This limitation is baked into the symplectic manifold $(\mathcal{T}^*\mathcal{P}^+(\mathcal{M}), \omega)$, which enforcing volume-preserving dynamics, can only accommodate mass variation by rigidly coupling $z^t$ and $v^t$ within a fixed bound.
Notably, \ac{CUOT} overcomes the symplectic structural limitations by exploiting contact geometry to decouple transport from mass growth, offering a principled framework for more accurate, controlled adaptation to mass changes.\looseness-1

\subsection{Contact Hamiltonian Structure of the Non-Conservative Optimal Transport}
\label{app:non-conservative_OT}
Our formulation is inspired by \citet{testa2026contact}, who reformulated the conservative Hamiltonian OT problem~\eqref{eq:schrodinger_bridge} via a contact framework to let energy vary during transport, yielding more flexible maps with non-constant velocity $v^t$ that better capture real-world processes involving energy dissipation and external interactions,
\begin{subequations}
\label{eq:contact_schrödinger_bridge_iclr}
\begin{align}
    \min_{v^t} \ &J(v^t, \rho^t)=\int_0^1  \partial_t z^t \, dt; \label{eq:CSB_iclr:J}\\ \text{s.t.} \ \ & \partial_t z^t = \int_\mathcal{M} \frac{1}{2} \Vert v^t(x) \Vert^2 \rho^t(x) \, dx - \gamma z^t;
    \label{eq:CSB_iclr:z}\\ &\partial_t\rho^t(x) +  \nabla_x \cdot \big(\rho^t(x)\,v^t(x)\big) = \varepsilon\Delta_x \rho^t(x); \label{eq:CSB_iclr:FP}\\ &\rho^{t_m} = \rho_{m}, \, \forall m \in \{ 0, \dots, M \}. \label{eq:CSB_iclr:BC}
\end{align}
\end{subequations}
Here, $z^t$ represents a scalar Lagrangian action and $\gamma \in \mathbb{R}$ is a constant damping factor.
Our problem~\eqref{eq:contact_schrodinger_bridge} extends their framework in two fundamental ways by introducing (i) a spatially varying action field through the quadratic term $\frac{1}{2} z^t(x)^2$ in equation~\eqref{eq:contact_z}, instead of the scalar linear action $z^t$ in~\eqref{eq:CSB_iclr:z}, and (ii) a corresponding source/sink term $w_z \rho^t(x) z^t(x)$ in equation~\eqref{eq:contact_rho}, which is missing in~\eqref{eq:CSB_iclr:FP}.
These differences reflect a different interpretation of the contact action. In \citet{testa2026contact}, $z^t$ is a time-dependent, non-learnable scalar obtained by integrating the dynamics generated by the learned velocity field. Through the relation~\eqref{eq:CSB_iclr:z}, it acts as a global damping or amplification variable that modulates the energy of the dynamics according to a prescribed coefficient $\gamma$, thereby allowing non-conservative dynamics with a predefined energy trend.
In contrast, we reinterpret $z^t(x)$ as a local mass growth-rate field learned from the data, allowing the energy of the transport map to adapt to the geometry of the space. To this end, we introduce a quadratic term $\frac{1}{2} z^t(x)^2$ to obtain a source/sink term $w_z \rho^t(x) z^t(x)$ in equation~\eqref{eq:contact_rho} via the contact Hamiltonian equations, ensuring a theoretically consistent evolution of the density. Consequently, while their approach is restricted to balanced OT problems, our method extends to unbalanced OT and adapts the velocity field $v^t$ to the local data density,refining the integration step size in high-density regions rather than relying on an a priori schedule for $v^t$ based on a prescribed damping coefficient $\gamma$.

Moreover, the formal derivation of problem~\eqref{eq:contact_schrodinger_bridge} diverges from that of \citet{testa2026contact} by employing a different basis on the contact manifold (the Legendre-variant coordinates). A discussion of this alternative representation, alongside the formal mathematical proofs for our derivation, is provided in Appendices \ref{app:geometry_probability_space} and \ref{app:Contact-sb_demons}.
\section{Implementation Details}
\label{app:implementation}

\subsection{Architecture}
The transport and growth networks $v_\theta$ and $z_\theta$ share a common backbone: a time-conditioned multilayer perceptron (MLP) with a sinusoidal Fourier feature encoding of the input. Concretely, given a particle position $x^{s_k}_n \in \mathbb{R}^d$ and a scalar variable $s_k \in [0, 1]$, the concatenated input $[x^{s_k}_n, s_k] \in \mathbb{R}^{d+1}$ is lifted to a higher-dimensional representation via a multi-scale Fourier encoding,
\begin{equation}
\begin{split}
\phi([x^{s_k}_n, s_k]) = \bigl[[x^{s_k}_n,\, s_k,\, \sin(2^0 [x^{s_k}_n, s_k]),\, \cos(2^0 [x^{s_k}_n, s_k]),\, \ldots &,\\ 
    \sin(2^{F-1} [x^{s_k}_n, s_k]),\, \cos(2^{F-1} [x^{s_k}_n, s_k]&)\bigr],
\end{split}
\end{equation}
where $F$ denotes the number of frequency bands. The encoded input $\phi([x^{s_k}_n, s_k])$ is then processed by a shared trunk consisting of $L$ hidden layers of width $H$, with \texttt{SiLU} activations and \texttt{LayerNorm} normalization applied after each hidden layer to prevent mode collapse in the learned vector fields. 
The trunk output is then passed to two separate linear heads: a velocity head producing $v_\theta(x^{s_k}_n, s_k) \in \mathbb{R}^d$, and a growth head producing $z_\theta(x^{s_k}_n, s_k) \in \mathbb{R}$. The full architecture is 
summarized as
\begin{equation}
    h = \mathrm{Trunk}\bigl(\phi([x^{s_k}_n, s_k])\bigr), \qquad
    v_\theta(x^{s_k}_n, s_k) = W_v h, \qquad
    z_\theta(x^{s_k}_n, s_k) = W_g h,
\end{equation}
where $W_v \in \mathbb{R}^{d \times H}$ and $W_g \in \mathbb{R}^{1 \times H}$ are the learned head weights. The number of trunk layers $L$, hidden width $H$, and number of frequency bands $F$ are treated as hyperparameters and tuned per dataset (see Table~\ref{tab:hyperparameters}).

\subsection{Discrete optimal equations}
Given an estimated discrete geodesic $\{(x_n^{s_k}, w_n^{s_k})\}_{k=0}^K$,
the contact optimality conditions~\eqref{eq:contact_hamiltonian_dynamics} can be deterministically discretized by evaluating all fields at the current sample positions. 
Since all distributional distances in our solver are computed with entropy-regularized Sinkhorn divergences, the entropy contribution associated with the diffusion term is already incorporated by the distance evaluation. Consequently, in the discretized dynamics we consider the limit $\varepsilon \to 0$.
Moreover, the solver integrates the dynamics along the particle trajectories, so each transition $k \to k+1$ advances the trajectory in time and updates the particle position. Therefore, all derivatives appearing in equations~\eqref{eq:contact_hamiltonian_dynamics} are evaluated along the same discrete path using forward finite differences, $Df=(f_n^{s_{k+1}} - f_n^{s_{k}})/{\tau}$, where $D$ denotes the corresponding derivative along the trajectory (time derivative for scalar fields and directional derivative for spatial quantities).
Under this trajectory-based discretization, equations~\eqref{eq:contact_S} and~\eqref{eq:contact_z} directly yield equations~\eqref{eq:discrete_S} and~\eqref{eq:discrete_z}, respectively.
Equation~\eqref{eq:discrete_nu} is then obtained by taking the spatial derivative of equation~\eqref{eq:contact_S} along the particle trajectory and using,
\begin{equation}
\frac{1}{2}\nabla_x \|\nabla_x S^t(x)\|^2 = \frac{1}{2}\nabla_x \|v^t(x)\|^2 = v^t(x)\nabla_x v^t(x) \approx \nu_n^{s_k} \frac{\nu_n^{s_{k+1}} - \nu_n^{s_k}}{\tau}.
\end{equation}
Crucially, the resulting discrete equations~\eqref{eq:discrete_optimality} depend exclusively on the growth network $z_\theta$: given the current estimate of $z_\theta$, they can be integrated forward along the geodesic to obtain the contact Wasserstein metric $g^\mathcal{C} = \sqrt{\mathcal{K}}\, g^{\mathcal{W}_2}$. This metric then governs the correction of the geodesic estimate via the transport network $v_\theta$, through the arc-length minimization objective~\eqref{eq:loss-opt}. Consequently, $z_\theta$ and $v_\theta$ are trained in a mutually consistent manner: $z_\theta$ determines the geometry against which $v_\theta$ is optimized, while $v_\theta$ updates the particle trajectories along which $z_\theta$ is subsequently re-evaluated, tightening the contact Hamiltonian constraints at each training step.
The parameter $w_z$ must be negative to ensure that $z_n^{s_k}$ decays toward zero over the optimization interval $[t_0, t_M]$, thereby regularizing the learned initial condition $z_\theta(x^{s_0}, s_0)$ and reducing its magnitude irrespective of the sign of the growth rate. This mechanism governs the temporal evolution of $z_\theta(x^{s_k}, s_k)$ without imposing spatial constraints. Consequently, since the growth rate is only relevant where the distribution has support, the optimizer can initialize $z_\theta$ so that, under the induced dynamics, it attains suitable values when the distribution reaches those regions.
The discrete geodesic $\{(x_n^{s_k}, w_n^{s_k})\}_{k=0}^K$ is computed via explicit Euler integration of the learned vector fields over $K$ uniform steps of step size $\tau = (s_1 - s_0) / K$. The full particle trajectory 
$\{x_n^{s_k}\}_{k=0}^K$ and weight trajectory $\{w_n^{s_k}\}_{k=0}^K$ are stored at every integration step and used to evaluate the loss~\eqref{eq:loss}.
At each training iteration, the optimality variables $\{(\nu_n^{s_k}, z_n^{s_k})\}_{k=0}^K$ are computed from the current growth network $z_\theta$ by forward integration of the discretized contact Hamiltonian dynamics~\eqref{eq:discrete_optimality}. The variables are initialized as $\nu_n^{s_0} = 1$ and $S_n^{s_0} = 0$ for all particles.

\subsection{Training}
\begin{algorithm}[t!]
\caption{Training procedure for the CuOT method.}
\label{alg:training}
\begin{algorithmic}[1]
\Require Marginal samples $\{x^m_n\}_{n=1}^{N_m} \sim \rho^{m}$ for $m = 0, \ldots, M$; 
         integration step $\tau$.
\Ensure  Trained transport and growth networks $v_\theta$, $z_\theta$.
\State   Initialize network parameters $\theta$ for $v_\theta$ and $z_\theta$.
\For{each training epoch}
    \For{each mini-batch}
        \State Sample a batch $\{x^0_n\}_{n=1}^{N} \sim \rho^{0}$ from the initial marginal.
        \State Propagate the discrete geodesic $\{(x_n^{s_k}, w_n^{s_k})\}_{k=0}^K$ 
               forward via~\eqref{eq:discrete_geodesic}.
        \State Compute the optimality variables 
               $\{(\nu_n^{s_k}, z_n^{s_k})\}_{k=0}^K$ 
               from the current $z_\theta$ via~\eqref{eq:discrete_optimality}.
        \State Sample reference batches $\{x^m_n\}_{n=1}^{N_m} \sim \rho^{m}$ 
               for $m = 1, \ldots, M$, and evaluate the loss~\eqref{eq:loss}.
        \State Update $\theta$ via gradient descent on~\eqref{eq:loss}.
    \EndFor
\EndFor
\end{algorithmic}
\end{algorithm}
\paragraph{Optimization.}
The training procedure is summarized in Algorithm~\ref{alg:training}.
All models are trained using the Adam optimizer with weight decay $10^{-4}$ and automatic mixed precision enabled for memory efficiency on GPU. Gradient norms are clipped to a maximum of $1.0$ at each update step to stabilize training. At each iteration, a batch of $B \times N_b$ particles is pre-sampled from the full dataset at the start of each epoch and partitioned into $N_b$ mini-batches of size $B$, each 
processed as an independent gradient step. All particles are initialized with unit mass $w_n^{s_0}=1$, and the relative total mass at each observed marginal $\rho^m$ is computed from the empirical cell counts, providing the reference mass targets for the feasibility loss.

\paragraph{Loss function.}
The total training objective combines an optimality term and a feasibility term \eqref{eq:loss}, where $w_{\text{opt}}$ and $w_{\text{fea}}$ are scalar weights balancing the two objectives (see Table~\ref{tab:hyperparameters}). The optimality term enforces minimization of the contact Wasserstein arc-length. It is computed as the sum of a geodesic energy term and a mass regularity term as shown in equation \eqref{eq:loss-opt},
\begin{equation*}
\ell_\text{opt} = \sum_{k=0}^{K-1} \sum_{n=0}^{N-1} \bigg({
({\nu_n^k})^2w_n^{s_k}\,
\Vert x_n^{s_{k+1}}-x_n^{s_{k}}\Vert^2 
}+ {\Vert z_n^{s_{k+1}}- z_\theta (x^{s_k}_n, s_{k+1})\Vert^2}\bigg).
\end{equation*}
The geodesic energy $({\nu_n^k})^2w_n^{s_k}\,
\Vert x_n^{s_{k+1}}-x_n^{s_{k}}\Vert^2 
$ computed as $(\overline{\nu}_n^{s_k})^2 w_n^{s_k} \, (\overline{d}^{s_k}_n)^2$, with $\overline{\nu}_n^{s_k}$ and $\overline{d}^{s_k}_n$ denoting the column-normalized scalar speeds and squared particle displacements at step $k$, respectively. The column-wise normalization
\begin{equation}
    \bar{u}_{k} = u_k \cdot \frac{K}{\sum_{k'} u_{k'} + \varepsilon}
\end{equation}
is applied to both $\nu_n^{s_k}$ and $d_n^{s_k}$ before multiplication, ensuring that the energy is scale-invariant across integration steps and robust to vanishing gradients. The product $(\overline{\nu}_n^{s_k})^2 w_n^{s_k} \, (\overline{d}^{s_k}_n)^2$ implements the conformal scaling of the displacement by the contact Wasserstein metric $g^{\mathcal{C}} = \sqrt{\mathcal{K}}\, g^{\mathcal{W}_2}$: steps with higher kinetic energy (large $\nu^{s_k}_n$) are penalized more heavily, aligning the learned velocity magnitude with the optimality condition for the given growth rate $z_\theta$.

The feasibility term enforces consistency of the predicted trajectory with the observed marginals $\{\rho^m\}_{m=0}^M$ at the designated integration steps $\{s_{k_m}\}$. The loss \eqref{eq:loss-fea} comprises two components,
\begin{equation*}
\ell_\text{fea} =\sum_{m=1}^{M} 
{
\left(d^{\mathcal{W}_2}(w^{s_m},x^{s_m},w^m,x^m)
+ w_{\text{mass}}\Vert W^{s_m}-W^m\Vert^2 \right)
}.
\end{equation*}
where the weighted Wasserstein distance $d^{\mathcal{W}_2}$ is practically computed as the Sinkhorn approximation between the predicted weighted ensemble $\{(x_n^{s_{k_m}}, \tilde{w}_n^{s_{k_m}})\}$ and the reference marginal $\rho^m$, computed via the \texttt{geomloss} library~\citep{feydy2019interpolating}. Here $\tilde{w}_n^{s_{k_m}} = w_n^{s_{k_m}} / \sum_n w_n^{s_{k_m}}$ denotes the normalized predicted weights used as a discrete probability measure. The mass term $\Vert W^{s_m}-W^m\Vert^2$ penalizes the deviation of the predicted total mass $W^{s_m}$ from the reference mass $W^m$. At the integration steps corresponding to observed marginals, the particle positions are detached from the computational graph, so that the geodesic is constrained to interpolate the observed marginals as fixed waypoints without back-propagating through them.

This loss formulation is consistent with our uOT baselines \citep{zhang2025learning, sun2025variational}, which also combine forward simulation with feasibility and optimality objectives. The key difference lies in the enforcement of the optimality conditions. Existing methods regularize the learned dynamics by minimizing Monte Carlo estimates of the Hamilton-Jacobi-Bellman residual, introducing additional objectives whose optimization may become unstable \citep{buzun2026hota}. In contrast, our formulation discretizes and simulates the optimality system itself. Since the learned variables have a direct contact-geometric interpretation, they can be evolved jointly with the transport dynamics and their corresponding action can be evaluated directly along the trajectory. As a result, the loss retains the variational interpretation of the continuous Contact-uOT problem while avoiding residual minimization. The empirical consequences of this design choice are discussed in the following stability analysis. \looseness-1

\subsection{Stability Considerations}\label{app:stability}

The CuOT solver discretizes the geodesic flow $\{(x_n^{s_k}, w_n^{s_k})\}_{k=0}^K$ on the 
Wasserstein manifold $\mathcal{P}^+(\mathcal{M})$ and can be interpreted as a shooting direct method~\citep{betts1998survey} for the optimal control problem of finding parameters $\theta^*$ minimizing the loss $\ell = w_{\text{opt}}\,\ell_{\text{opt}} + w_{\text{fea}}\,\ell_{\text{fea}}$ \eqref{eq:loss}. We show that each gradient update does not destabilize training and converges toward a local minimum.

\begin{proposition}
The gradient $\nabla_\theta \ell_{\emph{fea}}$ is Lipschitz-continuous in $\theta$.
\end{proposition}
\begin{proof}
The feasibility loss $\ell_{\text{fea}}$ comprises two terms,

\emph{(1) Wasserstein term.} The weighted Wasserstein distance $d^{\mathcal{W}_2}$ is computed via the Sinkhorn approximation, whose solution corresponds to the optimal transport plan $\pi^m_{il}(\theta)$ between the predicted ensemble $\{(x_n^{s_{k_m}}, {w}_n^{s_{k_m}})\}$ and the reference marginal $\{(x_n^{m}, {w}_n^{m})\}$. 
Since $\pi^m_{il}(\theta)$ solves a linear program with feasibility constraint $\sum_{il} \pi^m_{il} = 1$, and its dependence on $\theta$ enters through the predicted positions and weights, which are Lipschitz-continuous functions of $\theta$ by 
construction, the transport plan satisfies,
\begin{equation}
    \left\Vert \pi^m_{il}(\theta') - \pi^m_{il}(\theta) \right\Vert \leq L_\pi 
    \left\Vert \theta' - \theta \right\Vert,
\end{equation}
for a scalar constant $L_\pi > 0$. Consequently, $\nabla_\theta \pi^m_{il}(\theta)$ is Lipschitz-continuous.\looseness-1

\emph{(2) Mass term.} The total mass $W^{s_m} = \sum_n w_n^{s_m}$ depends nonlinearly on $\theta$ through the growth network $z_\theta$, so the map $\theta \mapsto W^{s_m}$ is in general non-convex. However, the squared mass deviation $\Vert W^{s_m} - 
W^m\Vert^2$ is locally convex in a neighborhood of any local minimizer $\theta^*$, since the Hessian $\nabla^2_\theta \Vert W^{s_m} - W^m \Vert^2 \succeq 0$ in that neighborhood. Therefore $\nabla_\theta \Vert W^{s_m} - W^m \Vert^2$ is locally Lipschitz-continuous.\looseness-1

Combining (1) and (2), $\nabla_\theta \ell_{\text{fea}}$ is locally 
Lipschitz-continuous in $\theta$.
\end{proof}

\begin{proposition}
The gradient $\nabla_\theta \ell_{\emph{opt}}$ is locally Lipschitz-continuous in $\theta$ under mild smoothness assumptions on the learned fields.
\end{proposition}

\begin{proof}
We show that $\nabla_{\theta} \ell_{\text{opt}}$ is locally Lipschitz continuous by establishing this property for the dynamical variables $\nu_{n}^{s_k}$, $S_{n}^{s_k}$, and $z_{n}^{s_k}$ via induction on $k$.

The initial conditions $\nu_{n}^{s_0}$, $S_{n}^{s_0}$, and $z_{n}^{s_0}$ are fixed and independent of $\theta$. Consequently, their gradients with respect to $\theta$ are identically zero and trivially locally Lipschitz. The growth network $z_{\theta}(x_{n}^{s_k}, s_k)$ employs smooth activations, therefore, both $z_{\theta}$ and $\nabla_{\theta} z_{\theta}$ are locally Lipschitz. The update equations~\eqref{eq:discrete_optimality} are rational functions of $z_{\theta}$ and the step-$k$ variables. Since rational operations preserve local Lipschitz continuity, the gradients of the updated variables $(\nu_{n}^{s_{1}}, S_{n}^{s_{1}}, z_{n}^{s_{1}})$ are also locally Lipschitz. Via induction, the gradients of $(\nu_{n}^{s_{k}}, S_{n}^{s_{k}}, z_{n}^{s_{k}})$ are locally Lipschitz for all $k$.\looseness-1

The loss $\ell_{\text{opt}}$ is a finite sum of products involving these dynamical variables and the squared displacements $\|x_{n}^{s_{k+1}} - x_{n}^{s_k}\|^2$. The particle positions depend on $\theta$ through the velocity field $v_{\theta}$, which is locally Lipschitz. Since finite sums and products of locally Lipschitz functions remain locally Lipschitz, it follows that $\nabla_{\theta} \ell_{\text{opt}}$ is locally Lipschitz continuous.
\end{proof}

The parameter update rule
\begin{equation}
    \theta_{(j+1)} = \theta{(j)} - \alpha\,\nabla_\theta \ell(\theta_{(j)}),
\end{equation}
with step size $\alpha > 0$, is a gradient descent step on the total loss $\ell$. Since $\nabla_\theta \ell = w_{\text{opt}}\,\nabla_\theta \ell_{\text{opt}} + w_{\text{fea}}\,\nabla_\theta \ell_{\text{fea}}$ is locally Lipschitz-continuous by Propositions 1 and 2, and the loss is bounded below by zero, the gradient descent iterates $\{\theta_{(j)}\}_{j \geq 0}$ satisfy the sufficient decrease condition
\begin{equation}
    \ell(\theta_{(j+1)}) \leq \ell(\theta_{(j)}) - 
    \frac{\alpha}{2}\left\Vert\nabla_\theta \ell(\theta_{(j)})\right\Vert^2,
\end{equation}
for sufficiently small $\alpha$. Each update therefore strictly decreases the loss 
unless $\nabla_\theta \ell = 0$, guaranteeing convergence to a stationary point 
$\theta^*$. Geometrically, $\theta^*$ corresponds to a local coordinate chart of the 
tangent bundle $\mathcal{T}\mathcal{P}^+(\mathcal{M})$ that is consistent with the 
observed marginal constraints $\{\rho^m\}_{m=0}^M$.

\paragraph{Stability of shooting versus HJB residual minimization.}
An alternative approach to solving the optimal control problem~\eqref{eq:contact_schrodinger_bridge} is to directly parameterize the dual potential (value function) $S_\theta(x^{s_k}_n, s_k)$ with a neural network and minimize the residual of the associated Hamilton-Jacobi-Bellman (HJB) equation~\eqref{eq:contact_S}, as explored in~\citet{neklyudov2023action, neklyudov2023computational, sun2025variational, buzun2026hota}. This strategy requires to learning the optimality conditions, rather than enforcing them directly, with important implications for both stability and numerical tractability.

In our formulation, the discretized optimality conditions~\eqref{eq:discrete_optimality} are enforced explicitly along predicted trajectories, reducing the problem to evaluating scalar quantities along sample paths ${(x_n^{s_{k_m}}, w_n^{s_{k_m}})}$ that are already available. As a result, training only requires backpropagation through $\nabla_\theta z_\theta$, leading to a comparatively well-conditioned optimization procedure. In contrast, minimizing the HJB residual necessitates differentiating through mixed second-order terms $\nabla^2_{x\theta} S_\theta$, which are notoriously ill-conditioned and often induce vanishing or exploding gradients~\citep{han2018solving, rackauckas2020universal}. Moreover, the HJB residual admits a constant degenerate solution $S_\theta(x^{s_k}_n, s_k)=C$ that trivially minimize the loss while encoding no transport information, thereby risking collapse of the optimization~\citep{wang2022is}. In this setting, methods that derive the drift solely from $\nabla_x S_\theta$ tend to produce noisy approximations \citep{neklyudov2023computational, sun2025variational, buzun2026hota}. When no additional optimal transport solver is used to anchor the learned dynamics to the prescribed marginals, the problem further inherits the limitations of collocation-based schemes, including the curse of dimensionality and the need for extensive domain sampling~\citep{neklyudov2023computational}. These issues are exacerbated by the intrinsic regularity properties of optimal transport value functions, which frequently develop non-differentiable shocks. Accurately capturing such viscosity solutions is challenging for standard neural network parameterizations and typically requires substantial regularization~\citep{han2018solving, rackauckas2020universal}.

Taken together, these considerations highlight a key advantage of the proposed shooting formulation: by enforcing optimality conditions directly along trajectories, it avoids both the degeneracies and the numerical instabilities inherent to HJB residual minimization.

\subsection{Computational Complexity and Practical Considerations}\label{app:timecomplexity}
At each iteration of the geodesic optimization procedure (Alg.~\ref{alg:training}), we sample $N$ points $x_n^{s_0} \in \mathcal{M}$ of dimension $D$ from the initial marginal $\rho^0$. We then compute the discrete geodesic $\{(x_n^{s_k}, w_n^{s_k})\}$ using equations \eqref{eq:discrete_geodesic}, along with the optimality variables $\{(\nu_n^{s_k}, S_n^{s_k}, z_n^{s_k})\}$ via equations \eqref{eq:discrete_optimality}. Each integration step requires $7$ evaluations of the neural networks $v_\theta$ and $z_\theta$, implemented as MLPs with $L$ layers and hidden width $W$. For $K$ integration steps, the computational cost of this forward pass scales as $\mathcal{O}(7 \, N \, D \, K \, L \, W)$.
The loss function $\ell$ in equation~(\ref{eq:loss}) requires $M$ evaluations of the Wasserstein distance between sample batches. We employ the \href{https://www.kernel-operations.io/geomloss/}{\texttt{geomloss}} library~\citep{feydy2019interpolating}, which uses the Sinkhorn algorithm with time complexity $\mathcal{O}(I_{\text{sh}} M D N \log N)$, where $I_{\text{sh}}$ is the number of Sinkhorn iterations until convergence~\citep{feydy2019interpolating}.
Combining both contributions, and noting that in practice $N\gg D, K, L, W$, the overall time complexity becomes,
\begin{equation}
    \mathcal{O}(7 \, N \, D \, K \, L \, W) + \mathcal{O}(I_{\text{sh}} D N \log N) 
    \approx \mathcal{O}(I_{\text{sh}} D N \log N).
\end{equation}
Overall, the method exhibits favorable scaling properties: the complexity grows linearly with the data dimension $D$ and nearly linearly with the batch size $N$. Additionally, it depends only weakly on the number of marginals and avoids costly outer iterative procedures required by alternative approaches.

\subsection{Geodesic Parameterization: Finer discretization in high-variability regions}\label{app:geodesic-parameterization}
The result of integration equation \eqref{eq:discrete_geodesic}, ${(x_n^{s_{k_m}}, w_n^{s_{k_m}})}$, discretizes the geodesic $\rho^s$ on $\mathcal{P}^+(\mathcal{M})$ into a piecewise-linear path segmented at nodes $\{ \rho_\theta^{s_k}\}_{k=0}^K$, 
thereby defining a specific sampling of points along the geodesic.
By default, minimizing the curve energy induced by the Wasserstein metric $g^{\mathcal{W}_2}$,
\begin{equation}
    d^2_{g^{\mathcal{W}_2}}(\rho^0, \rho^M) = \sum_{k=1}^K d^2_{\mathcal{W}_2}(\rho^{s_k}_\theta,\rho^{s_{k-1}}_\theta),
\end{equation}
corresponds to a uniform arc-length discretization $s$, where each segment has approximately the same length, i.e., $d^2_{\mathcal{W}_2}(\rho^{s_k}_\theta,\rho^{s_{k-1}}_\theta) \approx C$. 
The relationship between the arc-length discretization $s$ and the time discretization $t$, which describes the progression along the geodesic with respect to the time variable, is given by $ds = \sqrt{\mathcal{K}} \, dt$, where $\mathcal{K}$ denotes the kinetic energy of the system whose dynamics is interpreted as a geodesic.
In the energy-conserving case, $\mathcal{K}$ remains constant, and the time and arc-length discretizations differ only by a constant rescaling of the geodesic length. Nevertheless, the discretization nodes $\{ \rho_\theta^{s_k}\}_{k=0}^K$ remain unchanged, uniformly distributed in space and in time. In contrast, when modeling an energy-varying system whose associated trajectory projects onto a geodesic, the curve energy induced by the contact metric $g^\mathcal{C}$ takes the form, \looseness-1
\begin{equation}
    d^2_{g^\mathcal{C}}(\rho^0, \rho^M) =\sum_{k=1}^{K} \mathcal{K}^{s_k} \, d_{\mathcal{W}_2}^2({\rho}_\theta^{s_k}, {\rho}_\theta^{s_{k-1}}).
\end{equation}
This optimization enforces an approximately uniform geodesic parameterization under $g^\mathcal{C}$, resulting in a non-uniform node spacing $\Delta s$ on the Wasserstein manifold, as $\Delta s = d^2_{\mathcal{W}_2}(\rho^{s_k}_\theta,\rho^{s_{k-1}}_\theta) \approx \frac{C}{\mathcal{K}^{s_k}}$. As follows from equations~\eqref{eq:contact_hamiltonian_dynamics}, the scaling term $\mathcal{K}$ increases with the growth rate $z^t$. Consequently, regions with higher density concentration induce a larger effective metric. This, in turn, leads to a geodesic discretization that allocates more nodes in these areas, effectively assigning finer resolution to regions of higher variability and complexity. 
\begin{table}[t]
\centering
\small
\caption{Model size and training time across methods and experimental settings. Memory is reported in MB and time in minutes.}
\vspace{1mm}
\setlength{\tabcolsep}{3pt}
\begin{tabular}{lccccc}
\toprule
Dataset / Autoencoder & OTCFM  & TIGON & DeepRUOT & VarRUOT & \textbf{CuOT (Ours)} \\
\midrule
\multicolumn{6}{c}{\textit{Model Size (million of parameters)}} \\
\midrule
Simulation \citep{zhang2025learning}   & 0.2   & 0.2  & 0.2   & 0.5   & 0.5   \\
Mouse \citep{weinreb2020lineage}       & 0.2   & 0.2  & 0.2   & 0.5   & 0.8   \\
EMT \citep{cook2020context}            & 0.2  & 0.2 & 0.2  & 0.5  & 0.2  \\
Pancreatic \citep{veres2019charting}   & 0.5  & 0.5 & 0.5  & 0.8  & 1  \\
Zebrafish \citep{farrell2018single}     & 1  & 1  & 1  & 0.8  & 1  \\
\multicolumn{6}{l}{{FFHQ} \citep{karras2019style}} \\
\quad ALAE encoder \citep{pidhorskyi2020adversarial} & 12 & 1 & 3 & 1 & 3 \\
\quad e4e encoder \citep{richardson2021encoding} & 18 & 3 & 14 & 5 & 15 \\
\multicolumn{6}{l}{{UCF101} \citep{soomro2012ucf101}} \\
\quad DC encoder \citep{chen2025deep} & 18 & 3 & 14 & 5 & 18 \\
\quad TiTok encoder \citep{yu2024an}  & 18 & 3 & 14 & 5 & 18 \\
\midrule
\multicolumn{6}{c}{\textit{Training Time (minutes)}} \\
\midrule
Simulation \citep{zhang2025learning}   & 10  & 300  & 5  & 15  & 3  \\
Mouse \citep{weinreb2020lineage}       & 10  & 300  & 5  & 15  & 3  \\
EMT \citep{cook2020context}            & 10  & 300  & 5  & 30  & 10  \\
Pancreatic \citep{veres2019charting}   & 15  & 300  & 25  & 75  & 15  \\
Zebrafish \citep{farrell2018single}     & 20  & 300  & 30  & 80  & 15  \\
\multicolumn{6}{l}{{FFHQ} \citep{karras2019style}} \\
\quad ALAE encoder \citep{pidhorskyi2020adversarial} & 20  & 400  & 60  & 115  & 25  \\
\quad e4e encoder \citep{richardson2021encoding} & 60  & 300   & 120  & 300   & 300  \\
\multicolumn{6}{l}{{UCF101} \citep{soomro2012ucf101}} \\
\quad DC encoder \citep{chen2025deep} & 60 & 400 & 60 & 240 & 200 \\
\quad TiTok encoder \citep{yu2024an}  & 90 & 300  & 120 & 300  & 300  \\
\bottomrule
\end{tabular}
\label{tab:memory}
\end{table}
\begin{table}[t]
    \centering
    \caption{Parameter configurations for each experimental setting. 
    $K$: number of integration steps; $L$: number of hidden layers in the shared trunk network used by $v_\theta$ and $z_\theta$;
    $F$: number of sinusoidal frequencies; $H$: hidden feature dimension; 
    $B$: batch size; $N_b$: number of batches per iteration; 
    $N_{\text{iter}}$: number of training iterations; $\eta$: learning rate;
    $\varepsilon$: entropy level in Sinkhorn algorithm.
    $w_{(\cdot)}$: weighting coefficients. Additional details on these parameters are provided in Appendix \ref{app:implementation}.}
    \vspace{1mm}
    \setlength{\tabcolsep}{2pt}
    \resizebox{\textwidth}{!}{
    \begin{tabular}{lcccccccccccccc}
        \toprule
        Dataset / Autoencoder & Dim. & $K$ & $L$ & $F$ & $H$ & $B$ & $N_b$ & $N_{\text{iter}}$ & $\eta$ & $\varepsilon$ & $w_z$ & $w_m$ & $w_g$ & $w_{\text{mass}}$ \\
        \midrule
        Simulation \citep{zhang2025learning} & 2   & 20  & 3 & 3 & 512  & 512  & 1  & 2500  & 1e-5 & 0.005 & -1 & 10 & 0.01 & 10  \\
        Mouse \citep{weinreb2020lineage}     & 2   & 20  & 4 & 3 & 512  & 512  & 1  & 1500  & 1e-5 & 0.005 & -1 & 10 & 0.01 & 1   \\
        EMT \citep{cook2020context}           & 10  & 21  & 3 & 3 & 256  & 512  & 4  & 2500  & 1e-5 & 0.005 & -1 & 10 & 0.3  & 1   \\
        Pancreatic \citep{veres2019charting} & 30  & 21  & 4 & 4 & 512  & 1000 & 12 & 500   & 1e-5 & 0.005 & -1 & 1  & 0.3  & 100 \\
        Zebrafish \citep{farrell2018single}  & 50  & 20  & 4 & 3 & 512  & 1000 & 7  & 500   & 5e-4 & 0.005 & -1 & 1  & 0.1  & 1   \\
        \multicolumn{15}{l}{{FFHQ} \citep{karras2019style}} \\
        \quad ALAE encoder \citep{pidhorskyi2020adversarial} & 512 & 20  & 4 & 3 & 512  & 1000 & 5  & 10000 & 1e-5 & 0.005 & -1 & 1  & 0.1  & 5   \\
        \quad e4e encoder \citep{richardson2021encoding} & 9216 & 20  & 3 & 3 & 2048  & 1000 & 5  & 10000 & 3e-5 & 0.005 & -1 & 1  & 0.01  & 500   \\
        \multicolumn{15}{l}{{UCF101} \citep{soomro2012ucf101}} \\
        \quad DC encoder \citep{chen2025deep}  & 2048 & 180 & 3 & 5 & 2048 & 101 & 1 & 15000 & 3e-5 & 0.005 & -1 & 1 & 0.01 & 500 \\
        \quad TiTok encoder \citep{yu2024an}  & \textit{d}  & 180 & 3 & 5 & 2048 & 101 & 1 & 15000 & 3e-5 & 0.005 & -1 & 1 & 0.01 & 500 \\
        \bottomrule
    \end{tabular}
    }
    \label{tab:hyperparameters}
\end{table}

\section{Extended Results}\label{app:results}
Details on the experimental settings adopted in each experiments are provided in Table \ref{tab:hyperparameters}. 
The synthetic ablation study (Appendix~\ref{app:ablations}) and the single-cell experiments (Appendix~\ref{app:single-cell}) are conducted on a local machine equipped with an NVIDIA GeForce RTX 5090 GPU (32 GB memory, CUDA 13.0, driver 580.126.09). Methods are evaluated over 20 runs with different random seeds.
The dataset-pairing (Appendix~\ref{app:dataset-pairing}) and image-to-video generation experiments (Appendix~\ref{app:image-to-video}) are run on a shared LSF-managed GPU cluster using the \texttt{batch\_a100} queue. Each run is allocated a single GPU in exclusive mode, together with 1 CPU core, 32 GB of system memory, and CUDA 12.6 support. Both Child $\to$ Adult and Adult $\to$ Child transports are evaluated over 10 independent retrainings each, while for image-to-video generation each method is retrained 10 times per action class.

\subsection{Ablation Studies and Analysis of Mode Collapse}
\label{app:ablations}

To isolate the individual sources of performance gains in the proposed CuOT framework, evaluate the impact of adaptive step-size selection, and quantitatively assess resilience to soft mode collapse, we perform a detailed empirical study on a synthetic dataset with known mode correspondences and mass dynamics. \looseness-1

The source distribution consists of three 2D Gaussian modes (variance $\sigma = 0.15$) centered at $[0.0, 1.0]$, $[-0.8, 0.5]$, and $[0.8, 0.5]$, each with $400$ samples (total $N=1200$). The target distribution comprises three distinct shapes centered at $[0.0, 1.4]$ (Swiss roll, $N=600$), $[-1.2, 0.7]$ (concentric rings, $N=600$), and $[1.2, 0.7]$ (pinwheel, $N=200$). 
Under correct mode-to-mode matching, the first two source modes undergo mass expansion (requiring positive growth rates, $z^t > 0$), whereas the third mode undergoes mass decay (requiring a negative growth rate, $z^t < 0$). This setup intentionally misaligns the transport direction with the growth gradient for the third mode to directly target the failure modes of standard WFR-based formulations.

Performance is evaluated across global transport, mass conservation, and mode-level fidelity using, 2-Wasserstein ($\mathcal{W}_2$) and 1-Wasserstein ($\mathcal{W}_1$) distances between generated and ground-truth target distributions; absolute error in total reconstructed target mass; the fraction of source samples transported to incorrect target modes (Leakage); the chamfer distance from ground-truth target samples to generated samples (CD); and individual $W_2$ metrics for each of the three target modes $[M_1, M_2, M_3]$ to quantify minority mode recovery.
The results of the comparison are resumed in Table \ref{tab:synthetic_ablation}.

We compare the complete CuOT model against five baselines and structural variants:
\begin{enumerate}
    \item {CuOT (Complete):} The proposed full contact geometry formulation with spatial-temporal growth dynamics $z_\theta(x^{s_k}, s_k)$ and adaptive step-size allocation.
    \item {CuOT w/ Time-Only Growth:} Replaces $z_\theta(x^{s_k}, s_k)$ with a spatially uniform, time-dependent growth rate parameter $z_\theta(s_k)$.
    \item {CuOT w/ WFR Geometry:} Replaces the Herglotz variational uOT problem \eqref{eq:contact_schrodinger_bridge} with the WFR uOT \eqref{eq:fisher-rao-schrodinger_bridge} and applies the geodesic-based discretization described in Section \ref{sec:implementation}. The discrete FP equation \eqref{eq:discrete_geodesic} and the feasibility loss term \eqref{eq:loss-fea} remain unchanged, but the discrete optimality conditions \eqref{eq:discrete_optimality} now discretize the WFR conditions \eqref{eq:FR_hamiltonian_dynamics}, and the optimality loss optimizes the WFR geodesic objective coupled with growth regularization:
    \begin{equation*}
        \ell_{\mathrm{opt}} = \frac{1}{2} (v_n^{s_k})^2 w_n^{s_k} d^{\mathcal{W}_2} + \| z_n^{s_k} - z_\theta(x_n^{s_k}, s_k) \|^2.
    \end{equation*}
    \item {CuOT w/o Adaptive Step:} Retains the growth regularization term from the contact formulation but uses standard 2-Wasserstein geodesic minimization for trajectory parameterization:
    \begin{equation*}
        \ell_{\mathrm{opt}} = (d^{\mathcal{W}_2})^2 + \| z_n^{s_k} - z_\theta(x_n^{s_k}, s_k) \|^2.
    \end{equation*}
    This isolates the contact-metric step adaptation but preserves  mass variation dynamics.
    \item {{VarRUOT}:} Standard variational WFR baseline.
    \item {{OTCFM}:} OT-based baseline assisted by an independent global mass network.
\end{enumerate}
Comparing {CuOT} (Complete) with {CuOT} w/o Adaptive Step directly isolates the impact of dynamic temporal resolution allocation.
Although both variants achieve near-identical mass reconstruction errors ($0.0006$ vs. $0.0009$) and mode leakage ($0.2\%$), removing the adaptive step degrades overall transport accuracy ($\mathcal{W}_2$: $0.10 \to 0.13$; $\mathcal{W}_1$: $0.08 \to 0.10$). Inspecting per-mode target errors reveals that these performance gains are concentrated in regions undergoing mass expansion: {CuOT} (Complete) yields per-mode $\mathcal{W}_2$ errors of $[0.09, 0.10, 0.13]$, compared to $[0.13, 0.14, 0.13]$ for the non-adaptive variant.
This difference arises because the non-adaptive variant distributes discrete trajectory points uniformly due to the standard 2-Wasserstein geodesic objective. In contrast, the contact metric dynamically reduces step sizes in regions with positive local mass growth ($z_\theta > 0$), allocating higher temporal resolution to high-density target regions (Modes 1 and 2). Shrinking regions (Mode 3) retain standard resolution, maintaining consistent accuracy ($\mathcal{W}_2 = 0.13$).

WFR-based methods suffer from mode leakage, improperly assigning samples to distant target modes when transport directions conflict with local growth dynamics. This behavior is demonstrated in Table~\ref{tab:synthetic_ablation}, where {VarRUOT} and CuOT w/ WFR Geometry exhibit $6.5\%$ and $10.4\%$ mode leakage, respectively. Examining per-mode $\mathcal{W}_2$ errors for {VarRUOT} ($[0.15, 0.16, 0.20]$) reveals progressive accuracy degradation on the third mode, which requires negative growth, because samples are instead transported toward other directions.
As growth regularization vanishes, leakage decreases to 0, but WFR geometry induces positive growth for all modes and and high CD ($>0.3$), causing samples to cover only partial modes and leading to mode collapse. The time-only growth variant cannot capture spatially varying dynamics, yielding the worst performance, as the target requires simultaneous growth and decay across different regions.
The OTCFM method with the global mass network successfully minimizes global mass error ($0.0001$), but as it keeps sample weights constant it fails to prevent misrouting, resulting in the highest mode leakage ($17.7$ \%).

\subsection{Single-Cell Dynamics Prediction}
\label{app:single-cell}
The benchmarks considered in the experiments range from a controlled synthetic setting to four real-world datasets of increasing dimensionality and complexity. All datasets share the common structure of temporal snapshots of single-cell gene expression data: each marginal $\rho^m$ is a collection of cells measured at a given time point, represented as points in a high-dimensional feature space, with the total cell count varying across time points due to proliferation and death. 
As summarized in the dataset-specific tables below, the Wasserstein distance between successive marginals, which the learned transport network $v_\theta$ must replicate, is not proportional to the growth rate gradient $\nabla_x z^t$, systematically violating the geometric coupling assumption underlying WFR-based reconstruction methods.\looseness-1

\paragraph{Three-gene simulation~\citep{zhang2025learning} (2D).}
This synthetic dataset simulates stochastic gene expression dynamics for a system of three genes incorporating self-activation, mutual inhibition, and external activation. The two-dimensional representation corresponds to the two leading principal genes of the gene expression trajectory. The initial distribution splits into two clusters, one of which remains quiescent while the other evolves through the state space, with a monotonically decreasing distance between successive marginals and an approximately constant growth rate gradient. The dataset comprises $M = 4$ observed marginals with three intermediate marginals included in training.
\vspace{-5mm}
\begin{table}[H]
\footnotesize
    \centering
    \caption{Details on the three-gene simulated dataset \citep{zhang2025learning}.}
    \begin{tabular}{ccccccc}
        \toprule
        Time & Mass & Growth Rate $z$& $\nabla z$ & N samples & $\mathcal{W}_2$ Distance & Distance \% \\ 
        \midrule
        0 & 1.000 & 10.5\% & - & 400 & 0.5185 & 33.9\% \\
        1 & 1.105 & 19.9\% & 9.4\%  & 442 & 0.5117 & 33.5\% \\
        2 & 1.325 & 30.2\% & 10.3\% & 530 & 0.3193 & 20.9\% \\
        3 & 1.725 & 40.4\% & 10.2\% & 690 & 0.1783 & 11.7\% \\
        4 & 2.422 & - &  - & 969 & - & - \\ 
        \bottomrule
    \end{tabular}
    \label{tab:3gene_dataset}
\end{table}
\vspace{-5mm}
\paragraph{Mouse hematopoiesis~\citep{weinreb2020lineage} (2D).}
This dataset tracks blood stem cells in a mouse as they commit to one of two specialized blood cell lineages over time. The two-dimensional representation corresponds to the two principal axes of the differentiation process. The dataset comprises $M = 2$ observed marginals plus the initial condition, with 
a single intermediate marginal included in training.
\vspace{-5mm}
\begin{table}[H]
\footnotesize
    \centering
    \caption{Details on the mouse blood hematopoiesis dataset \citep{weinreb2020lineage}.}
    \begin{tabular}{ccccccc}
        \toprule
        Time & Mass & Growth Rate $z$& $\nabla z$ & N samples & $\mathcal{W}_2$ Distance & Distance \% \\ 
        \midrule
        0 & 1.000 & 164.6\% & -  & 1429 & 0.8041 & 61.11\% \\
        1 & 2.646 & 53.1\%  & -111.5\% & 3781 & 0.5117 & 38.89\% \\
        2 & 4.050 & - & - & 5788 & - & - \\ 
        \bottomrule
    \end{tabular}
    \label{tab:mouse_dataset}
\end{table}
\vspace{-5mm}
\begin{figure}[H]
\centering

\begin{subfigure}{0.24\textwidth}
    \centering
    \caption{\textbf{CuOT (Ours)}}
    \includegraphics[width=\linewidth]{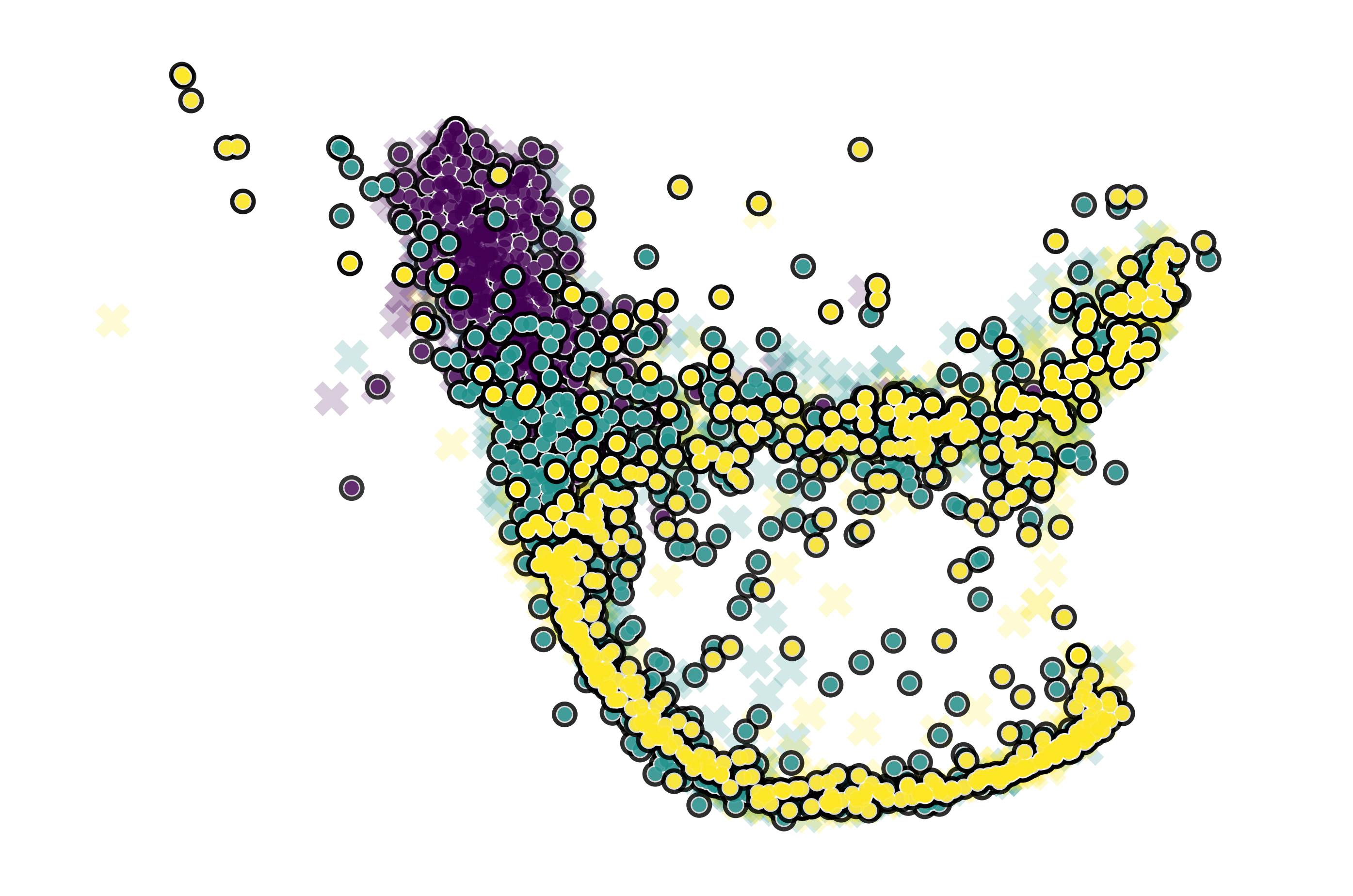}
\end{subfigure}
\hfill
\begin{subfigure}{0.24\textwidth}
    \centering
    \caption{VarRUOT}
    \includegraphics[width=\linewidth]{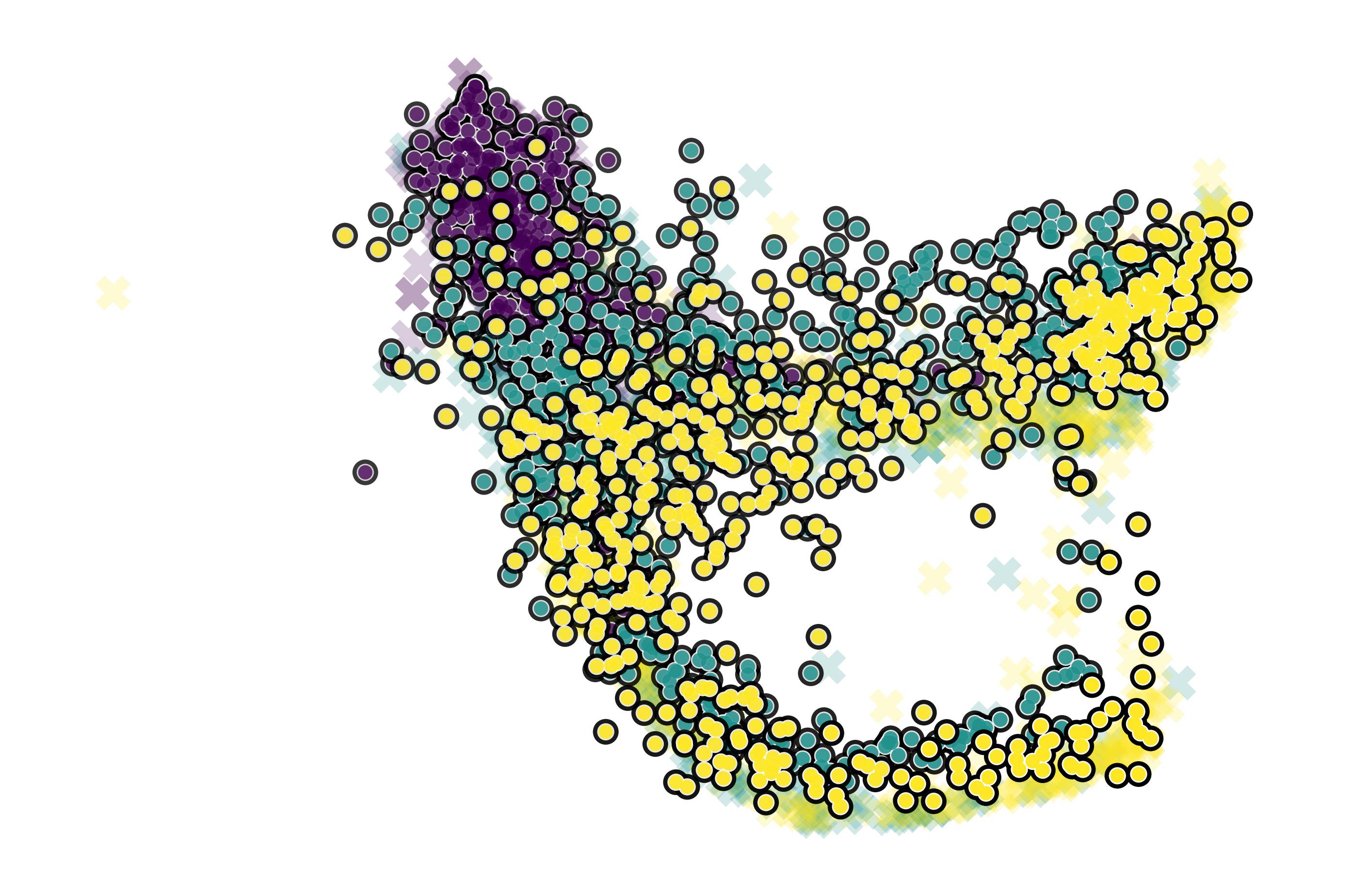}
\end{subfigure}
\hfill
\begin{subfigure}{0.24\textwidth}
    \centering
    \caption{DeepRUOT}
    \includegraphics[width=\linewidth]{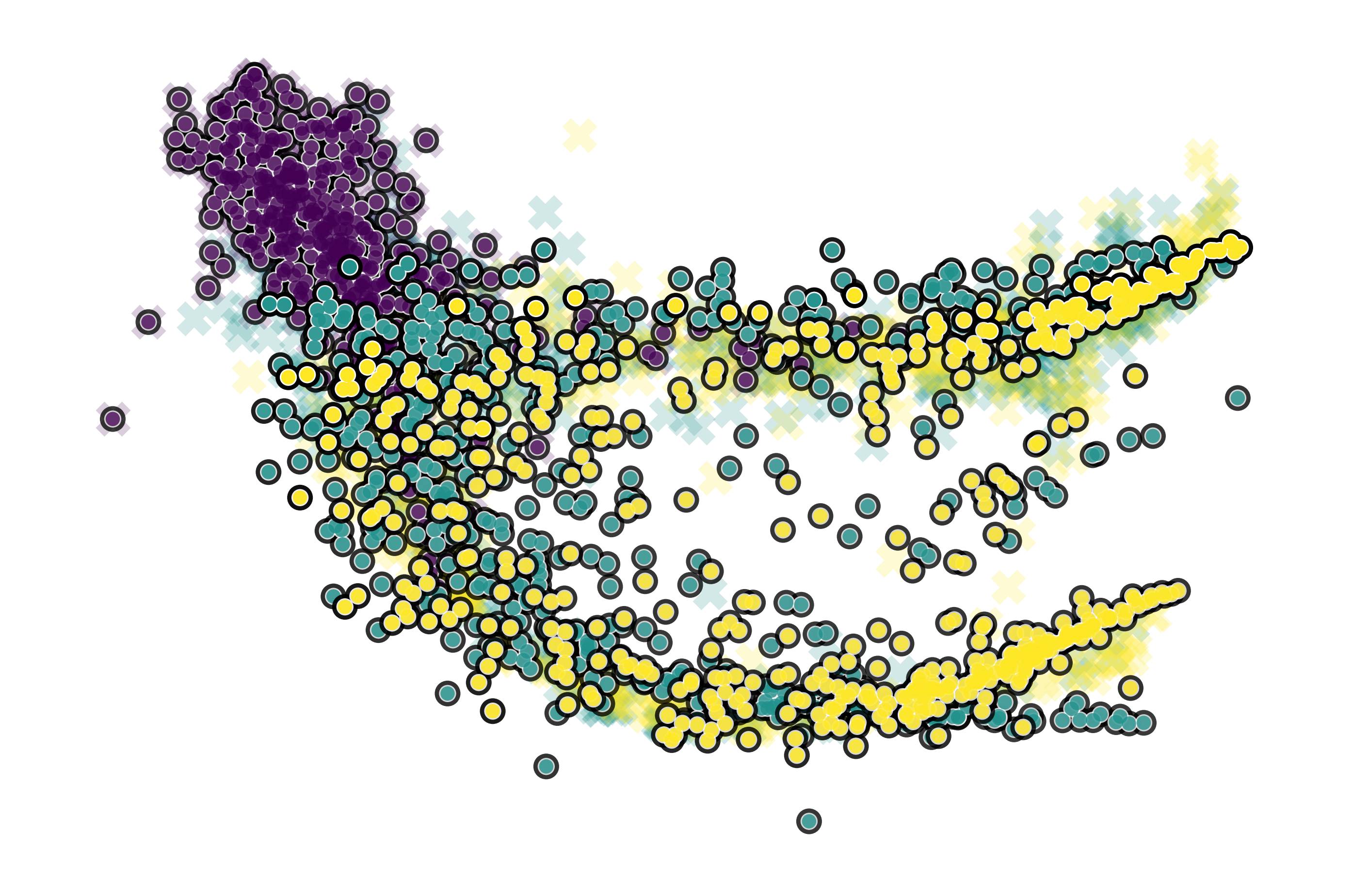}
\end{subfigure}
\hfill
\begin{subfigure}{0.24\textwidth}
    \centering
    \caption{OTCFM}
    \includegraphics[width=\linewidth, height=2.25cm]{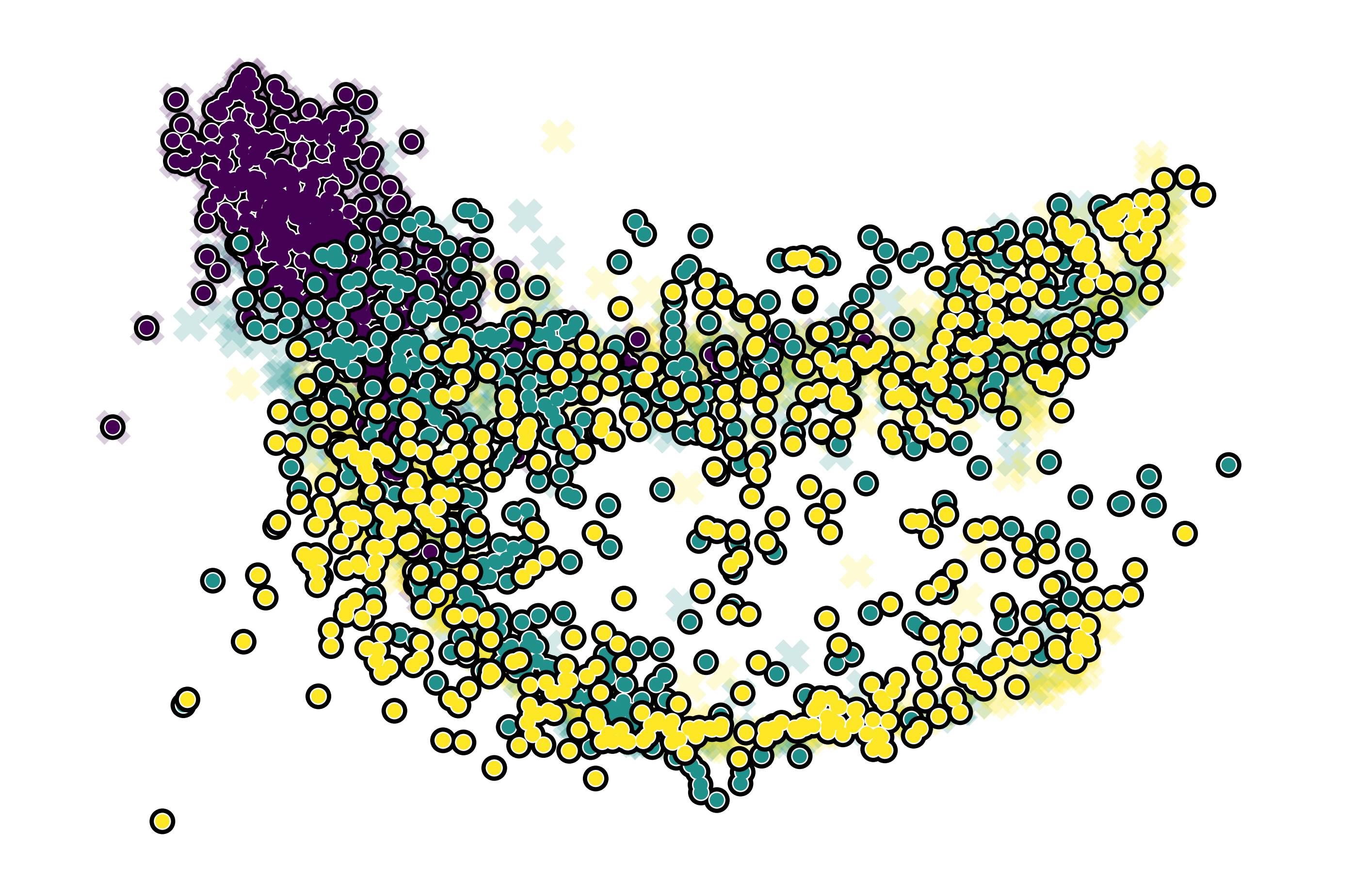}
\end{subfigure}

\vspace{0.0cm}

\begin{subfigure}{0.24\textwidth}
    \centering
    \includegraphics[width=\linewidth]{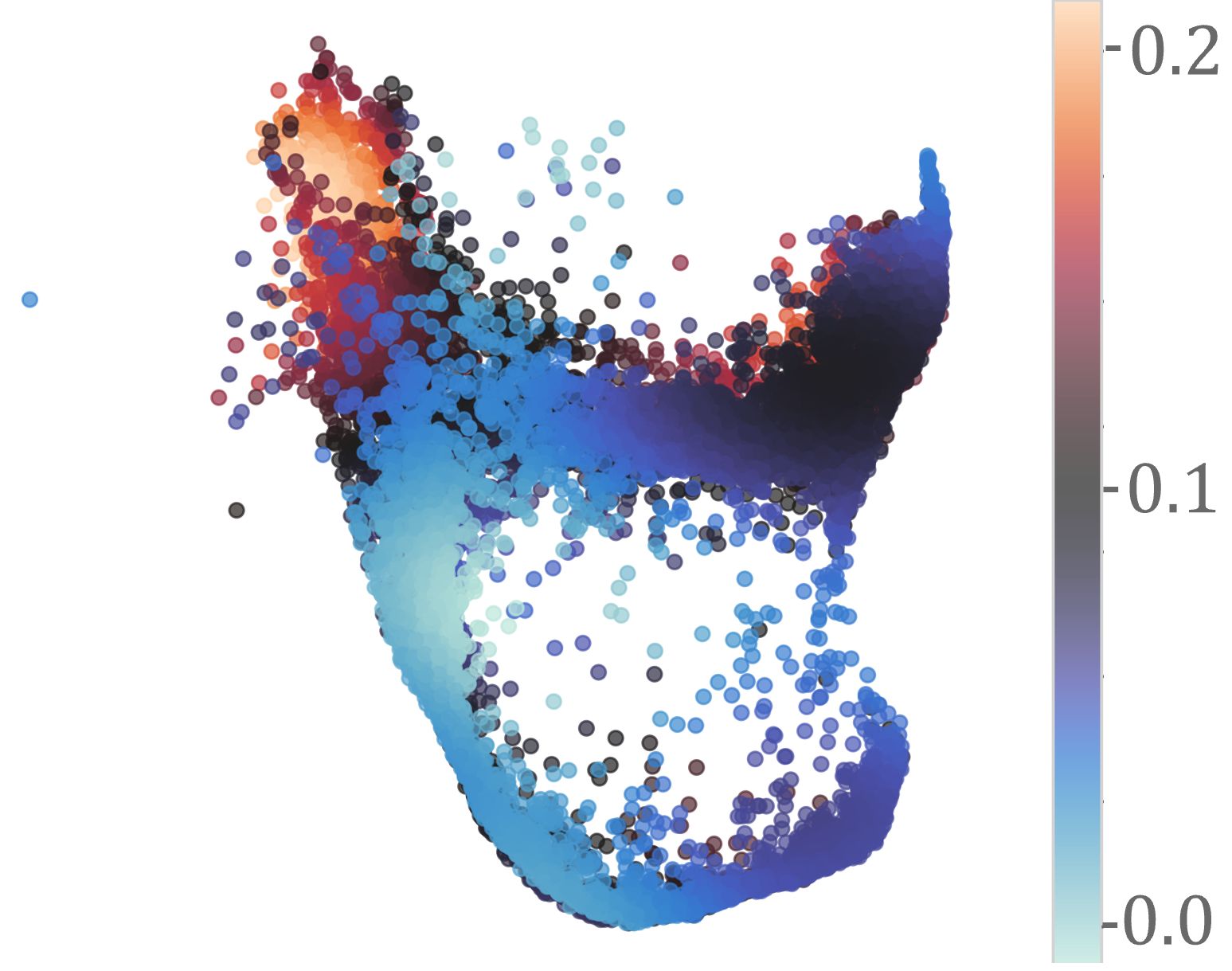}
\end{subfigure}
\hfill
\begin{subfigure}{0.24\textwidth}
    \centering
    \includegraphics[width=\linewidth]{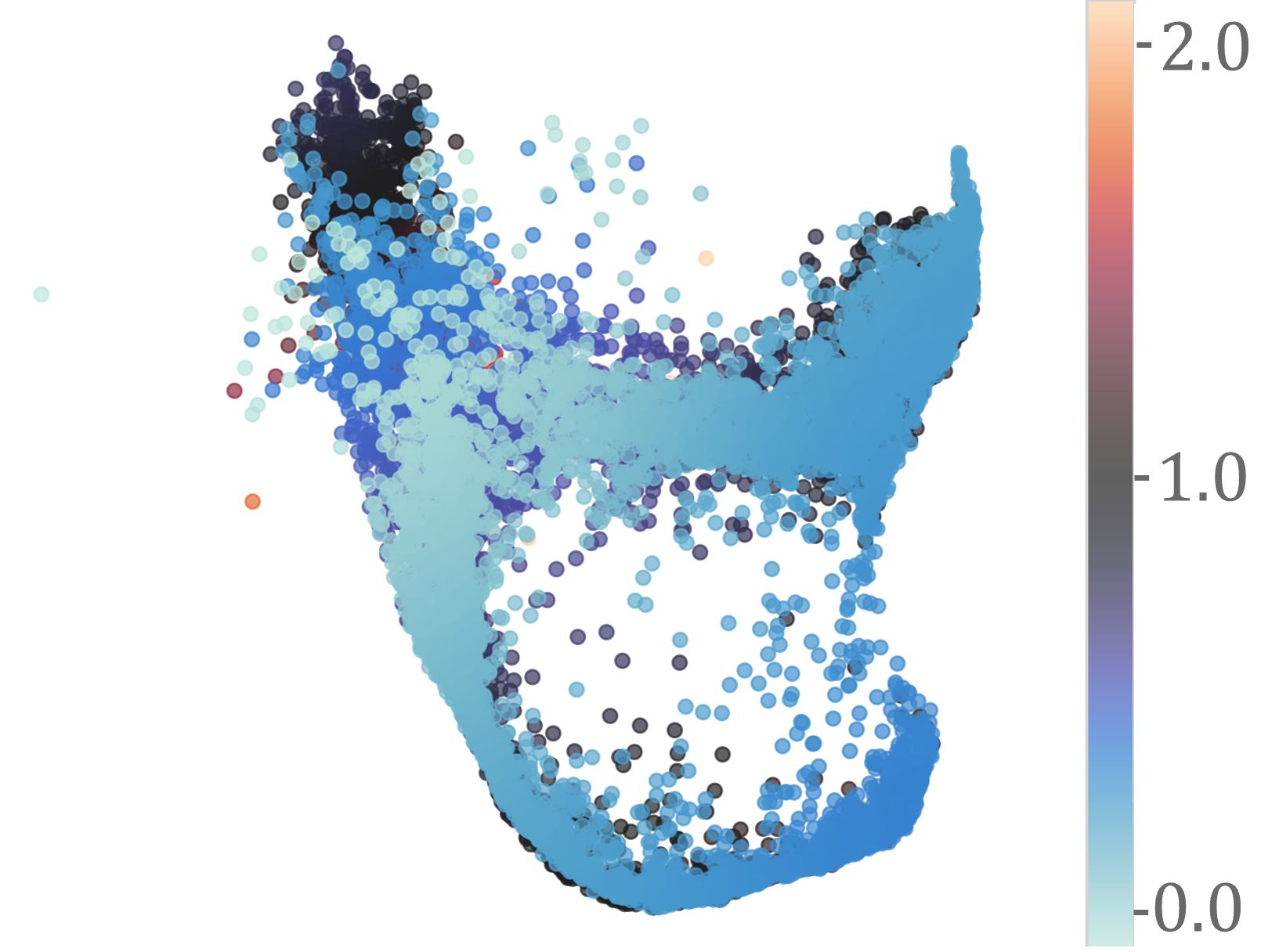}
\end{subfigure}
\hfill
\begin{subfigure}{0.24\textwidth}
    \centering
    \includegraphics[width=\linewidth]{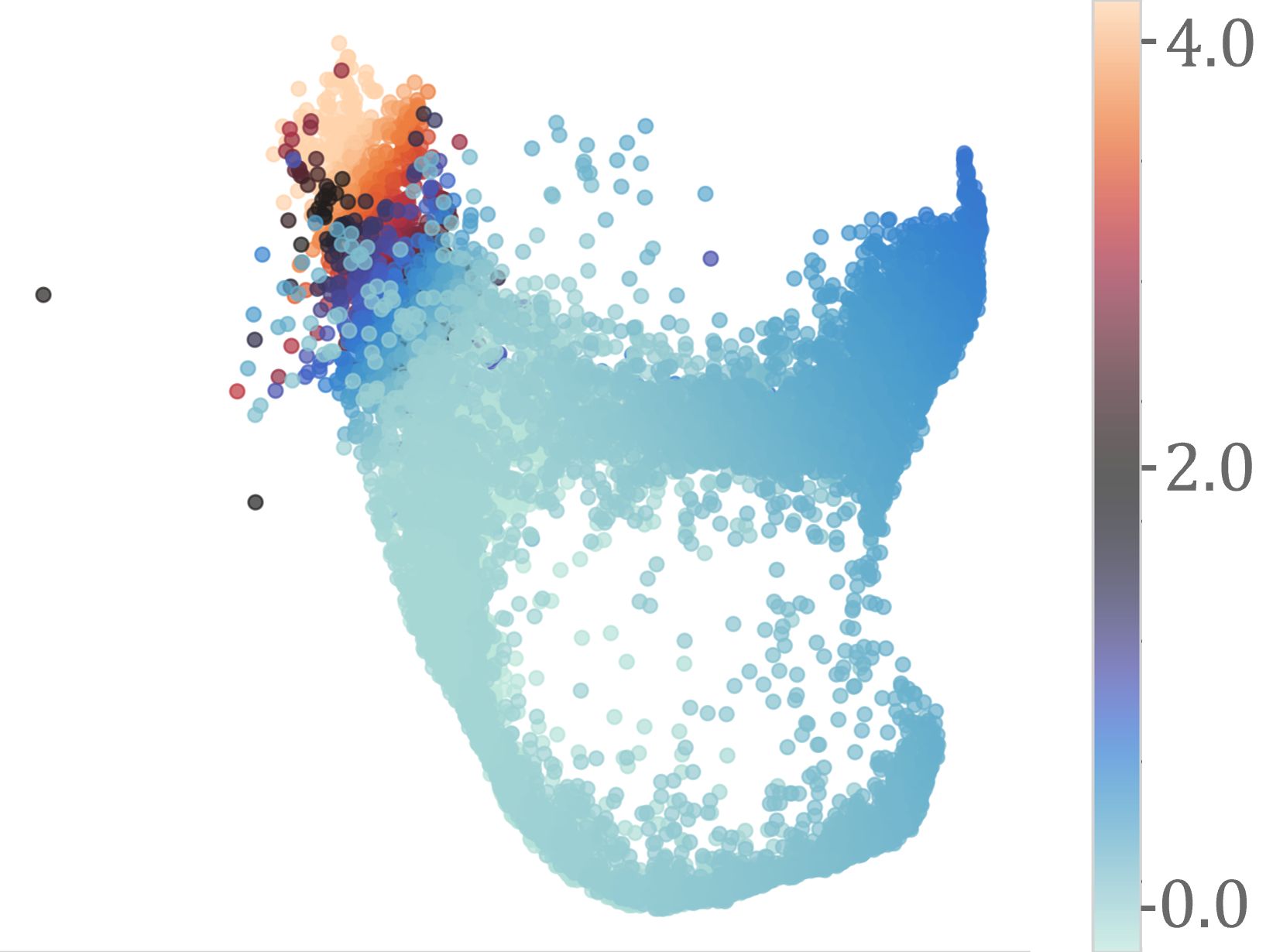}
\end{subfigure}
\hfill
\begin{subfigure}{0.24\textwidth}
    \centering
    \includegraphics[width=\linewidth]{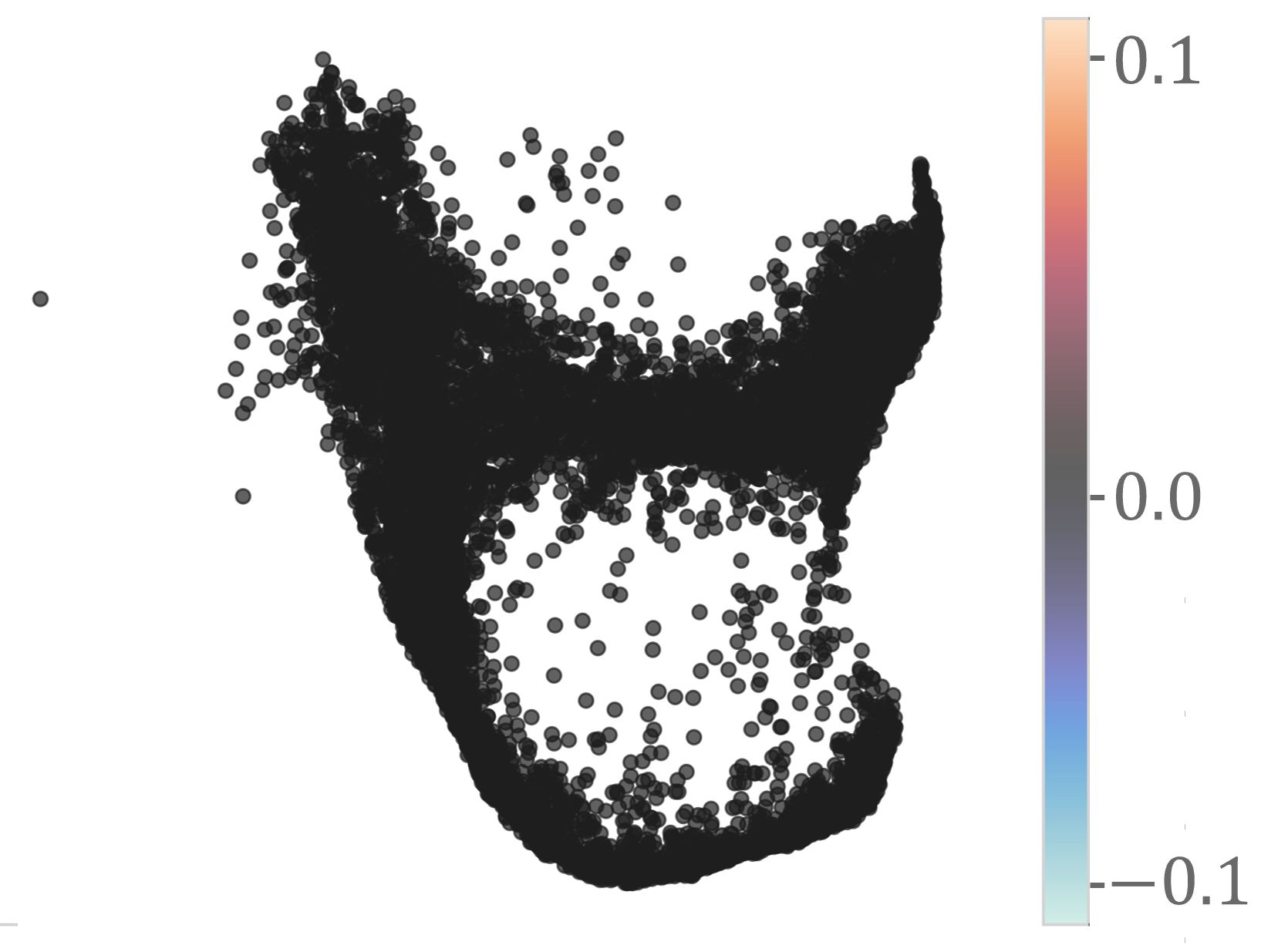}
\end{subfigure}
\caption{Qualitative analysis on the mouse blood
hematopoiesis dataset \citep{weinreb2020lineage}. Predicted dynamics $\big(t_0$\tball{v3_1},\, $t_1$\tball{v3_2},\, $t_2$\tball{v3_3}$\big)$ (top) and growth rate (bottom).}
\label{fig:mouse}
\end{figure}
\vspace{-5mm}
\begin{table}[H]
\centering
\caption{Quantitative analysis on the mouse blood
hematopoiesis dataset \citep{weinreb2020lineage}.}
\small
\setlength{\tabcolsep}{2pt}
\resizebox{\textwidth}{!}{
\begin{tabular}{lcccccc}
\toprule
& \multicolumn{3}{c}{$t=1$} & \multicolumn{3}{c}{$t=2$} \\
\cmidrule(lr){2-4} \cmidrule(lr){5-7}
Model 
& $\mathcal{W}_1$ & $\mathcal{W}_2$ & Mass
& $\mathcal{W}_1$ & $\mathcal{W}_2$ & Mass \\
\midrule

OTCFM \citep{tong2023improving} 
& 0.1657$_{\pm 0.0271}$ & 0.1795$_{\pm 0.0183}$ & 2.7090$_{\pm 0.0000}$
& 0.1670$_{\pm 0.0268}$ & 0.1820$_{\pm 0.0208}$ & 9.3048$_{\pm 0.0000}$ \\

TIGON \citep{sha2024reconstructing}
& 0.3392$_{\pm 0.0294}$ & 0.3975$_{\pm 0.0538}$ & 1.3125$_{\pm 0.3903}$
& 0.4530$_{\pm 0.0544}$ & 0.5474$_{\pm 0.0364}$ & 5.1929$_{\pm 0.7556}$ \\

DeepRUOT \citep{zhang2025learning}
& $\underline{0.1254_{\pm 0.0063}}$ & 0.1442$_{\pm 0.0052}$ & 0.1391$_{\pm 0.0336}$
& $\underline{0.1137_{\pm 0.0074}}$ & $\underline{0.1292_{\pm 0.0064}}$ & 0.6741$_{\pm 0.1275}$ \\

VarRUOT \citep{sun2025variational}
& $\underline{0.1211_{\pm 0.0048}}$ & $\underline{0.1399_{\pm 0.0051}}$ & $\underline{0.0048_{\pm 0.0010}}$
& $\underline{0.1226_{\pm 0.0080}}$ & $\underline{0.1349_{\pm 0.0067}}$ & $\underline{0.0089_{\pm 0.0055}}$ \\

\textbf{CuOT (Ours)}
& $\bm{0.0745_{\pm 0.0095}}$ & $\bm{0.0765_{\pm 0.0071}}$ & $\bm{0.0003_{\pm 0.0003}}$
& $\bm{0.0561_{\pm 0.0100}}$ & $\bm{0.0550_{\pm 0.0071}}$ & $\bm{0.0019_{\pm 0.0021}}$ \\
\bottomrule
\end{tabular}
}
\label{tab:mouse}
\end{table}
\vspace{-5mm}
\paragraph{Epithelial--mesenchymal transition (EMT)~\citep{cook2020context} (10D).}
This dataset captures the process by which cancer cells switch from a stationary, tightly-packed epithelial state to a mobile, loosely-connected mesenchymal state, a key step in tumor metastasis and invasion. The 10-dimensional feature space corresponds to the leading principal components of the gene expression profile. The transition induces a pronounced shift in the distribution support across time points, accompanied by moderate mass variation, and the dataset includes $M = 3$ observed marginals with two intermediate marginals used in training.
\vspace{-5mm}
\begin{table}[H]
\footnotesize
    \centering
    \caption{Details on the Emt dataset \citep{cook2020context}.}
    \begin{tabular}{ccccccc}
        \toprule
        Time & Mass & Growth Rate $z$& $\nabla z$ & N samples & $\mathcal{W}_2$ Distance & Distance \% \\ 
        \midrule
        0 & 1.000 & 53.4\%  & - & 577 & 3.8210 & 61.19\% \\
        1 & 1.534 & -11.0\% & -64.4\% & 885 & 1.2108 & 19.39\% \\
        2 & 1.366 & 12.0\%  & 23.0\%  & 788 & 1.2122 & 19.41\% \\
        3 & 1.530 & - & - & 883 & - & - \\ 
        \bottomrule
    \end{tabular}
    \label{tab:emt_dataset}
\end{table}
\vspace{-5mm}
\begin{figure}[H]
\centering

\begin{subfigure}{0.24\textwidth}
    \centering
    \caption{\textbf{CuOT (Ours)}}
    \includegraphics[width=\linewidth]{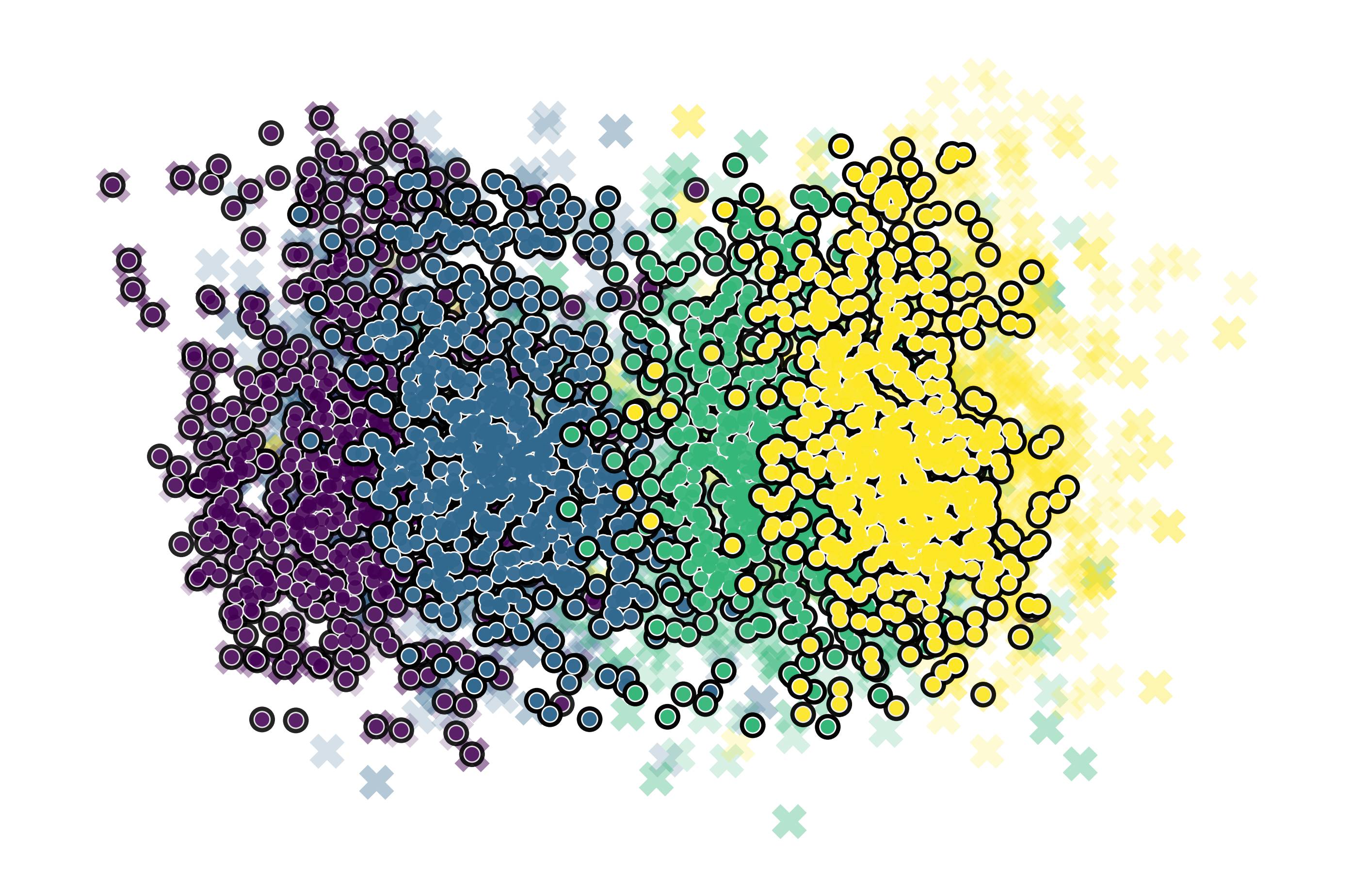}
\end{subfigure}
\hfill
\begin{subfigure}{0.24\textwidth}
    \centering
    \caption{VarRUOT}
    \includegraphics[width=\linewidth]{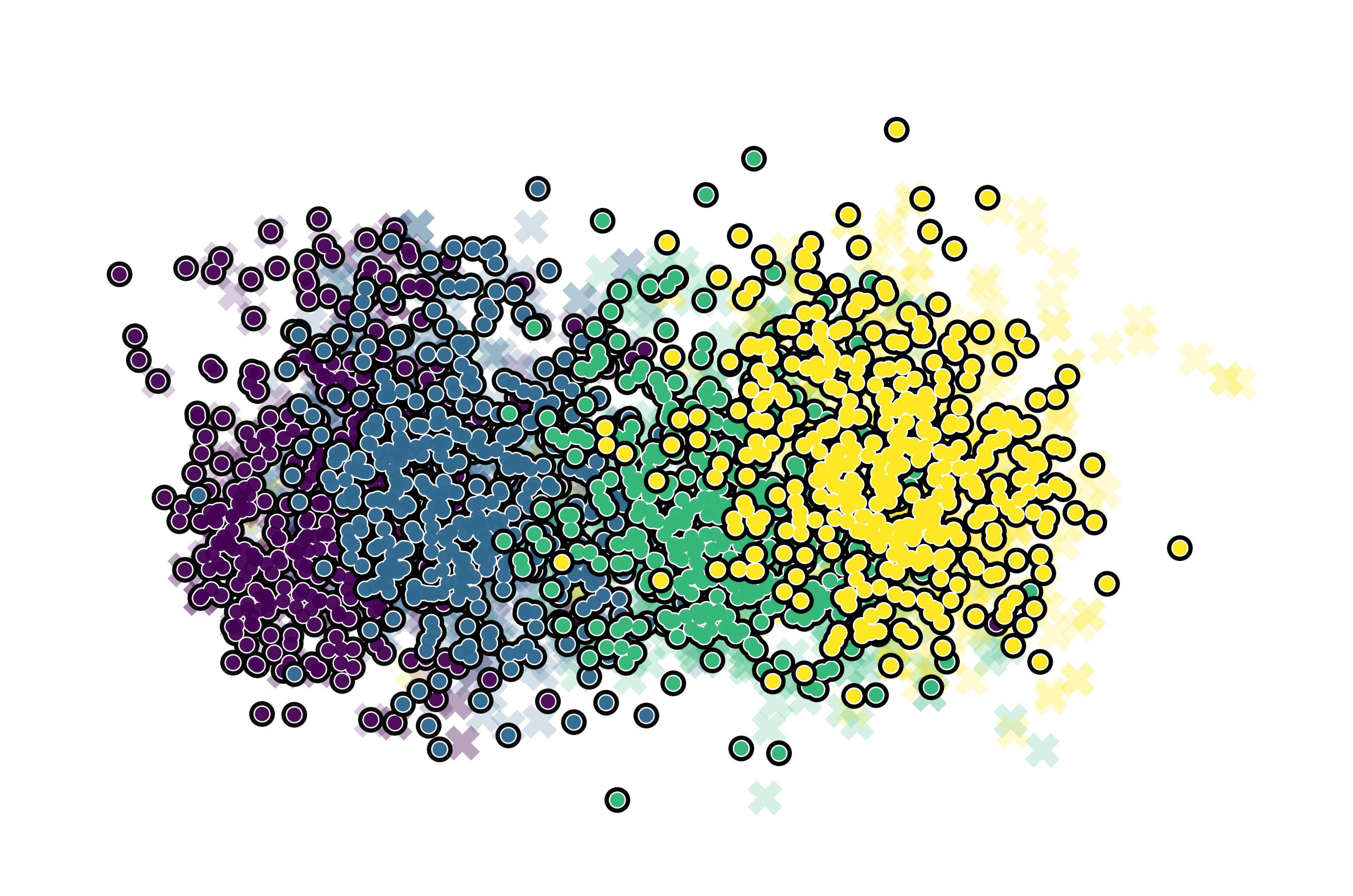}
\end{subfigure}
\hfill
\begin{subfigure}{0.24\textwidth}
    \centering
    \caption{DeepRUOT}
    \includegraphics[width=\linewidth]{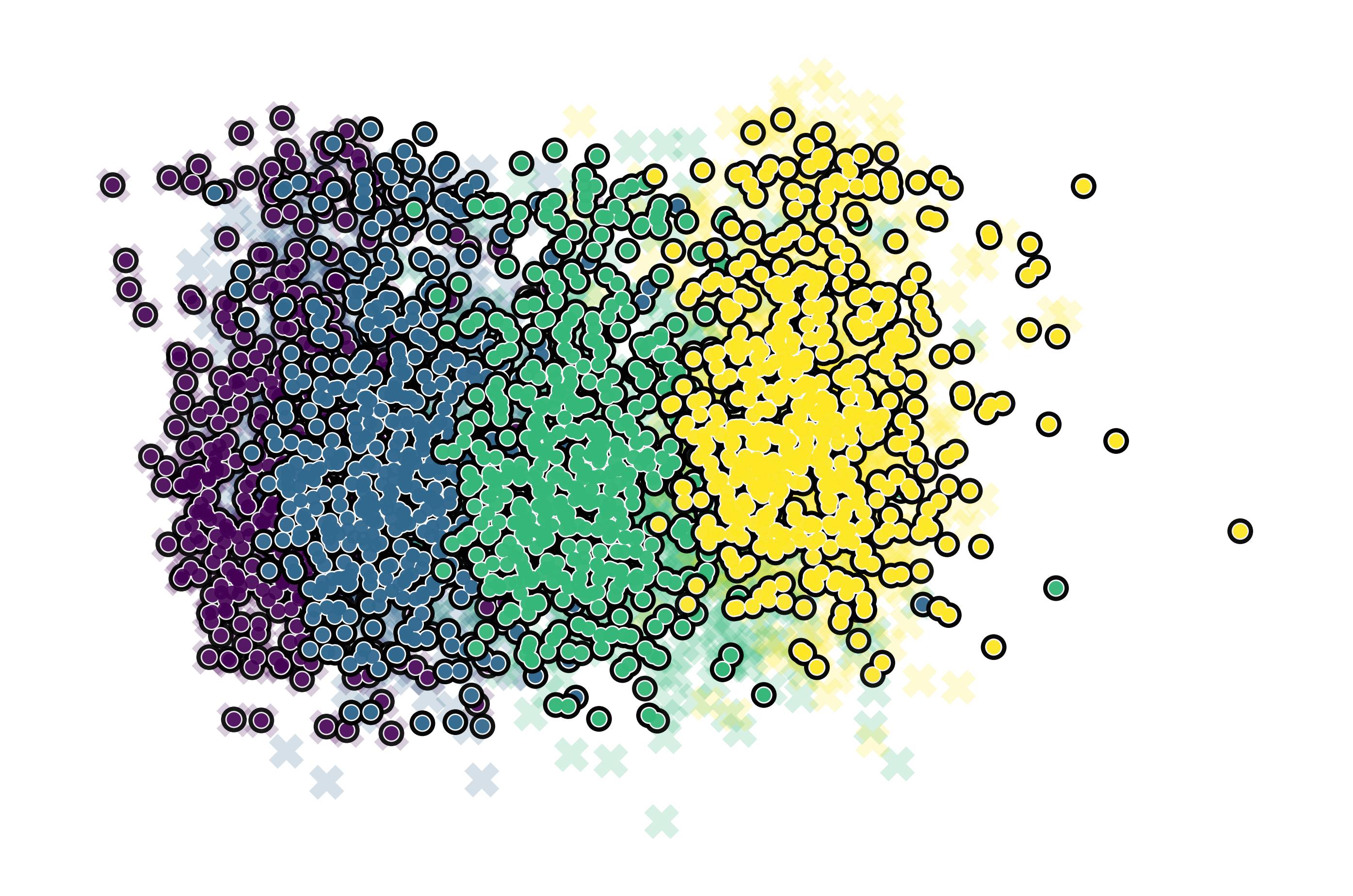}
\end{subfigure}
\hfill
\begin{subfigure}{0.24\textwidth}
    \centering
    \caption{OTCFM}
    \includegraphics[width=\linewidth, height=2.25cm]{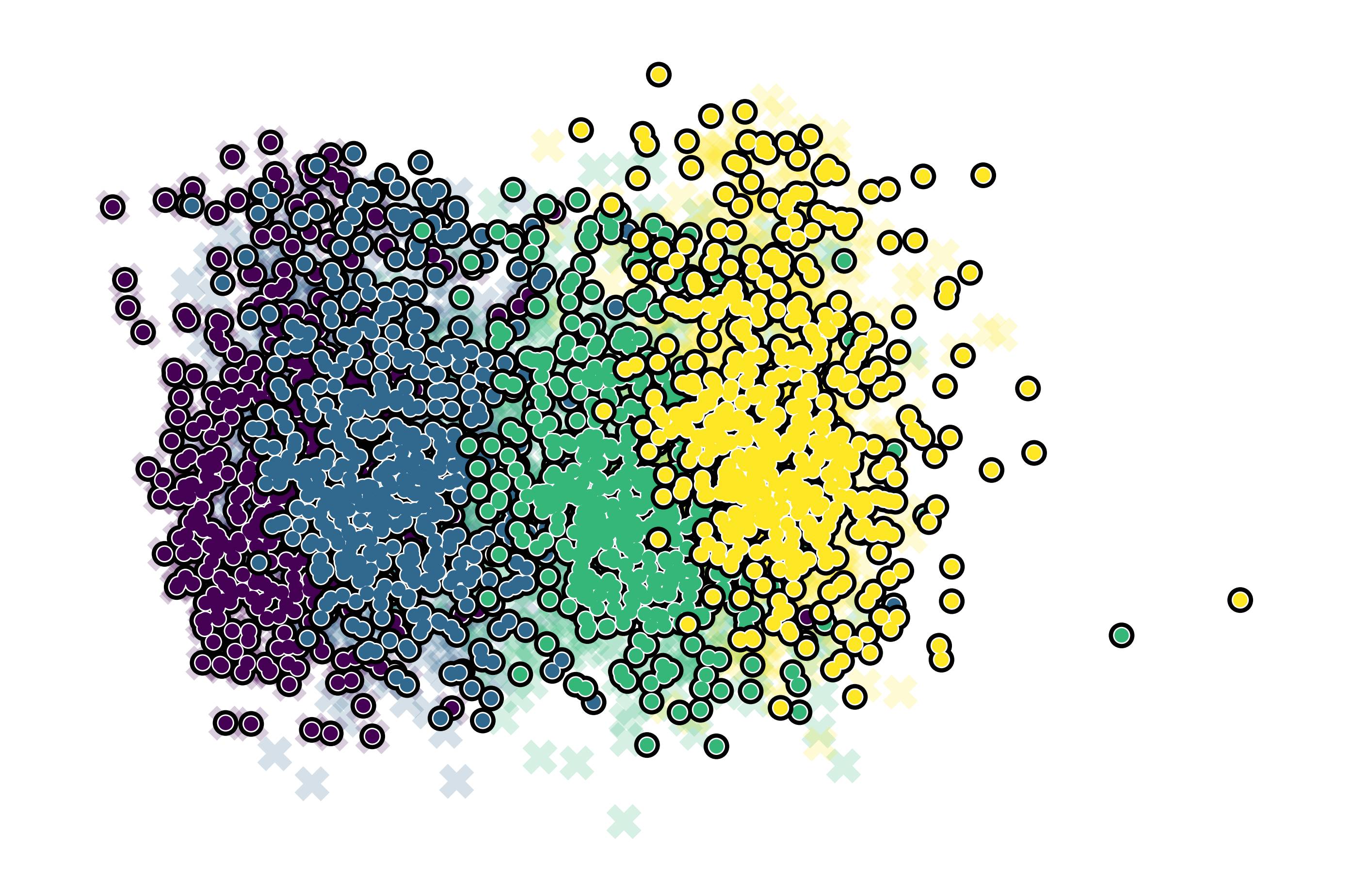}
\end{subfigure}

\vspace{0.0cm}

\begin{subfigure}{0.24\textwidth}
    \centering
    \includegraphics[width=\linewidth]{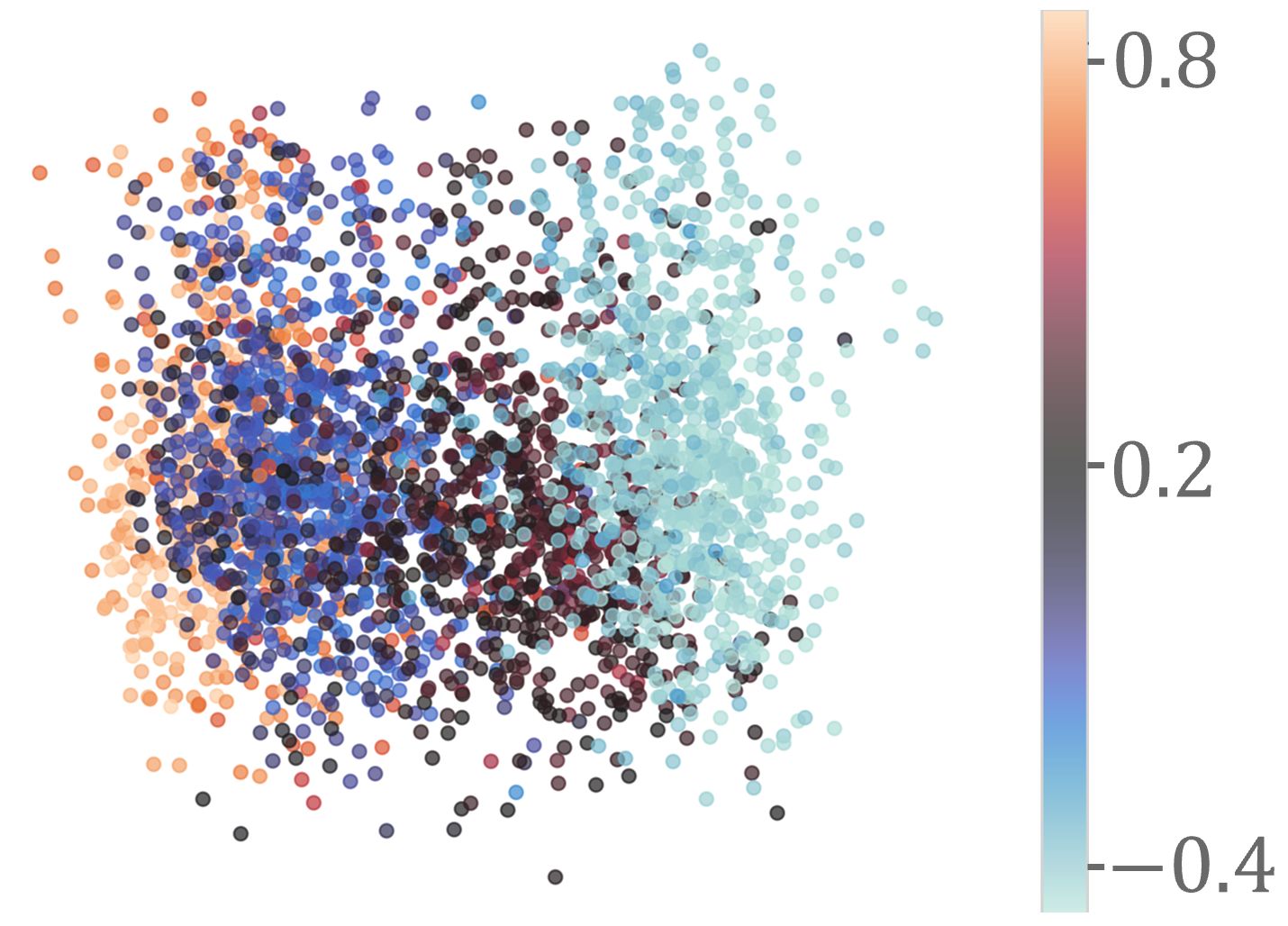}
\end{subfigure}
\hfill
\begin{subfigure}{0.24\textwidth}
    \centering
    \includegraphics[width=\linewidth]{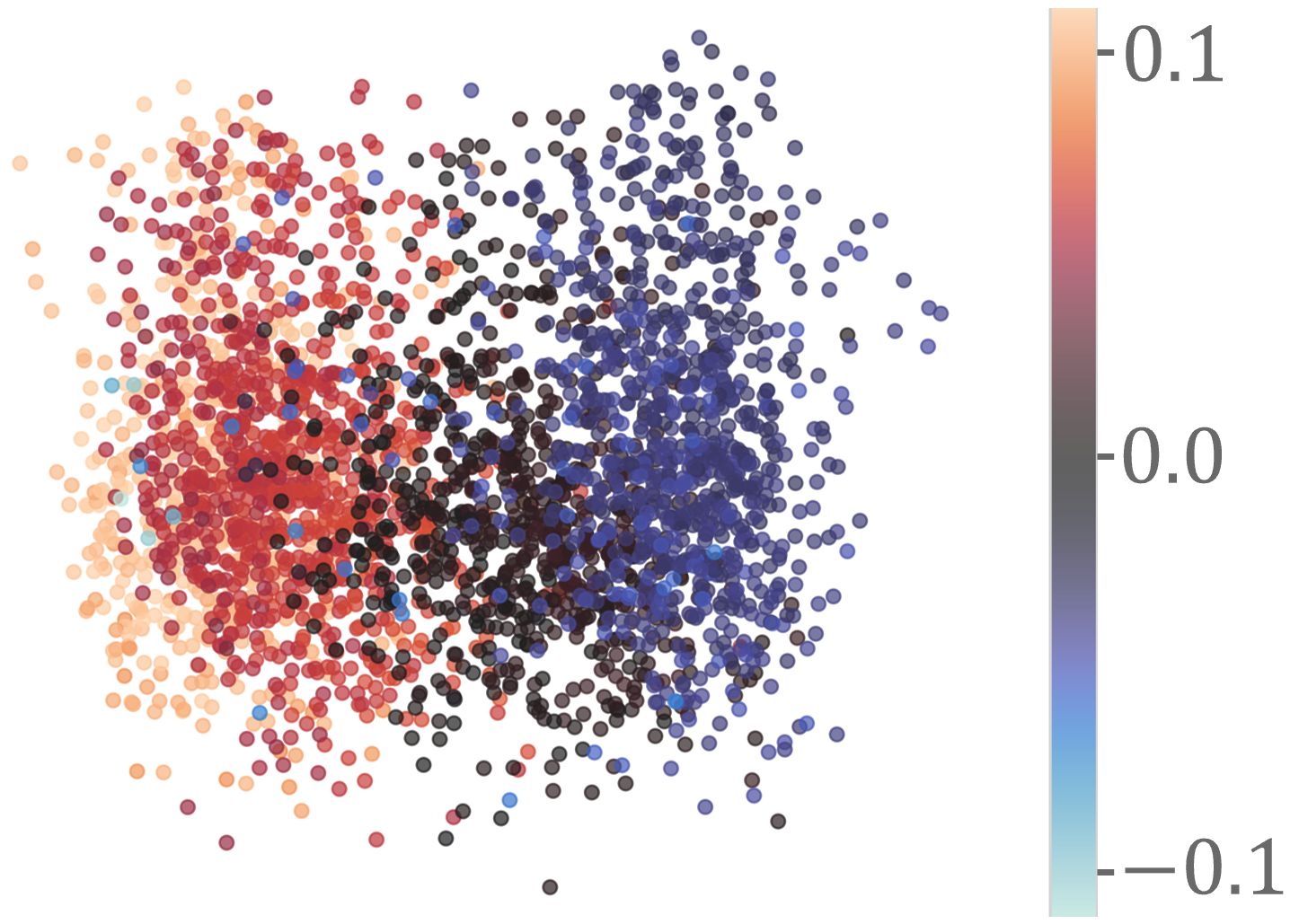}
\end{subfigure}
\hfill
\begin{subfigure}{0.24\textwidth}
    \centering
    \includegraphics[width=\linewidth]{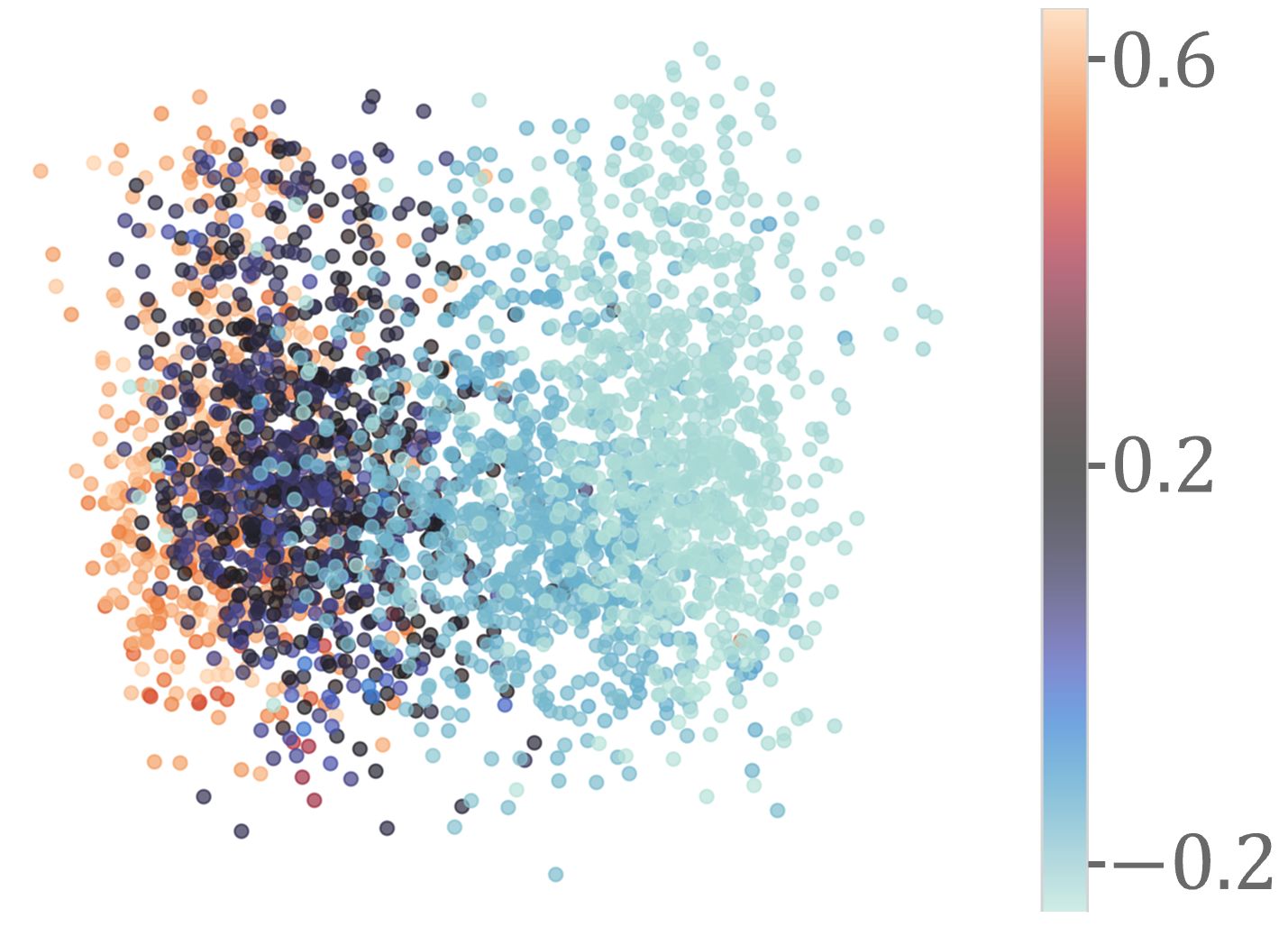}
\end{subfigure}
\hfill
\begin{subfigure}{0.24\textwidth}
    \centering
    \includegraphics[width=\linewidth]{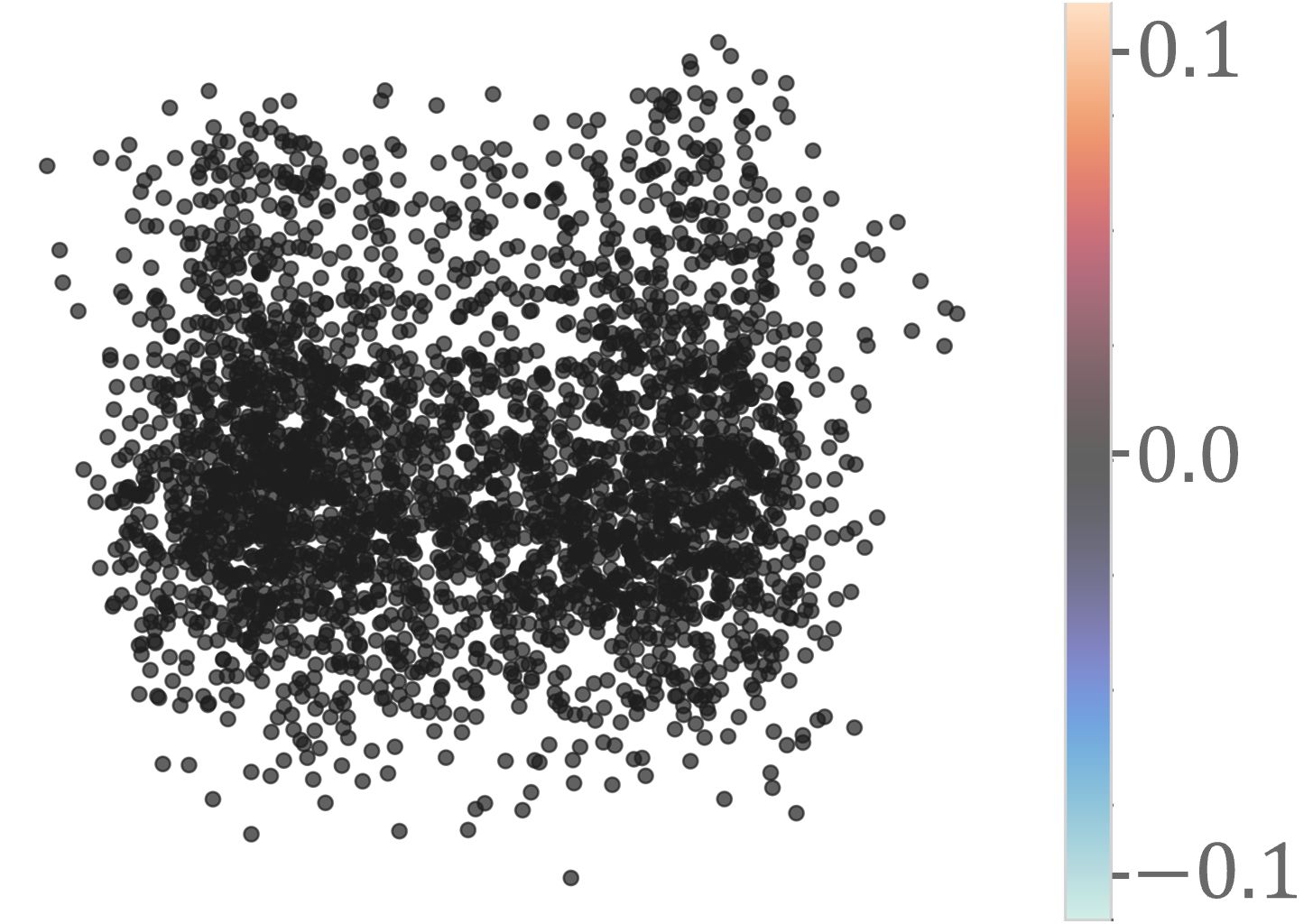}
\end{subfigure}
\caption{Qualitative analysis on the EMT dataset \citep{cook2020context}. Predicted dynamics $\big(t_0$\tball{v4_1},\, $t_1$\tball{v4_2},\, $t_2$\tball{v4_3},\, $t_3$\tball{v4_4}$\big)$ (top) and growth rate (bottom).}
\label{fig:emt}
\end{figure}
\vspace{-5mm}
\begin{table}[H]
\centering
\caption{Quantitative analysis on the EMT dataset \citep{cook2020context}.}
\small
\setlength{\tabcolsep}{2pt}
\resizebox{\textwidth}{!}{
\begin{tabular}{lccccccccc}
\toprule
& \multicolumn{3}{c}{$t=1$} & \multicolumn{3}{c}{$t=2$} & \multicolumn{3}{c}{$t=3$} \\
\cmidrule(lr){2-4} \cmidrule(lr){5-7} \cmidrule(lr){8-10}
Model 
& $\mathcal{W}_1$ & $\mathcal{W}_2$ & Mass
& $\mathcal{W}_1$ & $\mathcal{W}_2$ & Mass
& $\mathcal{W}_1$ & $\mathcal{W}_2$ & Mass \\
\midrule

OTCFM \citep{tong2023improving} 
& $\underline{0.1831_{\pm .0006}}$ & $\underline{0.2502_{\pm .0007}}$ & $0.2849_{\pm .0000}$
& $0.1948_{\pm .0007}$ & ${0.2662_{\pm .0007}}$ & $0.1337_{\pm .0000}$
& ${0.2026_{\pm .0009}}$ & ${0.2744_{\pm .0008}}$ & $0.2812_{\pm .0000}$ \\

TIGON \citep{sha2024reconstructing}
& $0.2705_{\pm .0008}$ & $0.3209_{\pm .0009}$ & $0.2377_{\pm .0010}$
& $0.3377_{\pm .0011}$ & $0.3552_{\pm .0010}$ & $0.2885_{\pm .010}$
& $0.3751_{\pm .0012}$ & $0.3935_{\pm .0011}$ & $0.3126_{\pm .0012}$ \\

DeepRUOT \citep{zhang2025learning}
& $0.2664_{\pm .0023}$ & $0.3654_{\pm .0022}$ & $0.3632_{\pm .0029}$
& $0.2367_{\pm .0024}$ & $0.3260_{\pm .0025}$ & $0.2777_{\pm .0032}$
& $0.3901_{\pm .0028}$ & $0.5513_{\pm .0030}$ & $0.5533_{\pm .0034}$ \\

VarRUOT \citep{sun2025variational}
& $0.1846_{\pm .0018}$ & $0.2533_{\pm .0021}$ & $\underline{0.2234_{\pm .0016}}$
& $\bm{0.1901}_{\pm .0017}$ & $\underline{0.2602_{\pm .0027}}$ & $\underline{0.0577_{\pm .0015}}$
& $\underline{0.1952_{\pm .0020}}$ & $\underline{0.2637_{\pm .0027}}$ & $\underline{0.1152_{\pm .0019}}$ \\

\textbf{CuOT (Ours)}
& $\bm{0.1782_{\pm .0016}}$ & $\bm{0.2424_{\pm .0022}}$ & $\bm{0.0011_{\pm .0005}}$
& $\bm{0.1889_{\pm .0024}}$ & $\bm{0.2581_{\pm .0033}}$ & $\bm{0.0017_{\pm .0014}}$
& $\bm{0.1841_{.0025}}$ & $\bm{0.2487_{\pm .0031}}$ & $\bm{0.0015_{\pm .0013}}$ \\
\bottomrule
\end{tabular}
}
\label{tab:emt}
\end{table}
\vspace{-5mm}
\paragraph{Pancreatic $\beta$-cell differentiation~\citep{veres2019charting} (30D).}
This dataset describes the \emph{in vitro}maturation of human stem cells into insulin-secreting pancreatic $\beta$-cells, a process of direct relevance to diabetes research. The 30-dimensional feature space is obtained via UMAP dimensionality reduction of the full gene expression profile. The dataset presents significant variability across time points, partly attributable to experimental sampling noise inherent to the \emph{in vitro} protocol, and strong mass variation reflecting heterogeneous differentiation rates. It comprises $M = 7$ observed marginals with six intermediate marginals included in training, making it the most densely observed benchmark.
\vspace{-5mm}
\begin{table}[H]
\footnotesize
    \centering
    \caption{Details on the pancreatic $\beta$-cell differentiation dataset \citep{veres2019charting}.}
    \begin{tabular}{ccccccc}
        \toprule
        Time & Mass & Growth Rate $z$ & $\nabla z$ & N samples & $\mathcal{W}_2$ Distance & Distance \% \\ 
        \midrule
        0 & 1.000 & -0.2\%   & -  & 6190  & 5.8191   & 0.709\%  \\
        1 & 0.998 & -42.4\%  & -42.2\%  & 6177  & 181.0527 & 22.062\% \\
        2 & 0.575 & 250.6\%  & 293.0\%  & 3562  & 341.0221 & 41.555\% \\
        3 & 2.016 & -48.1\%  & -298.7\% & 12477 & 141.7822 & 17.277\% \\
        4 & 1.047 & -1.1\%   & 47.0\%   & 6480  & 4.8370   & 0.589\%  \\
        5 & 1.035 & -6.2\%   & -5.1\%   & 6404  & 17.1638  & 2.091\%  \\
        6 & 0.971 & -33.9\%  & -27.7\%  & 6013  & 128.9814 & 15.717\% \\
        7 & 0.642 & - & - & 3971  & - & - \\ 
        \bottomrule
    \end{tabular}
    \label{tab:veres_dataset}
\end{table}
\vspace{-5mm}
\begin{figure}[H]
\centering
\begin{subfigure}{0.24\textwidth}
    \centering
    \caption{\textbf{CuOT (Ours)}}
    \includegraphics[width=\linewidth]{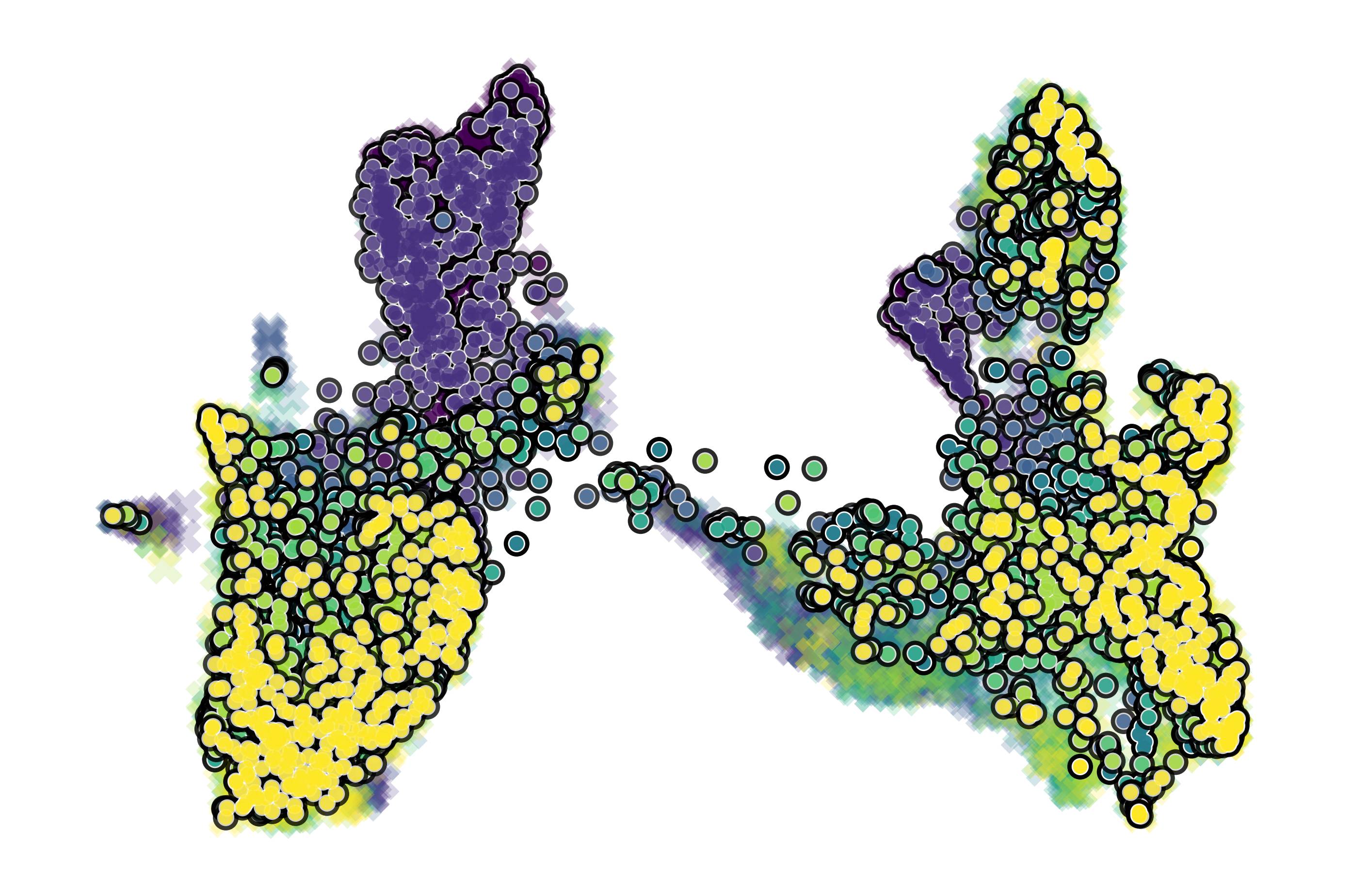}
\end{subfigure}
\hfill
\begin{subfigure}{0.24\textwidth}
    \centering
    \caption{VarRUOT}
    \includegraphics[width=\linewidth]{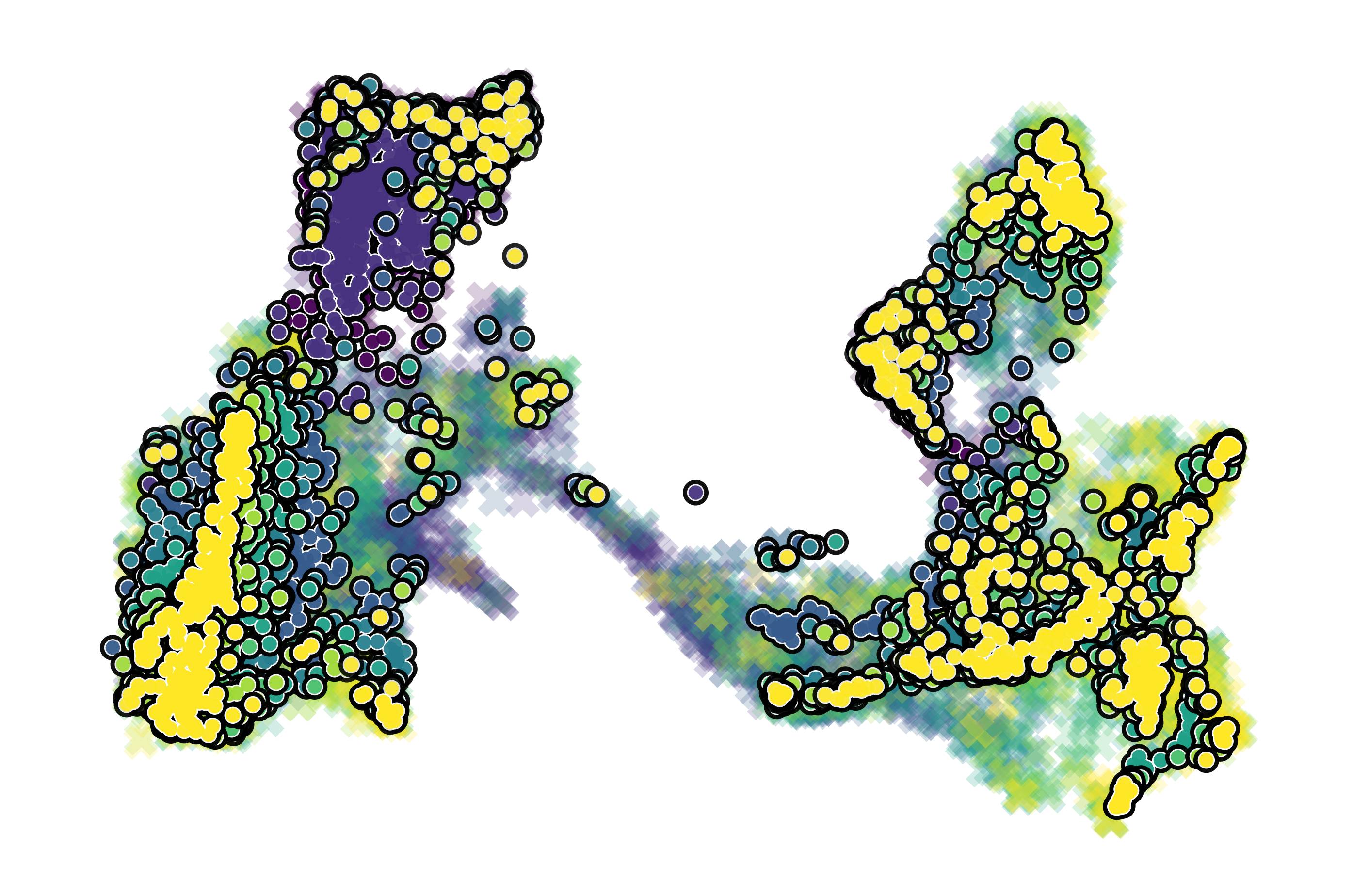}
\end{subfigure}
\hfill
\begin{subfigure}{0.24\textwidth}
    \centering
    \caption{DeepRUOT}
    \includegraphics[width=\linewidth]{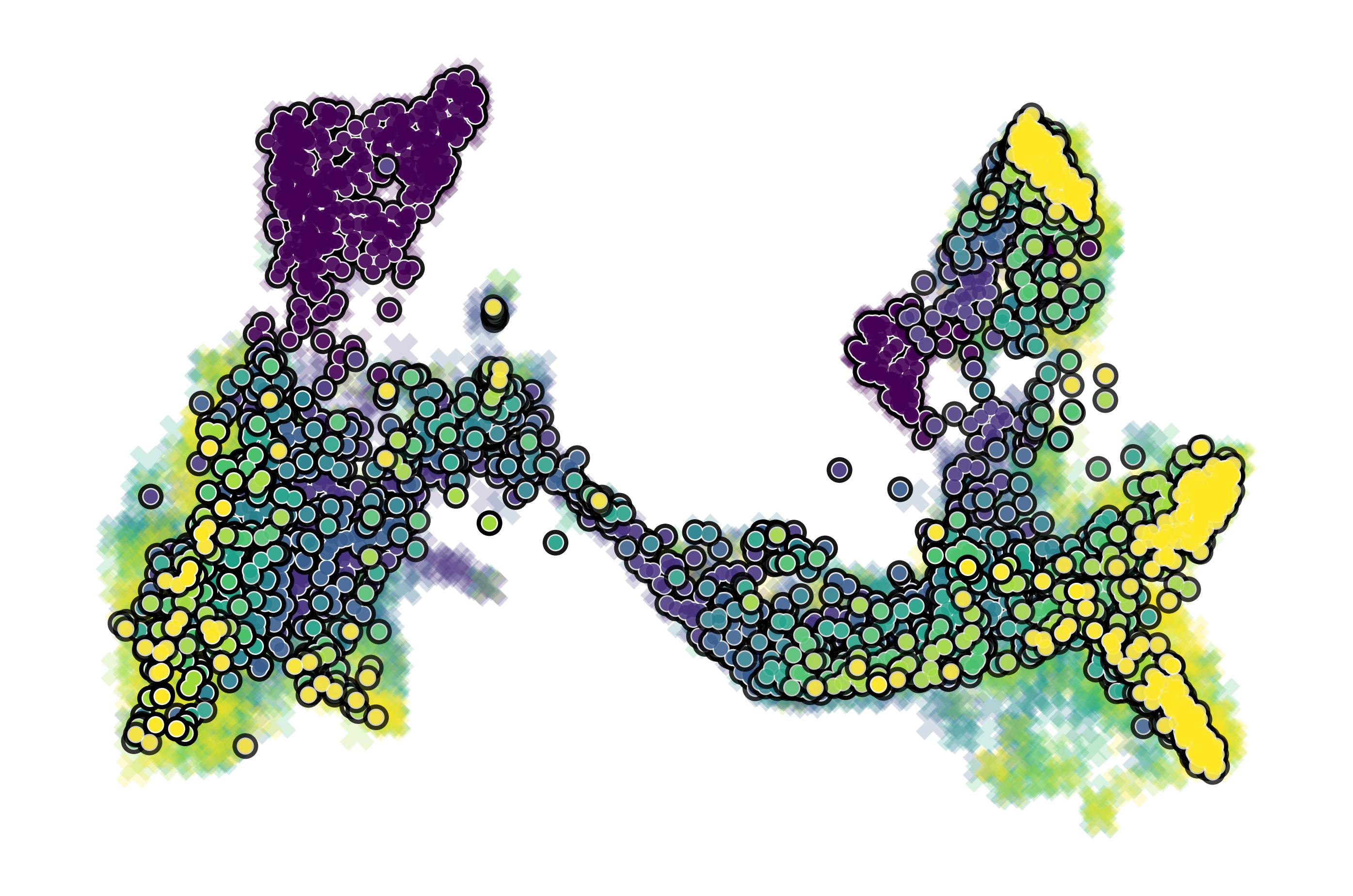}
\end{subfigure}
\hfill
\begin{subfigure}{0.24\textwidth}
    \centering
    \caption{OTCFM}
    \includegraphics[width=\linewidth, height=2.25cm]{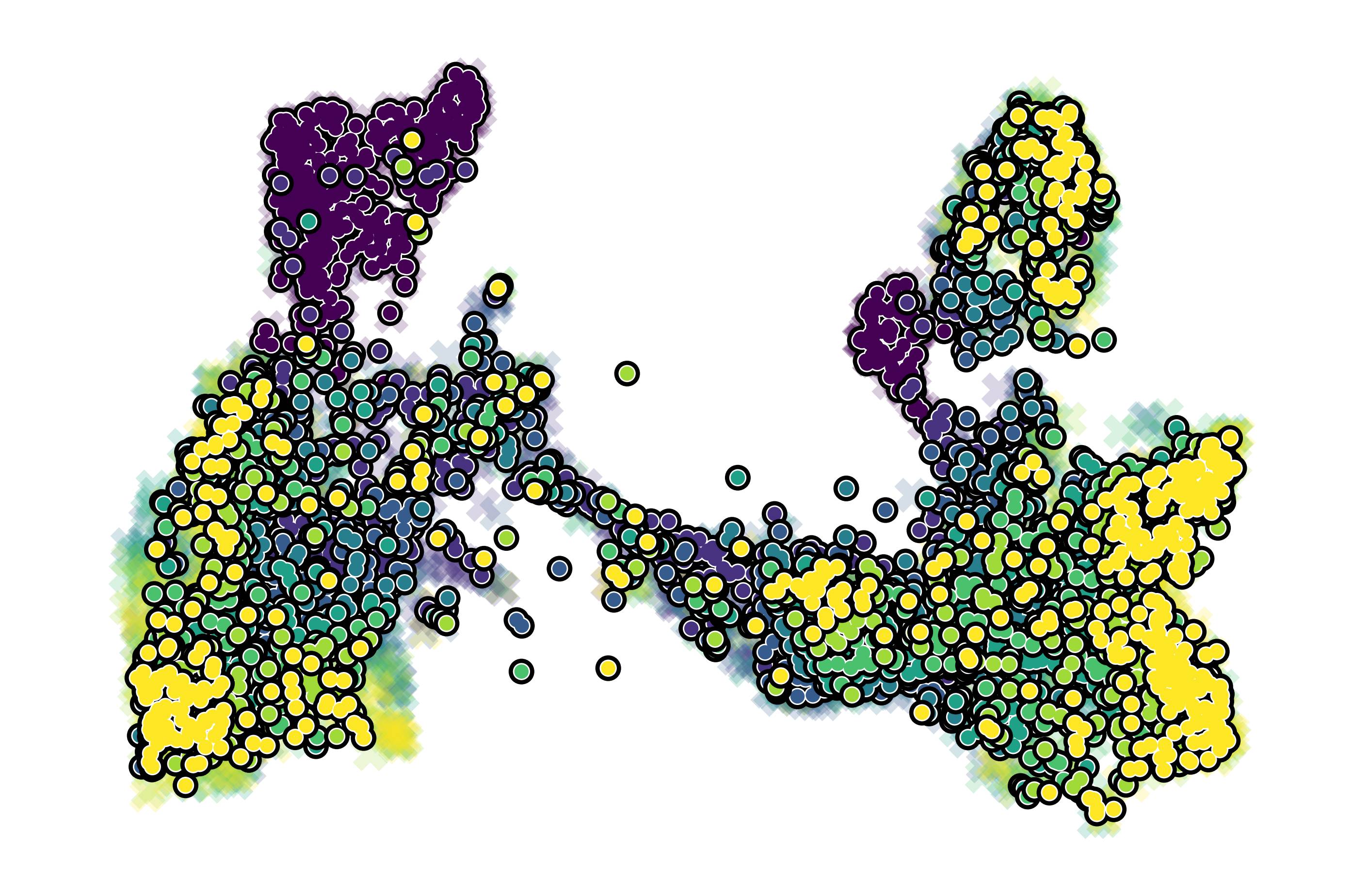}
\end{subfigure}

\vspace{0.0cm}

\begin{subfigure}{0.24\textwidth}
    \centering
    \includegraphics[width=\linewidth]{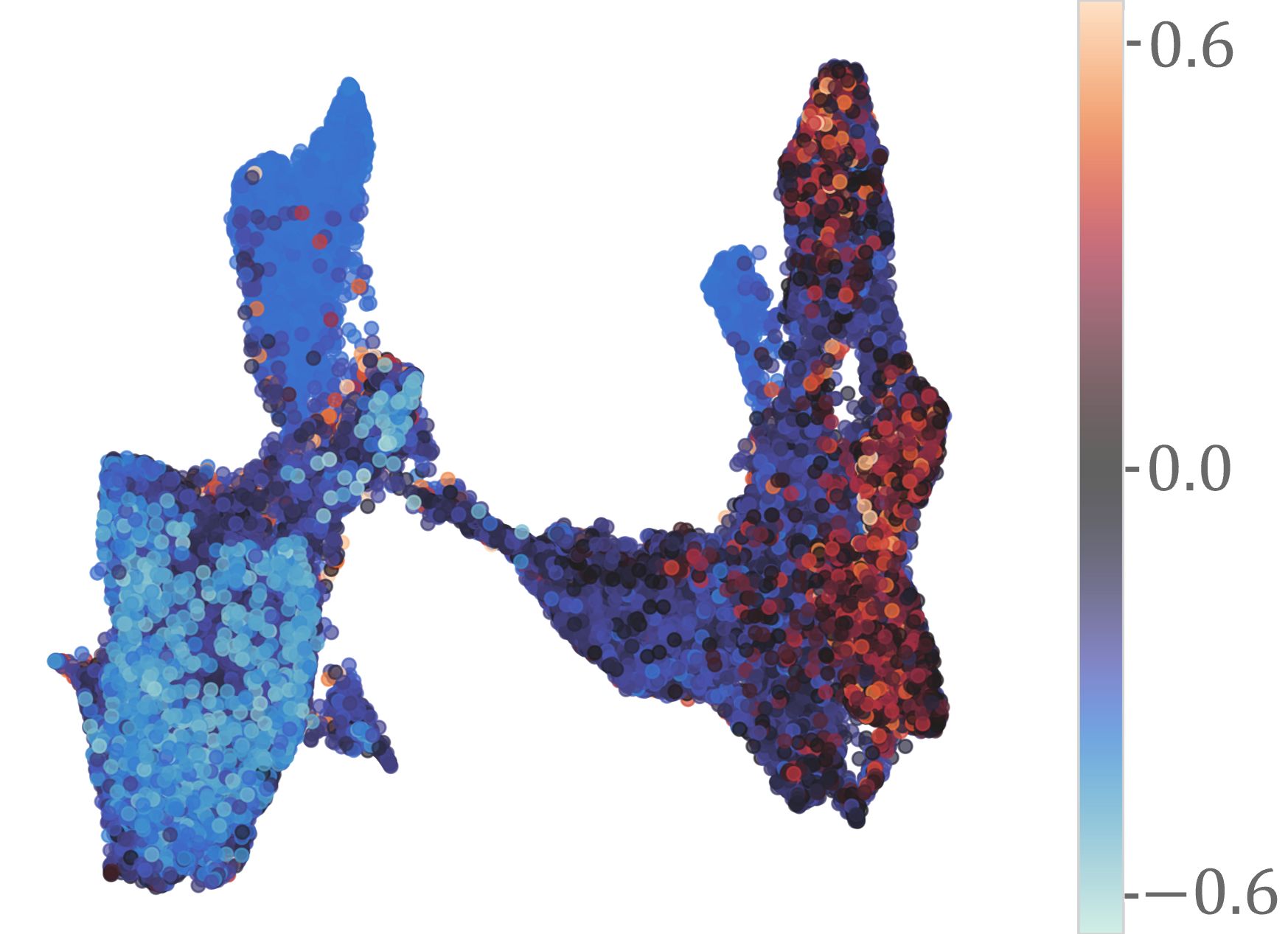}
\end{subfigure}
\hfill
\begin{subfigure}{0.24\textwidth}
    \centering
    \includegraphics[width=\linewidth]{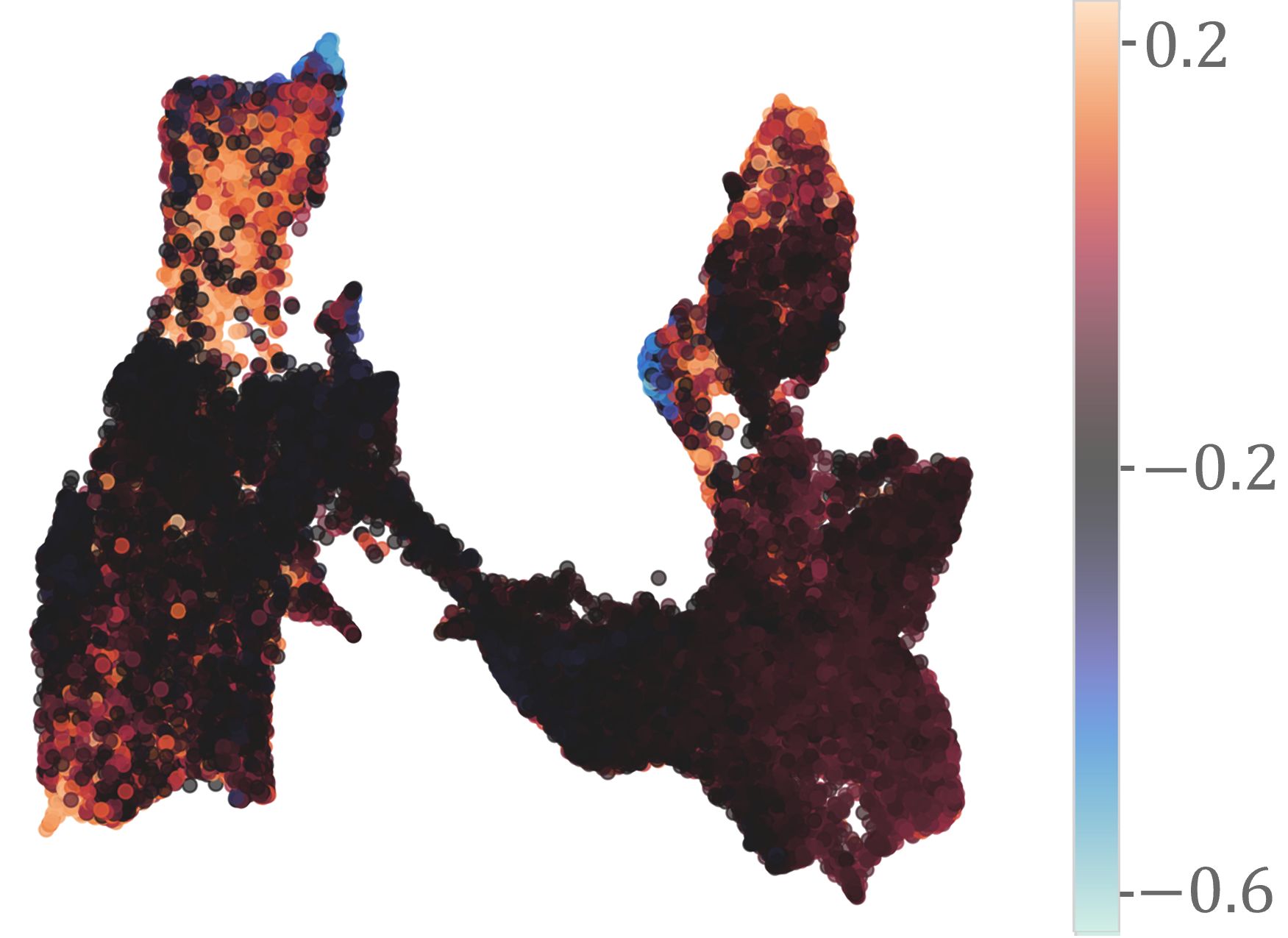}
\end{subfigure}
\hfill
\begin{subfigure}{0.24\textwidth}
    \centering
    \includegraphics[width=\linewidth]{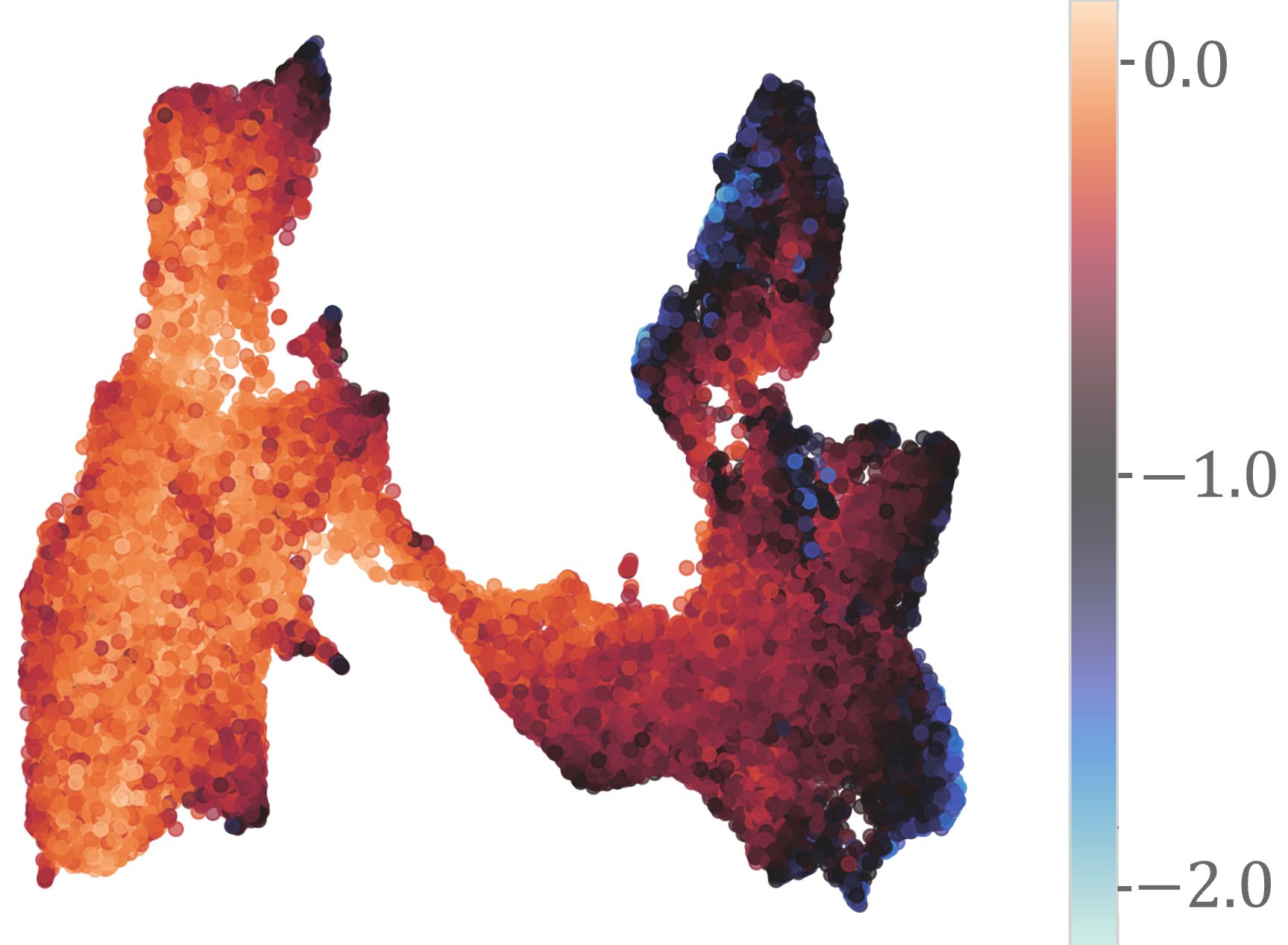}
\end{subfigure}
\hfill
\begin{subfigure}{0.24\textwidth}
    \centering
    \includegraphics[width=\linewidth]{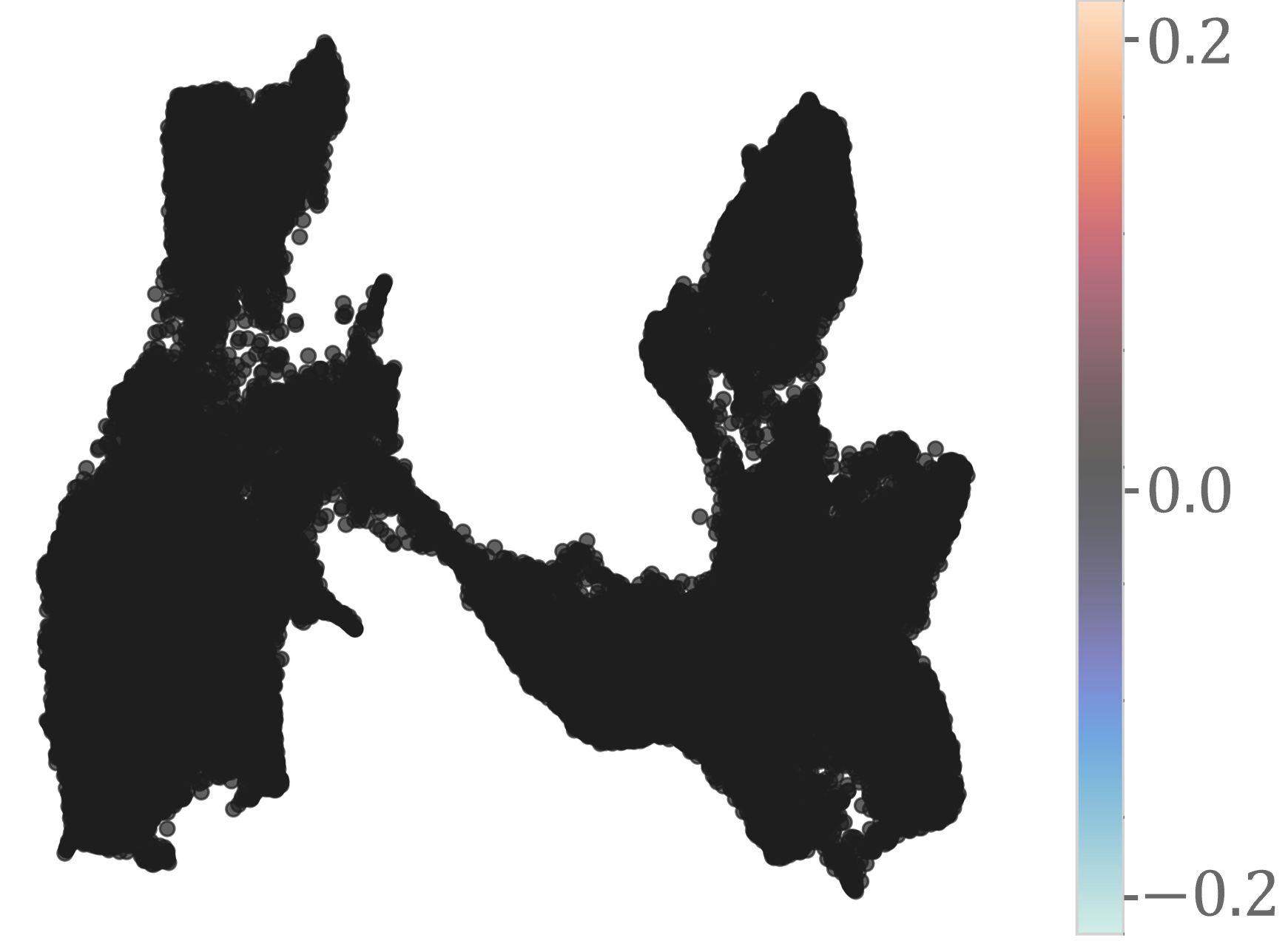}
\end{subfigure}
\caption{Qualitative analysis on the pancreatic $\beta$-cell differentiation dataset \citep{veres2019charting}. Predicted dynamics $\big(t_0$\tball{v8_1},\, $t_1$\tball{v8_2},\, $t_2$\tball{v8_3},\, $t_3$\tball{v8_4},\, $t_4$\tball{v8_5},\, $t_5$\tball{v8_6},\, $t_6$\tball{v8_7},\, $t_7$\tball{v8_8}$\big)$ (top) and growth rate (bottom).}
\label{fig:veres}
\end{figure}
\vspace{-8mm}
\begin{table}[H]
\footnotesize
\centering
\caption{Quantitative analysis on the pancreatic $\beta$-cell differentiation dataset \citep{veres2019charting}.}
\resizebox{\textwidth}{!}{
\begin{tabular}{lccc}
\toprule
 & \multicolumn{3}{c}{$t = 1,2,3,4,5,6,7$ (average)} \\
\cmidrule(lr){2-4}
Method & $\mathcal{W}_1$ & $\mathcal{W}_2$ & Mass \\
\midrule
OTCFM \citep{tong2023improving} & $\underline{7.3684_{\pm 1.3049}}$ & $\underline{5.5259_{\pm 1.0255}}$ & $\underline{0.1921_{\pm 0.3775}}$ \\
TIGON \citep{sha2024reconstructing} & $127.3335_{\pm 39.4399}$ & $16.6406_{\pm 2.4754}$ & $1.0401_{\pm 0.8187}$ \\
DeepRUOT \citep{zhang2025learning} & $11.2069_{\pm 13.3341}$ & $\underline{5.1074_{\pm 5.3032}}$ & $0.5323_{\pm 0.6582}$ \\
VarRUOT \citep{sun2025variational} & $10.2294_{\pm 12.2393}$ & $\underline{4.6187_{\pm 4.8103}}$ & $\underline{0.1897_{\pm 0.3517}}$ \\
\textbf{CuOT (Ours)} & $\bm{4.1082_{\pm 5.6363}}$ & $\bm{2.5572_{\pm 3.0217}}$ & $\bm{0.0889_{\pm 0.1064}}$ \\
\bottomrule
\end{tabular}
}
\label{tab:veres}
\end{table}
\vspace{-8mm}
\begin{table}[h!]
\centering
\caption{Quantitative analysis on the pancreatic $\beta$-cell differentiation dataset \citep{veres2019charting}.
}
\setlength{\tabcolsep}{4pt}
\resizebox{\textwidth}{!}{
\begin{tabular}{llccccccc}
\toprule
Model & Metric & $t=1$ & $t=2$ & $t=3$ & $t=4$ & $t=5$ & $t=6$ & $t=7$ \\
\midrule

\multirow{3}{*}{OTCFM \citep{tong2023improving}}
& $\mathcal{W}_1$
& ${5.3684}_{\pm .0916}$ & $\bm{6.3219_{\pm .0936}}$ & $\bm{6.7157_{\pm .1134}}$ & ${7.5628}_{\pm .1133}$ & ${7.9461}_{\pm .1123}$ & ${8.6920}_{\pm .1120}$ & $\underline{8.6920_{\pm .1152}}$ \\
& $\mathcal{W}_2$
& ${3.9929}_{\pm .0172}$ & $\bm{4.6971_{\pm .0513}}$ & $\bm{4.9720_{\pm .0491}}$ & ${5.6603}_{\pm .0408}$ & ${5.9798}_{\pm .0436}$ & ${6.5607}_{\pm .0343}$ & $\underline{6.8187_{\pm .0478}}$ \\
& Mass
& $\bm{0.0000_{\pm .0000}}$ & ${0.1802}_{\pm .0000}$ & ${1.0316}_{\pm .0000}$ & $\underline{{0.0022}_{\pm .0000}}$ & $\underline{0.0012_{\pm .0000}}$ & $\bm{0.0008_{\pm .0000}}$ & $\underline{0.1285_{\pm .0000}}$ \\

\midrule

\multirow{3}{*}{TIGON \citep{sha2024reconstructing}}
& $\mathcal{W}_1$
& ${105.8188}_{\pm .5371}$ & ${94.6576}_{\pm .7118}$ & ${134.7391}_{\pm .8179}$ & ${127.8763}_{\pm .8631}$ & ${180.0941}_{\pm .8922}$ & ${173.7854}_{\pm .8194}$ & ${74.3633}_{\pm .9463}$ \\
& $\mathcal{W}_2$
& ${18.5314}_{\pm .1283}$ & ${14.3638}_{\pm .1343}$ & ${19.9549}_{\pm .4356}$ & ${16.1346}_{\pm .3465}$ & ${15.8342}_{\pm .4329}$ & ${18.5540}_{\pm .4133}$ & ${13.1114}_{\pm .4520}$ \\
& Mass
& ${0.0088}_{\pm .0008}$ & ${0.46382}_{\pm .0982}$ & ${0.5629}_{\pm .0967}$ & ${0.6783}_{\pm .00983}$ & ${1.4551}_{\pm .0924}$ & ${1.9310}_{\pm .0916}$ & ${2.1807}_{\pm .1002}$ \\

\midrule

\multirow{3}{*}{DeepRUOT \citep{zhang2025learning}}
& $\mathcal{W}_1$
& $\underline{0.1757_{\pm .0001}}$ & ${31.8596}_{\pm .0320}$ & ${22.7228}_{\pm .0525}$ & ${0.4565}_{\pm .0011}$ & ${0.3750}_{\pm .0030}$ & ${0.4328}_{\pm .0005}$ & ${22.4263}_{\pm .0545}$ \\
& $\mathcal{W}_2$
& $\underline{0.2583_{\pm .0001}}$ & ${12.2500}_{\pm .0122}$ & ${9.6641}_{\pm .0229}$ & ${0.9371}_{\pm .0033}$ & ${0.8706}_{\pm .0062}$ & ${0.8383}_{\pm .0043}$ & ${10.9334}_{\pm .0110}$ \\
& Mass
& ${0.0258}_{\pm .0000}$ & $\bm{0.0238_{\pm .0000}}$ & ${2.0108}_{\pm .0015}$ & ${0.3471}_{\pm .0016}$ & ${0.5010}_{\pm .0029}$ & ${0.5659}_{\pm .0037}$ & ${0.2518}_{\pm .0027}$ \\

\midrule

\multirow{3}{*}{VarRUOT \citep{sun2025variational}}
& $\mathcal{W}_1$
& ${0.2518}_{\pm .0006}$ & ${30.3818}_{\pm .4328}$ & ${21.6480}_{\pm .1574}$ & $\underline{0.3557_{\pm .0043}}$ & $\underline{0.2827_{\pm .0041}}$ & $\underline{0.2764_{\pm .0026}}$ & ${18.4096}_{\pm .3291}$ \\
& $\mathcal{W}_2$
& ${0.3479}_{\pm .0006}$ & ${11.6851}_{\pm .1553}$ & ${9.0374}_{\pm .0239}$ & $\underline{0.8773_{\pm .0134}}$ & $\underline{0.6406_{\pm .0144}}$ & $\underline{0.5941_{\pm .0087}}$ & ${9.1489}_{\pm .0627}$ \\
& Mass
& $\bm{0.0000_{\pm .0000}}$ & ${0.1842}_{\pm .0035}$ & ${1.0129}_{\pm .0060}$ & $\bm{0.0014_{\pm .0005}}$ & $\underline{0.0009_{\pm .0007}}$ & $\underline{0.0014_{\pm .0017}}$ & $\underline{0.1270_{\pm .0267}}$ \\

\midrule

\multirow{3}{*}{\textbf{CuOT (Ours)}}
& $\mathcal{W}_1$
& $\bm{0.1225_{\pm .0023}}$ & $\underline{13.6947}_{\pm .0247}$ & $\underline{9.9932}_{\pm .0361}$ & $\bm{0.2247_{\pm .0055}}$ & $\bm{0.2158_{\pm .0050}}$ & $\bm{0.1552_{\pm .0042}}$ & $\bm{8.3567_{\pm .0368}}$ \\
& $\mathcal{W}_2$
& $\bm{0.1196_{\pm .0011}}$ & $\underline{7.1560_{\pm .0187}}$ & $\underline{5.4073_{\pm .0239}}$ & $\bm{0.6094_{\pm .0025}}$ & $\bm{0.4372_{\pm .0028}}$ & $\bm{0.5715_{\pm .0037}}$ & $\bm{5.9001_{\pm .0102}}$ \\
& Mass
& $\underline{0.0148_{\pm .0009}}$ & $\underline{0.1644_{\pm .0084}}$ & $\bm{0.2840_{\pm .0088}}$ & ${0.1172}_{\pm .0087}$ & $\bm{0.0000_{\pm .0001}}$ & ${0.0424}_{\pm .0032}$ & $\bm{0.0000_{\pm .0001}}$ \\

\bottomrule
\end{tabular}
}
\label{tab:extended_veres}
\end{table}
\newpage
\paragraph{Zebrafish embryogenesis~\citep{farrell2018single} (50D).}
This dataset captures the early development of a zebrafish embryo, during which a population of initially undifferentiated cells diversifies into 25 distinct specialized cell types over approximately 9 hours. The 50-dimensional feature space is obtained via UMAP reduction of the full transcriptome. This is the highest-dimensional single-cell benchmark, characterized by rapid and highly heterogeneous proliferation across branching cell lineages, and presents the most challenging mass variation profile. The dataset comprises $M = 6$ observed marginals with five intermediate marginals included in training.
\vspace{-5mm}
\begin{table}[H]
\footnotesize
    \centering
    \caption{Details on the zebrafish embryogenesis dataset \cite{farrell2018single}.}
    \begin{tabular}{ccccccc}
        \toprule
        Time & Mass & Growth Rate $z$ & $\nabla z$ & N samples & $\mathcal{W}_2$ Distance & Distance \% \\ 
        \midrule
        0 & 1.000 & 82.3\%  & -       & 3136 & 43944.6406 & 41.364\% \\
        1 & 1.823 & -10.3\% & -92.6\% & 5716 & 1106.7998  & 6.564\%  \\
        2 & 1.635 & 20.5\%  & 30.8\%  & 5127 & 3699.7798  & 12.002\% \\
        3 & 1.970 & -11.9\% & -32.4\% & 6178 & 2166.1155  & 9.183\%  \\
        4 & 1.735 & 30.8\%  & 42.7\%  & 5442 & 8879.2041  & 18.593\% \\
        5 & 2.269 & -15.4\% & -46.2\% & 7114 & 3879.7812  & 12.290\% \\
        6 & 1.919 & -       & -       & 6018 & -          & -        \\ 
        \bottomrule
    \end{tabular}
    \label{tab:zebra_dataset}
\end{table}
\vspace{-5mm}
\begin{table}[h!]
\centering
\caption{Quantitative analysis on the zebrafish embryogenesis dataset \citep{farrell2018single}.
}
\setlength{\tabcolsep}{4pt}
\resizebox{\textwidth}{!}{
\begin{tabular}{llcccccc}
\toprule
Model & Metric & $t=1$ & $t=2$ & $t=3$ & $t=4$ & $t=5$ & $t=6$ \\
\midrule

\multirow{3}{*}{OTCFM \citep{tong2023improving}}
& $\mathcal{W}_1$
& ${4.0287}_{\pm .0151}$ & ${4.9408}_{\pm .0165}$ & ${5.7038}_{\pm .0535}$ & ${7.1590}_{\pm .2742}$ & ${8.4766}_{\pm .3243}$ & ${10.7044}_{\pm .3525}$ \\
& $\mathcal{W}_2$
& $\underline{3.4801_{\pm .0105}}$ & ${3.9062}_{\pm .0057}$ & ${4.4139}_{\pm .0640}$ & ${5.3939}_{\pm .2031}$ & ${6.2450}_{\pm .3327}$ & ${7.1066}_{\pm .3944}$ \\
& Mass
& ${0.6768}_{\pm 0.0000}$ & ${0.4030}_{\pm .0000}$ & ${0.9409}_{\pm .0000}$ & ${0.5407}_{\pm .0000}$ & ${1.6090}_{\pm .0000}$ & ${0.8445}_{\pm .0000}$ \\

\midrule

\multirow{3}{*}{TIGON \citep{sha2024reconstructing}}
& $\mathcal{W}_1$
& ${6.8247}_{\pm .2426}$ & ${6.6419}_{\pm .3035}$ & ${7.7523}_{\pm .4264}$ & ${8.4695}_{\pm .4572}$ & ${9.1028}_{\pm .4876}$ & ${10.7916}_{\pm .4237}$ \\
& $\mathcal{W}_2$
& ${5.5471}_{\pm .1314}$ & ${5.3589}_{\pm .1298}$ & ${5.9637}_{\pm .4412}$ & ${7.1284}_{\pm .3519}$ & ${9.8475}_{\pm .4371}$ & ${9.5468}_{\pm .4186}$ \\
& Mass
& ${0.5733}_{\pm .0134}$ & ${0.3638}_{\pm .0257}$ & ${0.6629}_{\pm .0380}$ & ${0.4783}_{\pm .0192}$ & ${0.4551}_{\pm .0202}$ & ${0.6310}_{\pm .0418}$ \\

\midrule

\multirow{3}{*}{DeepRUOT \citep{zhang2025learning}}
& $\mathcal{W}_1$
& $\underline{3.7562_{\pm .0430}}$ & $\underline{3.9570_{\pm .1302}}$ & $\underline{4.2795_{\pm .01279}}$ & $\underline{4.9962_{\pm .1367}}$ & $\underline{5.7769_{\pm .1698}}$ & $\underline{6.9029_{\pm .1804}}$ \\
& $\mathcal{W}_2$
& ${3.5913}_{\pm .0332}$ & $\underline{3.5195_{\pm .0916}}$ & $\underline{3.8945_{\pm .1382}}$ & $\underline{4.3707_{\pm .1471}}$ & $\underline{4.9568_{\pm .1653}}$ & $\underline{6.2108_{\pm .1690}}$ \\
& Mass
& $\underline{0.4842_{\pm .0131}}$ & $\underline{0.0930_{\pm .0103}}$ & $\underline{0.1680_{\pm .0127}}$ & $\underline{0.0078_{\pm .0020}}$ & $\underline{0.0189_{\pm .0094}}$ & $\underline{0.3214_{\pm .0452}}$ \\

\midrule

\multirow{3}{*}{VarRUOT \citep{sun2025variational}}
& $\mathcal{W}_1$
& ${4.6608}_{\pm .0712}$ & ${4.4166}_{\pm .2412}$ & ${4.6337}_{\pm .3522}$ & ${5.1508}_{\pm .3319}$ & $\underline{5.7901_{\pm .7662}}$ & $\underline{6.8333_{\pm .6958}}$ \\
& $\mathcal{W}_2$
& ${4.4311}_{\pm .0290}$ & ${4.0036}_{\pm .3274}$ & ${4.3000}_{\pm .3541}$ & ${4.6924}_{\pm .3099}$ & ${5.1680}_{\pm .3128}$ & $\underline{6.2095_{\pm .2757}}$ \\
& Mass
& $\underline{0.4744_{\pm .0114}}$ & ${0.1152}_{\pm .0189}$ & ${0.3929}_{\pm .0234}$ & ${0.1639}_{\pm .0340}$ & ${0.9363}_{\pm .0394}$ & ${0.4324}_{\pm .0496}$ \\

\midrule

\multirow{3}{*}{\textbf{CuOT (Ours)}}
& $\mathcal{W}_1$
& $\bm{3.1730_{\pm .0356}}$ & $\bm{2.6959_{\pm .0828}}$ & $\bm{2.8354_{\pm .1037}}$ & $\bm{3.2731_{\pm .1002}}$ & $\bm{3.4931_{\pm .1696}}$ & $\bm{4.1428_{\pm .1790}}$ \\
& $\mathcal{W}_2$
& $\bm{3.0379_{\pm .0296}}$ & $\bm{2.3572_{\pm .0653}}$ & $\bm{2.8226_{\pm .1378}}$ & $\bm{3.2909_{\pm .1579}}$ & $\bm{3.08173_{\pm .1473}}$ & $\bm{3.6236_{\pm .1342}}$ \\
& Mass
& $\bm{0.0345_{\pm .0201}}$ & $\bm{0.0373_{\pm .0298}}$ & $\bm{0.0168_{\pm .0257}}$ & $\bm{0.0107_{\pm .0063}}$ & $\bm{0.0257_{\pm .0174}}$ & $\bm{0.0203_{\pm .0148}}$ \\

\bottomrule
\end{tabular}
}
\label{tab:extended_zebrafish}
\end{table}

\newpage
\subsection{Unmatched Datasets Pairing}
\label{app:dataset-pairing}
The \href{https://github.com/NVlabs/ffhq-dataset}{Flickr-Faces-HQ (FFHQ)} dataset is a high-resolution ($1024 \times 1024$) collection of human faces that exhibits a remarkably wide range of visual variations \citep{karras2019style}. We follow the experimental protocol adopted by \cite{gushchin2024light}, using their publicly available datasets from \small{\url{https://github.com/ngushchin/LightSB}}\normalsize.
We split the dataset into training (first $60K$ images) and testing (last $10K$ images) subsets. Each subset is further partitioned into age groups, specifically \textit{adults} and \textit{children}, using the per-image age labels provided by the \href{https://github.com/DCGM/ffhq-features-dataset}{FFHQ-Features} dataset \citep{karras2019style}, which supplies facial attribute annotations for the full FFHQ collection.
These two groups correspond to the reference marginal distributions $\rho^m$, yielding an unbalanced, unpaired matching problem: the training split contains approximately $45K$ adult samples and $15K$ child samples. 
The goal of this experiment is to synthesize age-transformed images of the same individual. Specifically, given an adult image, the model generates a realistic depiction of that person as a child, and conversely, given a child image, it produces a plausible adult version of the same individual.
To train our solvers using the MLP architectures common across all baseline methods, we operate within the latent space of pre-trained autoencoders. To evaluate sensitivity to the choice of latent representation, we consider two encoder backbones: (i) the ALAE encoder \citep{pidhorskyi2020adversarial}, which compresses a $1024 \times 1024$ image into a $512$-dimensional vector, and (ii) the e4e encoder \citep{richardson2021encoding}, which maps images into the StyleGAN latent space by generating a distinct $512$-dimensional style code for each of the $18$ generator layers. This $18$-fold increase in feature dimensionality ($9216$ dimensions in total) necessitates scaling up the baseline model capacities accordingly. During inference, an input image is mapped to the latent space, transformed via the learned transport map, and reconstructed through the corresponding decoder to produce the age-transformed result. Full hyperparameter configurations for both setups are detailed in Table~\ref{tab:hyperparameters}.

The experiment defines two complementary transport directions. The Adult $\to$ Child task requires accurate early-stage dynamics in a high-density source distribution to ensure correct transport to the right mode in the lower-density target mode. Because the adult distribution is roughly three times larger than the child distribution, correctly learning the initial transport direction is fundamental to retrieve the correct equivalent mode in the target. Conversely, the Child $\to$ Adult task demands high-fidelity reconstructions and correct coverage of the broader adult distribution support, as the model must generate plausible, diverse adult appearances from a compact source.

We evaluate all methods using FID, and the latent-space Wasserstein distances $\mathcal{W}_1$ and $\mathcal{W}_2$, together with a mass discrepancy term.
FID is computed directly in pixel space using the standard InceptionV3 feature extractor
(2048-dimensional pool layer), following the formulation,
\begin{equation}
\label{eq:fid}
    \mathrm{FID} = \|\mu_r - \mu_g\|^2 + \mathrm{Tr}\!\left(\Sigma_r + \Sigma_g
    - 2\left(\Sigma_r \Sigma_g\right)^{1/2}\right),
\end{equation}
where $(\mu_r, \Sigma_r)$ and $(\mu_g, \Sigma_g)$ are the mean and covariance of the InceptionV3 activations computed over the real and generated test images, respectively.
The matrix square root is computed via Schur decomposition, with the real part taken whenever the result contains negligible imaginary components due to numerical precision.

\begin{figure}[H]
\centering
\begin{tikzpicture}
    \node[anchor=south west, inner sep=0] (grid) at (0,0)
    {\includegraphics[width=0.8\linewidth]{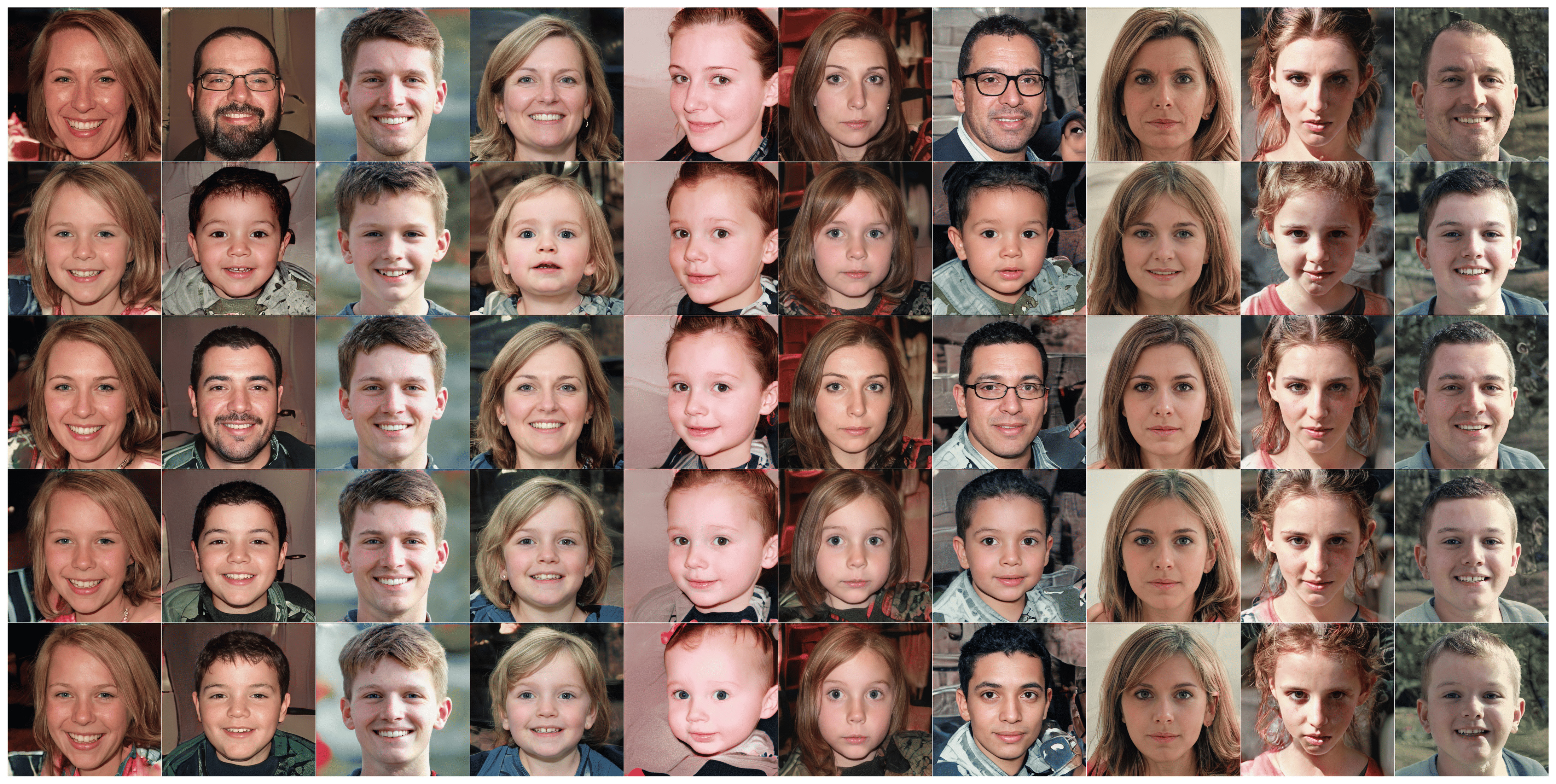}};
    \node[left] at (-0.0,5.0) { \footnotesize Input};
    \node[left] at (-0.0,3.9) {{ \footnotesize CuOT { (Ours)}}};
    \node[left] at (-0.0,2.8) {\footnotesize VarRUOT};
    \node[left] at (-0.0,1.7) {\footnotesize DeepRUOT};
    \node[left] at (-0.0,0.6) {\footnotesize OTCFM};
\end{tikzpicture}
\vspace{-2mm}
\caption{Adult $\to$ Child image generation experiment on the FFHQ dataset \citep{karras2019style}, ALAE encoder \citep{pidhorskyi2020adversarial}. \looseness-1}
\label{fig:ffhq_a2c}
\end{figure}

\begin{figure}[H]
\centering
\begin{tikzpicture}
    \node[anchor=south west, inner sep=0] (grid) at (0,0)
    {\includegraphics[width=0.9\linewidth]{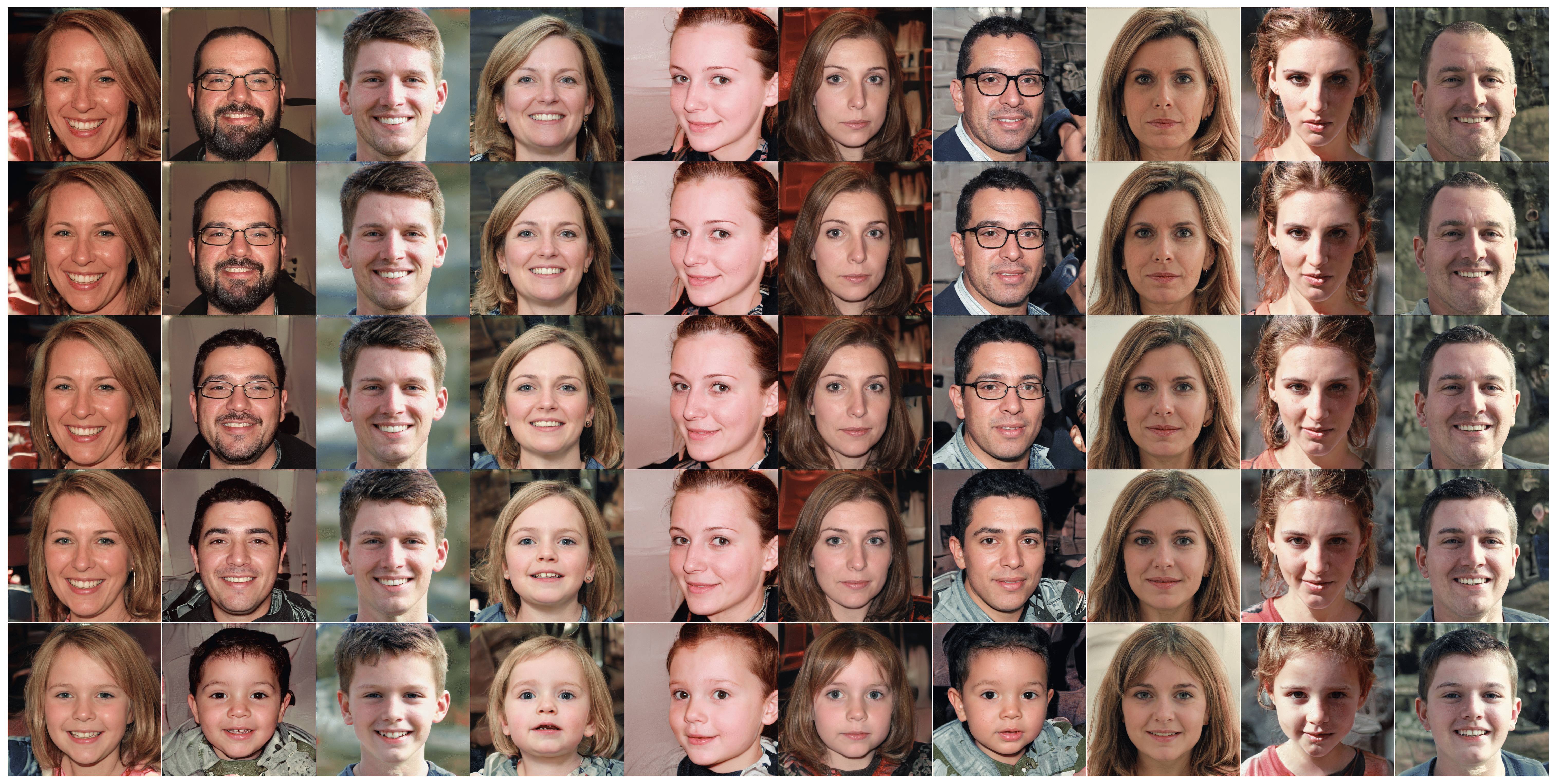}};
    \node[left] at (-0.2,5.7) {$t_0$};
    \node[left] at (-0.2,4.4) {$t_1$};
    \node[left] at (-0.2,3.2) {$t_2$};
    \node[left] at (-0.2,1.9) {$t_3$};
    \node[left] at (-0.2,0.7) {$t_4$};
\end{tikzpicture}
\caption{Visualization of the bridge computed on the FFHQ dataset \citep{karras2019style} by our CuOT method (ALAE encoder \citep{pidhorskyi2020adversarial}).}
\label{fig:ffhq_contactubs}
\end{figure}

\begin{figure}[H]
\centering
\begin{tikzpicture}
    \node[anchor=south west, inner sep=0] (grid) at (0,0)
    {\includegraphics[width=0.9\linewidth]{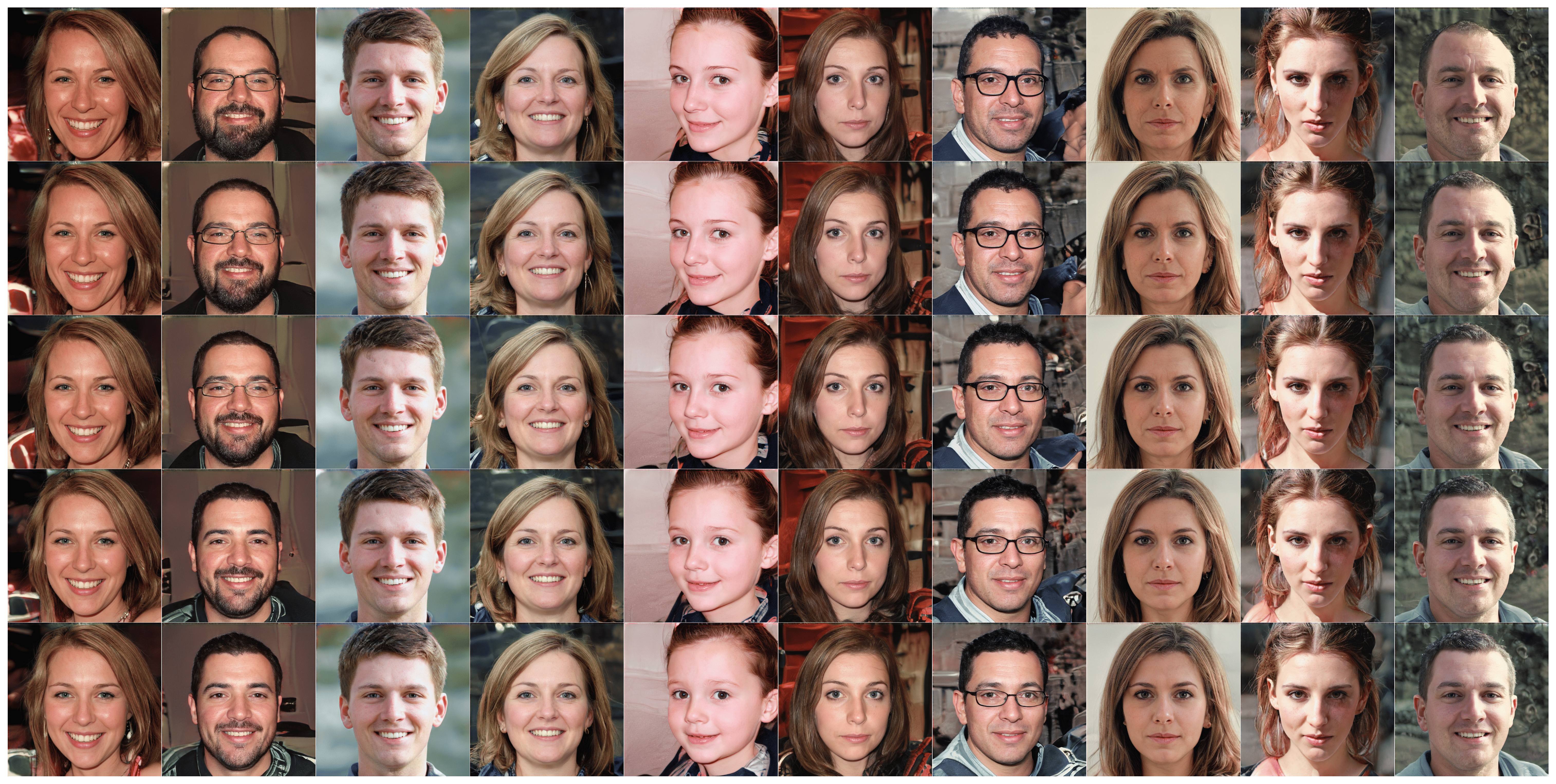}};
    \node[left] at (-0.2,5.7) {$t_0$};
    \node[left] at (-0.2,4.4) {$t_1$};
    \node[left] at (-0.2,3.2) {$t_2$};
    \node[left] at (-0.2,1.9) {$t_3$};
    \node[left] at (-0.2,0.7) {$t_4$};
\end{tikzpicture}
\caption{Visualization of the bridge computed on the FFHQ dataset \citep{karras2019style} by the VarRUOT method \cite{sun2025variational} (ALAE encoder \citep{pidhorskyi2020adversarial}).}
\label{fig:ffhq_varruot}
\end{figure}

\begin{figure}[H]
\centering
\begin{tikzpicture}
    \node[anchor=south west, inner sep=0] (grid) at (0,0)
    {\includegraphics[width=0.9\linewidth]{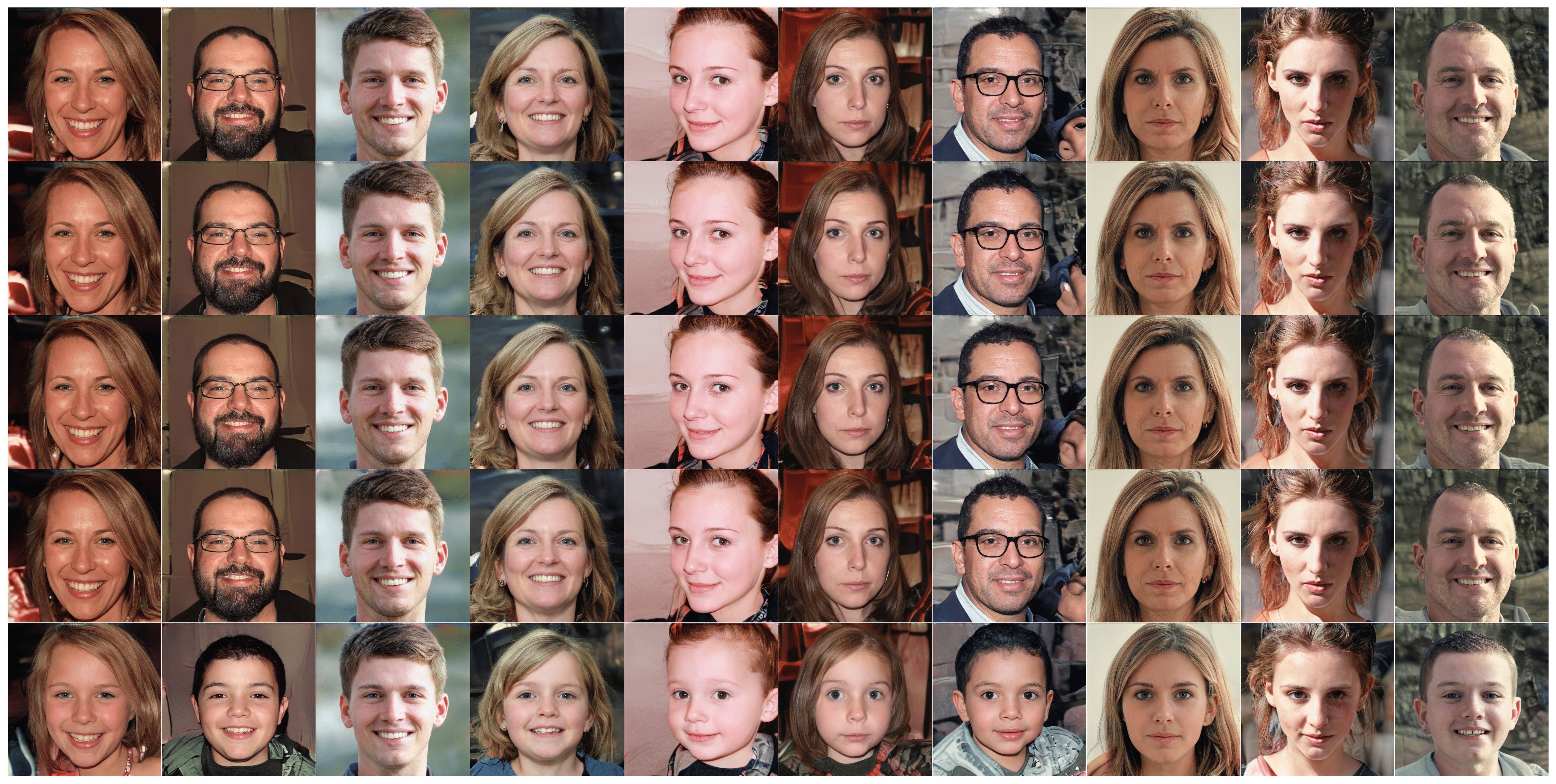}};
    \node[left] at (-0.2,5.7) {$t_0$};
    \node[left] at (-0.2,4.4) {$t_1$};
    \node[left] at (-0.2,3.2) {$t_2$};
    \node[left] at (-0.2,1.9) {$t_3$};
    \node[left] at (-0.2,0.7) {$t_4$};
\end{tikzpicture}
\caption{Visualization of the bridge computed on the FFHQ dataset \citep{karras2019style} by the DeepRUOT method \cite{zhang2025learning} (ALAE encoder \citep{pidhorskyi2020adversarial}).}
\label{fig:ffhq_deep}
\end{figure}

\begin{figure}[H]
\centering
\begin{tikzpicture}
    \node[anchor=south west, inner sep=0] (grid) at (0,0)
    {\includegraphics[width=0.9\linewidth]{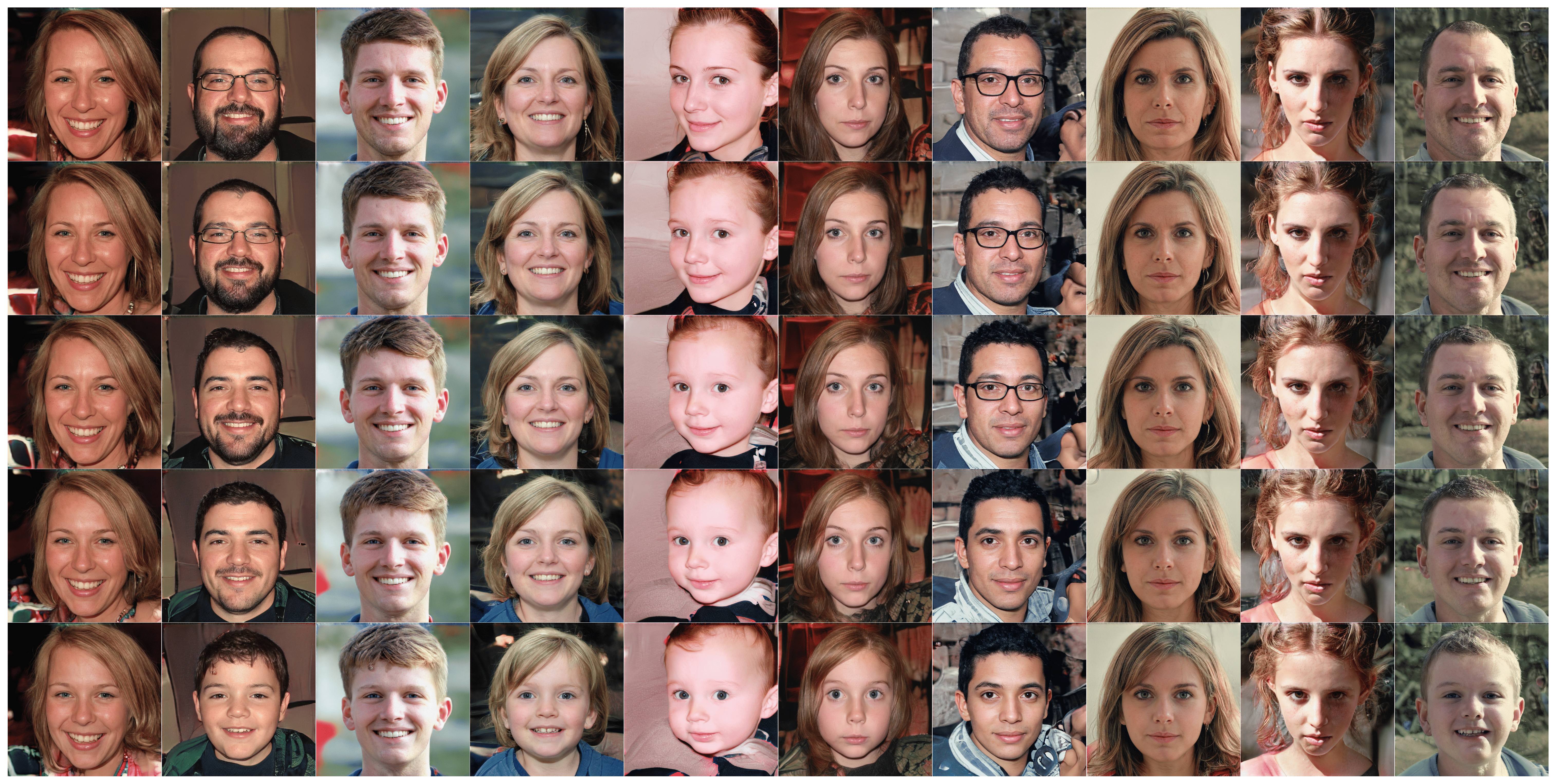}};
    \node[left] at (-0.2,5.7) {$t_0$};
    \node[left] at (-0.2,4.4) {$t_1$};
    \node[left] at (-0.2,3.2) {$t_2$};
    \node[left] at (-0.2,1.9) {$t_3$};
    \node[left] at (-0.2,0.7) {$t_4$};
\end{tikzpicture}
\caption{Visualization of the bridge computed on the FFHQ dataset \citep{karras2019style} by the OTCFM method \cite{tong2023improving} (ALAE encoder \citep{pidhorskyi2020adversarial}).}
\label{fig:ffhq_otcfm}
\end{figure}

\begin{figure}[H]
\centering
\begin{tikzpicture}
    \node[anchor=south west, inner sep=0] (grid) at (0,0)
    {\includegraphics[width=0.9\linewidth]{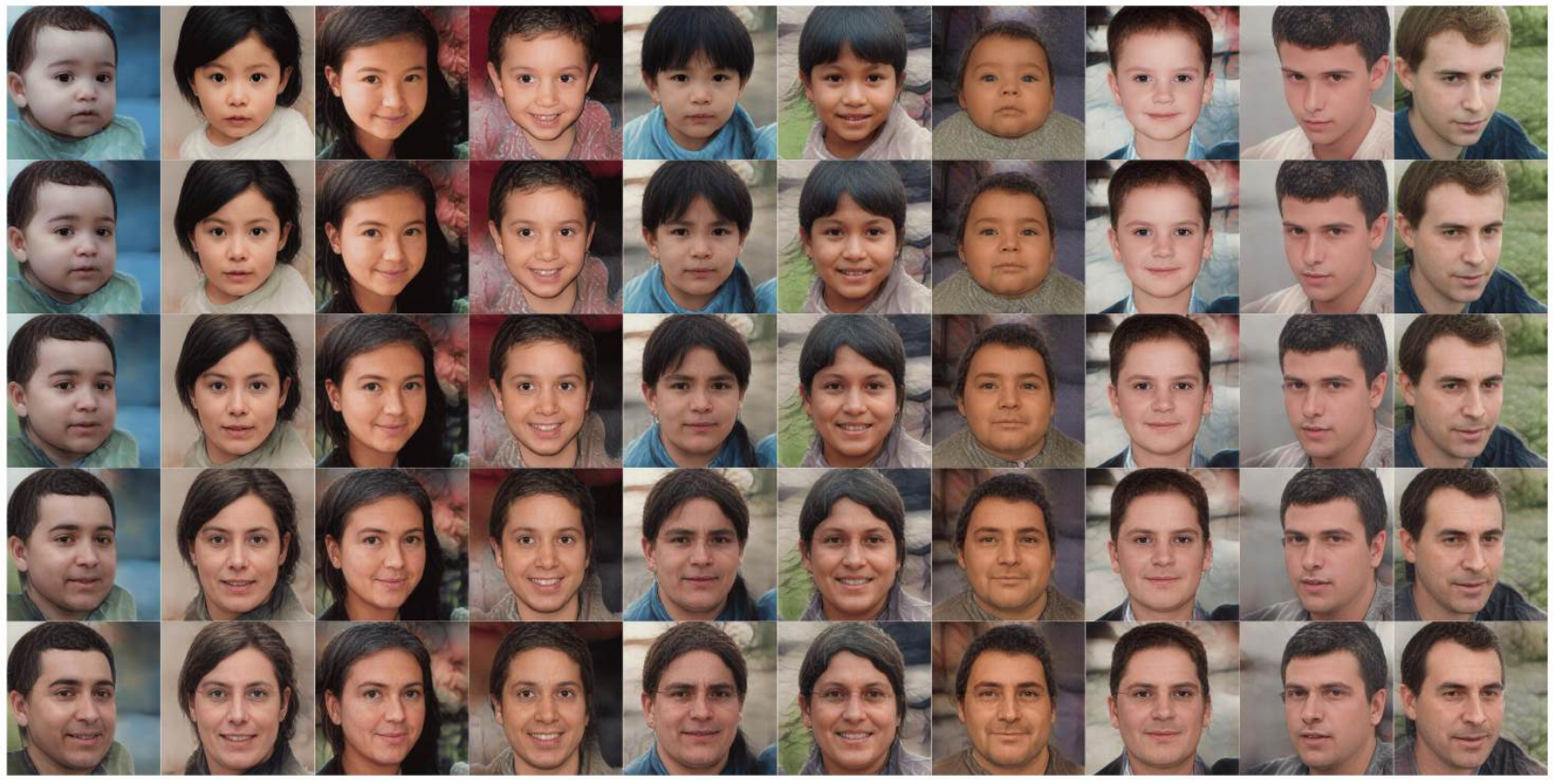}};
    \node[left] at (-0.2,5.7) {$t_0$};
    \node[left] at (-0.2,4.4) {$t_1$};
    \node[left] at (-0.2,3.2) {$t_2$};
    \node[left] at (-0.2,1.9) {$t_3$};
    \node[left] at (-0.2,0.7) {$t_4$};
\end{tikzpicture}
\caption{Visualization of the bridge computed on the FFHQ dataset \citep{karras2019style} by our CuOT method (e2e encoder \citep{richardson2021encoding}).}
\label{fig:ffhq_e2e}
\end{figure}

Quantitative results for both transport directions are reported separately in Tables~\ref{tab:ffhq} and~\ref{tab:ffhq2} for the ALAE latent space, and in Tables~\ref{tab:ffhq_ablation1} and~\ref{tab:ffhq_ablation2} for the e2e latent space.
Table~\ref{tab:ffhq_avg} of the main paper reports the combined average across directions for the ALAE latent space. In this setting, while balanced OT approaches yield larger latent distances from the target distribution than WFR-based uOT methods, they achieve the second-best FID score, which better measures mode coverage in the reconstructed images. As shown in Figs.~\ref{fig:ffhq_a2c} and~\ref{fig:ffhq_c2a}, higher latent distances correlate with lower image quality.
Nevertheless, balanced OT methods still correctly capture target modes and provide full distribution coverage, producing images that broadly reflect the intended age group. In contrast, WFR-based uOT methods frequently fail to complete the transport effectively: despite achieving lower latent distances, they exhibit significantly higher FID scores, indicating that mass-growth bias causes distributional collapse in pixel space rather than faithful age transformation. Due to contact geometry, CuOT achieves state-of-the-art performance across all metrics, with improved target coverage and sample quality in both directions.
Qualitative examples of the full transport bridge computed by our CuOT method and each baseline are presented in Figures~\ref{fig:ffhq_contactubs}--\ref{fig:ffhq_otcfm}. These figures also illustrate how CuOT allocates more integration steps in high-density regions of the source distribution, following the finer geodesic discretization induced by the contact Wasserstein metric $g^C = \sqrt{K}\,g^{W_2}$ (see Appendix~\ref{app:geodesic-parameterization}). A compact side-by-side qualitative comparison at the endpoint is shown in Figures~\ref{fig:ffhq_a2c} and~\ref{fig:ffhq_c2a}.
In the higher-dimensional latent space produced by the e2e encoder, CuOT maintains its competitive advantage, exhibiting only a modest performance drop relative to the lower-dimensional setup. Among the baselines, OTCFM is the sole method that scales reasonably well, whereas other WFR-based uOT frameworks suffer severe performance degradation. These results demonstrate both the adaptability of our framework to distinct latent architectures and its superior scalability compared to state-of-the-art baselines. Figure \ref{fig:ffhq_e2e} visualize the bridge decoded from the the e2e late space.\looseness-1
\vspace{-3mm}
\begin{table}[H]
\footnotesize{
\centering
\caption{Adult $\to$ Child. FFHQ dataset \citep{karras2019style}, ALAE encoder \citep{pidhorskyi2020adversarial}. \looseness-1}
\resizebox{\textwidth}{!}{
\begin{tabular}{lcccc}
\toprule
Method & FID & $\mathcal{W}_1$ & $\mathcal{W}_2$ & Mass \\
\midrule
OTCFM \citep{tong2023improving} & $\underline{6.3081_{\pm 0.0910}}$ & $14.3921_{\pm 0.1162}$ & $10.2690_{\pm 0.1007}$ & $0.6066_{\pm 0.0000}$ \\
TIGON \citep{sha2024reconstructing} & $23.1358_{\pm{0.7232}}$ & $16.3305_{\pm 0.1795}$ & $11.2321_{\pm 0.1468}$ & $0.8012_{\pm 0.0572}$ \\
DeepRUOT \citep{zhang2025learning} & $6.9551_{\pm{0.3167}}$ & $\underline{{13.7117}_{\pm 0.1228}}$ & $\underline{{9.7566}_{\pm 0.1198}}$ & $\underline{{0.0181}_{\pm 0.0067}}$ \\
VarRUOT \citep{sun2025variational} & $30.0459_{\pm{0.5180}}$ & $14.0375_{\pm 0.1266}$ & $10.0279_{\pm 0.1121}$ & $0.9190_{\pm 0.2783}$ \\
\textbf{CuOT (Ours)} & $\bm{5.9340_{\pm 0.1216}}$ & $\bm{12.4156_{\pm 0.0989}}$ & $\bm{8.8793_{\pm 0.0947}}$ & $\bm{0.0086_{\pm 0.0035}}$ \\
\bottomrule
\end{tabular}
\label{tab:ffhq}
}}
\end{table}
\vspace{-6mm}
\begin{table}[H]
\footnotesize{
\centering
\caption{Child $\to$ Adult. FFHQ dataset \citep{karras2019style}, ALAE encoder \citep{pidhorskyi2020adversarial}. \looseness-1}
\resizebox{\textwidth}{!}{
\begin{tabular}{lcccc}
\toprule
Method & FID & $\mathcal{W}_1$ & $\mathcal{W}_2$ & Mass \\
\midrule
OTCFM \citep{tong2023improving} & $\underline{4.7533_{\pm 0.0721}}$ & $14.5779_{\pm 0.1012}$ & $10.3942_{\pm 0.1186}$ & $12.4037_{\pm 0.0000}$ \\
TIGON \citep{sha2024reconstructing} & $20.6358_{\pm{0.1834}}$ & $16.5892_{\pm 0.1795}$ & $11.8064_{\pm 0.1510}$ & $0.7877_{\pm 0.0819}$ \\
DeepRUOT \citep{zhang2025learning} & $6.1756_{\pm{0.0435}}$ & $\underline{{13.8212}_{\pm 0.0952}}$ & $\underline{{9.8473}_{\pm 0.0917}}$ & $\underline{{0.0023}_{\pm 0.0096}}$ \\
VarRUOT \citep{sun2025variational} & $24.3997_{\pm{0.0093}}$ & $15.5000_{\pm 0.1263}$ & $11.0904_{\pm 0.1220}$ & $0.8507_{\pm 0.3332}$ \\
\textbf{CuOT (Ours)} & $\bm{4.4188_{\pm 0.0306}}$ & $\bm{12.3927_{\pm 0.1170}}$ & $\bm{8.8354_{\pm 0.1063}}$ & $\bm{0.0000_{\pm 0.0000}}$ \\
\bottomrule
\end{tabular}
\label{tab:ffhq2}
}}
\end{table}
\begin{table}[H]
\footnotesize{
\centering
\caption{Adult $\to$ Child. FFHQ dataset \citep{karras2019style}, e2e encoder \citep{richardson2021encoding}. \looseness-1}
\resizebox{\textwidth}{!}{
\begin{tabular}{lcccc}
\toprule
Method & FID & $\mathcal{W}_1$ & $\mathcal{W}_2$ & Mass \\
\midrule
OTCFM \citep{tong2023improving} & \underline{$10.2854_{\pm 0.2088}$} & \underline{$20.3649_{\pm 0.3954}$} & \underline{$16.0835_{\pm 0.19013}$} & $0.6066_{\pm 0.0000}$ \\
TIGON \citep{sha2024reconstructing} & $48.2374_{\pm 0.4570}$ & $29.02231_{\pm 0.3294}$ & $20.9152_{\pm 0.3492}$ & $1.8263_{\pm 0.1899}$ \\
DeepRUOT \citep{zhang2025learning} & $41.2531_{\pm 0.1992}$ & $28.0340_{\pm 0.2265}$ & $21.5633_{\pm 0.2068}$ & \underline{$0.0839_{\pm 0.0092}$} \\
VarRUOT \citep{sun2025variational} & $53.6016_{\pm 0.0724}$ & $30.5293_{\pm 0.2545}$ & $24.3990_{\pm 0.3351}$ & $1.0479_{\pm 0.3780}$ \\
\textbf{CuOT (Ours)} & $\bm{8.9612_{\pm 0.1035}}$ & $\bm{17.0263_{\pm 0.0935}}$ & $\bm{13.1258_{\pm 0.1593}}$ & $\bm{0.01983_{\pm 0.0148}}$ \\
\bottomrule
\end{tabular}
\label{tab:ffhq_ablation1}
}}
\end{table}
\begin{table}[H]
\footnotesize{
\centering
\caption{Child $\to$ Adult. FFHQ dataset \citep{karras2019style}, e2e encoder \citep{richardson2021encoding}. \looseness-1}
\resizebox{\textwidth}{!}{
\begin{tabular}{lcccc}
\toprule
Method & FID & $\mathcal{W}_1$ & $\mathcal{W}_2$ & Mass \\
\midrule
OTCFM \citep{tong2023improving} & \underline{$8.4863_{\pm 0.1204}$} & \underline{$18.3142_{\pm 0.2071}$} & \underline{$13.5799_{\pm 0.1860}$} & $12.4032_{\pm 0.0000}$ \\
TIGON \citep{sha2024reconstructing} & $42.3115_{\pm 0.3207}$ & $28.6456_{\pm 0.3737}$ & $20.1813_{\pm 0.3662}$ & $1.3548_{\pm 0.1207}$ \\
DeepRUOT \citep{zhang2025learning} & $38.1583_{\pm 0.1911}$ & $27.4958_{\pm 0.2104}$ & $19.8897_{\pm 0.2186}$ & \underline{$0.0631_{\pm 0.0114}$} \\
VarRUOT \citep{sun2025variational} & $55.6621_{\pm 0.0875}$ & $32.1010_{\pm 0.2637}$ & $24.7603_{\pm 0.2958}$ & $1.2267_{\pm 0.4270}$ \\
\textbf{CuOT (Ours)} & $\bm{7.8332_{\pm 0.0909}}$ & $\bm{15.1472_{\pm 0.1326}}$ & $\bm{11.0913_{\pm 0.1222}}$ & $\bm{0.0107_{\pm 0.0140}}$ \\
\bottomrule
\end{tabular}
\label{tab:ffhq_ablation2}
}}
\end{table}
\vspace{-5mm}
\vspace{-7mm}
\vspace{-7mm}
\vspace{-7mm}

\subsection{Image-to-Video Experiment}
\label{app:image-to-video}
We evaluate the scalability of CuOT on temporal video reconstruction using the \href{https://www.crcv.ucf.edu/data/UCF101.php}{UCF101} dataset \citep{soomro2012ucf101}, a benchmark of $101$ human action classes recorded at $320 \times 240$ pixel resolution. We select three action classes: \textit{BreastStroke}, \textit{BlowingCandles}, and \textit{CuttingInKitchen}, each comprising respectively of $101$, $105$ and $104$ videos. Within each class, the temporal sequence of frames defines the reference marginal distributions $\{\rho^m\}_{m=0}^{M}$, with $M = 60$ intermediate marginals corresponding to uniformly sampled discrete time steps along the video.
Each marginal $\rho^m$ is thus represented by the ensemble of frames $\{x^{m}_n\}$ at time step $m$ across all videos in the class, providing the empirical distribution at that point in time. This formulation casts video reconstruction as a multi-marginal transport problem. 
Because the frame count per video remains constant across time steps, mass variation in this setting is negligible. Consequently, the primary objective of this experiment is not to test CuOT's mass-unbalancing capabilities, as in our other benchmark evaluations, but rather to demonstrate its scalability and stability when interpolating across a large number of marginal distributions in high-dimensional, structured data spaces.
\looseness-1

To mitigate computational overhead, all solvers are trained directly within the latent space of pre-trained autoencoders. To evaluate method performance across varying latent space topologies, we consider two distinct encoder backbones: (i) the Deep Compression Autoencoder \citep{chen2025deep}, which maps each $320 \times 240$ frame into a compact $32 \times 8 \times 8 = 2048$-dimensional feature map, and (ii) the TiTok encoder \citep{yu2024an}, which compresses frames into a $128 \times 32$ ($4096$-dimensional) sequence. Although TiTok yields a higher-dimensional representation, flattening the 2D spatial grid structure simplifies feature alignment for the transport networks, allowing us to maintain identical model capacities across both setups. During inference, an initial frame is encoded into the latent space, propagated along the predicted geodesic via the learned transport map, and decoded at each time step to reconstruct the video sequence. In both settings, we set the number of integration steps to $K = 180$, providing $K/M = 3$ discretization steps per observed marginal interval to ensure high temporal resolution. Full network and training configurations are detailed in Table~\ref{tab:hyperparameters}.

This experiment poses two distinct challenges beyond those of the FFHQ pairing task. First, the high ambient dimensionality ($D = 2048$) stresses scalability, and the $32 \times 8 \times 8$ grid structure preserves spatial information, meaning the data do not lie in a simple vector space.
Second, the large number of intermediate marginals ($M = 60$) requires the learned transport map to remain consistent across a long temporal horizon, preventing error accumulation along the predicted trajectory.
These conditions expose the structural limitations of WFR-based methods, whose growth-rate bias compounds across marginals and leads to trajectory collapse in this high-dimensional regime.

We report three metrics. The FID is computed in
the ambient pixel space after decoding, comparing the full set of reference frames against the full set of reconstructed frames across the entire video. We use the standard InceptionV3 feature extractor (2048-dimensional pool layer), following the formulation in equation \eqref{eq:fid} (Appendix~\ref{app:dataset-pairing}).
This metric measures the ability of each method to recover the distributional support and diversity of the video content. The latent Wasserstein distances $\mathcal{W}_1$ and $\mathcal{W}_2$ are evaluated time-step by time-step in the latent space: for each $m \in \{0,\ldots,M\}$, the empirical distribution of predicted latent codes is compared against the reference marginal $\rho^m$, and the reported value is averaged over all marginals. This per-timestep criterion directly probes whether each method correctly tracks the sequence of marginal distributions along the trajectory, penalising both early divergence and late-stage drift.
The mass metric was intentionally omitted from the evaluation as the transport problem does not present mass variation.

\begin{figure}[H]
\centering
\begin{tikzpicture}
    \node[anchor=south west, inner sep=0] (grid) at (0,0)
    {\includegraphics[width=0.86\linewidth]{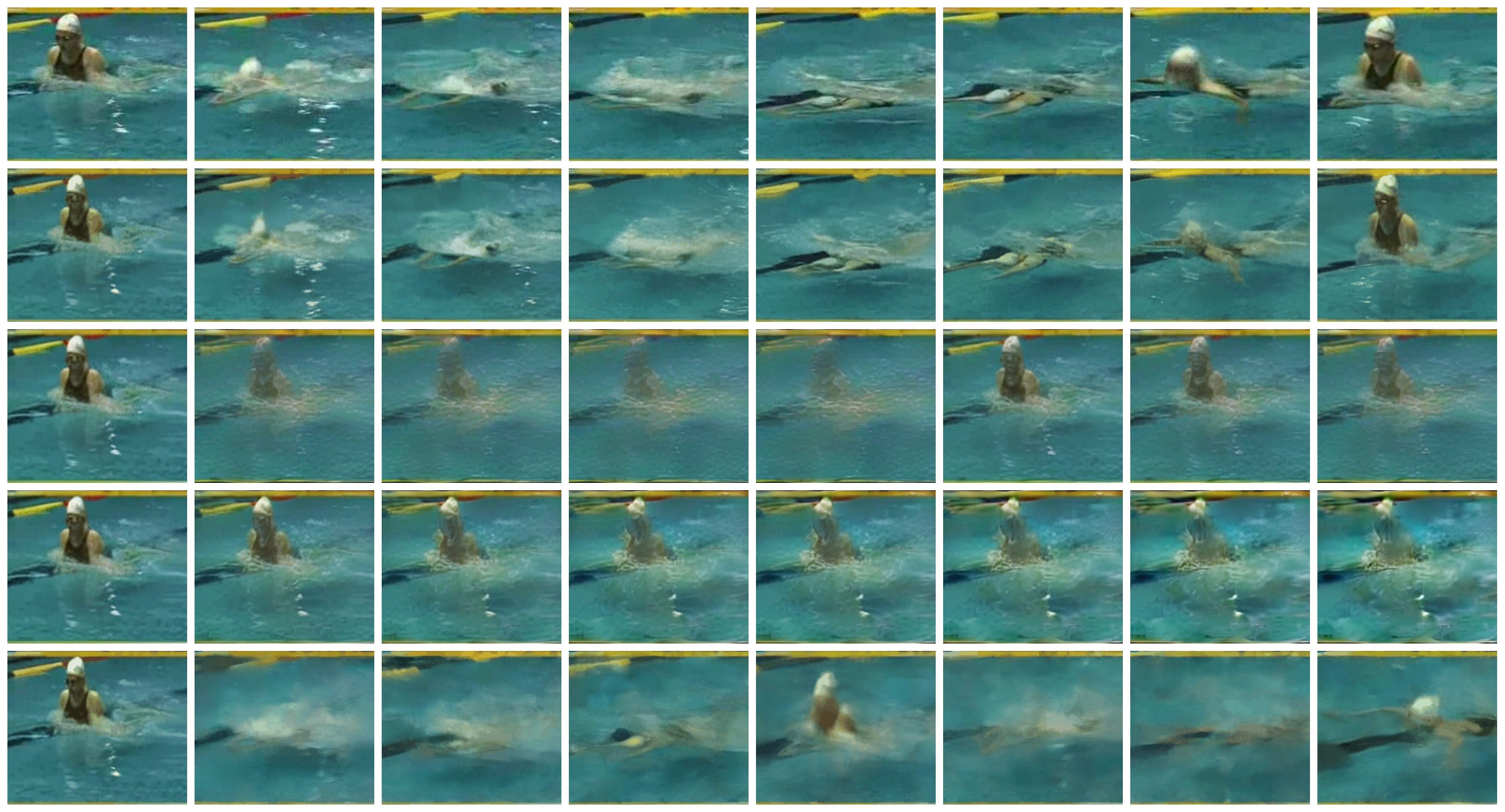}};
    \node[left] at (-0.0,5.7) { \footnotesize Reference};
    \node[left] at (-0.0,4.5) {{\footnotesize \textbf{CuOT (Ours)}}};
    \node[left] at (-0.0,3.2) {\footnotesize VarRUOT};
    \node[left] at (-0.0,1.9) {\footnotesize DeepRUOT};
    \node[left] at (-0.0,0.7) {\footnotesize OTCFM};
\end{tikzpicture}
\vspace{-2mm}
\caption{\textit{BreastStroke} class, UCF101 \citep{soomro2012ucf101}, CB encoder \citep{chen2025deep}.}
\label{fig:ucf101_breast}
\end{figure}

\begin{figure}[H]
\centering
\begin{tikzpicture}
    \node[anchor=south west, inner sep=0] (grid) at (0,0)
    {\includegraphics[width=0.86\linewidth]{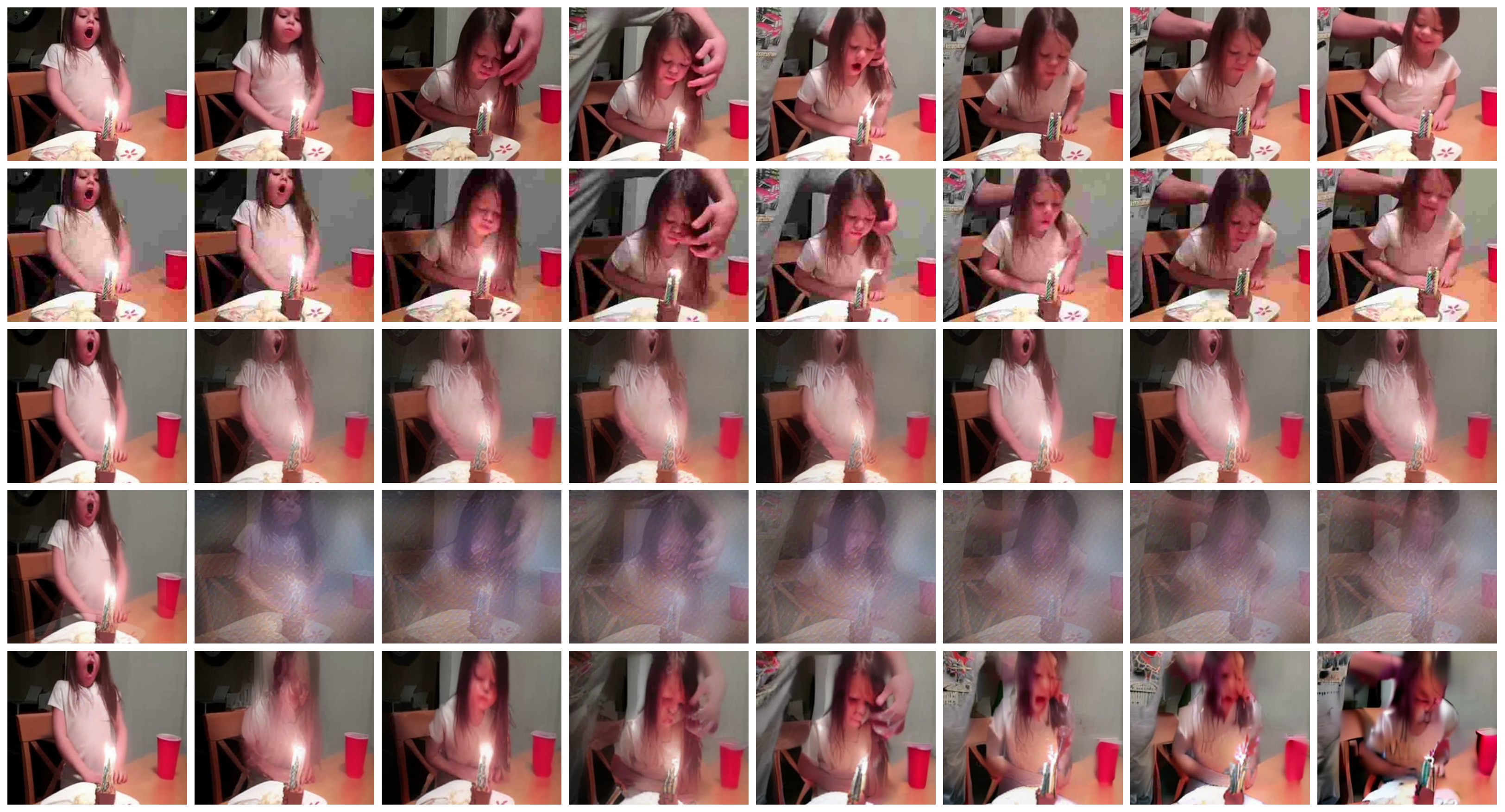}};
    \node[left] at (-0.0,5.7) { \footnotesize Reference};
    \node[left] at (-0.0,4.5) {{\footnotesize \textbf{CuOT (Ours)}}};
    \node[left] at (-0.0,3.2) {\footnotesize VarRUOT};
    \node[left] at (-0.0,1.9) {\footnotesize DeepRUOT};
    \node[left] at (-0.0,0.7) {\footnotesize OTCFM};
\end{tikzpicture}
\vspace{-2mm}
\caption{\textit{BlowingCandles} class, UCF101 \citep{soomro2012ucf101}, CB encoder \citep{chen2025deep}.}
\label{fig:ucf101_candles}
\end{figure}

\begin{figure}[H]
\centering
\includegraphics[width=\textwidth]{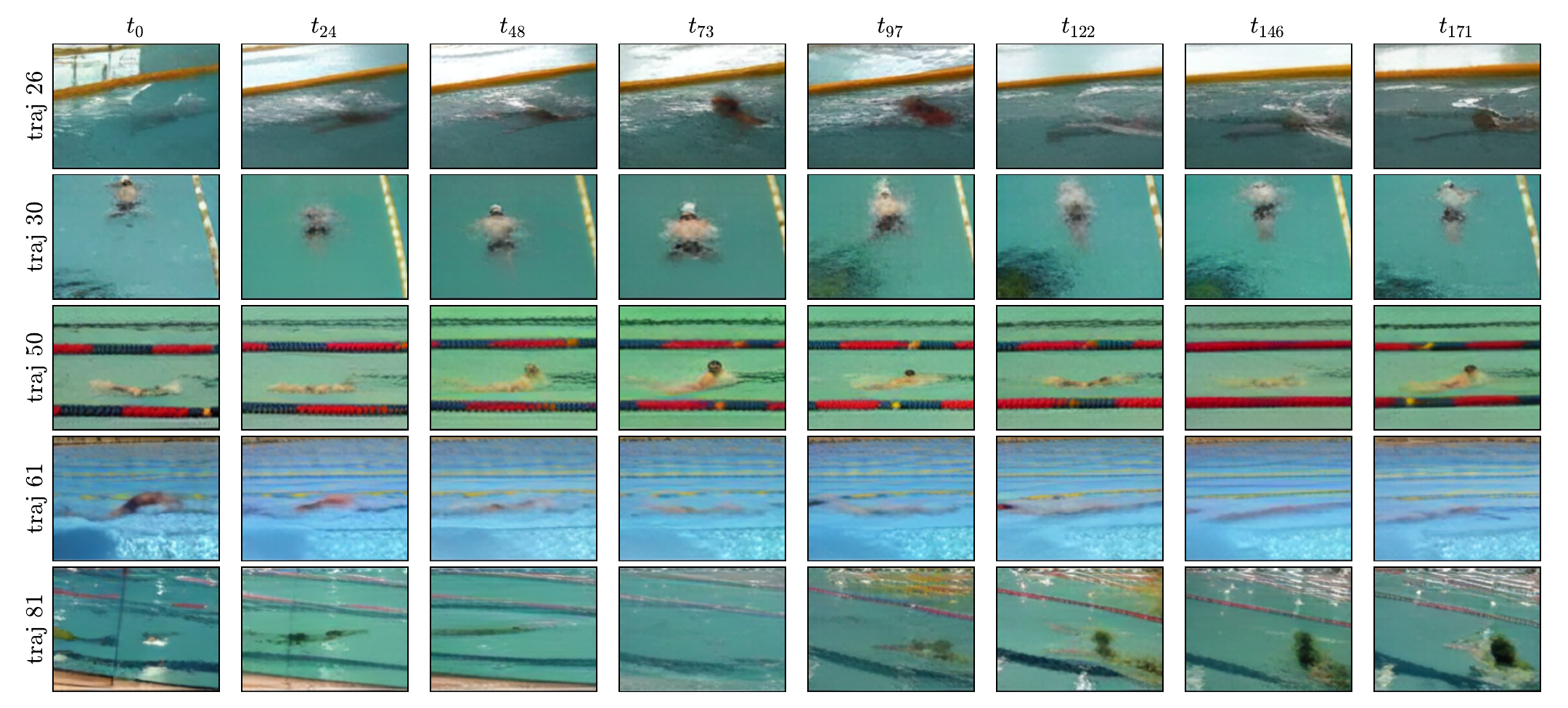}
\vspace{-2mm}
\caption{\textit{BreastStroke} class, UCF101 \citep{soomro2012ucf101}, TiTok encoder \citep{yu2024an}.}
\label{fig:ucf101_titok}
\end{figure}

Figures~\ref{fig:ucf101_breast}--\ref{fig:ucf101_candles} display uniformly sampled frame sequences reconstructed from the CB latent space across each action class. Tables~\ref{tab:breast}--\ref{tab:cutting} detail the corresponding per-class quantitative results, while aggregate metrics are summarized in Table~\ref{tab:ucf101}. Additionally, Table~\ref{tab:titok} presents the ablation results using the TiTok latent space, with corresponding decoded video trajectories across various initial frames visualized in Figure~\ref{fig:ucf101_titok}.

CuOT consistently achieves the best scores across all metrics, classes and latent representations, by a wide margin. Given the stringent requirement of aligning many intermediate marginals, this highlights the stability and effectiveness of the proposed geodesic-based resolution algorithm in high-dimensional multi-marginal settings. WFR-based baselines are typically not demonstrated on benchmarks with dimensionalities higher than $50$, and they struggle significantly on this task. In this class of methods,  \citet{zhang2025learning} represents the strongest option, but still performs roughly an order of magnitude worse. 
This gap likely reflects the difficulty of jointly parameterizing multiple variables and the need to enforce the HJB optimality equation in the loss (notoriously unstable \citep{buzun2026hota}), rather than directly simulating transport (provable stable in Appendix \ref{app:stability}). In this class we also find the method performing worst, \citet{sun2025variational}. Its strict enforcement of $v^t(x) = \nabla_x z^t(x)$ yields a noisy and unstable signal in the high-dimensional latent spaces, causing collapse to near-static trajectories where samples remain close to $\rho^0$ and exhibit minimal temporal evolution.
\citet{tong2023improving} has been demonstrated to scale to higher-dimensional datasets like CIFAR-10, but it is fundamentally optimized for two marginals. Because it relies heavily on flow matching, adapting it to multiple marginals requires iterative reference refinement techniques to update the ground-truth matching along the learned path. This introduces significant computational overhead and drastically slows down convergence compared to our framework. While computationally efficient (similarly to \citet{sha2024reconstructing}), its reliance on density estimation rather than explicit OT matching does not scale: without re-anchoring to reference marginals, discretization errors accumulate over the $M = 60$ steps, leading to drift and higher variance.

CuOT consistently achieves the best performance across all metrics, action classes, and latent representations by a wide margin. Given the stringent requirement of aligning numerous intermediate marginals, these results highlight the stability and effectiveness of our proposed geodesic-based resolution algorithm in high-dimensional multi-marginal settings. 
Existing WFR-based uOT baselines are typically evaluated only on low-dimensional benchmarks ($\le 50$ dimensions) and struggle significantly on this task. Among these, DeepRUOT \citep{zhang2025learning} represents the strongest option, yet it performs roughly an order of magnitude worse than CuOT. This performance gap stems from the inherent difficulty of jointly parameterizing multiple field variables and enforcing the HJB optimality loss, a process known to be notoriously unstable \citep{buzun2026hota}, as opposed to directly simulating transport maps (provably stable; see Appendix~\ref{app:stability}). Within the same family, VarRUOT \citep{sun2025variational} exhibits the poorest performance: its strict constraint $v^t(x) = \nabla_x z^t(x)$ produces noisy gradient signals in high-dimensional latent spaces, causing trajectories to collapse into near-static states that remain anchored close to $\rho^0$ with minimal temporal evolution.
Flow-matching baselines like OTCFM \citep{tong2023improving} have been shown to scale to higher-dimensional image datasets (e.g., CIFAR-10), but they are fundamentally optimized for two-marginal problems. Adapting OTCFM to multi-marginal trajectories requires iterative reference refinement to update ground-truth matchings along the learned path, introducing heavy computational overhead and substantially slowing convergence. \looseness-1

\begin{table}[H]
\centering
\footnotesize{
\caption{\textit{BreastStroke} class, UCF101 \citep{soomro2012ucf101}, CB encoder \citep{chen2025deep}.}
\begin{tabular}{lccc}
\toprule
Method & FID & $\mathcal{W}_1$ & $\mathcal{W}_2$ \\
\midrule
OTCFM    & $186.4736_{\pm 2.1483}$ & $21.6347_{\pm 1.5824}$ & $14.8247_{\pm 1.2436}$ \\
TIGON    & $318.4247_{\pm 2.8364}$ & $26.8347_{\pm 1.9628}$ & $17.4236_{\pm 1.4827}$ \\
DeepRUOT & $\underline{132.9642_{\pm 1.4471}}$ & $\underline{17.9528_{\pm 1.0326}}$ & $\underline{12.5841_{\pm 0.8735}}$ \\
VarRUOT  & $310.7648_{\pm 2.9365}$ & $24.1739_{\pm 1.8421}$ & $15.8127_{\pm 1.2638}$ \\
\textbf{CuOT (Ours)} & $\bm{16.4875_{\pm 0.6842}}$ & $\bm{9.5832_{\pm 0.5217}}$ & $\bm{7.3124_{\pm 0.4631}}$ \\
\bottomrule
\end{tabular}
\label{tab:breast}
}
\end{table}
\vspace{-4mm}
\vspace{-4mm}
\begin{table}[H]
\centering
\footnotesize{
\caption{\textit{BlowingCandles} class, UCF101 \citep{soomro2012ucf101}, CB encoder \citep{chen2025deep}..}
\begin{tabular}{lccc}
\toprule
Method & FID & $\mathcal{W}_1$ & $\mathcal{W}_2$ \\
\midrule
OTCFM    & $224.3847_{\pm 2.7624}$ & $28.4736_{\pm 2.0847}$ & $19.3628_{\pm 1.6247}$ \\
TIGON    & $352.1736_{\pm 3.9248}$ & $36.4736_{\pm 2.5847}$ & $23.1847_{\pm 1.8436}$ \\
DeepRUOT & $\underline{143.9126_{\pm 1.9637}}$ & $\underline{23.4632_{\pm 1.5426}}$ & $\underline{16.7315_{\pm 1.2084}}$ \\
VarRUOT  & $340.6289_{\pm 3.8427}$ & $33.7814_{\pm 2.4365}$ & $21.9438_{\pm 1.7421}$ \\
\textbf{CuOT (Ours)} & $\bm{20.4368_{\pm 0.8135}}$ & $\bm{13.2147_{\pm 0.7426}}$ & $\bm{9.3826_{\pm 0.5184}}$ \\
\bottomrule
\end{tabular}
\label{tab:candles}
}
\end{table}
\vspace{-4mm}
\vspace{-4mm}
\begin{table}[H]
\centering
\footnotesize{
\caption{\textit{CuttingInKitchen} class, UCF101 \citep{soomro2012ucf101}, CB encoder \citep{chen2025deep}..}
\begin{tabular}{lccc}
\toprule
Method & FID & $\mathcal{W}_1$ & $\mathcal{W}_2$ \\
\midrule
OTCFM    & $214.7362_{\pm 2.5824}$ & $26.8436_{\pm 1.9473}$ & $17.9473_{\pm 1.4836}$ \\
TIGON    & $338.9362_{\pm 3.2647}$ & $29.7647_{\pm 2.1436}$ & $19.8347_{\pm 1.6827}$ \\
DeepRUOT & $\underline{137.9231_{\pm 1.7326}}$ & $\underline{20.5843_{\pm 1.2874}}$ & $\underline{13.8417_{\pm 1.0635}}$ \\
VarRUOT  & $325.7062_{\pm 3.2148}$ & $26.3447_{\pm 2.1046}$ & $17.3431_{\pm 1.5627}$ \\
\textbf{CuOT (Ours)} & $\bm{18.2106_{\pm 0.8043}}$ & $\bm{10.8039_{\pm 0.6235}}$ & $\bm{7.8058_{\pm 0.5172}}$ \\
\bottomrule
\end{tabular}
\label{tab:cutting}
}
\end{table}
\vspace{-4mm}
\begin{figure}[H]
\centering
\begin{tikzpicture}
    \node[anchor=south west, inner sep=0] (grid) at (0,0)
    {\includegraphics[width=0.86\linewidth]{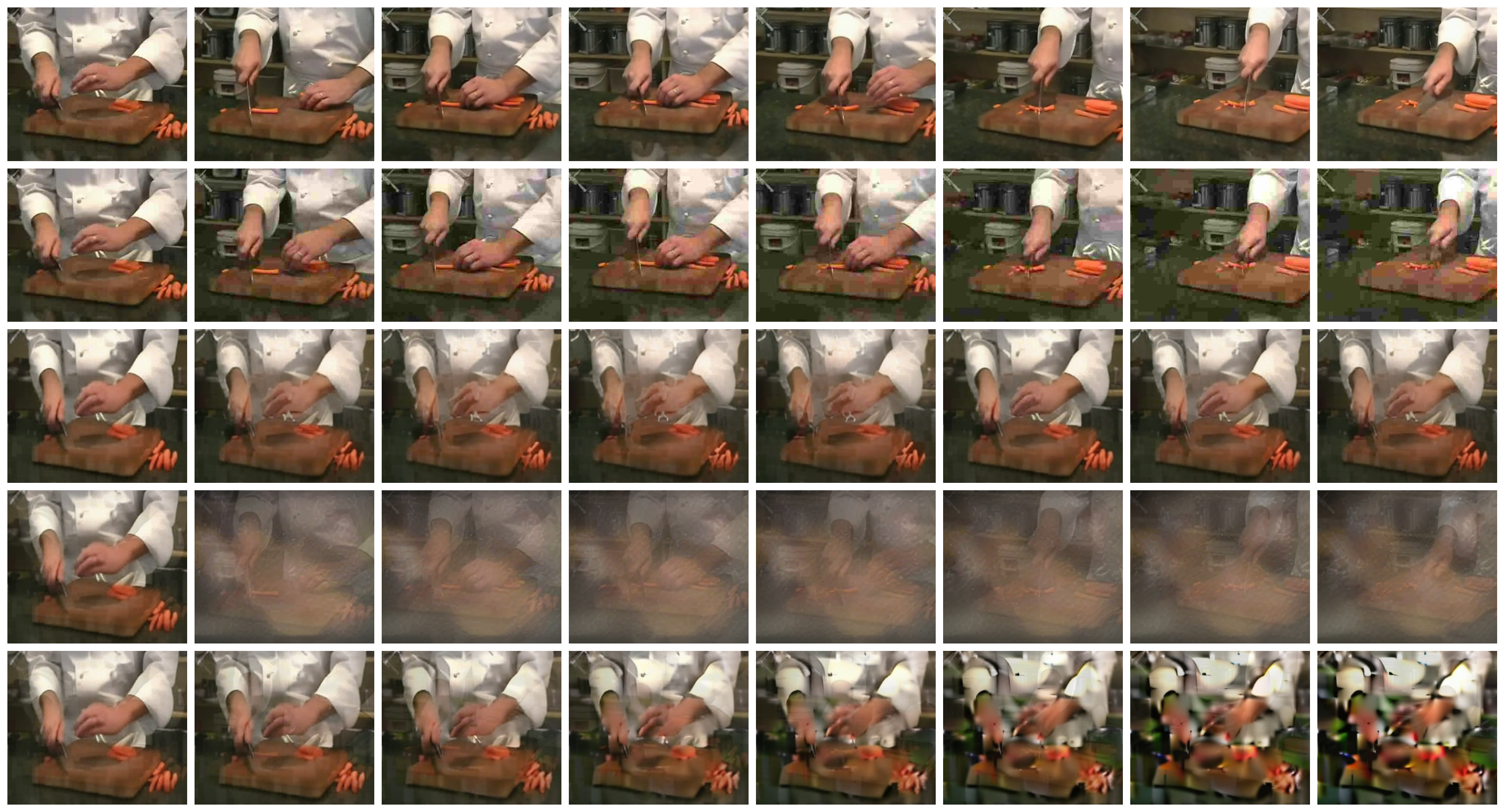}};
    \node[left] at (-0.0,5.7) { \footnotesize Reference};
    \node[left] at (-0.0,4.5) {{\footnotesize \textbf{CuOT (Ours)}}};
    \node[left] at (-0.0,3.2) {\footnotesize VarRUOT};
    \node[left] at (-0.0,1.9) {\footnotesize DeepRUOT};
    \node[left] at (-0.0,0.7) {\footnotesize OTCFM};
\end{tikzpicture}
\vspace{-2mm}
\caption{\textit{CuttingInKitchen} class, UCF101 \citep{soomro2012ucf101}, CB encoder \citep{chen2025deep}.}
\label{fig:ucf101_cutting}
\end{figure}
\begin{table}[H]
\centering
\footnotesize{
\caption{\textit{BreastStroke} class, UCF101 \citep{soomro2012ucf101}, TiTok encoder \citep{yu2024an}.}
\begin{tabular}{lccc}
\toprule
Method & FID & $\mathcal{W}_1$ & $\mathcal{W}_2$ \\
\midrule
OTCFM    & \underline{$92.4136_{\pm 2.9927}$} & \underline{$18.2705_{\pm 1.8718}$} & \underline{$15.7947_{\pm 1.3327}$} \\
TIGON    & $361.0779_{\pm 3.0314}$ & $39.5178_{\pm 2.0123}$ & $25.0515_{\pm 2.0329}$ \\
DeepRUOT & $103.87_{\pm 2.1189}$ & $19.3748_{\pm 1.9142}$ & $17.7740_{\pm 1.0970}$ \\
VarRUOT  & $312.1575_{\pm 3.4569}$ & $28.2140_{\pm 2.5769}$ & $23.5365_{\pm 1.5311}$ \\
\textbf{CuOT (Ours)} & $\bm{25.5301_{\pm 1.0747}}$ & $\bm{11.4517_{\pm 0.6784}}$ & $\bm{7.1934_{\pm 0.5704}}$ \\
\bottomrule
\end{tabular}
\label{tab:titok}
}
\end{table}
\vspace{-4mm}
\newpage
\section{Broader Impacts}\label{app:impacts}

CuOT introduces contact geometry as a unifying framework for optimal transport, bridging physical transport (described by Wasserstein geometry) and the information-theoretic structure of data distributions (described by Fisher--Rao geometry). This constitutes a significant theoretical advance in the study of the geometry of generative processes, as prior Wasserstein--Fisher--Rao approaches typically interpolate between these distinct regimes, whereas our method provides a generalized lift that avoids restrictive assumptions.

Our framework is particularly suited for modeling stochastic dynamics in the natural sciences. 
By removing the need for expert-specified growth priors~\citep{chizat2022trajectory, lavenant2024toward, pariset2023unbalanced, ventre2023trajectory, schiebinger2019optimal}, CuOT facilitates more accurate reconstruction of cell-fate dynamics and disease progression from sparse, static observations. 
This reduces reliance on costly longitudinal single-cell experiments and supports faster progress in areas such as regenerative medicine and oncology by extracting high-resolution developmental information from standard datasets.
The proposed framework improves unpaired domain adaptation, particularly in translating synthetic data to real clinical imagery \citep{wang2024structure, bortoli2021diffusion}. 
This enhances the robustness and generalizability of diagnostic models across varying acquisition conditions and clinical environments. 
CuOT also ensures that underrepresented data modes are preserved, potentially reducing bias in generative models used for decision-making or synthetic data augmentation.
While CuOT improves reliability, deployment in medical settings requires careful validation to ensure that the decoupling of mass growth does not introduce artifacts that could affect clinical decisions.

On the implementation front, CuOT efficiency reduces the total computational budget, and the associated carbon footprint required for large-scale distribution alignment tasks compared to more resource-intensive quadratic-complexity alternatives, summarized in Table~\ref{tab:comparison}.

\section{Limitations and Future Directions}\label{app:limitations}

The experimental validation of CuOT has been conducted exclusively using MLP architectures for both the transport network $v_\theta^t$ and the growth network $z_\theta^t$. MLPs treat input coordinates as mutually independent, and are therefore well-suited to the vector-valued data encountered in the single-cell experiments and in highly-compressed image latent spaces. However, they do not explicitly capture inter-element dependencies, such as spatial correlations in images or sequential structure in time series, that are critical in domains where the data geometry cannot be faithfully represented by a coordinate-wise factorization. Architectures such as Transformers and U-Nets represent the state of the art in structured sequence modeling. While the contact-geometric formulation of CuOT is architecture-agnostic and does not preclude the use of such backbones, their integration with the geodesic discretization scheme of Eqs.~\eqref{eq:discrete_geodesic}--\eqref{eq:discrete_optimality} has not been investigated, and the scalability and reconstruction quality of CuOT in these high-structure regimes remains to be studied.

As an Optimal Transport (OT) formulation designed for interpolating between distribution pairs and multiple marginals, CuOT is not directly comparable to specialized image-to-video models \citep{ni2023conditional, wang2024framebridge}. Interpolation-based approaches, which treat each video frame as a discrete sample, are inherently less efficient than architectures that represent an entire video as a single latent point. By utilizing the initial image as a conditioning factor, these specialized models optimize higher-dimensional but less constrained generative dynamics, resulting in a more scalable optimization framework.

CuOT enables the adaptation of generative sampling steps to the varying data density of the interpolated marginals, mitigating overshooting effects along the transport path.
An interesting structural parallel arises with recent work investigating the geometry of diffusion-based generation~\citep{karczewski2026the}, which operates in a more restricted setting by connecting data points within a single trained model via geodesics under the Fisher--Rao metric, rather than bridging arbitrary unpaired marginals. Both frameworks share the underlying principle that step sizes should reflect the local geometry of the data distribution rather than ambient Euclidean structure. Nevertheless, an explicit characterization of the link between the contact geometry introduced here and the Fisher--Rao geometry, and more broadly, the relationship between contact dynamics on the probability manifold and information-geometric metrics on latent spaces, remains an open direction for future work.
\end{document}